\documentclass[10pt,journal,compsoc]{IEEEtran}

\ifCLASSOPTIONcompsoc
  \usepackage[nocompress]{cite}
\else
  \usepackage{cite}
\fi

\ifCLASSINFOpdf

\else

\fi

\usepackage{times}
\usepackage{epsfig}
\usepackage{graphicx}
\usepackage{amsmath}
\usepackage{amssymb}

\usepackage[utf8]{inputenc} 
\usepackage[T1]{fontenc}    
\usepackage{url}            
\usepackage{booktabs}       
\usepackage{amsfonts}       
\usepackage{nicefrac}       
\usepackage{microtype}      
\usepackage{algorithm}
\usepackage{algorithmic}
\usepackage{array}
\usepackage{bm}
\usepackage{multirow}
\usepackage{color}

\usepackage{caption}
\usepackage{subcaption}

\def\E{{\rm E}}

\def\KL{{\rm KL}}

\def\P{P_{\rm data}}

\def\tY{\tilde{Y}}
\def\tn{\tilde{n}}
\def\hY{\hat{Y}}

\begin{document}

\title{Learning Energy-Based 3D Descriptor Networks \\ for Volumetric Shape Synthesis and Analysis}

\title{Generative VoxelNet: Learning Energy-Based Models for 3D Shape Synthesis and Analysis}

\author{Jianwen Xie *, Zilong Zheng *, Ruiqi Gao, Wenguan Wang, \\ Song-Chun Zhu,~\IEEEmembership{Fellow, IEEE,}
        and~Ying Nian Wu
\IEEEcompsocitemizethanks{
\IEEEcompsocthanksitem J. Xie is with the Cognitive Computing Lab, Baidu Research, Bellevue, WA 98004, USA. E-mail: jianwen@ucla.edu
\IEEEcompsocthanksitem Z. Zheng is with the Department of Computer Science, University of California, Los Angeles, CA 90095, USA. E-mail: z.zheng@ucla.edu
\IEEEcompsocthanksitem R. Gao is with the Department of Statistics, University of California, Los Angeles, CA 90095, USA. E-mail: ruiqigao@ucla.edu
\IEEEcompsocthanksitem W. Wang is with ETH Zürich, 8092 Zürich, Switzerland. E-mail: wenguanwang.ai@gmail.com
\IEEEcompsocthanksitem S.-C. Zhu is with Tsinghua University and Peking University, Beijing, China. E-mail: sczhu@stat.ucla.edu
\IEEEcompsocthanksitem Y. N. Wu is with the Department of Statistics, University of California, Los Angeles, CA 90095, USA. E-mail: ywu@stat.ucla.edu 
\IEEEcompsocthanksitem * indicates equal contributions.
}
}

\markboth{}%
{Xie \MakeLowercase{\textit{et al.}}: Bare Demo of IEEEtran.cls for Computer Society Journals}

\IEEEtitleabstractindextext{%

\begin{abstract}
3D data that contains rich geometry information of objects and scenes is valuable for understanding 3D physical world. With the recent emergence of large-scale 3D datasets, it becomes increasingly crucial to have a powerful 3D generative model for 3D shape synthesis and analysis. This paper proposes a deep 3D energy-based model to represent volumetric shapes. The maximum likelihood training of the model follows an ``analysis by synthesis'' scheme. The benefits of the proposed model are six-fold: first, unlike GANs and VAEs, the model training does not rely on any auxiliary models; second, the model can synthesize realistic 3D shapes by Markov chain Monte Carlo (MCMC); third, the conditional model can be applied to 3D object recovery and super resolution; fourth, the model can serve as a building block in a multi-grid modeling and sampling framework for high resolution 3D shape synthesis; fifth, the model can be used to train a 3D generator via MCMC teaching; sixth, the unsupervisedly trained model provides a powerful feature extractor for 3D data, which is useful for 3D object classification. Experiments demonstrate that the proposed model can generate high-quality 3D shape patterns and can be useful for a wide variety of 3D shape analysis.
\end{abstract}

\begin{IEEEkeywords}
Deep generative models; Energy-based models; Langevin dynamics; Volumetric shape synthesis; Generative VoxelNet; Cooperative learning; Multi-grid sampling.
\end{IEEEkeywords}}

\maketitle

\IEEEdisplaynontitleabstractindextext

\IEEEpeerreviewmaketitle

\IEEEraisesectionheading{\section{Introduction}\label{sec:introduction}}

\subsection{Background and motivation}
\IEEEPARstart{U}{nderstanding} 3D world is a core mission of computer vision as many vision-based perception systems in a wide range of real-world application scenarios, such as autonomous navigation \cite{geiger2012we}, home robotics \cite{oh2002development}, and augmented/virtual reality \cite{azuma1997survey, azuma2001recent}, immensely rely on the ability to accurately parse, reconstruct and interact with the physical 3D environment, as well as the ability to simultaneously detect, recognize, reason, and predict the behaviors of agents within the 3D environment.

The rapid innovations and advancements in 3D sensing technologies, especially 3D data acquisition devices (e.g., structured-light 3D scanners \cite{geng2011structured} and time-of-flight cameras \cite{hansard2012time}), as well as the recent emergence of large-scale 3D datasets (e.g., NYU depth\cite{silberman2012indoor}, SUN3D \cite{xiao2013sun3d}, SUN RGB-D\cite{song2015sun}, ModelNet \cite{wu20153d},  and ShapeNet \cite{chang2015shapenet}), have tremendously facilitated the accessibility of 3D data. 
With full geometry information of the 3D sensed objects and scenes, some computer vision tasks, such as classification, localization, segmentation, and alignment, will become much easier than in 2D image space, therefore 3D data is a very useful and valuable asset in the field of computer vision. The amazing success of deep learning models along with the increased availability of 3D data have been encouraging the computer vision community to investigate the generalization of deep learning technologies to 3D data. 

There exist various types of representations of 3D data, such as multi-view data, RGB-D data, volumetric data, 3D point clouds, and 3D meshes. Different types of 3D representations may have different data structures and geometric properties \cite{ahmed2018deep}. In this paper, we only focus on volumetric representation of 3D shape, where each 3D object is characterized as a regular voxel grid in the three-dimensional space. The value of a voxel represents opacity on a scale of [0, 1], with 0 indicating ``fully transparent'' and 1 indicating ``fully opaque''.

\subsection{Statistical models of 3D volumetric  shapes}
Recently, 
some interesting attempts \cite{girdhar2016learning, su2015multi,qi2016volumetric} have been made on object recognition and synthesis based on volumetric 3D shape data. From the perspective of statistical modeling, the existing 3D models can be grouped into two main categories: (1) 3D discriminators, such as Voxnet \cite{maturana2015voxnet}, which aim to learn a mapping from 3D voxel input to semantic labels for the purpose of 3D object classification and recognition, and (2) 3D generators, such as 3D-GAN\cite{3dgan}, which are in the form of latent variable models that assume that the 3D voxel signals are generated by some latent variables. The training of discriminators usually relies on big data with annotations and is accomplished by a direct minimization of the prediction errors, while the training of the generators  learns a mapping from the latent space to 3D voxel data space. 

The generator model, while useful for synthesizing 3D shape patterns, involves a challenging inference step (i.e., sampling from the posterior distribution) in maximum likelihood learning, therefore variational inference \cite{kingma2013auto} and adversarial learning \cite{goodfellow2014generative, radford2015unsupervised, 3dgan} methods are commonly used, where an extra network is incorporated into the learning algorithm to get around the difficulty of the posterior inference. 

The past few years have witnessed impressive progress on developing  discriminator models and generator models for 3D shape data, however, there has not been much work in the literature on modeling 3D shape data based on energy-based models. \cite{zhu2003statistical} calls this type of models the descriptor models, because the models describe the data based on bottom-up descriptive features learned from the data. The focus of the present paper is to develop a volumetric 3D energy-based model for voxelized shape data. It can be considered an alternative to 3D-GAN \cite{3dgan} for 3D shape generation. 

\subsection{Energy-based models for 3D volumetric shapes}
Specifically, we present a novel framework for probabilistic modeling of volumetric shape patterns  by combining the merits of energy-based model \cite{Lecun2006} and 3D voxel convolutional neural network \cite{maturana2015voxnet}. The model is a probability density function directly defined on voxelized shape signal, and the model is in the form of an energy-based model, where the feature statistics or the energy function is defined by a bottom-up 3D voxel ConvNet (VoxelNet) that maps the 3D voxel signal to the features. We call the proposed model the generative VoxelNet, because the model learns the VoxelNet in a generative manner.  

The training of the proposed model via maximum likelihood estimation follows an ``analysis by synthesis'' scheme \cite{grenander2007pattern}, where, within each iteration of the learning algorithm, we synthesize examples by Markov chain Monte Carlo (MCMC) sampling method, and then update the model parameters based on the difference between the observed training examples and the synthesized examples. This process can be interpreted as an alternating mode seeking and mode shifting process.

 Different from the variational inference or adversarial learning, the proposed model does not need to incorporate an extra inference network or an adversarial discriminator in the learning process.  The learning and sampling processes are guided by the same set of parameters of a single model, which makes it a particularly natural and statistically rigorous framework for probabilistic 3D shape modeling. 

Modeling 3D shape data by a probability density function provides distinctive advantages: First, it is able to synthesize realistic 3D shape patterns by sampling examples from the distribution via MCMC, such as Langevin dynamics \cite{neal2011mcmc}. Second, the model can be modified to be a conditional model, which is useful for 3D object recovery and 3D object super resolution. Specifically, a conditional probability density function that maps the corrupted (or low resolution) 3D object to the recovered (or high resolution) 3D object is trained, and then the 3D recovery (or 3D super resolution) can be achieved by sampling from the learned conditional distribution given the corrupted or low resolution 3D object as the conditional input. Third, the model can be used in a cooperative training scheme \cite{xie2016cooperative}, as opposed to adversarial training and variational inference, to train a 3D generator model via MCMC teaching. The training of 3D generator in such a scheme is stable and does not encounter mode collapse issue. Fourth, the model can be used in a multi-grid modeling and sampling framework \cite{gao2018learning} for high resolution 3D synthesis. Fifth, the model is useful for semi-supervised learning. After learning the model from unlabeled data, the learned features can be used to train a classifier on the labeled data.

We show that the proposed energy-based generative VoxelNet can be used to synthesize realistic 3D shape patterns, and its conditional version is useful for 3D object recovery and 3D object super resolution. The multi-grid modeling and sampling of the generative VoxelNet can generate high resolution 3D patterns. The 3D generator trained by the generative VoxelNet in a cooperative scheme carries semantic information about 3D objects. The feature maps trained by the generative VoxelNet in an unsupervised manner are useful for 3D object classification.   

\subsection{Related work}

\textit{3D object synthesis.} Researchers in the fields of graphics and vision have studied the 3D object synthesis problems \cite{carlson1982algorithm, blanz1999morphable, chaudhuri2011probabilistic, kalogerakis2012probabilistic, xu2012fit}. For example, component-based 3D object synthesis methods, e.g., \cite{chaudhuri2011probabilistic} and \cite{kalogerakis2012probabilistic}, propose to use a graphical model for individual 3D objects that encodes semantic and geometric relationships among shape components. These models can be trained on a given collection of 3D shapes and then used to generate new objects by recombining the training object components. \cite{xu2012fit} proposes a set evolution method to evolve a set of diverse objects to fit user's preference for generating sets of 3D novel shapes. However, most of these object synthesis methods are nonparametric and based on meshes or skeletons, and they generate new patterns by retrieving and merging parts from an existing CAD database. The rapid development of generative models has also enabled researchers to utilize parametric model-based approaches, such as \cite{wu20153d,3dgan,tatarchenko2017octree,achlioptas2017learning,khan2019unsupervised,chen2019learning,huang20193d,gao2019sdm}, for 3D object synthesis. Our framework represents voxelized 3D objects by proposing an energy-based model, which is a parametric probability distribution defined on the 3D shape domain. The model can learn from a repository of 3D shapes. 3D object synthesis can be achieved by running MCMC, such as Langevin dynamics, to draw samples from the learned model, without borrowing components from the training shape repository. Since the space of the 3D shapes is of high dimension, the MCMC-based synthesis is very challenging, but conceptually novel.

\textit{Deep learning for volumetric data.} Recently, the vision community has witnessed the success of deep learning, and researchers have used  the models in the field of deep learning for the sake of voxel-based object synthesis and analysis. For example, 3D ShapeNets\cite{wu20153d} represents a 3D volumetric shape by a probability distribution of binary variables on a 3D voxel grid, using a convolutional deep belief network.\cite{maturana2015voxnet} and \cite{qi2016volumetric} adopt a supervised volumetric ConvNet for 3D object recognition. 
\cite{3dgan} proposes the 3D deep convolutional generative adversarial net (3D-GAN) to synthesize 3D objects by leveraging 3D voxel ConvNets (VoxelNets) and generative adversarial nets (GANs). \cite{tatarchenko2017octree} proposes a deep convolutional decoder that can generate high resolution 3D volumetric shapes in a computationally efficient way by using an octree representation. Building on the early work of \cite{tu2007learning}, recently \cite{jin2017introspective, lazarow2017introspective, lee2018wasserstein} have developed an introspective learning method to learn the energy-based model, where the energy function is discriminatively learned. A 3D version of Introspective Neural Networks for 3D volumetric shape modeling is studied in \cite{huang20193d}. 
Our 3D model is also powered by the VoxelNets. It incorporates a bottom-up VoxelNet that captures volumetric feature statistics of the 3D shapes for defining the probability density of shape in the 3D voxel grid, and learns the parameters of the VoxelNet by an ``analysis by synthesis'' scheme.

\textit{Energy-based models for synthesis.} 
The energy-based model specifies a probability distribution of the signal, which is based on an energy function that extracts some descriptive feature statistics from the signal. 
Our model is related to the following energy-based models. The FRAME (Filters, Random field, And Maximum Entropy) \cite{zhu1998filters} model was developed for modeling stochastic textures. 
 It is a Markov random field model or an energy-based model of stationary spatial processes. The energy function of the distribution is the sum of translation-invariant potential functions that are one-dimensional non-linear transformations of
linear Gabor filter responses. The sparse FRAME model \cite{xie2015learning, xie2016inducing} generalizes the original FRAME model to modeling object patterns that are non-stationary in the spatial domain. The model is a sparse and inhomogeneous version of original FRAME, where the potential functions are location specific, and they are non-zero only at selected locations.
Inspired by the successes of deep convolutional neural networks (CNNs or ConvNets), \cite{lu2015learning} proposes a deep FRAME model, where the linear filters used in the original FRAME model are replaced by the non-linear filters at a certain convolutional layer of a pre-trained deep ConvNet. Such filters can capture more complicated patterns than linear Gabor filters. We refer readers to \cite{wu2018sparse} for a comprehensive tutorial on FRAME models. Instead of using filters from a pre-trained ConvNet, \cite{XieLuICML} learns the ConvNet filters from the observed data by maximum likelihood estimation. The resulting model is an energy-based generative ConvNet, which can be considered a recursive multi-layer generalization of the original FRAME model. 
\cite{nijkamp2019learning} and\cite{nijkamp2019anatomy} further study learning the energy-based generative ConvNet with a non-convergent non-persistent short-run MCMC. \cite{du2019implicit} adopts a residual network \cite{he2016deep} to parameterize the energy function. The energy-based spatial-temporal generative ConvNet\cite{xie2017synthesizing,xie2019learning} is a generalization of \cite{XieLuICML} by adding the temporal dimension for the purpose of modeling and synthesizing dynamic patterns, such as dynamic textures and action patterns.
While previous models focus on modeling 2D images or image sequences, our paper studies an energy-based model with the energy function parameterized by the VoxelNet for 3D volumetric shape patterns. We call the model the generative VoxelNet. We then further generalize the model to a conditional version for 3D shape recovery and super resolution, as well as a multi-grid version for high resolution synthesis. This paper is an expanded version of our conference paper \cite{xie2018learning}. 

\textit{3D object classification.} Even though our paper studies energy-based generative modeling of 3D volumetric shapes, the features learned from the model can be used for classification. We review methods of 3D shape classification below. 
The computer vision and graphics
community has developed various 3D shape descriptors, such as Light Field descriptor \cite{chen2003visual} and Spherical Harmonic descriptor \cite{kazhdan2003rotation}. Standard classifiers can be easily trained on those descriptors for 3D classification. Another direction is to apply established 2D discriminative ConvNet to the 2D information extracted from the
3D shapes. For example, \cite{sinha2016deep} proposes to covert 3D shapes into geometry images, on which standard 2D ConvNet is directly adopted for classification. \cite{shi2015deeppano, sfikas2017exploiting} propose to classify 3D shapes by applying standard 2D ConvNet to their panoramic \cite{papadakis2010panorama} view images. Besides, researchers also directly design end-to-end 3D discriminative neural networks that are suitable for 3D raw data. For example, \cite{simonovsky2017dynamic} applies graph convolutional neural network to 3D point cloud classification. \cite{maturana2015voxnet} proposes a 3D volumetric ConvNet for 3D classification. 
 The following approaches learn generative 3D representations in unsupervised manners, and trained classifiers on the features extracted from the learned models along with class labels. 3D ShapeNets \cite{wu20153d} trains a convolutional deep belief network from 3D volumetric shapes. The features learned from 3D ShapeNets are discriminatively fine-tuned with class labels for classification. \cite{sharma2016vconv} learns a convolutional volumetric auto-encoder from 3D shapes and uses the learned embedding for classification.   \cite{3dgan, khan2019unsupervised} train 3D generative adversarial nets, and then trains classifiers on the representations learned by the discriminators. \cite{huang20193d} learns a 3D Wasserstein introspective neural network \cite{lee2018wasserstein}, and trains a linear SVM on top of the features extracted from the model. Our paper also learns generative features for 3D classification. The major difference between ours and those mentioned above is that our features are extracted by the bottom-up energy function of the energy-based model, which is parameterized by a VoxelNet, and trained via MCMC-based maximum likelihood estimation.

\subsection{Contributions}
(1) We propose a 3D deep energy-based model that we call generative VoxelNet to model 3D voxel patterns by combining the 3D voxel ConvNets \cite{maturana2015voxnet} and the energy-based generative ConvNets \cite{XieLuICML}. The proposed model provides an explicit probability distribution defined on the 3D voxel grid, and its training does not rely on any auxiliary models. 
(2) For theoretical understanding, we present a mode seeking and mode shifting interpretation of the ``analysis by synthesis'' learning process of the model, and an adversarial interpretation of the zero temperature limit of the learning process. (4) We propose a conditional learning framework based on the proposed 3D energy-based model for recovery tasks, such as 3D shape super resolution and 3D object recovery. (5) We  propose  metrics that can be useful for evaluating 3D generative models. (6) We propose a multi-grid energy-based modeling and sampling framework for high resolution 3D volumetric shape synthesis. (7) A 3D cooperative training scheme is provided as an alternative to the adversarial learning method or the variational inference approach to train 3D generators. (8) The proposed model outperforms the other unsupervised baseline 3D models for volumetric data (e.g., 3D-VAE, 3D-GAN, and 3D-WINN), and obtains state-of-the-art performance in terms of both synthesis quality and classification accuracy. 

\subsection{Organization}
The rest of this paper is structured as follows. In section \ref{sec:model}, we first propose a deep energy-based model for 3D volumetric object patterns that we call the generative VoxelNet, by leveraging previous advances on 3D voxel convolutional networks \cite{maturana2015voxnet} and the energy-based generative ConvNets \cite{XieLuICML}. We then present an interpretation of the learning process of the model by explaining it as a mode seeking and mode shifting dynamics. Section \ref{sec:condition} presents a conditional version of the generative VoxelNet model for 3D shape recovery. Section \ref{sec:multigrid} describes a multi-grid energy-based modeling and sampling strategy for high resolution 3D volumetric shape synthesis. In section \ref{sec:coopnets}, we show how to learn a top-down ConvNet (3D generator) as a sampler simultaneously with the generative VoxelNet, so that (1) the MCMC simulation of the generative VoxelNet can be initialized by the learned sampler, and (2) the 3D generator can be trained by the generative VoxelNet via MCMC teaching. Section \ref{sec:exp} presents extensive qualitative and quantitative experiments to evaluate the proposed models.     

\section{Energy-based generative VoxelNet}
\label{sec:model} 
\subsection{Probability density for 3D volumetric shapes}
Let $Y$ be a 3D volumetric shape defined on the 3D voxel grid.
The generative VoxelNet model is a 3D deep convolutional energy-based model defined on $Y$, which is in the form of exponential tilting of a reference distribution \cite{XieLuICML}: 
\begin{eqnarray} 
   p(Y; \theta) = \frac{1}{Z(\theta)} \exp\left[ f(Y; \theta)\right] p_0(Y), 
\label{equ:model}
\end{eqnarray}
 where $p_0(Y)$ is the reference distribution such as Gaussian white noise model, i.e.,
$ p_0(Y) \propto \exp \left( -{\|Y\|^2}/{2 s^2}\right)$ (Standard deviation $s$ is a hyperparameter. We shall set $s=0.5$). $f(Y; \theta)$ is the scoring function mapping $Y$ to a scalar by a bottom-up 3D voxel convolutional neural network (VoxelNet) that is a composition of $L_{f}$ layers of 3D volumetric convolutions, subsamplings, and non-linear rectifications (e.g., Rectified Linear
Unit (ReLU) \cite{krizhevsky2012imagenet}), as illustrated by the following diagram:
\begin{eqnarray}
& Y \rightarrow h^{(1)} \rightarrow \cdots  h^{(l-1)} \rightarrow h^{(l)} \rightarrow \cdots  h^{(L_{f})} \rightarrow f(Y;\theta), \nonumber
\end{eqnarray}
where $h^{(l)}$ is the hidden output  computed recursively by performing 3D convolution on $h^{(l-1)}$ followed by ReLUs. 
All the weight and bias parameters in this 3D ConvNet $f$ are denoted by $\theta$. The analytically intractable normalizing constant $Z(\theta) = \int  \exp\left[ f(Y; \theta)\right] p_0(Y) dY$ is used to reduce the unnormalized density function to a probability density function with total probability of one. The generative VoxelNet model $p(Y;\theta)$ defines explicit log-likelihood via $f(Y;\theta)$.
 
$p(Y;\theta)$ can be written in the form of
an energy-based model, i.e., $p(Y; \theta) = \frac{1}{Z(\theta)} \exp\left[ -{\cal E}(Y; \theta)\right]$. The energy function is
 \begin{eqnarray} 
   {\cal E}(Y; \theta) = \frac{\|Y\|^2}{2s^2} - f(Y; \theta). 
   \label{equ:E}
\end{eqnarray}
We may also take $p_0(Y)$ as uniform distribution within a bounded range, then ${\cal E}(Y; \theta) = - f(Y; \theta)$.  In this paper, we only consider using Gaussian distribution for the reference distribution $p_0(Y)$. The full potential of the energy-based model lies in the structure of the energy landscape defined by ${\cal E}(Y; \theta)$, which maps each of the observed examples to an energy value. This energy value serves as a measure of the ``goodness'' of a configuration of the input variable $Y$. A well trained model should assign lower energy values (i.e., high probabilities) to the observed examples than the unobserved examples. If the training data are highly varied, the learned energy landscape is likely to be multimodal.
The local energy minima \cite{hopfield1982neural} of equation (\ref{equ:E}) satisfy the Hopfield auto-encoder \cite{XieLuICML}
\begin{eqnarray} 
\frac{Y}{s^2}=\frac{\partial}{\partial Y} f(Y;\theta),
\end{eqnarray}
where $\frac{\partial}{\partial Y} f(Y; \theta)$ can be computed by feed forward computation (bottom-up encoding process) and back-propagation (top-down decoding process).
Such a model can be used for associative memory \cite{hopfield1982neural}, where given an incomplete, occluded, or corrupted memory (i.e., data in the high energy regions), diffusion along the energy manifold starting from the high energy memory can gradually recover the missing memory until it arrives at the low energy regions.   

 \subsection{Analysis by synthesis} 
 The maximum likelihood estimation (MLE) of the energy-based generative VoxelNet follows an ``analysis by synthesis'' scheme. Suppose we observe 3D training examples $\{Y_i, i = 1, ..., n\}$ from an unknown data distribution $P_{\text{data}}(Y)$. The MLE seeks to find $\theta$ to maximize the log-likelihood function
\begin{eqnarray}   
L(\theta) = \frac{1}{n} \sum_{i=1}^{n} \log p(Y_i; \theta). 
\end{eqnarray}  
The gradient of the $L(\theta)$ with respect to $\theta$ is 
  \begin{eqnarray} 
  L'(\theta) = \frac{1}{n} \sum_{i=1}^{n} \frac{\partial}{\partial \theta} f(Y_i; \theta) - \E_{\theta} \left[\frac{\partial}{\partial \theta} f(Y; \theta)\right],  \label{eq:lD}
\end{eqnarray} 
where  $\E_{\theta}$ denotes the expectation with respect to $p(Y; \theta)$. The expectation term in equation (\ref{eq:lD}) is due to
$\frac{\partial}{\partial \theta}  \log Z(\theta) = \E_{\theta}[\frac{\partial}{\partial \theta}  f(Y; \theta)]$, which is analytically intractable and has to be approximated by MCMC, such as Langevin  dynamics,  which iterates the following step: 
\begin{eqnarray}
   Y_{\tau+\Delta \tau} &=& Y_\tau - \frac{\Delta \tau}{2} \frac{\partial}{\partial Y}  {\cal E}(Y_\tau; \theta) + \sqrt{\Delta \tau}\epsilon_{\tau} \nonumber\\ 
   &=& Y_\tau - \frac{\Delta \tau}{2} \left[ \frac{Y_\tau}{s^2} - \frac{\partial}{\partial Y} f(Y_\tau; \theta) \right] + \sqrt{ \Delta \tau}\epsilon_{\tau},
    \label{eq:LangevinD}
\end{eqnarray}
where $\tau$ indexes the time steps of the Langevin dynamics, $\Delta \tau$ is the discretized step size, and $\epsilon_{\tau} \sim \mathcal{N}(0, I)$ is the Gaussian white noise term.  The Langevin dynamics consists of a  deterministic part, which is a gradient descent on the landscape defined by ${\cal E}(Y; \theta)$, and a stochastic part, which is a Brownian motion that helps  the chain to escape spurious local minima of the energy ${\cal E}(Y; \theta)$. A Metropolis-Hastings step can be added to correct for the finiteness of $\Delta \tau$. Comparing with Gibbs sampling which does not take into account the landscape geometry, the gradient-based Langevin sampling converges quickly and is computationally feasible.

Suppose we draw $\tilde{n}$ samples $\{ \tY_i, i=1,...,\tilde{n} \}$ from the distribution $p(Y;\theta)$ by running $\tilde{n}$ parallel chains of Langevin dynamics according to (\ref{eq:LangevinD}). The gradient of the log-likelihood $L{(\theta)}$ can be approximated by 
  \begin{eqnarray} 
  L'(\theta) \approx \frac{1}{n} \sum_{i=1}^{n} \frac{\partial}{\partial \theta} f(Y_i; \theta) - \frac{1}{\tilde{n}} \sum_{i=1}^{ \tilde{n} } \frac{\partial}{\partial \theta} f(\tilde Y_i; \theta),  \label{eq:lD2}
\end{eqnarray}
which is the difference between the observed examples and the synthesized examples. 
The above equation leads to the ``analysis by synthesis'' learning scheme. At $t$-th iteration, we sample $\tilde{Y}_i \sim p(Y;\theta^{(t)})$ for $i=1,...,\tilde{n}$. Then we update the model parameters via $\theta^{(t+1)} = \theta^{(t)} + \gamma_t L'(\theta^{(t)})$, where $\gamma_t$ is the learning rate.

If the sample size $n$ is large, the maximum likelihood estimator is equivalent to minimizing $\text{KL}(P_{\text{data}}(Y) \parallel p(Y;\theta))$, the Kullback-Leibler (KL) divergence from the data distribution $\P(Y)$ to the model distribution $p(Y;\theta)$. That is,
\begin{eqnarray}
&& \min_{\theta} \KL(\P(Y) \parallel p(Y;\theta)) \nonumber \\
&=& \min_{\theta}\{ \E_{\P}\left[\log \P(Y)\right] -  \E_{\P}\left[\log p(Y;\theta) \right] \}  \nonumber \\
&=& \max_{\theta} \E_{\P}\left[\log p(Y;\theta) \right]   \nonumber \\
&\approx & \max_{\theta} \frac{1}{n} \sum_{i=1}^{n} \log p(Y_i; \theta)
\label{eq:KL}
\end{eqnarray}

\subsection{Mode seeking and mode shifting} 

The $f(Y;\theta)$ or equivalently the energy function ${\cal E}(Y; \theta)$ parameterized by the VoxelNet can be flexible enough to create many local modes to fit $\P$ that is likely to be highly multimodal. The energy function is reshaped by the ``analysis by synthesis'' learning algorithm in order to put lower energy values on the observed examples than the unobserved examples. This process can be interpreted as density shifting or mode shifting. We can rewrite equation (\ref{eq:lD2}) in the form of  
 \begin{eqnarray} 
  L'(\theta) \approx \frac{\partial}{\partial \theta} \left[ \frac{1}{\tilde{n}} \sum_{i=1}^{ \tilde{n} }   {\cal E}(\tilde Y_i; \theta) - \frac{1}{n} \sum_{i=1}^{n}   {\cal E}(Y_i; \theta) \right]. \label{eq:lD3}
\end{eqnarray}
We define a value function
 \begin{eqnarray} 
 V(\{\tY_i\}; \theta) = \frac{1}{\tilde{n}} \sum_{i=1}^{ \tilde{n} }   {\cal E}(\tilde Y_i; \theta) - \frac{1}{n} \sum_{i=1}^{n}   {\cal E}(Y_i; \theta). \label{eq:V}
\end{eqnarray}
The equation (\ref{eq:lD3}) reveals that the gradient of the log-likelihood $L{(\theta)}$ coincides with the gradient of $V$. 

The sampling step in (\ref{eq:LangevinD}) can be interpreted as mode seeking, by finding low energy modes or high probability modes in the landscape defined by $ {\cal E}(Y;\theta)$ via stochastic gradient descent (Langevin dynamics) and placing the synthesized examples around the modes, so that $\frac{1}{\tilde{n}} \sum_{i=1}^{ \tilde{n} }   {\cal E}(\tilde Y_i; \theta)$ tends to be low. It seeks to decrease $V$.   

The learning step can be interpreted as mode shifting (as well as mode creating and mode sharpening) by shifting the low energy modes from the synthesized examples $\{\tY_i\}$ toward the observed examples $\{Y_i\}$ by modifying the energy function ${\cal E}(Y; \theta)$ in order to increase the difference between the synthesized statistics and the observed statistics, i.e., $\frac{1}{\tilde{n}} \sum_{i=1}^{ \tilde{n} }   {\cal E}(\tilde Y_i; \theta) - \frac{1}{n} \sum_{i=1}^{n}   {\cal E}(Y_i; \theta)$. It seeks to increase $V$. 

\begin{figure}
\centering	
\includegraphics[width=1\linewidth]{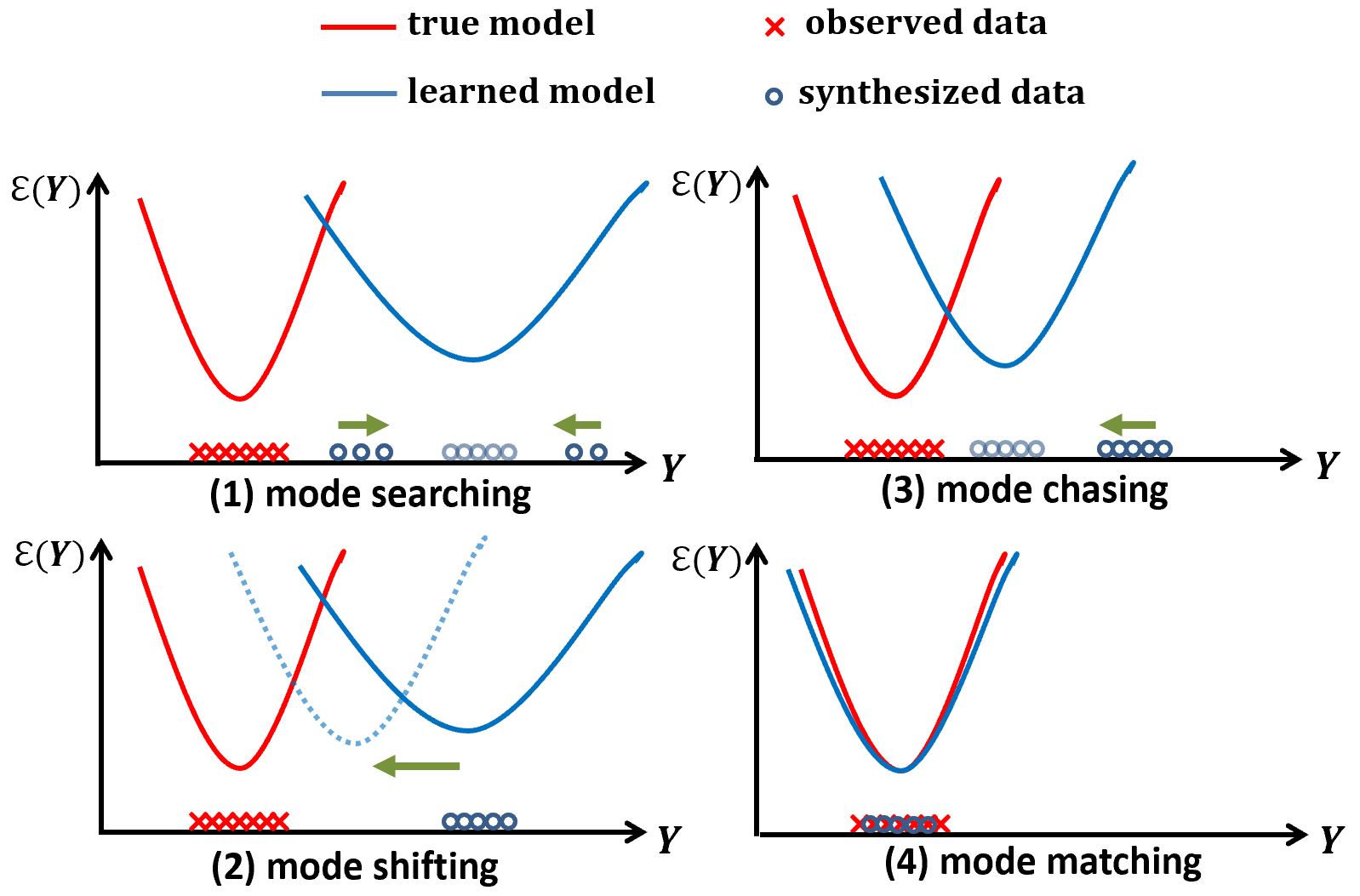}
\caption{An illustration of mode seeking and mode shifting. The red curves and blue curves represent the energy landscapes of the true model and the learned model, respectively. The red crosses indicate observed data, and the blue circles indicate the synthesized data. (1) mode searching by Langevin dynamics: finding low energy modes in the energy landscape of the model and placing the synthesized examples around the found modes. (2) model shifting (and sharpening) by model update: shifting the low energy modes toward the observed examples. (3) mode chasing: researching low energy modes in the updated energy landscape of the model. (4) mode matching: the modes in the learned energy landscape eventually match the ones in the data distribution after a few iterations of mode searching and shifting.}
\label{fig:mode_seeking}
\end{figure}

Even though the learning and sampling algorithm creates and shifts the modes of the energy landscape to the observed examples, there might be still numerous major modes that are not occupied by the observed examples, and these modes represent new examples that are considered similar to the observed examples. Accessing these modes corresponds to synthesizing new examples. While the maximum likelihood learning matches the average statistical properties between model and observed data, the ConvNet $f(Y;\theta)$ is sufficiently flexible and expressive to create modes to encode the highly varied 3D patterns.
An illustration of mode seeking and mode shifting is presented in Figure \ref{fig:mode_seeking}. The training algorithm of the generative VoxelNet is presented in Algorithm \ref{code:3D}.

\begin{algorithm}[h]
\caption{Generative VoxelNet}
\label{code:3D}
\begin{algorithmic}[1]
\REQUIRE ~~\\
(1) 3D training data $\{Y_i, i=1,...,n\}$; \\
(2) the number of Langevin steps $K$; \\
(3) the number of learning iterations $T$.

\ENSURE~~\\
(1) estimated parameters $\theta$; \\
(2) synthesized examples $\{\tY_i, i = 1, ..., \tilde{n}\}$. 

\item[]
\STATE Let $t\leftarrow 0$, randomly initialize $\theta^{(t)}$ and $\{\tY_i, i=1,...,\tilde{n}\}$ with Gaussian noise. 
\REPEAT 
\STATE \textbf{Mode seeking}: For each $i$, run $K$ steps of Langevin updates to revise $\tY_i$, i.e., starting from the current $\tY_i$, each step follows equation (\ref{eq:LangevinD}).\\
\STATE \textbf{Mode shifting}: Update $\theta^{(t+1)} = \theta^{(t)} + \gamma_t L'(\theta^{(t)}) $,  with learning rate $\gamma_t$, where $L'(\theta^{(t)})$ is computed according to equation (\ref{eq:lD2}). 
\STATE Let $t \leftarrow t+1$
\UNTIL $t = T$
\end{algorithmic}
\end{algorithm}

\subsection{Alternating back-propagation} 

Both mode seeking (sampling) and mode shifting (learning) steps involve the derivatives of $f(Y;\theta)$ with respect to $Y$ and $\theta$ respectively. Both derivatives can be computed efficiently by back-propagation, and share most of their steps. Specifically, for the bottom-up ConvNet structure $f(Y;\theta)$, the chain rule computation of $\partial h^{(l)}/\partial h^{(l-1)} $ for $l = 1, ..., L$ is shared in calculating $\partial f(Y; \theta)/\partial Y$ and $\partial f(Y; \theta)/\partial \theta$ in terms of coding. The algorithm is thus in the form of alternating back-propagation that iterates the following two steps: (1) Sampling back-propagation: Revise the synthesized examples by Langevin dynamics or gradient descent.
(2) Learning back-propagation: Update the model parameters
given the synthesized and the observed examples by gradient ascent. 

\subsection{Zero temperature limit} 
For notation simplicity, let $p_\theta$ be the model distribution, and let $p_\tau$ be the distribution of $Y_\tau$ of the Langevin dynamics, then $\KL(p_\tau\parallel p_\theta)$ decreases  to zero monotonically over $\tau$ according to the second law of thermodynamics \cite{cover2012elements}. In other words, the Langevin dynamics implements a variational approximation to $p_\theta$ with $p_\tau$. $\KL(p_\tau \parallel p_\theta) = -{\rm entropy}(p_\tau)+ \E_{p_\tau}[{\cal E}(Y_\tau; \theta)] + \log Z(\theta)$. The gradient descent part of the Langevin dynamics reduces the energy $\E_{p_\tau}[{\cal E}(Y_\tau; \theta)]$, while the noise term of the Langevin dynamics increases the entropy of $p_\tau$. 

We can add a temperature term to the model $p_T(Y; \theta) = \exp(- {\cal E}(Y; \theta)/T)/Z_T(\theta)$, where the original model corresponds to $T = 1$. At zero temperature limit as $T \rightarrow 0$, the Langevin sampling will become gradient descent where the noise term diminishes in comparison to the gradient descent term. The resulting algorithm approximately solves the minimax problem
  \begin{eqnarray} 
\max_{\theta} \min_{\{\tY_i\}}  V(\{\tY_i\}; \theta),  \label{eq:minimax}
\end{eqnarray}
with $\tY_i$ initialized from an initial distribution and approaching local modes of $V$.  We can regularize either the diversity of $\{\tY_i\}$ or the smoothness of ${\cal E}(Y; \theta)$.  This is an adversarial interpretation of the learning algorithm. It is also a generalized version of herding \cite{welling2009herding}. It is also related to Wasserstein GAN \cite{arjovsky2017wasserstein}, but the critic and the actor are the same model, i.e., the model itself is its own generator and critic.

In our experiments, we find that disabling the noise term of the Langevin dynamics in the later stage of the learning process often leads to better synthesis.  Ideally the learning algorithm should create a large number of local modes with similar low energies to capture the diverse observed examples as well as unseen examples.

\section{Conditional learning for recovery}
\label{sec:condition} 
 The conditional distribution $p(Y | C(Y) = c; \theta)$ can be derived from $p(Y; \theta)$. This conditional form of the generative VoxelNet can be used for recovery tasks such as inpainting and super resolution. In inpainting, $C(Y)$ consists of the visible part of $Y$. In super resolution, $C(Y)$ is the low resolution version of $Y$. For such tasks, we can learn the model from the fully observed training data $\{Y_i, i = 1, ..., n\}$ by maximizing the conditional log-likelihood 
 \begin{eqnarray}
     L(\theta) = \frac{1}{n} \sum_{i=1}^{n} \log p(Y_i \mid C(Y_i) = c_i; \theta), 
 \end{eqnarray} 
 where $c_i$ is the observed value of $C(Y_i)$. 
The learning and sampling algorithm is essentially the same as maximizing the original log-likelihood, except that in the Langevin sampling step, we need to sample from the conditional distribution, which amounts to fixing $C(Y_\tau)$ in the sampling process. 
The  zero temperature limit (with the noise term in the Langevin dynamics disabled) approximately solves the following minimax problem 
  \begin{eqnarray} 
\max_{\theta} \min_{\{\tY_i: C(\tY_i) = {c}_i\}}  V(\{\tY_i\}; \theta).  \label{eq:minimax1}
\end{eqnarray}

\section{3D Multi-grid modeling and sampling}
\label{sec:multigrid}

The maximum likelihood learning of the generative VoxelNet requires MCMC sampling, such as Langevin dynamics, of the 3D objects, which can be  expensive and time consuming, because it takes a long time to converge, if the learned $p(Y;\theta)$ is multimodal. This often happens because $\P$ is usually highly varied and multimodal. Therefore, it becomes the bottleneck and the difficulty of learning the energy-based model in the ``analysis by synthesis'' scheme. In this section, inspired by \cite{gao2018learning}, we will propose a 3D multi-grid modeling and sampling framework as a modified contrastive divergence to train generative VoxelNet models.

\subsection{Contrastive divergence}
Contrastive divergence (CD) method proposed in \cite{Hinton2002a} generates synthesized examples by finite steps of MCMC initialized from the observed examples, and updates the model parameters with the gradient computed by equation (\ref{eq:lD2}). The CD learning approximately minimizes the difference between two KL divergences:
\begin{eqnarray} 
\KL(\P\parallel p_{\theta}) - \KL(M_{\theta}\P \parallel p_{\theta}),  \label{eq:CD_KL}
\end{eqnarray} 
where $M_{\theta}$ is the transition kernel of the finite-step MCMC that samples from $p_{\theta}(Y)$, and $M_{\theta} \P(Y')=\int \P(Y)M_{\theta}(Y'|Y)dY$ is the distribution after running $M_{\theta}$ starting from $\P$.

The contrastive divergence (CD) learning method is related to score matching estimator \cite{Hyvrinen05estimationof,hyvarinen2007connections} and auto-encoder \cite{Vincent2010,Swersky2011,alain2014regularized}. This method enables the model to learn from large training datasets via mini-batch training.  However, the synthesized examples obtained by CD learning may be distant from fair samples of the current model, thus bringing in bias into the learned model parameters. 

An improvement of CD is persistent CD \cite{tieleman2008training}, where MCMC chains are still initialized from the observed examples at the initial epoch, but contrary to normal CD, in each subsequent epoch, the finite-step MCMC chains are initialized from the corresponding synthesized examples obtained at the previous epoch instead of starting over. However, it is still difficult for the persistent CD learning method to traverse different modes of the learned model.

\subsection{Modified contrastive divergence by multi-grid learning and sampling}

For a 3D volumetric shape $Y$, let $(Y^{(s)},s = 0,...,S)$ denote the multi-grid versions of $Y$, with $Y^{(0)}$ representing the minimal $1 \times 1 \times 1$ version of $Y$, and $Y^{(S)}$ representing $Y$. For each version $Y^{(s)}$, we can divide the 3D voxel grid into cubic blocks of $d \times d \times d$ voxels, and reduce each $d \times d \times d$ block into a single voxel by a down-scaling operation that averages the values of the $d \times d \times d$ voxels of $Y^{(s)}$ to obtain $Y^{(s-1)}$. Conversely, we can also define an up-scaling operation which expands each voxel of $Y^{(s-1)}$ into a $d \times d \times d$ block of constant value to obtain an up-scaled version $\check{Y}^{(s)}$ of $Y^{(s-1)}$. Note that the up-scaled $\check{Y}^{(s)}$ is not the same as the original $Y^{(s)}$ because the high resolution details are lost after down-scaling and up-scaling. The down-scaling mapping from $Y^{(s)}$ to $Y^{(s-1)}$ is a linear projection onto a set of orthogonal basis vectors, each of which is specified by a $d \times d \times d$ block, while the up-scaling operation is a pseudo-inverse of this linear mapping.  

Our multi-grid method learns a separate generative VoxelNet model at each grid. Let $p^{(s)}(Y^{(s)};\theta^{(s)})$ denote the generative VoxelNet at grid $s$. We can simply model $p^{(0)}$ by a one dimensional histogram of $Y^{(0)}$, which is pooled from the $1 \times 1 \times 1$ versions of $Y$. Within each iteration of the learning algorithm, for each observed 3D training example $Y_i$, we generate the corresponding synthesized examples at multiple grids, i.e., $(\tilde{Y}_{i}^{(s)}, s=0,...,S)$. Specifically, we initialize the finite-step MCMC sampling from the minimal $1 \times  1 \times 1$ version $Y_i^{(0)}$, and the up-scaled version of $\tilde{Y}^{(s-1)}_i$ sampled from the model $p^{(s-1)}(Y^{(s-1)}; \theta^{(s-1)})$ at the previous coarser grid serves to initialize the finite-step MCMC that samples from the model $p^{(s)}(Y^{(s)}; \theta^{(s)})$ at the subsequent finer grid.  The update of the model parameter $\theta^{(s)}$ for each grid follows equation (\ref{eq:lD2}), which is based on the difference between the synthesized ${\tilde{Y}^{(s)}}$ and the observed ${Y^{(s)}}$. Algorithm \ref{code:multigrid} presents a full description of the 3D multi-grid learning and sampling algorithm.

The learning gradient of model at grid $s$ approximately minimizes the difference between two KL divergences:
\begin{eqnarray} 
\KL(\P^{(s)}\parallel p^{(s)}_{\theta^{(s)}}) - \KL(M^{(s)}_{\theta^{(s)}} P^{(s)}_{\theta^{(s-1)}} \parallel p^{(s)}_{\theta^{(s)}}),  \label{eq:grid_KL}
\end{eqnarray} 
where $M^{(s)}_{\theta^{(s)}}$ is transition kernel of the finite-step Langevin
dynamics that samples from $p^{(s)}_{\theta^{(s)}}$. $P^{(s)}_{\theta^{(s-1)}}$ is the up-scaled version of the model $p^{(s-1)}_{\theta^{(s-1)}}$. That is, $P^{(s)}_{\theta^{(s-1)}}$ is the distribution of $\check{Y}^{(s)}$, which is the up-scaled version of $Y^{(s-1)}$ that follows distribution $p^{(s-1)}_{\theta^{(s-1)}}$.
$P^{(s)}_{\theta^{(s-1)}}$ is smoother than $p^{(s)}_{\theta^{(s)}}$, and the finite-step MCMC specified by $M^{(s)}_{\theta^{(s)}}$ will change $P^{(s)}_{\theta^{(s-1)}}$ into a distribution close to the target distribution $p^{(s)}_{\theta^{(s)}}$ by synthesizing details at the current resolution. Such a coarse-to-fine manner is related to super resolution discussed in Experiment \ref{Exp:sr}. Equation (\ref{eq:grid_KL}) can also be regarded as a modified version of CD learning in equation (\ref{eq:CD_KL}). 

The learned models are able to synthesize new examples from scratch, because we only need to initialize the MCMC by sampling from the one-dimensional histogram pooled from the $1 \times 1 \times 1$ versions of the training examples.

\begin{algorithm}[h]
\caption{3D multi-grid learning and sampling}
\label{code:multigrid}
\begin{algorithmic}[1]
\REQUIRE ~~\\
(1) 3D training data $\{Y^{(s)}_i, s=1,...,S; i=1,...,n\}$; \\
(2) the number of Langevin steps $K$; \\
(3) the number of learning iterations $T$.

\ENSURE~~\\
(1) estimated parameters $(\theta^{(s)}, s=1,...,S)$; \\
(2) synthesized examples $\{\tY^{(s)}_i, s=1,...,S; i = 1, ..., n\}$. 

\item[]
\STATE Let $t\leftarrow 0$. For $s = 1, ..., S$, initialize $\theta_{(t)}^{(s)}$ with Gaussian noise. 
\REPEAT
 \STATE For $i = 1,...,n$, initialize $\tilde{Y}^{
(0)}_i = Y^{(0)}_i$
\STATE For $s = 1,...,S$, initialize $\tilde{Y}^{(s)}_i$ as the up-scaled version
of $\tilde{Y}^{(s-1)}_i$, and run $K$ steps of the Langevin dynamics to evolve $\tilde{Y}^{(s)}_i$, each step following equation (\ref{eq:LangevinD}).\\
\STATE For $s = 1,...,S$, update $\theta^{(s)}_{(t+1)} = \theta^{(s)}_{(t)} + \gamma_t L'(\theta^{(s)}_{(t)}) $,  with learning rate $\gamma_t$, where $L'(\theta^{(s)}_{(t)})$ is computed according to equation (\ref{eq:lD2}). 
\STATE Let $t \leftarrow t+1$
\UNTIL $t = T$
\end{algorithmic}
\end{algorithm}

\section{Teaching 3D generator net}
\label{sec:coopnets}
We can let a 3D generator network \cite{goodfellow2014generative} learn from the MCMC sampling of the generative VoxelNet, so that the 3D generator network can be used as a much more efficient approximate non-iterative direct sampler of the generative VoxelNet. 

\subsection{3D generator model} 

The 3D generator model is a 3D non-linear multi-layer generalization of the traditional factor analysis model. The generator model has the following form
\begin{eqnarray} 
& Z \sim \mathcal{N}(0, I_d);\nonumber \\ 
 & Y = g(Z; \alpha)  + \epsilon; \epsilon \sim \mathcal{N}(0, \sigma^2 I_D), \label{eq:NFA}
 \end{eqnarray}
 where $Z$ is a $d$-dimensional vector of latent factors that follow $\mathcal{N}(0, I_d)$ independently, and the 3D object $Y$ is generated by first sampling $Z$ from its known prior distribution $\mathcal{N}(0, I_d)$ and then transforming $Z$ to the $D$-dimensional $Y$ by a top-down 3D deconvolutional network $g(Z; \alpha)$ plus the white noise $\epsilon \sim \mathcal{N}(0, \sigma^2 I_D)$, where $\sigma$ is the standard deviation and $d< D$. $g(Z; \alpha)$ is a composition of $L_{g}$ layers of 3D volumetric deconvolutions, upsamplings, and non-linear rectifications (e.g., ReLU), as illustrated by the following diagram: 
\begin{eqnarray}
&Z \rightarrow Z^{(L_{g})} \rightarrow \cdots Z^{(l+1)} \rightarrow Z^{(l)} \rightarrow \cdots   Z^{(1)} \rightarrow g(Z;\alpha), \nonumber
\end{eqnarray}
where $Z^{(l)}$ is the hidden output that is computed recursively by performing 3D deconvolution on $Z^{(l+1)}$ followed by ReLUs. 
$\alpha$ denotes all weight and bias  parameters of the 3D top-down ConvNet $g$. The generator model is a directed graphical model such that it can readily generate a new example $Y$ by ancestral sampling. 

The 3D generator model can be trained by a maximum likelihood algorithm studied in \cite{HanLu2016}, without resorting to any assisting networks. 
The training algorithm alternates two steps:
(1) For each observed example, inferring the latent factors by sampling from the current posterior distribution via MCMC, and (2) Updating the model parameters by a non-linear regression of the observed examples on their inferred latent factors, so that the learned parameters enable better reconstructions of the observed examples by the inferred latent factors. The algorithm is also an alternating back-propagation (ABP) algorithm, because both two steps can be powered by back-propagation. 

The ABP algorithm requires MCMC sampling to infer latent factors  in training 3D generator model. It is possible to avoid MCMC sampling by using either variational inference \cite{kingma2013auto} or adversarial learning \cite{goodfellow2014generative} where an auxiliary network is recruited to relieve the burden of MCMC-based iterative inference. 

As to variational learning, an inference model is simultaneously learned to replace the MCMC sampling of the latent factors. The inference model is an analytically tractable approximation of the posterior distribution. The objective of variational learning is an upper bound of the MLE objective. Consequently, the accuracy of the variational learning as an approximation to the MLE depends on the accuracy of the inference model as an approximation to the posterior distribution. 

Adversarial learning provides an alternative to maximum likelihood techniques. A discriminative model is recruited and simultaneously trained with the generator in a minimax two-player game, where the generator seeks to generate ``fake'' examples that can fool the discriminator, while the discriminator seeks to distinguish the ``fake'' examples and the real training examples. However, adversarial learning may commonly encounter mode collapse issue, where the generator only outputs samples from a single mode.

\subsection{MCMC teaching of 3D generator net} 

The 3D generator can be trained simultaneously with the generative VoxelNet  in a cooperative training scheme \cite{xie2016cooperative,xie2018cooperative}. The basic idea is to use the 3D generator to generate examples to initialize a finite-step Langevin dynamics for training the generative VoxelNet. In return, the 3D generator learns from how the Langevin dynamics changes the initial examples it generates. In this scheme, the generative VoxelNet is the teacher network, and the 3D generator model is the student network. The generative VoxelNet teaches the 3D generator via a finite-step MCMC. We call it MCMC teaching. In this cooperative learning process, the generative VoxelNet learns from the real data, while the 3D generator learns from the MCMC sampling of the generative VoxelNet.

Specifically, in each iteration, (1) We generate $Z_i$ from its known prior distribution $\mathcal{N}(0,I_d)$, and then generate the initial synthesized examples  by $\hY_i = g(Z_i; \alpha)  + \epsilon_i$ for $i=1,...,\tilde n$. (2) Starting from the initial examples $\{\hY_i\}$, we sample from the generative VoxelNet by running a finite number of steps of Langevin dynamics to obtain the revised synthesized examples $\{\tY_i\}$. (3) We then update the parameters $\theta$ of the generative VoxelNet based on $\{\tY_i\}$ according to equation (\ref{eq:lD2}), and update the parameters $\alpha$ of the 3D generator by gradient descent 
\begin{eqnarray}
\Delta \alpha \propto - \frac{\partial}{\partial \alpha} \left[\frac{1}{\tn} \sum_{i=1}^{\tilde n} \|\tY_i - g(Z_i; \alpha)\|^2\right]. \label{eq:alpha}
\end{eqnarray}
 We call it MCMC teaching because the revised examples $\{\tY_i\}$ generated by the finite-step MCMC are used to teach $g(Z; \alpha)$. 
For each $\tY_i$,  the vector of latent factors $Z_i$ is known to the 3D generator, so that there is no need to infer $Z_i$, and the learning becomes a much simpler supervised learning problem. Initially the 3D generator maps $\{Z_i\}$ to $\{\hat{Y}_i\}$. After MCMC teaching, the 3D generator seeks to map $\{Z_i\}$ to $\{\tY_i\}$. That is, the MCMC teaching enables the 3D generator to absorb and accumulate the MCMC transitions for sampling the generative VoxelNet, and shift its density from $\{\hat{Y}_i\}$ to $\{\tY_i\}$ for the sake of reproducing the effect of all the past MCMC transitions by one-step direct ancestral sampling. 

Algorithm \ref{code:3} presents a full description of the learning of the generative VoxelNet with a 3D generator as a sampler.

\subsection{Convergence analysis}    

We briefly discuss the convergence of the cooperative learning algorithm. In such a cooperative learning scheme, the training of the generative VoxelNet follows the modified contrastive divergence: 
\begin{eqnarray}
\theta_{t+1} = \arg \min_{\theta} [\KL(\P \parallel p_{\theta}) -\KL(M_{\theta_{t}}q_{\alpha} \parallel p_{\theta})], \label{eq:MCD_p}
\end{eqnarray}
where $M_{\theta_{t}}$ is the transition kernel of the finite-step MCMC samples from $p_{\theta_t}$ at $t$-th iteration, $q_{\alpha}$ is the distribution of the generator, and  $M_{\theta_{t}}q_{\alpha}$ denotes the marginal distribution obtained after running $M_{\theta_{t}}$ starting from $q_{\alpha}$.
The training of the 3D generator follows
\begin{eqnarray}
\alpha_{t+1} = \arg \min_{\alpha} \KL(M_{\theta}q_{\alpha_t} \parallel q_{\alpha}), \label{eq:MCD_g}
\end{eqnarray}
so that  the learned $q_{\alpha}$ eventually is the stationary distribution of $M_{\theta}$, i.e., $q_{\alpha} = M_{\theta}q_{\alpha}$. Since the stationary distribution of $M_{\theta}$ is nothing else than $p_{\theta}$, the learned $q_{\alpha}$ is equivalent to $p_{\theta}$. Consequently, the second KL-divergence in (\ref{eq:MCD_p}) will become zero, and the learned $\theta$ is to minimize the first KL-divergence, i.e., $\KL(\P \parallel p_{\theta})$. In other words, $\theta$ is a maximum likelihood estimate. The learned $\alpha$ in this way is also a maximum likelihood estimate, because $q_{\alpha}$ traces the $p_{\theta}$ that runs towards the data distribution, i.e., $q_{\alpha} \rightarrow p_{\theta} \rightarrow \P$.

\begin{algorithm}[h]
\caption{MCMC teaching of 3D generator net}
\label{code:3}
\begin{algorithmic}[1]

\REQUIRE~~\\
(1) 3D training examples $\{Y_i, i=1,...,n\}$;\\
(2) the numbers of Langevin steps $K$;\\
(3) the number of learning iterations $T$.

\ENSURE~~\\
(1) estimated parameters $\theta$ and $\alpha$;\\ (2) synthesized examples $\{\hY_i, \tY_i, i= 1, ..., \tilde{n}\}$.

\item[]
\STATE Let $t\leftarrow 0$, randomly initialize $\theta^{(t)}$ and $\alpha^{(t)}$ with Gaussian distribution.
\REPEAT 
\STATE {\bf Initializing mode seeking}: For $i = 1, ..., \tn$, generate $Z_i \sim \mathcal{N}(0, I_d)$, and generate $\hY_i = g(Z_i; \alpha^{(t)}) + \epsilon_i$. 
\STATE {\bf Mode seeking}: For $i = 1, ..., \tn$,  starting from $\hY_i$, run $K$ steps of Langevin  dynamics to obtain $\tY_i$,  each step 
following equation (\ref{eq:LangevinD}). 
\STATE {\bf Mode shifting}: Update $\theta^{(t+1)} = \theta^{(t)} + \gamma_t L'(\theta^{(t)})$,  where $L'(\theta^{(t)})$ is computed according to equation (\ref{eq:lD2}). 
\STATE {\bf Learning from mode seeking}: Update $\alpha^{(t+1)}$ according to equation (\ref{eq:alpha}).
\STATE Let $t \leftarrow t+1$
\UNTIL $t = T$
\end{algorithmic}
\end{algorithm}

\section{Experiments}
\label{sec:exp}

\textbf{Project page}: The code and more results can be found at 
\url{http://www.stat.ucla.edu/~jxie/3DEBM/}

In this section, we conduct experiments to evaluate the energy-based generative VoxelNet from different aspects. We first show qualitative results of the synthesized 3D objects, and then we quantitatively evaluate the synthesis quality by Inception score, softmax class probability, and classification error. Further, we show that the conditional generative VoxelNet can be useful for 3D object recovery and 3D super resolution. Additionally, the 3D generator taught by the generative VoxelNet in a cooperative scheme is analyzed from the perspectives of shape synthesis, 3D shape interpolation and arithmetic in the latent space. We also show that the generative VoxelNet trained in an unsupervised manner can be used as a feature extractor for training a classifier with annotated data. Lastly, we show experimental results regarding high resolution 3D shape synthesis by the 3D multi-grid energy-based sampling and modeling strategy.  

\subsection{3D object synthesis} 
\label{Exp:objectSynthesis}

\begin{figure*}[h]
	\centering	
\textbf{{\footnotesize obs1}}\hspace{6.5mm} \textbf{{\footnotesize obs2}} \hspace{6.5mm} \textbf{{\footnotesize obs3}} \hspace{6.5mm} \textbf{{\footnotesize syn1}} \hspace{6.5mm} \textbf{{\footnotesize syn2}} \hspace{6.5mm} \textbf{{\footnotesize syn3}} \hspace{6.5mm} \textbf{{\footnotesize syn4}} \hspace{6.5mm}
\textbf{{\footnotesize syn5}} \hspace{6.5mm} \textbf{{\footnotesize syn6}} \hspace{6.5mm} \textbf{{\footnotesize nn1}} \hspace{6.5mm}
\textbf{{\footnotesize nn2}} \hspace{6.5mm} \textbf{{\footnotesize nn3}} \hspace{6.5mm}
\textbf{{\footnotesize nn4}} \\
	\rotatebox{90}{\hspace{4mm}\textbf{{\footnotesize chair}}}	
	\includegraphics[height=.08\linewidth]{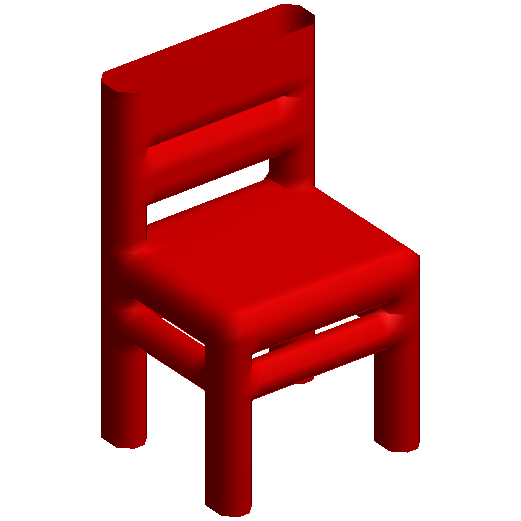} \hspace{-3mm} 
	\includegraphics[height=.08\linewidth]{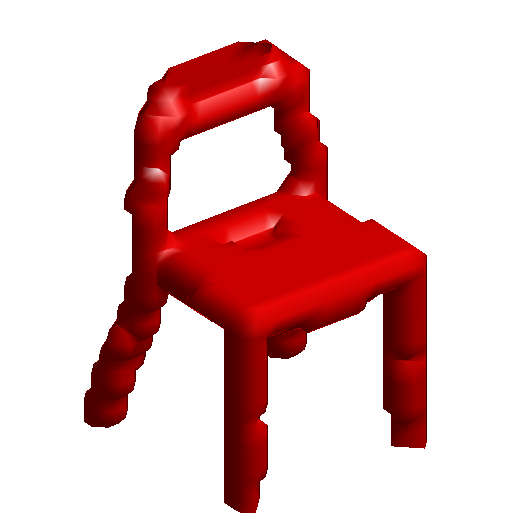} \hspace{-3mm} 
	\includegraphics[height=.08\linewidth]{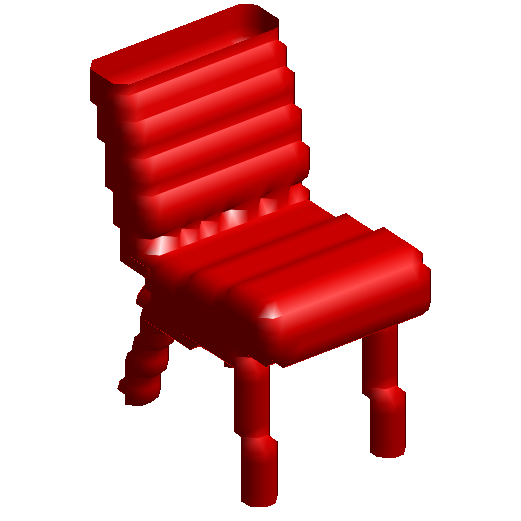} \hspace{-3mm}
    \includegraphics[height=.08\linewidth]{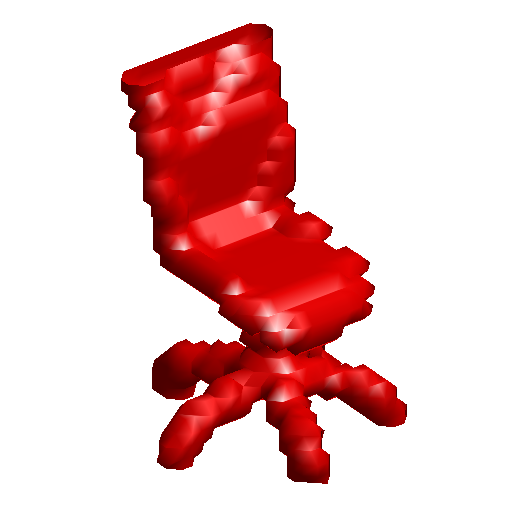} \hspace{-3mm}	
    \includegraphics[height=.08\linewidth]{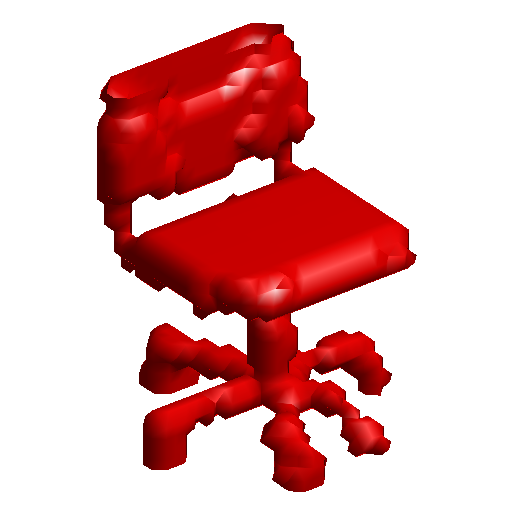} \hspace{-3mm}	
    \includegraphics[height=.08\linewidth]{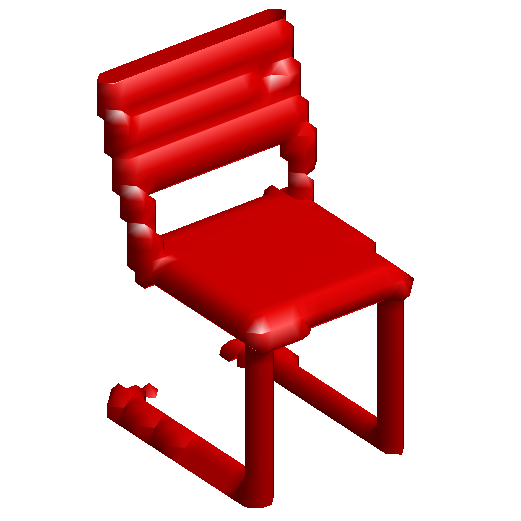} \hspace{-3mm}	
    \includegraphics[height=.08\linewidth]{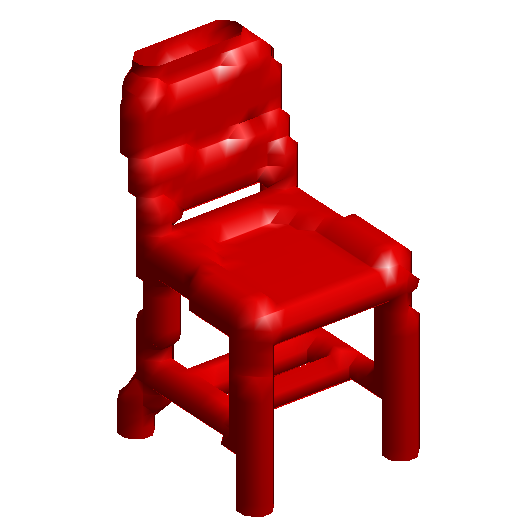} \hspace{-3mm}	
    \includegraphics[height=.08\linewidth]{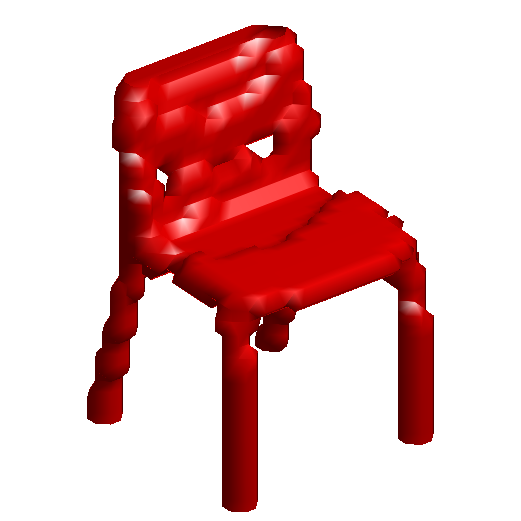} \hspace{-2mm}	
     \includegraphics[height=.08\linewidth]{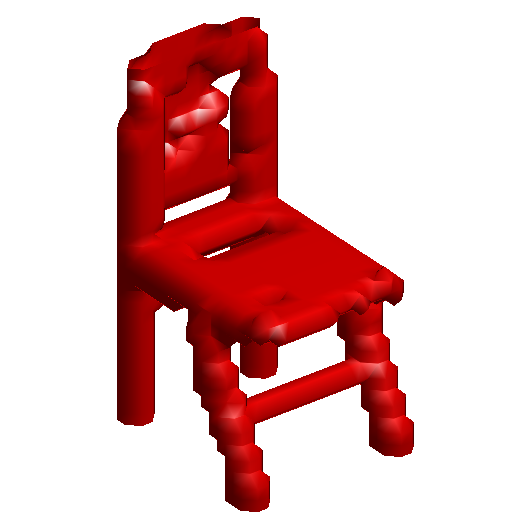}  \hspace{-2mm}   
     \includegraphics[height=.08\linewidth]{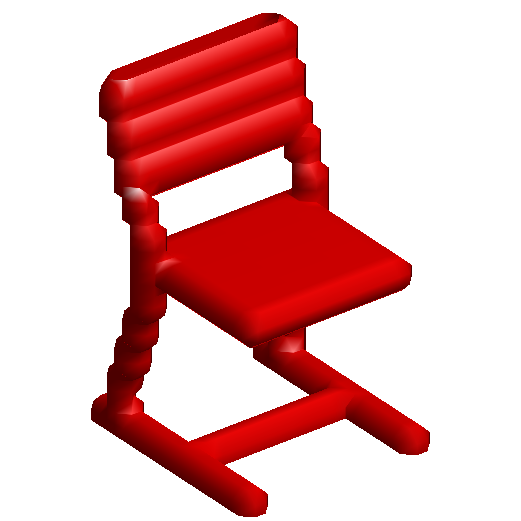} \hspace{-3mm}
     \includegraphics[height=.08\linewidth]{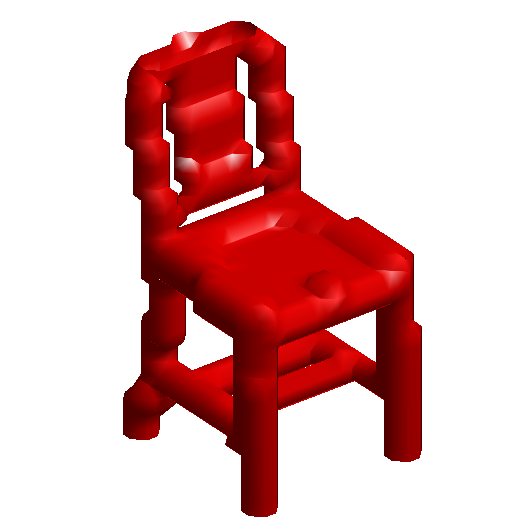}
      \hspace{-3mm}
     \includegraphics[height=.08\linewidth]{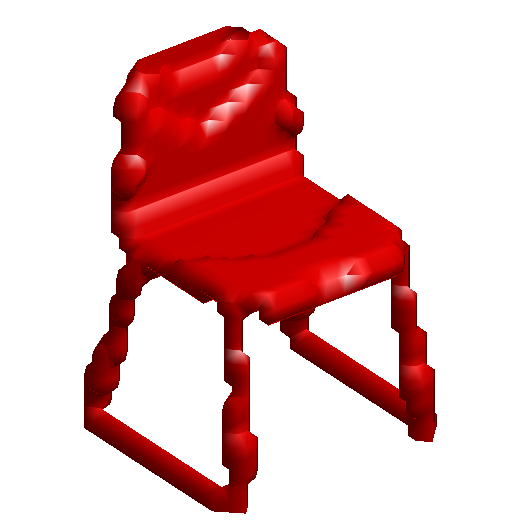} \hspace{-3mm}
     \includegraphics[height=.08\linewidth]{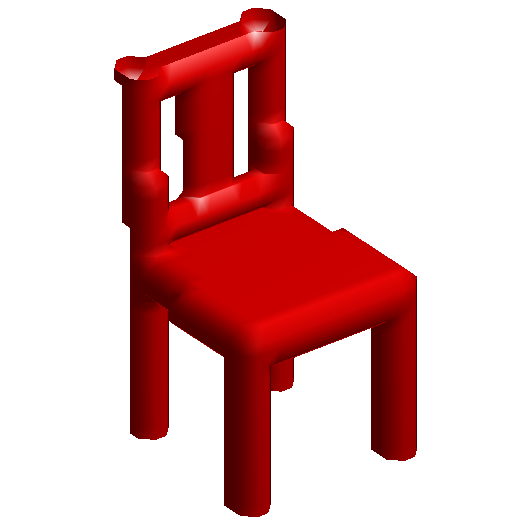}     
      \\ 
    \rotatebox[origin=l]{90}{\hspace{2mm} \textbf{{\footnotesize bed}}}
	\includegraphics[height=.07\linewidth]{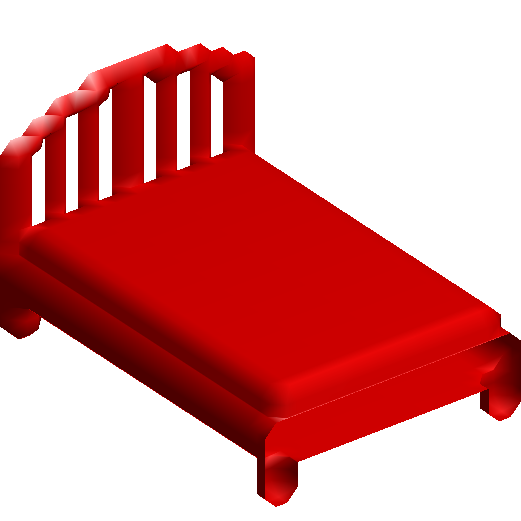} \hspace{-1mm}
	\includegraphics[height=.07\linewidth]{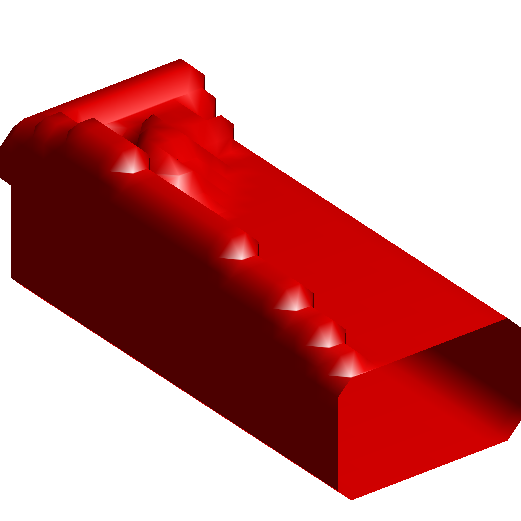} \hspace{-1mm}
	\includegraphics[height=.07\linewidth]{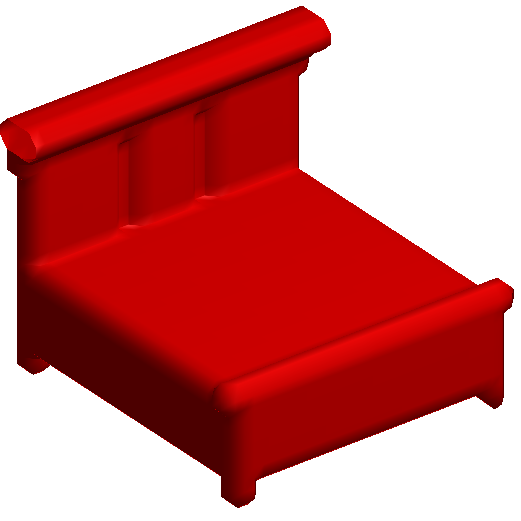}  
	\includegraphics[height=.07\linewidth]{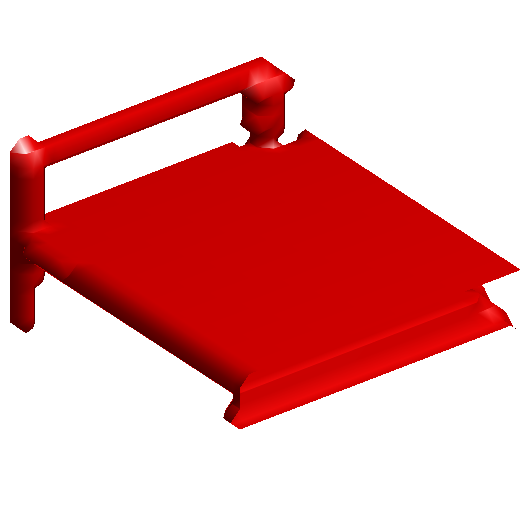}  \hspace{-1mm}
	\includegraphics[height=.07\linewidth]{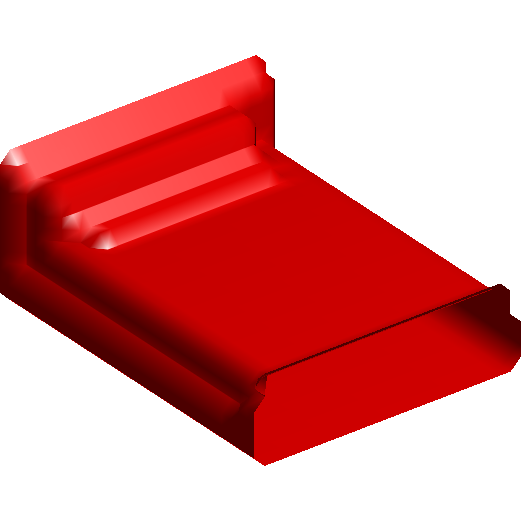}  \hspace{-1mm}
    \includegraphics[height=.07\linewidth]{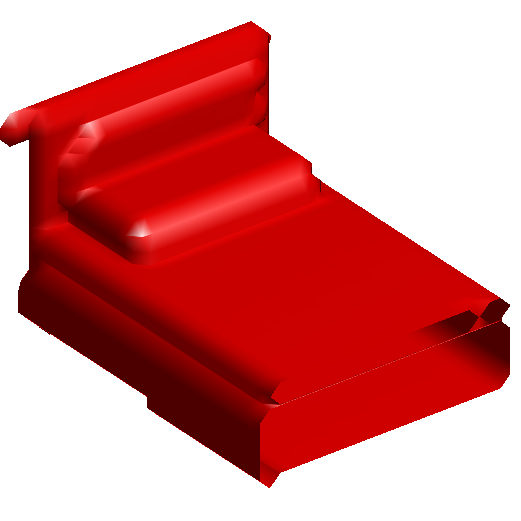}  \hspace{-1mm}
    \includegraphics[height=.07\linewidth]{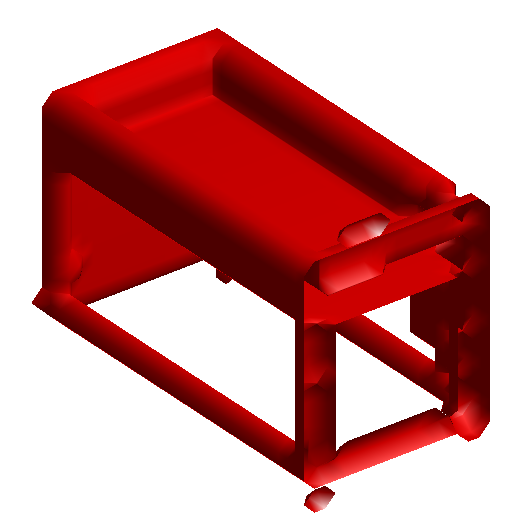}  \hspace{-1mm}
    \includegraphics[height=.07\linewidth]{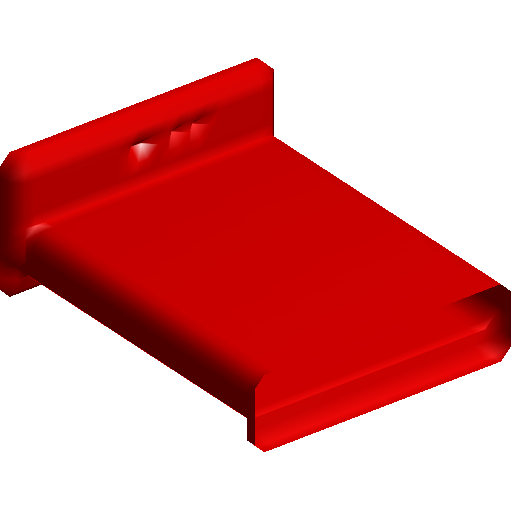} \hspace{-1mm}
    \includegraphics[height=.07\linewidth]{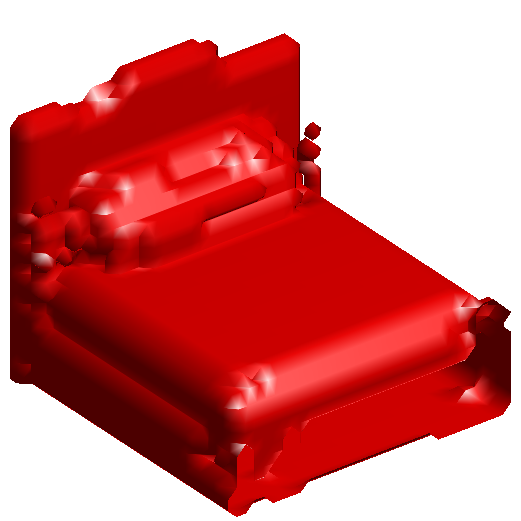}     
    \includegraphics[height=.07\linewidth]{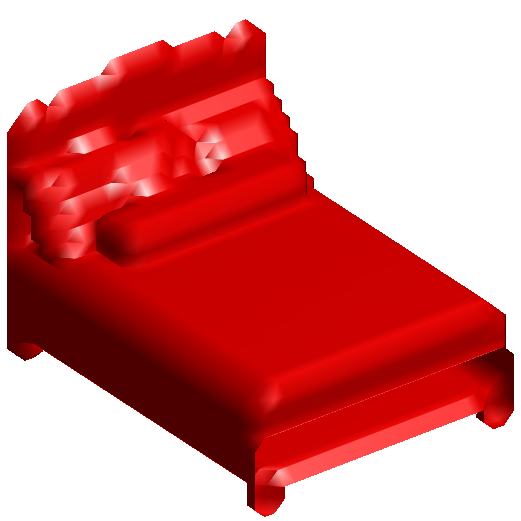}  \hspace{-1mm}
    \includegraphics[height=.07\linewidth]{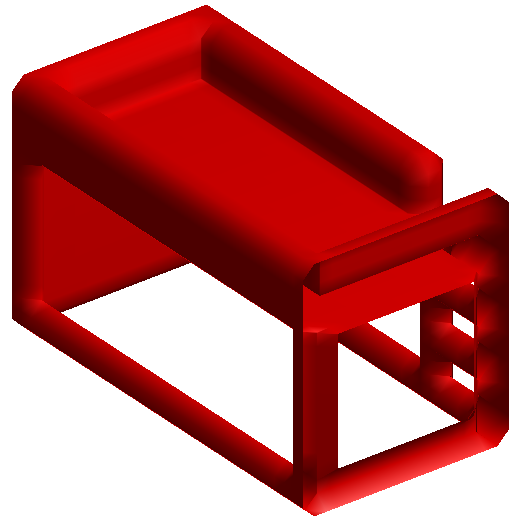}  \hspace{-1mm} 
    \includegraphics[height=.07\linewidth]{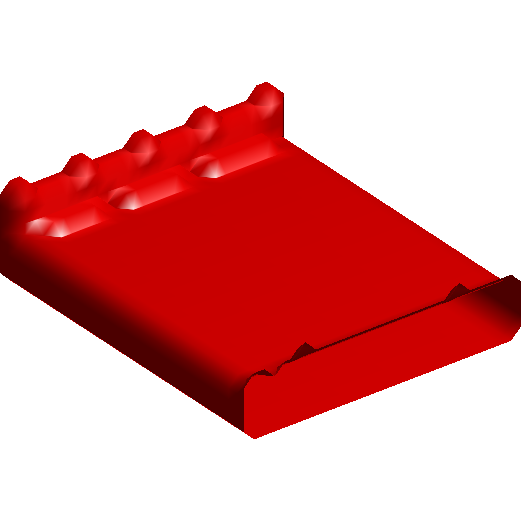} \hspace{-1mm}
    \includegraphics[height=.07\linewidth]{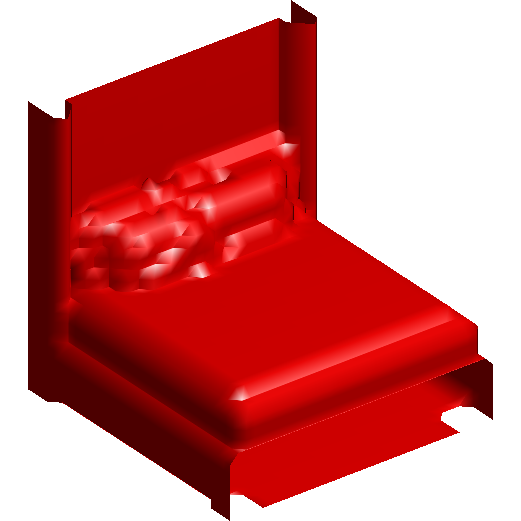}     
      \\
    \rotatebox[origin=l]{90}{\hspace{2mm} \textbf{{\footnotesize sofa}}}
    \includegraphics[height=.07\linewidth]{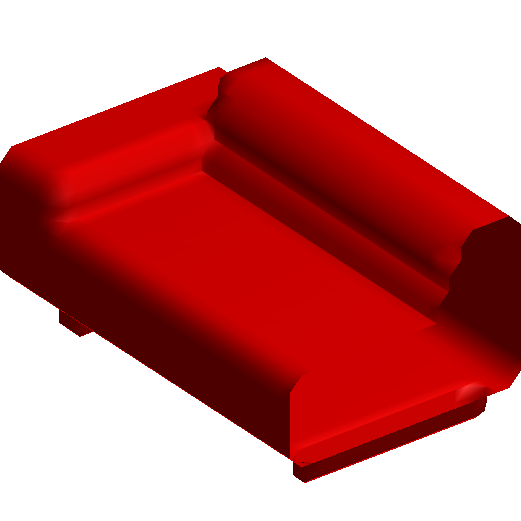}           \hspace{-1mm}
     \includegraphics[height=.07\linewidth]{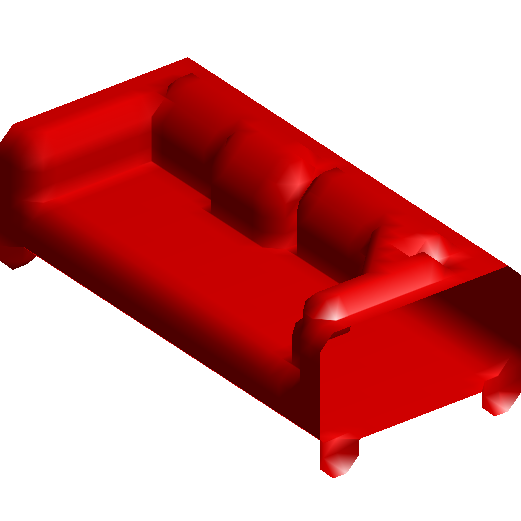}           \hspace{-1mm}
     \includegraphics[height=.07\linewidth]{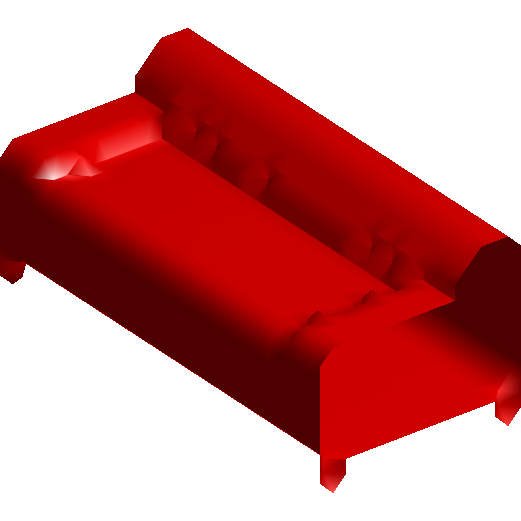}           
     \includegraphics[height=.07\linewidth]{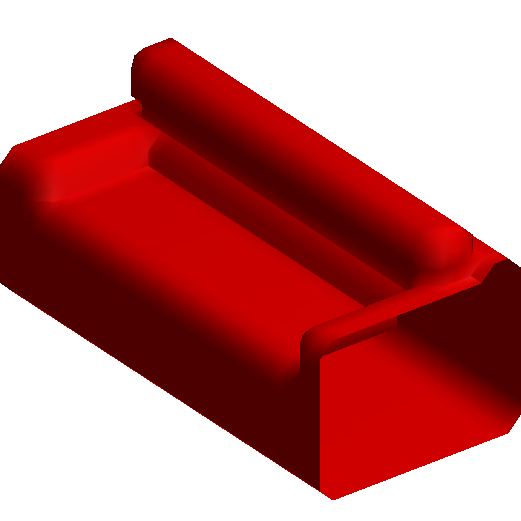}           \hspace{-1mm}  
      \includegraphics[height=.07\linewidth]{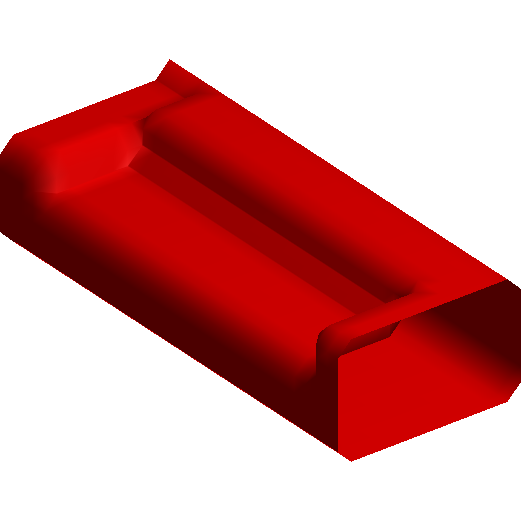}           \hspace{-1mm}    
      \includegraphics[height=.07\linewidth]{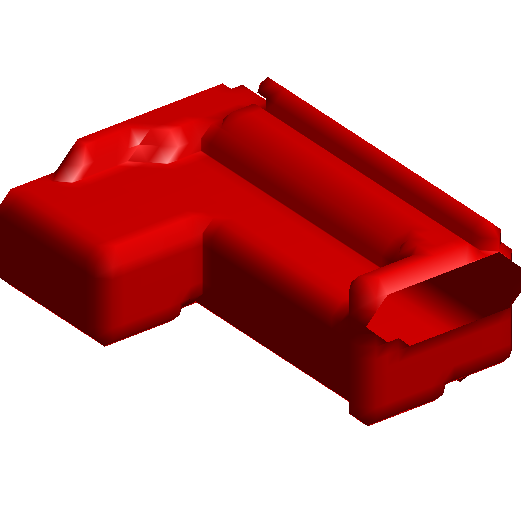}           \hspace{-1mm}
      \includegraphics[height=.07\linewidth]{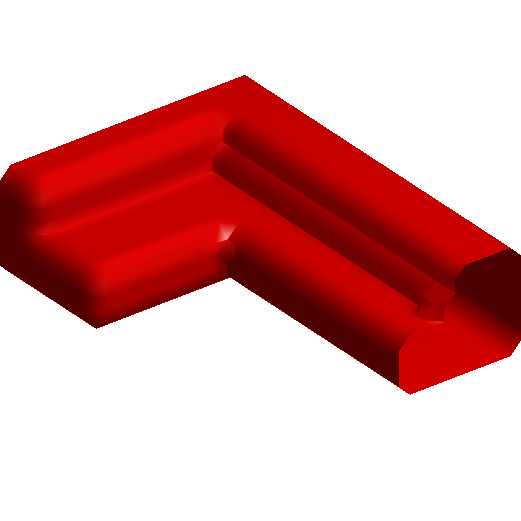}           \hspace{-1mm}     
      \includegraphics[height=.07\linewidth]{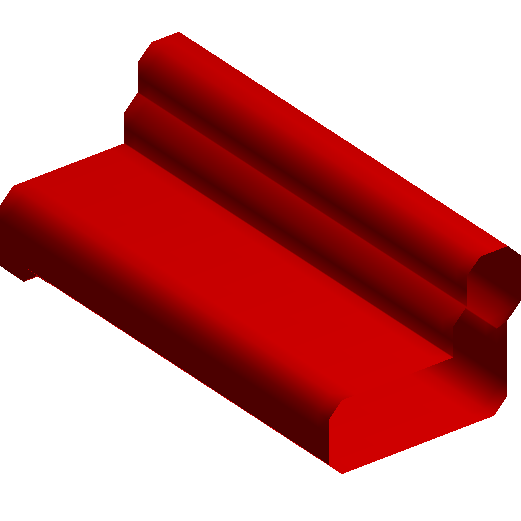}           \hspace{-1mm} 
      \includegraphics[height=.07\linewidth]{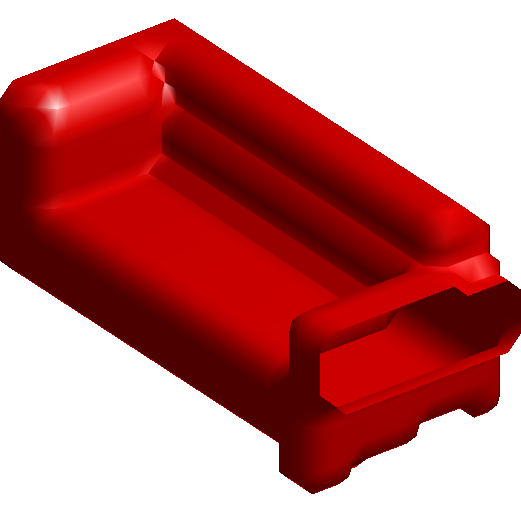}           
      \includegraphics[height=.07\linewidth]{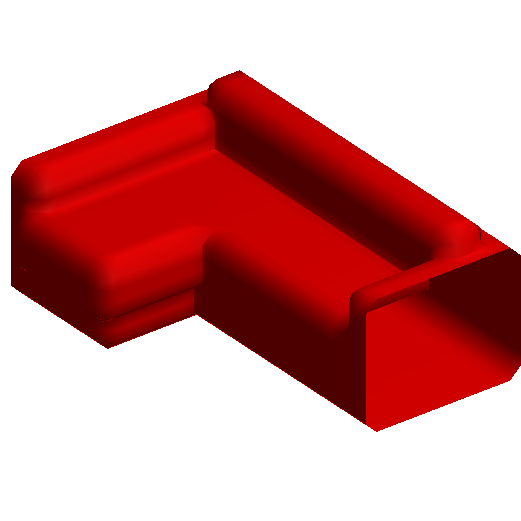}           \hspace{-1mm}
      \includegraphics[height=.07\linewidth]{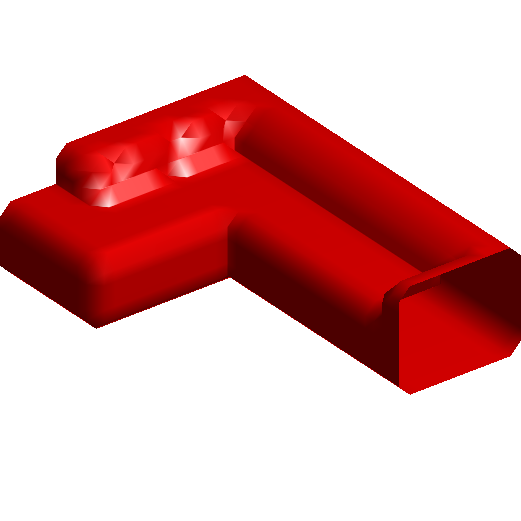}           \hspace{-1mm}     
      \includegraphics[height=.07\linewidth]{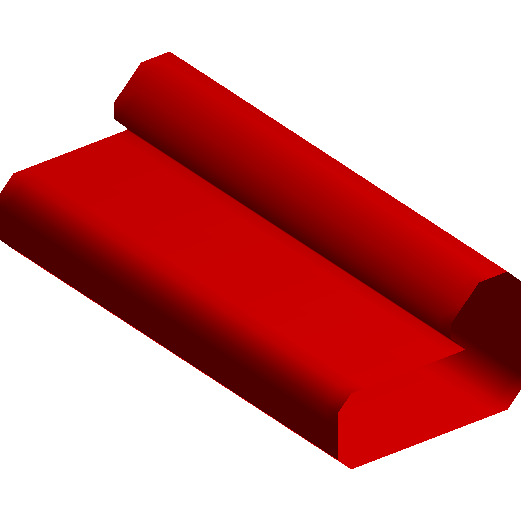}           \hspace{-1mm} 
      \includegraphics[height=.07\linewidth]{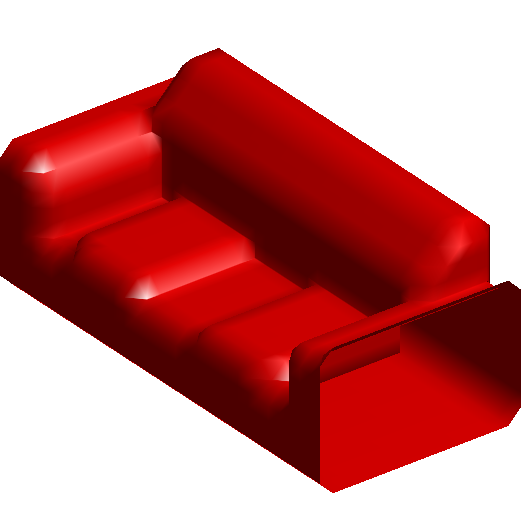}           \hspace{-1mm}     
      \\
     \rotatebox[origin=l]{90}{\hspace{2mm} \textbf{{\footnotesize table}}}
      \includegraphics[height=.07\linewidth]{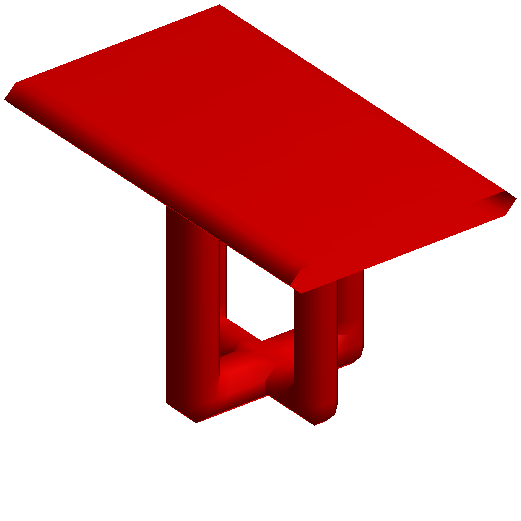}           \hspace{-1mm} 
      \includegraphics[height=.07\linewidth]{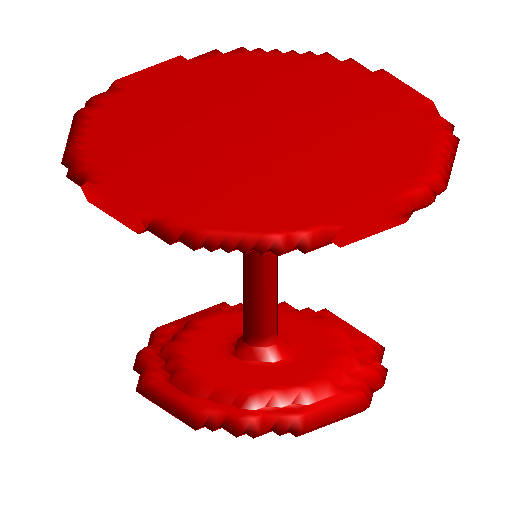}           \hspace{-1mm} 
       \includegraphics[height=.07\linewidth]{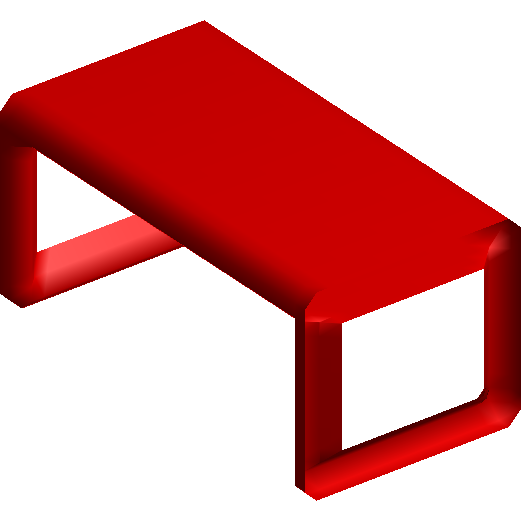}           
      \includegraphics[height=.07\linewidth]{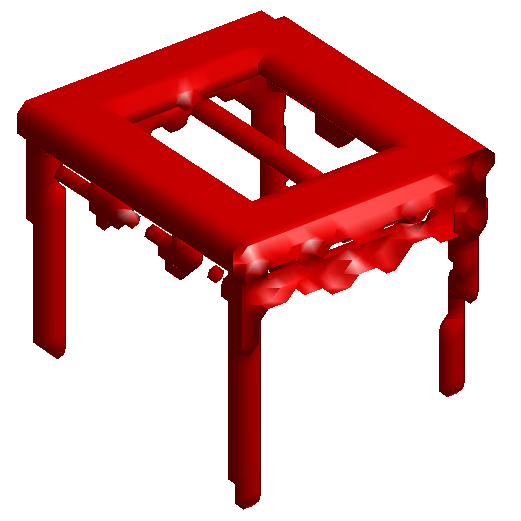}           \hspace{-1mm} 
     \includegraphics[height=.07\linewidth]{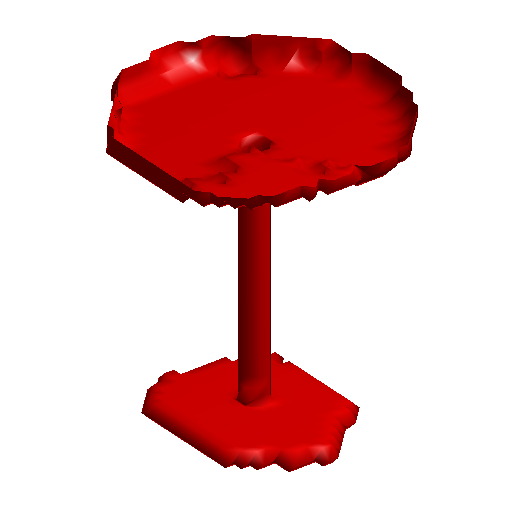}           \hspace{-1mm} 
     \includegraphics[height=.07\linewidth]{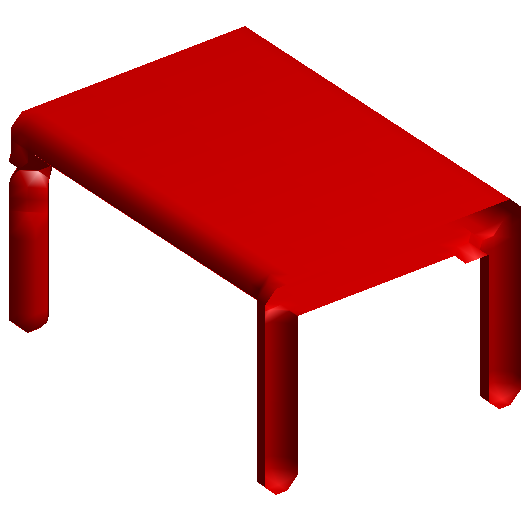}           \hspace{-1mm} 
      \includegraphics[height=.07\linewidth]{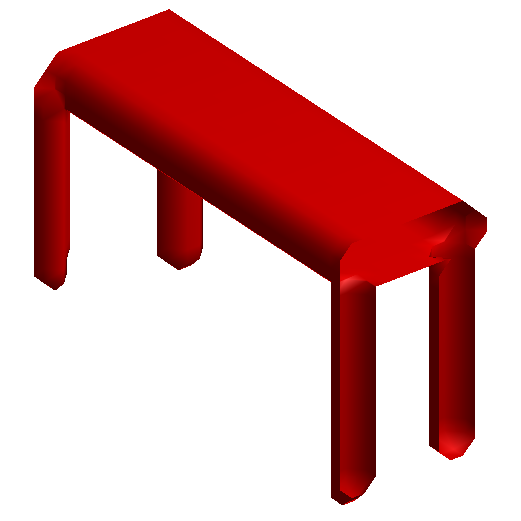}           \hspace{-1mm} 
      \includegraphics[height=.07\linewidth]{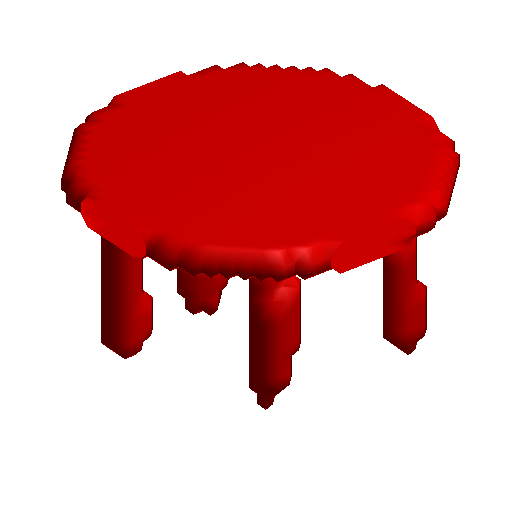}           \hspace{-1mm} 
     \includegraphics[height=.07\linewidth]{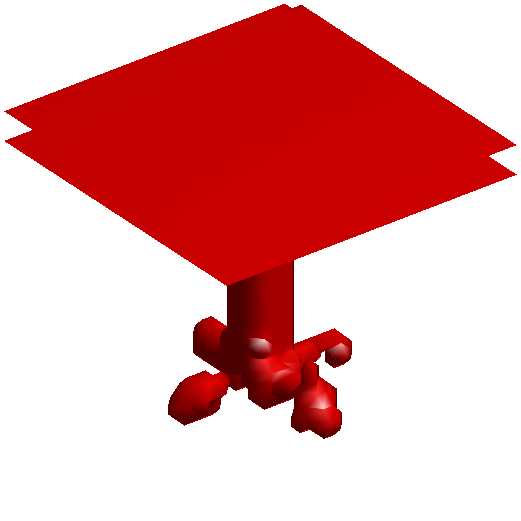}                
      \includegraphics[height=.07\linewidth]{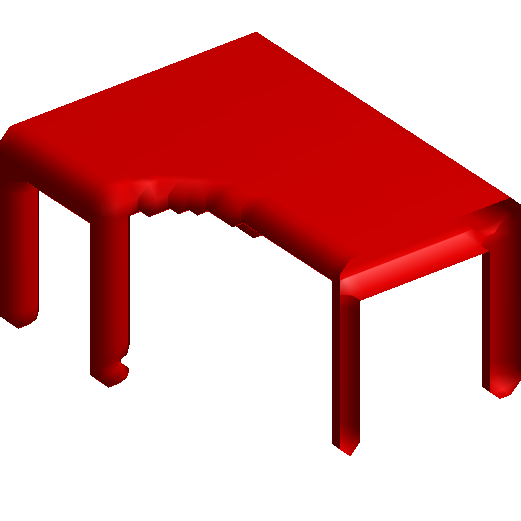}           \hspace{-1mm} 
      \includegraphics[height=.07\linewidth]{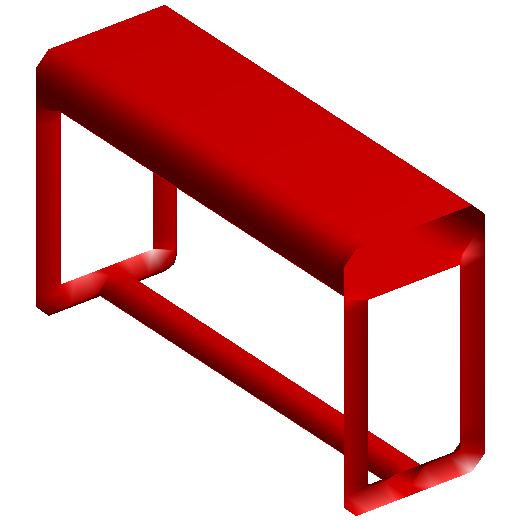}           \hspace{-1mm} 
     \includegraphics[height=.07\linewidth]{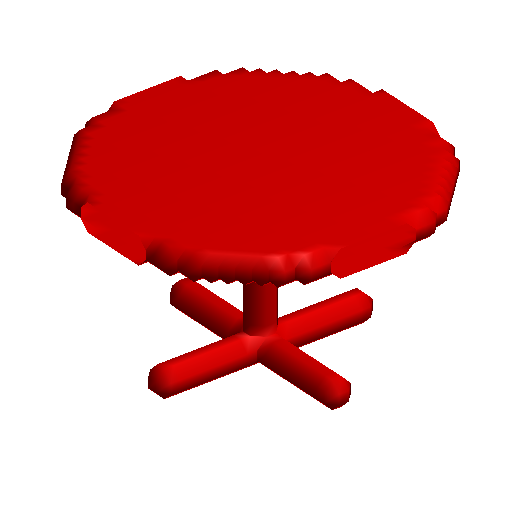}           \hspace{-1mm} 
    \includegraphics[height=.07\linewidth]{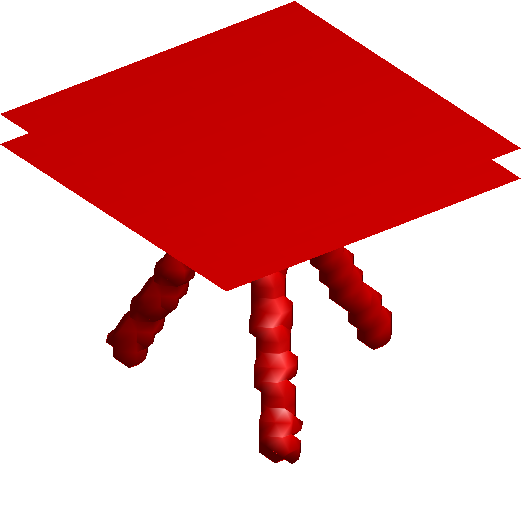}           \\
    \rotatebox[origin=l]{90}{\hspace{1mm}\textbf{{\footnotesize dresser}}}
    \includegraphics[height=.07\linewidth]{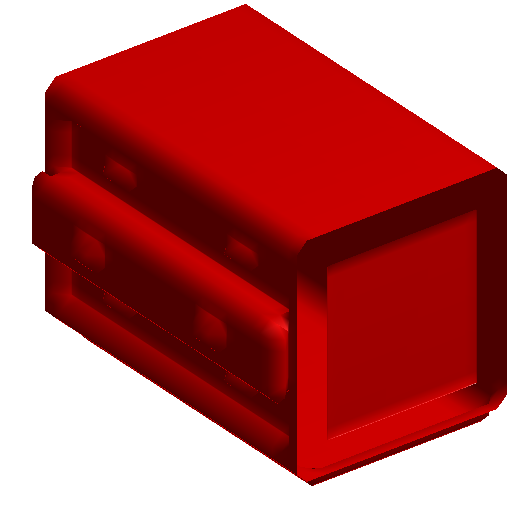}          \hspace{-1mm}  
     \includegraphics[height=.07\linewidth]{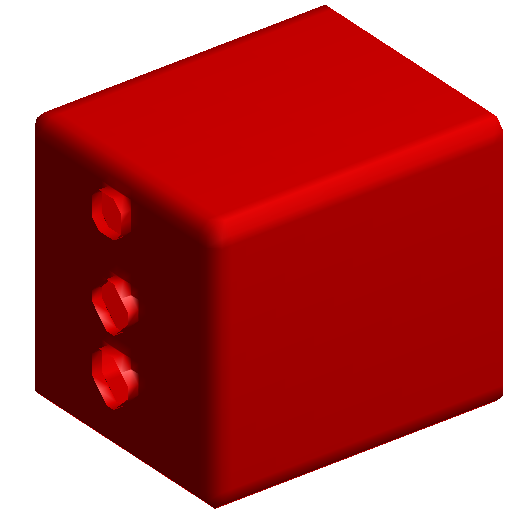}          \hspace{-1mm}  
      \includegraphics[height=.07\linewidth]{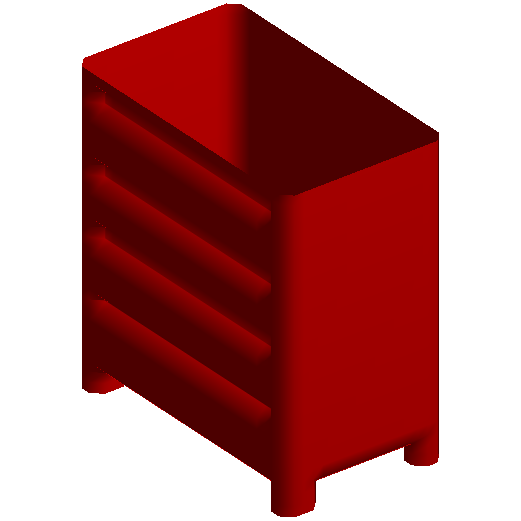}         
    \includegraphics[height=.07\linewidth]{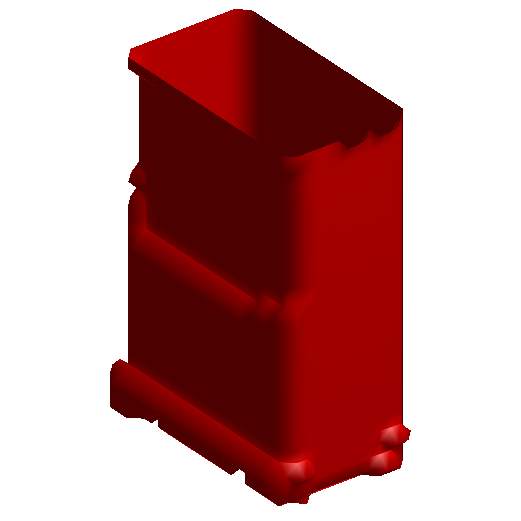}          \hspace{-1mm} 
     \includegraphics[height=.07\linewidth]{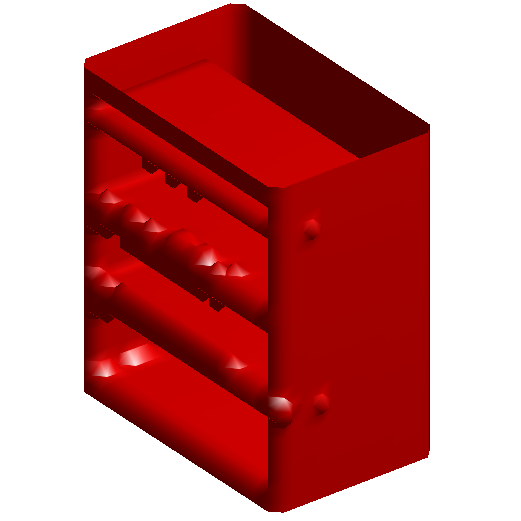}          \hspace{-1mm} 
     \includegraphics[height=.07\linewidth]{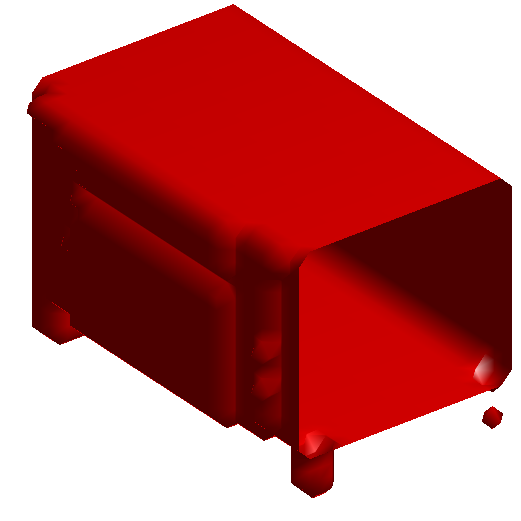}          \hspace{-1mm} 
     \includegraphics[height=.07\linewidth]{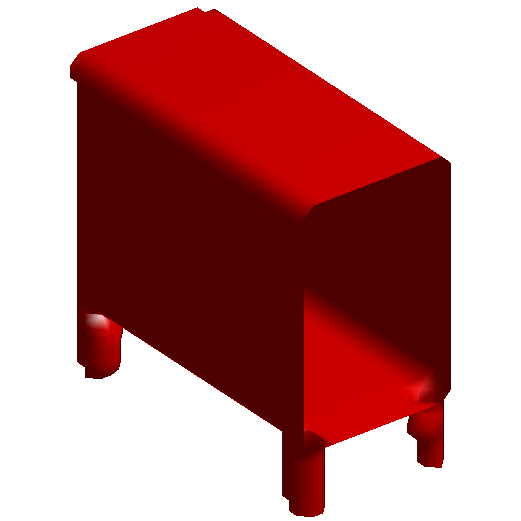}          \hspace{-1mm} 
     \includegraphics[height=.07\linewidth]{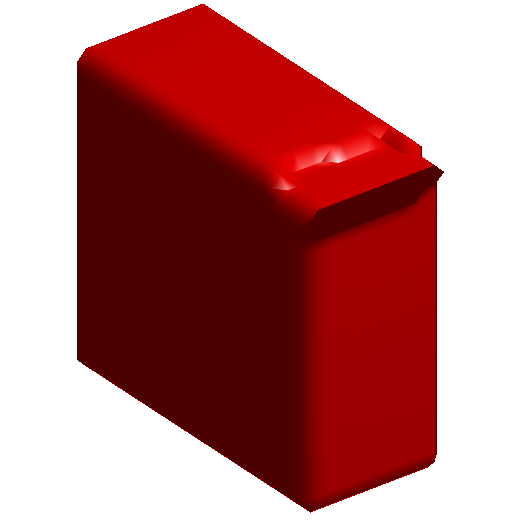}          \hspace{-1mm}  
     \includegraphics[height=.07\linewidth]{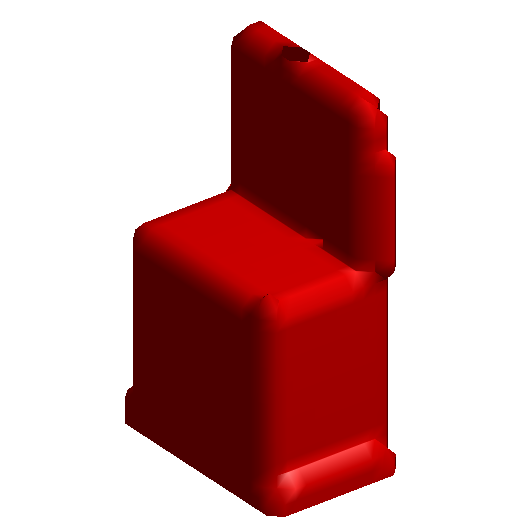}          
      \includegraphics[height=.07\linewidth]{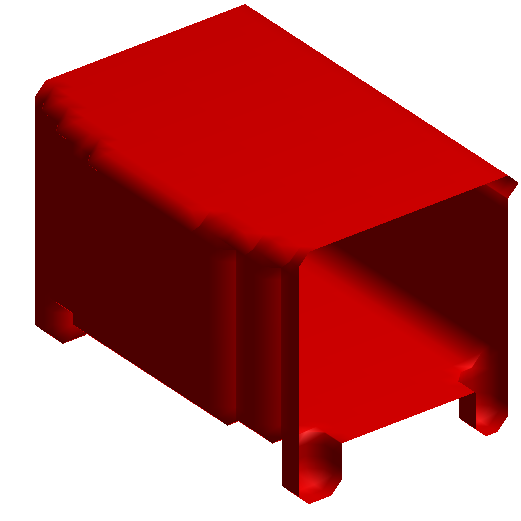}          \hspace{-1mm} 
     \includegraphics[height=.07\linewidth]{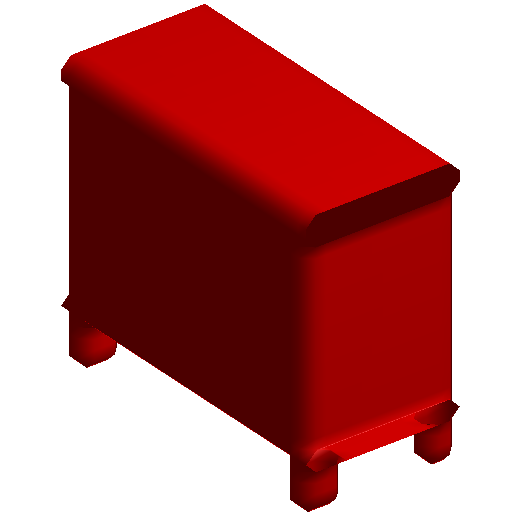}          \hspace{-1mm} 
     \includegraphics[height=.07\linewidth]{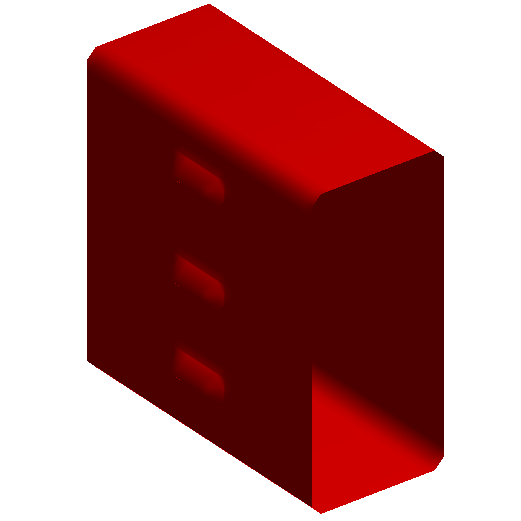}          \hspace{-1mm} 
     \includegraphics[height=.07\linewidth]{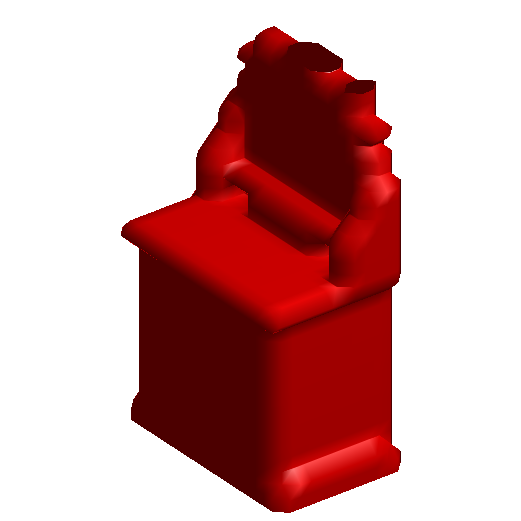}          \hspace{-1mm} \\
     \rotatebox[origin=l]{90}{\textbf{{\footnotesize night stand}}}
     \includegraphics[height=.07\linewidth]{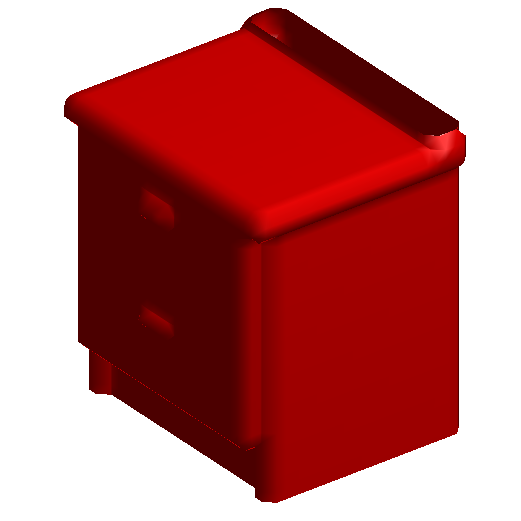}          \hspace{-1mm}  
     \includegraphics[height=.07\linewidth]{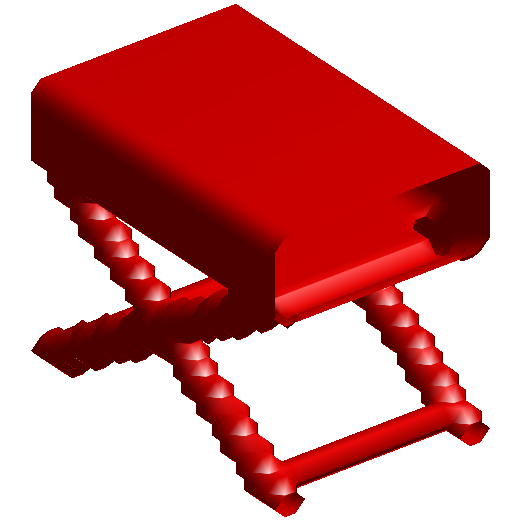}          \hspace{-1mm}  
     \includegraphics[height=.07\linewidth]{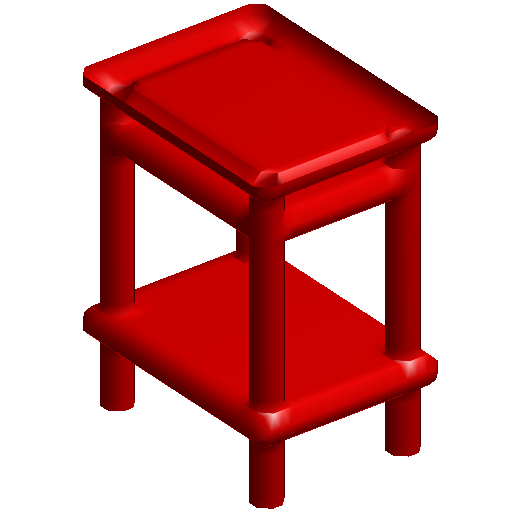}          \hspace{-1mm} 
     \includegraphics[height=.07\linewidth]{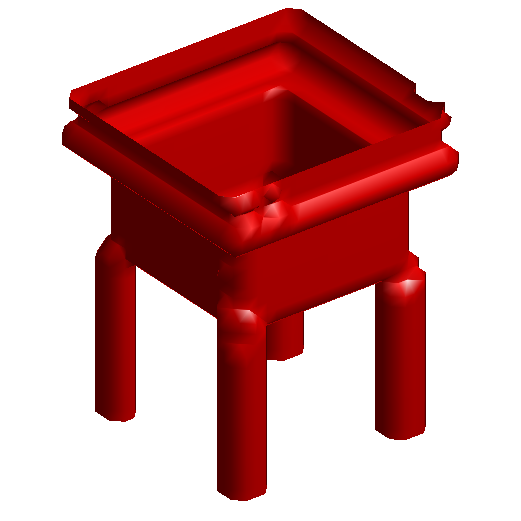}          \hspace{-1mm} 
       \includegraphics[height=.07\linewidth]{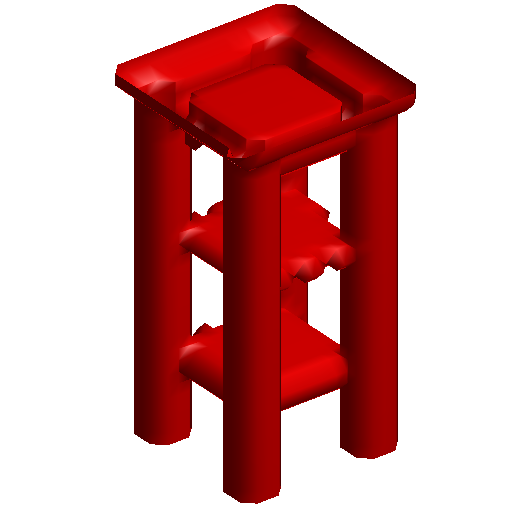}          \hspace{-1mm} 
      \includegraphics[height=.07\linewidth]{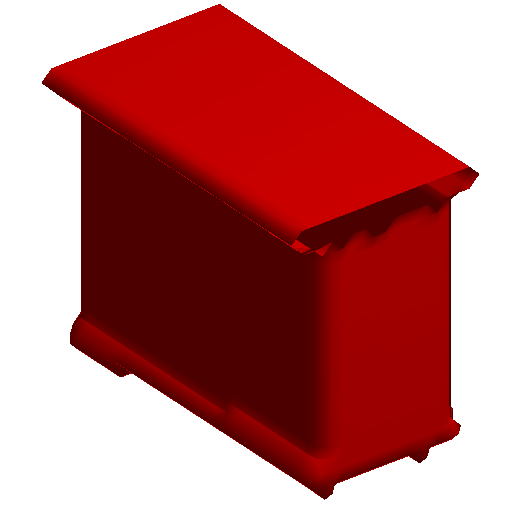}          \hspace{-1mm}  
       \includegraphics[height=.07\linewidth]{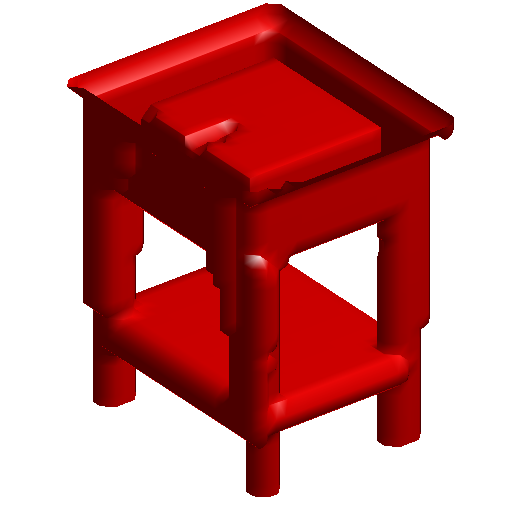}          \hspace{-1mm}  
     \includegraphics[height=.07\linewidth]{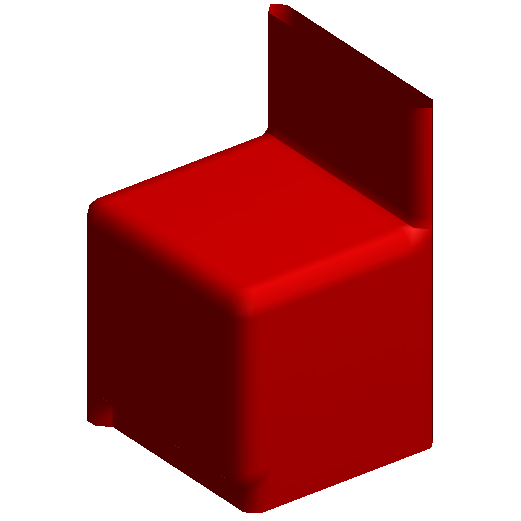}          \hspace{-1mm}  
     \includegraphics[height=.07\linewidth]{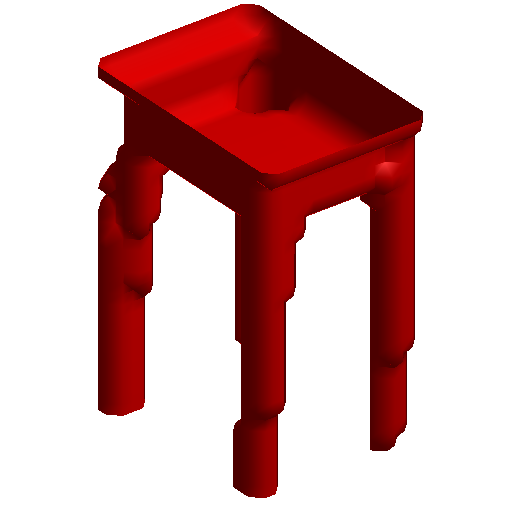}          \hspace{-1mm}
     \includegraphics[height=.07\linewidth]{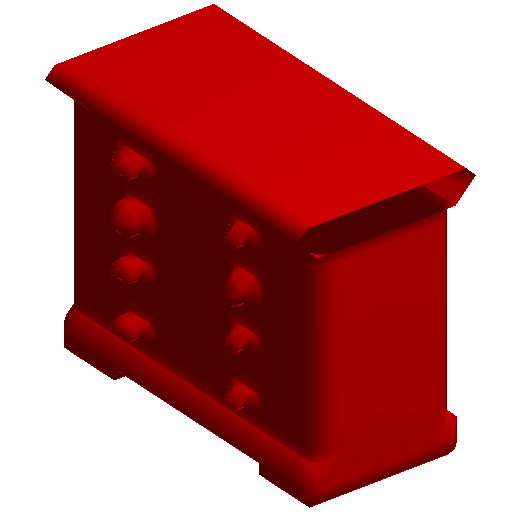}          \hspace{-1mm}   
     \includegraphics[height=.07\linewidth]{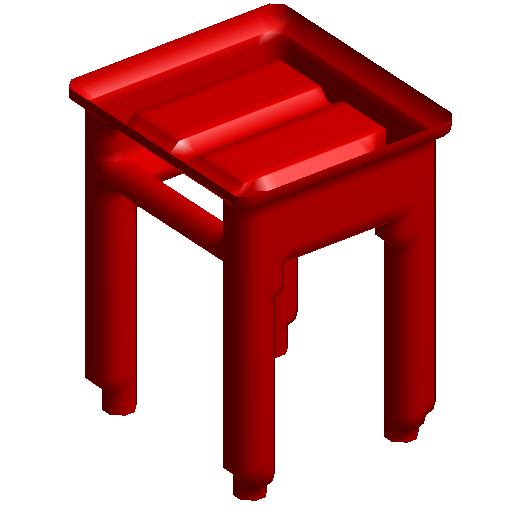}          \hspace{-1mm} 
     \includegraphics[height=.07\linewidth]{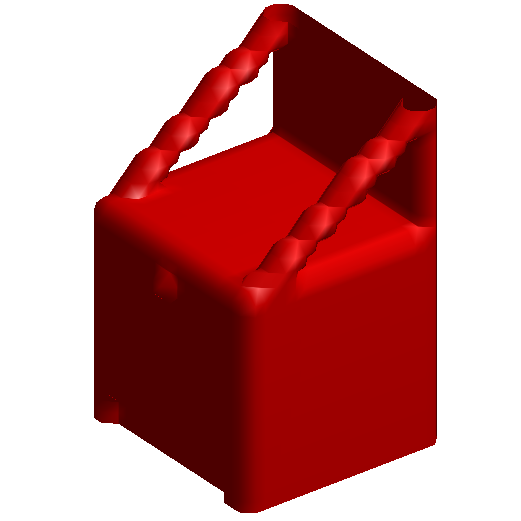}          \hspace{-1mm} 
      \includegraphics[height=.07\linewidth]{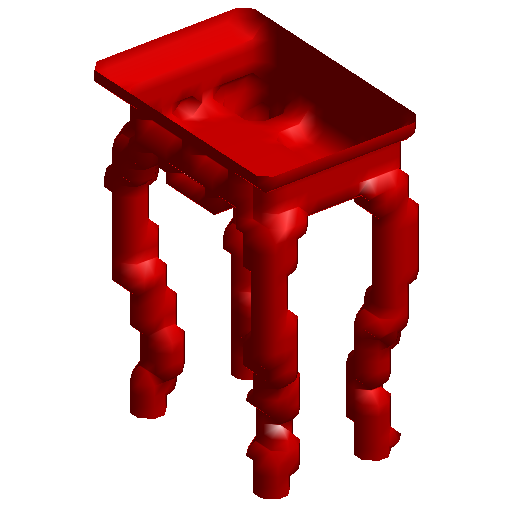}          \hspace{-1mm}\\
     \rotatebox[origin=l]{90}{\hspace{3mm}\textbf{{\footnotesize toilet}}}
        \includegraphics[height=.08\linewidth]{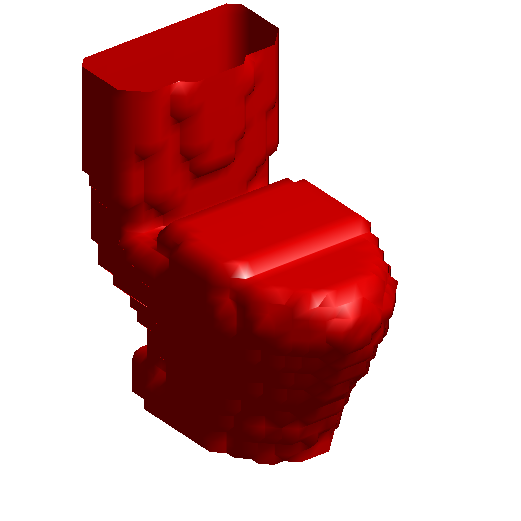} \hspace{-3.5mm}   
       \includegraphics[height=.08\linewidth]{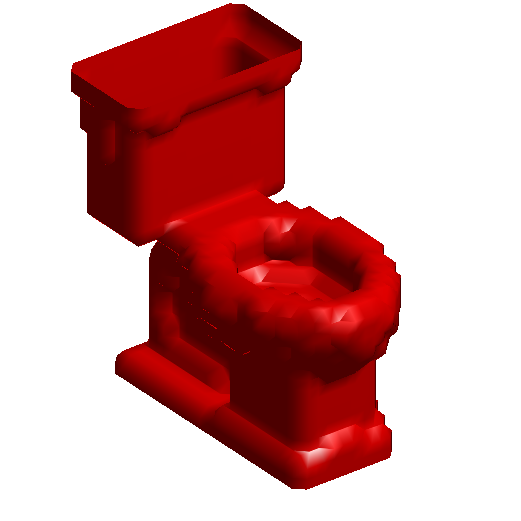}\hspace{-2.5mm}               
        \includegraphics[height=.08\linewidth]{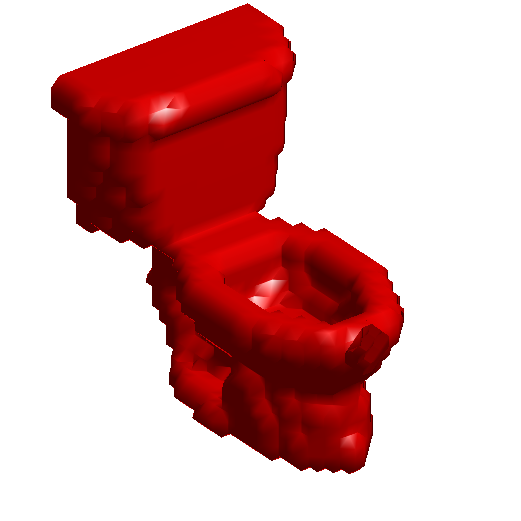}         \hspace{-2.5mm}  
       \includegraphics[height=.08\linewidth]{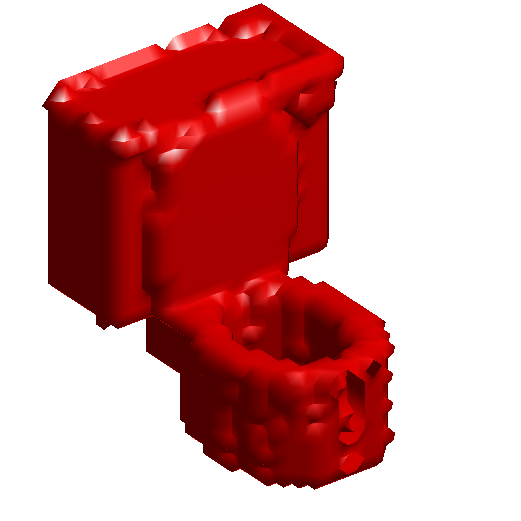}          \hspace{-2.5mm}   
     \includegraphics[height=.08\linewidth]{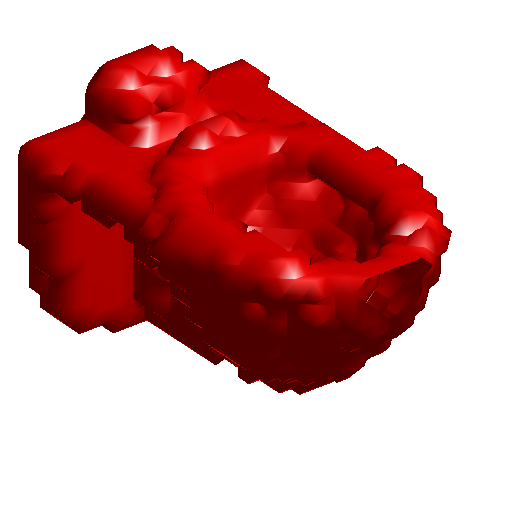}          \hspace{-2.5mm}   
      \includegraphics[height=.08\linewidth]{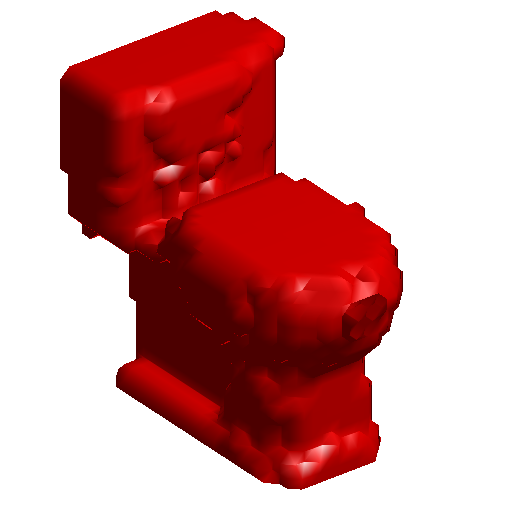}          \hspace{-2.5mm}   
     \includegraphics[height=.08\linewidth]{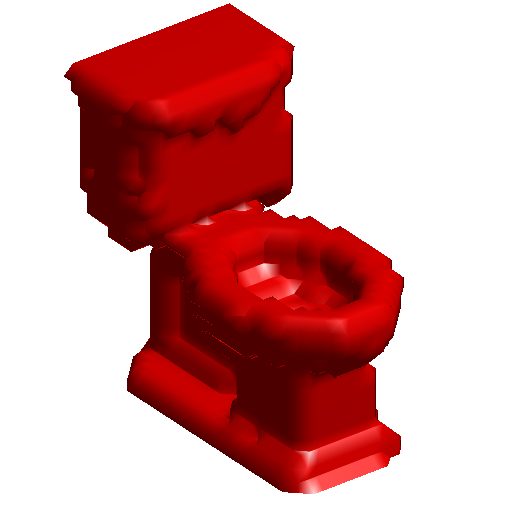}          \hspace{-3.5mm}   
     \includegraphics[height=.08\linewidth]{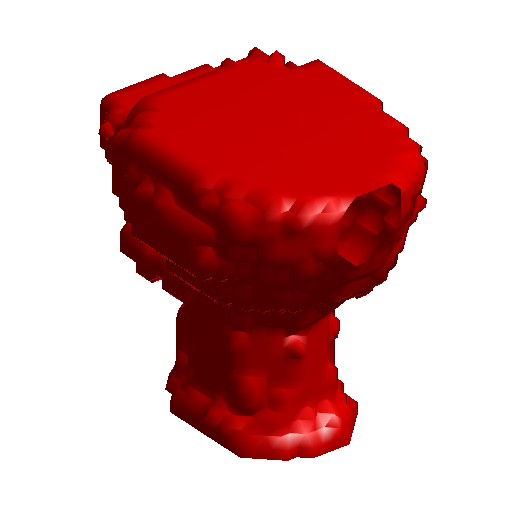}          \hspace{-2.5mm}   
     \includegraphics[height=.08\linewidth]{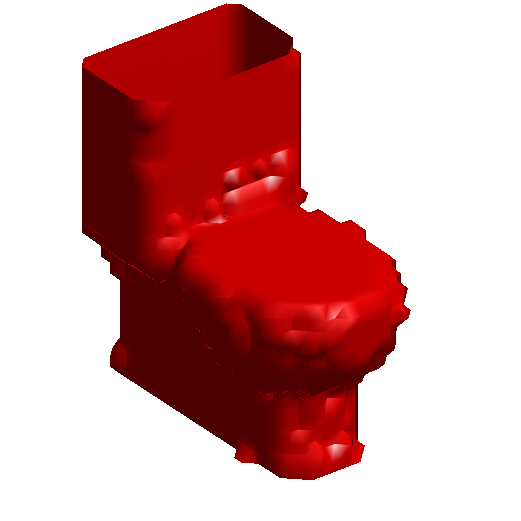}        \hspace{-2.5mm}        
     \includegraphics[height=.08\linewidth]{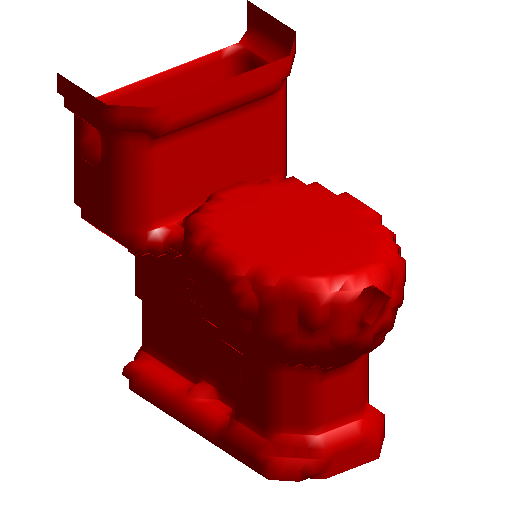}          \hspace{-3mm}   
     \includegraphics[height=.08\linewidth]{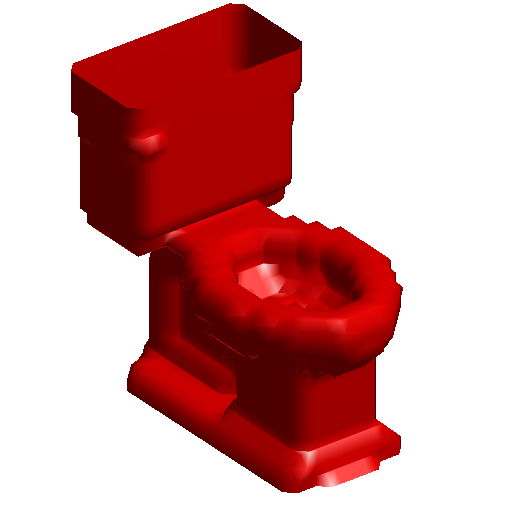}          \hspace{-3mm}   
     \includegraphics[height=.08\linewidth]{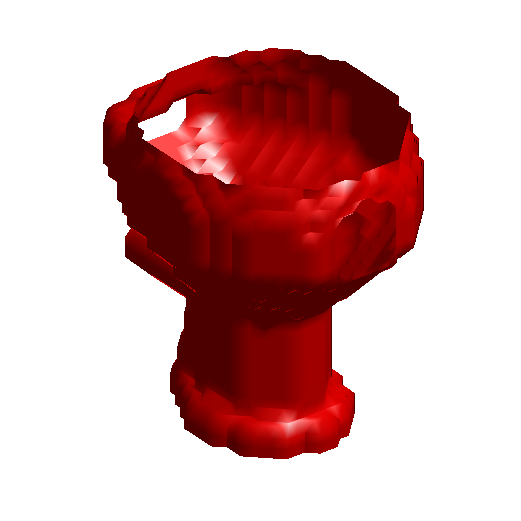}          \hspace{-2.5mm}  
     \includegraphics[height=.08\linewidth]{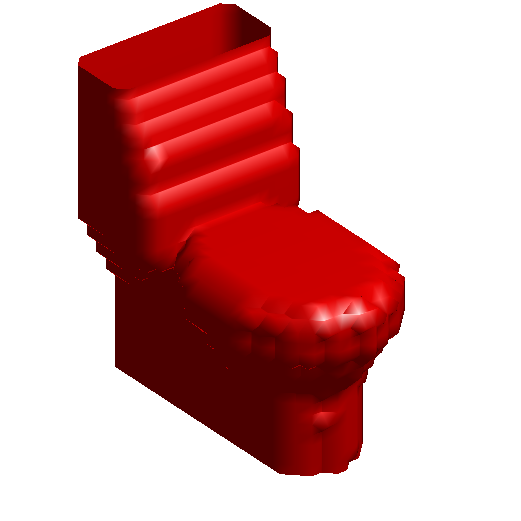}          \\
     \rotatebox[origin=l]{90}{\hspace{1mm}\textbf{{\footnotesize monitor}}}
     \includegraphics[height=.07\linewidth]{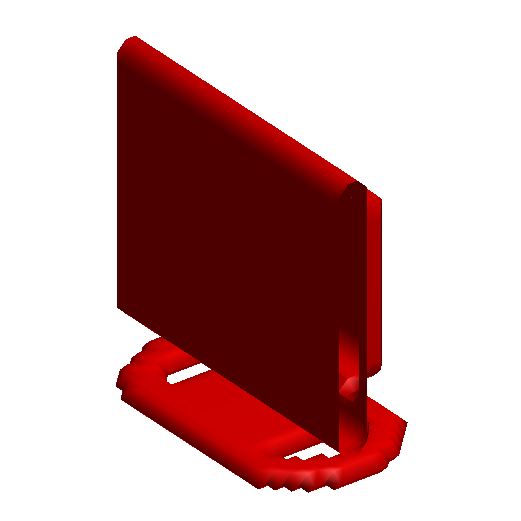}          \hspace{-1mm} 
     \includegraphics[height=.07\linewidth]{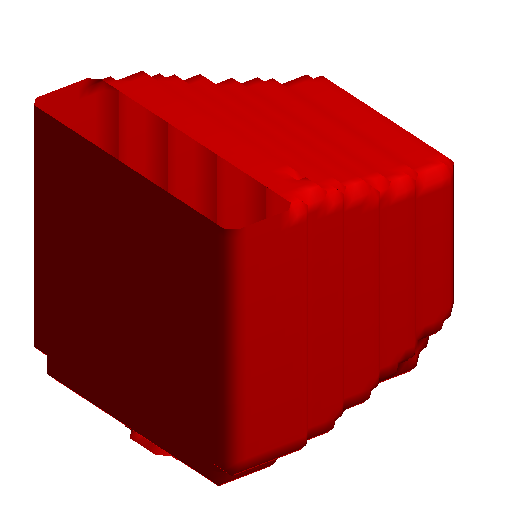}          \hspace{-1mm} 
      \includegraphics[height=.07\linewidth]{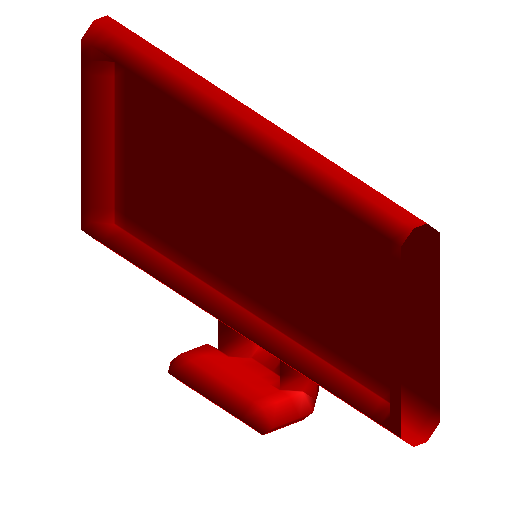}          
      \includegraphics[height=.07\linewidth]{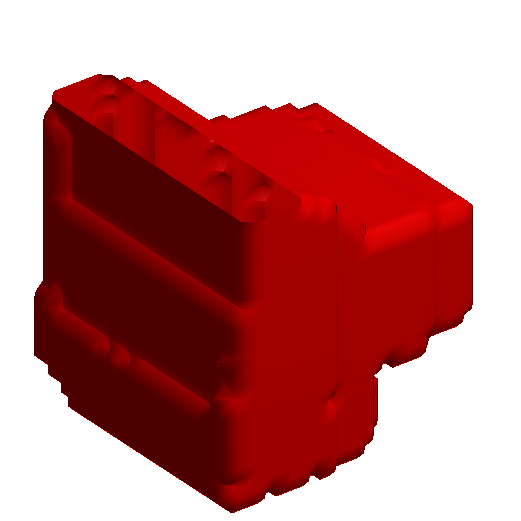}          \hspace{-1mm} 
     \includegraphics[height=.07\linewidth]{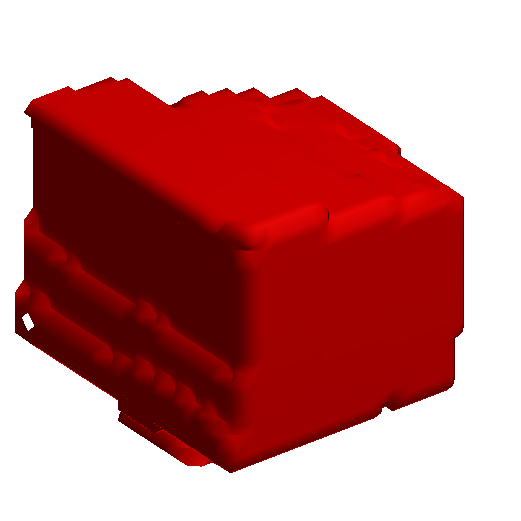}          \hspace{-1mm} 
     \includegraphics[height=.07\linewidth]{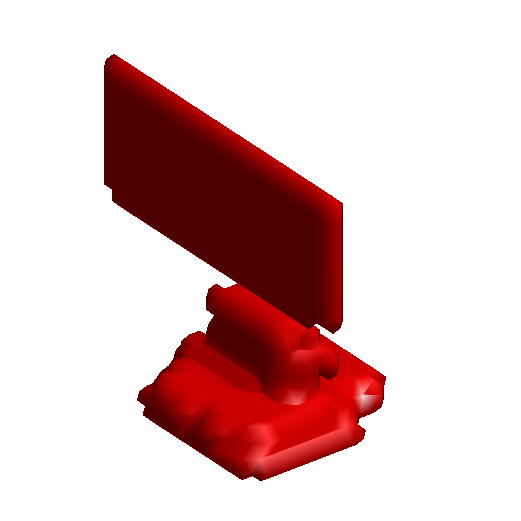}          \hspace{-1mm} 
     \includegraphics[height=.07\linewidth]{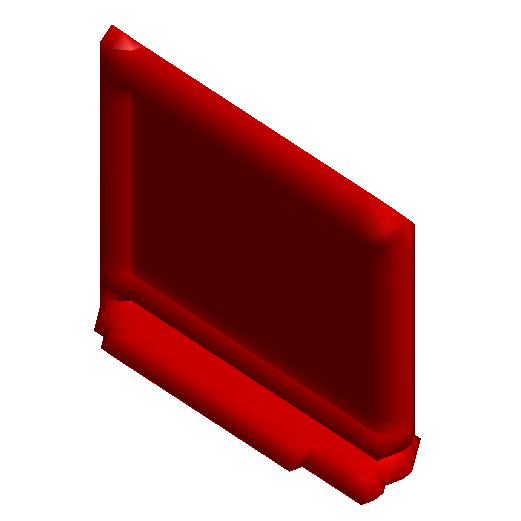}          \hspace{-1mm} 
     \includegraphics[height=.07\linewidth]{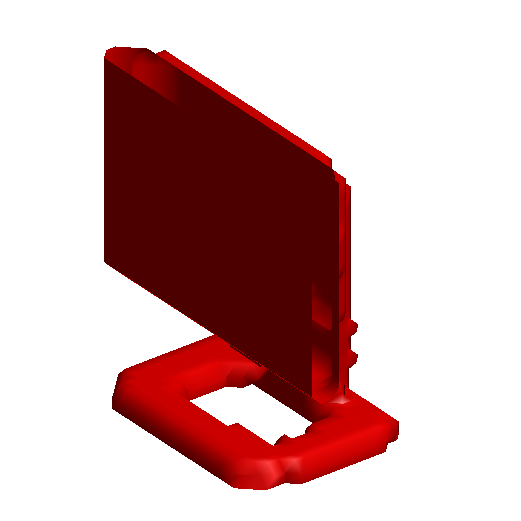}          \hspace{-1mm} 
     \includegraphics[height=.07\linewidth]{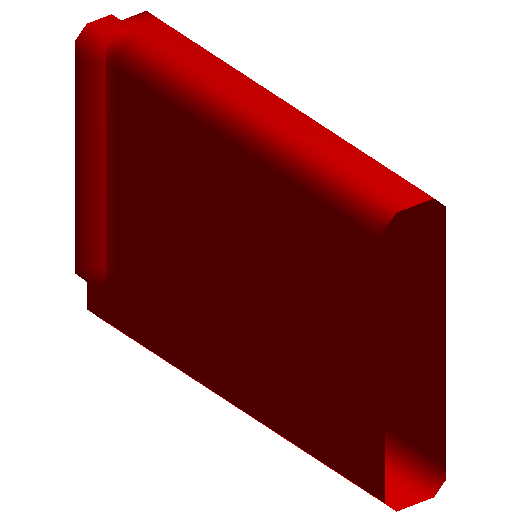} 
     \includegraphics[height=.07\linewidth]{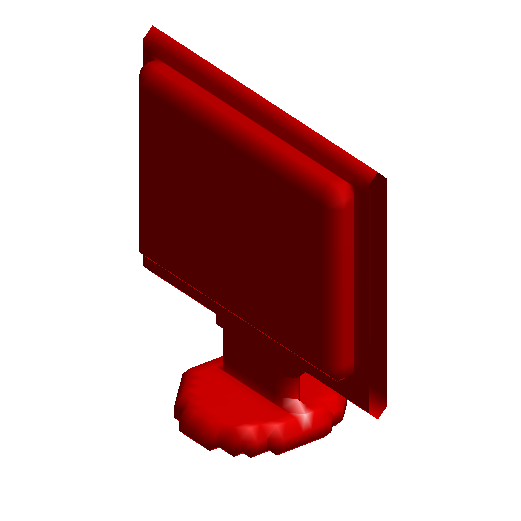}          \hspace{-1mm}
     \includegraphics[height=.07\linewidth]{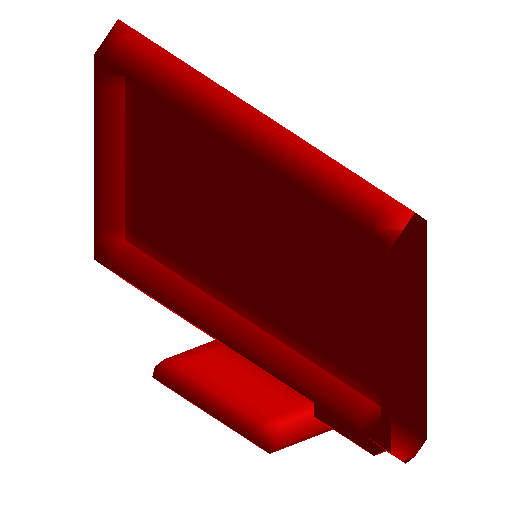}          \hspace{-1mm}         
     \includegraphics[height=.07\linewidth]{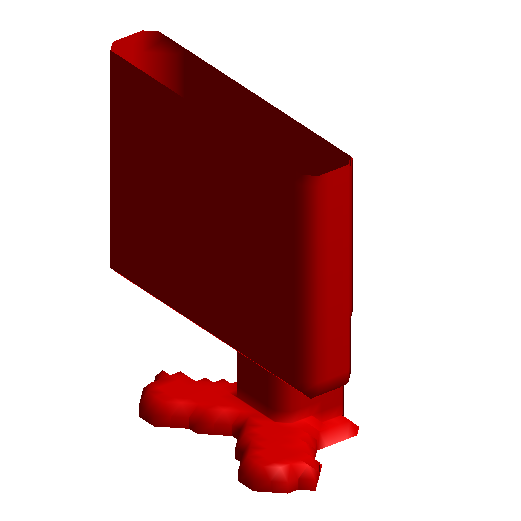}          \hspace{-1mm}
     \includegraphics[height=.07\linewidth]{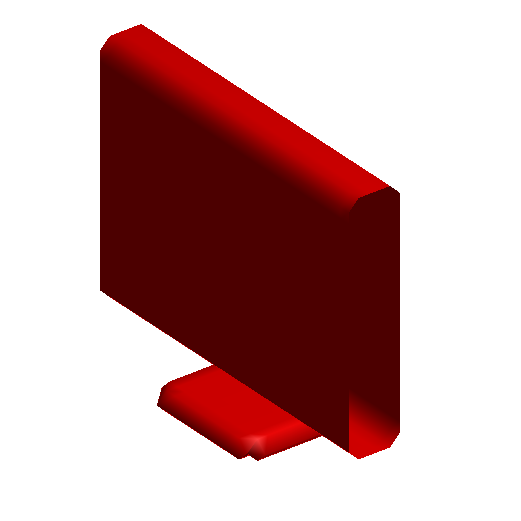}          \hspace{-1mm}
	\caption{Generating 3D objects. Each row displays one experiment, where the first three 3D objects are some observed examples, columns 4, 5, 6, 7, 8, and 9 are 6 of the synthesized 3D objects sampled from the learned model by Langevin dynamics. For the last four synthesized objects (shown in columns 6, 7, 8, and 9), their nearest neighbors retrieved from the training set are shown in columns 10, 11, 12, and 13.  }	
	\label{exp:synthesis}
\end{figure*}

We conduct experiments on learning generative VoxelNet via algorithm \ref{code:3D} to synthesize 3D objects of categories from ModelNet \cite{wu20153d}, which is a large-scale 3D CAD model dataset. Specifically, we use ModelNet10, a 10-category subset of ModelNet which is commonly used as a benchmark for 3D object analysis. The categories are chair, sofa, bathtub, toilet, bed, desk, table, nightstand, dresser, and monitor. The size of the training set for each category ranges from 100 to 700. To obtain voxel-based training data, we voxelize each CAD model in the ModelNet dataset by the following method: Each 3D mesh is represented as a binary 3D voxel grid, with 1 indicating the voxel is inside the mesh surface, and 0 indicating the voxel is outside the mesh.

For qualitative experiment, we learn one 3-layer generative VoxelNet for each object category in ModelNet10. The first layer has 200 $16\times16\times16$ filters with sub-sampling of 3, the second layer has 100 $6\times6\times6$ filters with sub-sampling of $2$, and the final layer is a fully connected layer with a single filter that covers the whole voxel grid. We add ReLU layers between  adjacent convolutional layers. We fix the standard deviation of the reference distribution of the model to be $s=0.5$. The number of Langevin dynamics steps in each learning iteration is $K$=20 and the step size $\Delta \tau =0.01$. We use Adam \cite{kingma2015adam} for optimization with $\beta_1=0.5$ and $\beta_2=0.999$. The learning rate is 0.001. The number of learning iterations is $3,000$. We disable the noise term in the Langevin step after $100$ iterations. The training data are of size $32 \times 32 \times 32$ voxels, whose values are 0 or 1. We pre-process the training data by subtracting the mean value of the dataset. We post-process the synthesized data by adding back the mean value and discretizing each voxel value into 0 or 1 by comparing it with a threshold 0.5. The mini-batch size is 20. The number of parallel chains for each batch is 25.

Figure \ref{exp:synthesis} displays the observed 3D objects randomly sampled from the training set, and the synthesized 3D objects generated by our models for categories chair, bed, sofa, table, dresser, night stand, toilet, and monitor. 
We visualize volumetric data via isosurfaces in our paper. To show that our model can synthesize new 3D objects beyond the training set, we compare the synthesized patterns with their nearest neighbors in the training set. The retrieved nearest neighbors are based on $\ell_2$ distance in the voxel space. As shown in Figure \ref{exp:synthesis}, our model can synthesize realistic 3D shape patterns, and the generated 3D objects are similar, but not identical, to the training set.

To quantitatively evaluate our model, we adopt the Inception score proposed by \cite{warde2016improving}, which uses a reference convolutional neural network to compute  
\begin{eqnarray} 
I(\{\tY_i,i=1,...,\tilde{n}\})=\exp\left( \E_{\tY}\left[ \KL( p(c|\tY)\parallel p(c))\right]\right), \nonumber 
\end{eqnarray} 
where $c$ denotes category, $\{\tY_i,i=1,...,\tilde{n}\}$ are synthesized examples sampled from the model, $p(c|\tY)$ is obtained from the output of the reference network, and $p(c)
\approx \frac{1}{\tilde{n}} \sum_{i=1}^{\tilde{n}} p(c|\tY_i)$. Both a low entropy conditional category distribution $p(c|\tY)$ (i.e., the  network classifies a given sample with high certainty) and a high entropy category distribution $p(c)$ (i.e., the network identifies a wide variety of categories among the generated samples) can lead to a high inception score. In our experiment, we use a state-of-the-art 3D multi-view ConvNet \cite{qi2016volumetric} trained on ModelNet dataset for 3D object classification as the reference network.

We firstly evaluate the generative VoxelNet learned from the training sets of 10-category  mixed 3D objects. Table \ref{tb:inception} reports the Inception scores of our model and some baseline models, including 3D ShapeNets \cite{wu20153d}, 3D-GAN \cite{3dgan},  3D-VAE \cite{brock2016generative}, 3D-WINN \cite{huang20193d}, and Primitive GAN \cite{khan2019unsupervised}.

We then evaluate the quality of the synthesized 3D shapes by the model learned from a single category by using two criteria: (a) average softmax class probability that the reference network assigns to the synthesized examples for the underlying category, and (b) classification error by the reference network, i.e., the probability that the underlying category does not belong to the categories with the top softmax probability. 
Table \ref{synthesisSingleExp} and \ref{synthesisSingleExp2} display the results for all 10 categories. It can be seen that our model generates 3D shape patterns with higher softmax class probabilities and lower classification errors than other baseline models. In other words, our model can generate more realistic examples than other methods, in the sense that the examples synthesized by our method are very vivid such that higher probabilities a state-of-the-art or well-trained classifier would assign for their underlying categories.

\begin{table}
\centering
\caption{Inception scores of different models learned from 10 3D object categories.}\label{tb:inception}
\begin{tabular}{|l|r|}
\hline 
Model &  Inception score \\ \hline \hline
3D ShapeNets \cite{wu20153d} & 4.126$\pm$0.193    \\ \hline
3D GAN \cite{3dgan}& 8.658$\pm$0.450     \\ \hline
3D VAE \cite{brock2016generative} & 11.015$\pm$0.420\\ \hline
3D WINN \cite{huang20193d} & 8.810$\pm$0.180\\ \hline
Primitive GAN \cite{khan2019unsupervised} & 11.520$\pm$0.330\\ \hline
generative VoxelNet (ours) &  \textbf{11.772$\pm$0.418} \\ \hline
\end{tabular}
\end{table}

\begin{table}[h]
\caption{Softmax class probability}\label{synthesisSingleExp}
\centering
\begin{tabular}{|c|c|c|c|c|}
\hline
category & ours   & \cite{3dgan}   & \cite{brock2016generative}  &  \cite{wu20153d}  \\ \hline \hline

bathtub    & \textbf{0.8348} & 0.7017 & 0.7190 & 0.1644 \\ \hline

bed     & \textbf{0.9202} & 0.7775 & 0.3963 & 0.3239 \\ \hline

chair   & \textbf{0.9920} & 0.9700  & 0.9892 & 0.8482 \\ \hline

desk      & \textbf{0.8203} & 0.7936 & 0.8145 & 0.1068  \\ \hline

dresser      & \textbf{0.7678} & 0.6314 & 0.7010 & 0.2166  \\ \hline

monitor & \textbf{0.9473} & 0.2493   & 0.8559 & 0.2767\\ \hline

night stand    & \textbf{0.7195} & 0.6853 & 0.6592 & 0.4969\\ \hline

sofa   & \textbf{0.9480} &  0.9276  & 0.3017 & 0.4888\\ \hline  

table    & \textbf{0.8910} & 0.8377 & 0.8751 &0.7902 \\ \hline
toilet   & \textbf{0.9701} &  0.8569  & 0.6943 & 0.8832\\ \hline \hline

Avg.   & \textbf{0.8811} &  0.7431   & 0.7006 & 0.4596\\ \hline 

\end{tabular}
\end{table}

\begin{table}[h]
\caption{Classification errors on the synthesized examples}
\label{synthesisSingleExp2}
\centering
\begin{tabular}{|c|c|c|c|c|}
\hline
& ours   & \cite{3dgan}   & \cite{brock2016generative}  &  \cite{wu20153d}  \\ \hline \hline
bathtub    & \textbf{0.1226} & 0.2642 & 0.2642 & 0.8208 \\\hline
bed     & \textbf{0.0583} & 0.1961 & 0.5981 & 0.6583 \\ \hline
chair   & \textbf{0.0050} & 0.0125  & 0.0100 & 0.1475 \\ \hline
desk      & \textbf{0.0165} & 0.1900 & 0.1700 &  0.8900 \\ \hline
dresser      & \textbf{0.0165} & 0.3000 & 0.2400 &  0.7950 \\ \hline
monitor & \textbf{0.0452} &  0.7484  & 0.1312 & 0.7204\\ \hline
night stand    & \textbf{0.2700} & 0.3000  & 0.3200 & 0.4800\\ \hline
sofa   & \textbf{0.0397} &  0.0471   & 0.6868 & 0.4912\\  
\hline
table    & \textbf{0.0800} & 0.1333 & 0.1067   & 0.1833 \\ \hline
toilet   & \textbf{0.0174} &  0.1233   & 0.2994 & 0.1047\\ \hline \hline
Avg.   & \textbf{0.0968} &  0.2315   & 0.2826 &  0.5291\\ \hline
\end{tabular}
\end{table}

\subsection{3D object recovery}
\label{Exp:objectRecovery}

We then test the conditional generative VoxelNet on the 3D object recovery task.  For each testing 3D object, we randomly corrupt some voxels of the 3D object. We then seek to recover the corrupted voxels by sampling from the conditional distribution $p(Y_M|Y_{\tilde{M}}; \theta)$ according to the learned model $p(Y; \theta)$, where $M$ and $\tilde{M}$ denote the corrupted and uncorrupted voxels, and $Y_M$ and $Y_{\tilde{M}}$ are the corrupted part and the uncorrupted part of the 3D object $Y$ respectively. The sampling of $p(Y_M|Y_{\tilde{M}}; \theta)$ is again accomplished by the Langevin dynamics, which is the same as the Langevin dynamics that samples from the full distribution $p(Y; \theta)$, except that we fix the uncorrupted part $Y_{\tilde{M}}$ and only update the corrupted part $Y_M$ throughout the Langevin dynamics. In the learning stage, we learn the model from the fully observed training 3D objects. To specialize the learned model to this recovery task, we learn the conditional distribution $p(Y_M|Y_{\tilde{M}}; \theta)$ directly. That is, in the learning stage, we also randomly corrupt each fully observed training 3D object $Y$, and run Langevin dynamics by fixing $Y_{\tilde{M}}$ to obtain the synthesized 3D object. The parameters $\theta$ are then updated by gradient ascent according to equation (\ref{eq:lD2}). Also see Algorithm \ref{code:3D_recovery} for a description. We try two different network architectures in this experiment. The first one is a 3-layer network, which is the same as the one used in Section \ref{Exp:objectSynthesis} for synthesis, and the other one is a 2-layer network consisting of 100 $16 \times 16 \times 16$ filters with sub-sampling of 3 in its first layer, followed by a fully connected layer with a single filer. The number of Langevin dynamics steps for recovery in each iteration is set to be $K=90$ and the step size is $\Delta \tau =0.0049$. The number of learning iterations is $1,000$. The size of the mini-batch is 50. The 3D training data are of size $32 \times 32 \times 32$ voxels. 

 After learning the model, we recover the corrupted voxels in each testing data $Y$ by sampling from $p(Y_M|Y_{\tilde{M}}; \theta)$ by running 90 Langevin dynamics steps. In the training stage, we randomly corrupt $70\%$ of each training 3D shape. In the testing stage, we experiment with the same percentage of corruption. We compare our methods with 3D-GAN and 3D ShapeNets. 

The original 3D-GAN does not naturally support shape recovery. To use the 3D-GAN algorithm to recover corrupted voxels, we firstly train the 3D-GAN model from the complete training data, and then use the learned 3D generator to predict the corrupted voxels from the observed ones by adding an extra inference step. Specifically, following \cite{HanLu2016}, we derive the posterior distribution $p(Z|Y;\alpha)$ from the generator model, and perform inference by sampling from the posterior distribution via Langevin dynamic to infer the vector of latent factors for each corrupted example. The unobserved voxels can then be recovered from their inferred factors by a direct top-down generation. Only those corrupted voxels get updated by the generated voxels. We adopt the same 3D-GAN architecture as in \cite{3dgan}.
 
\begin{figure}[h]
	\centering
	\rotatebox[origin=l]{90}{\hspace{0.8mm}\textbf{{\scriptsize original}}}
	\includegraphics[height=.139\linewidth]{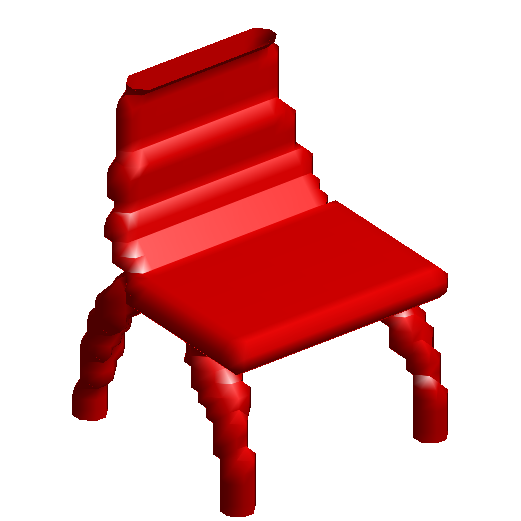} \hspace{-1.8mm}
	\includegraphics[height=.139\linewidth]{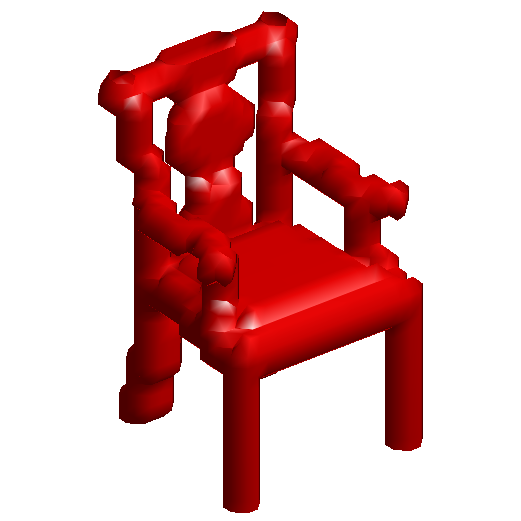} \hspace{-1.8mm}
    \includegraphics[height=.139\linewidth]{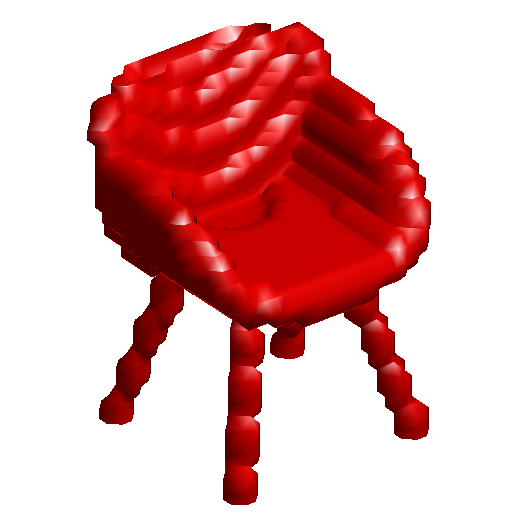} \hspace{-1.9mm}
    \includegraphics[height=.139\linewidth]{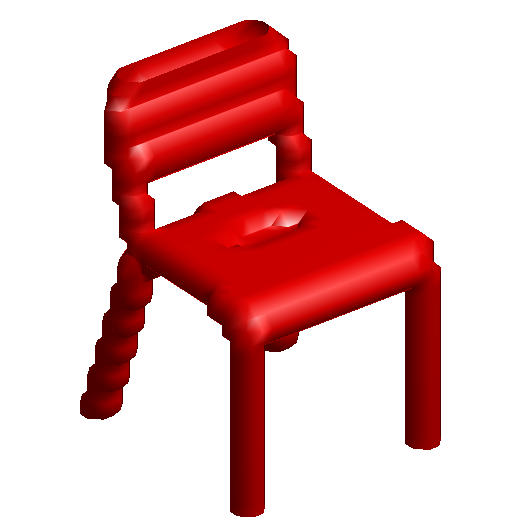} \hspace{-1.9mm}
    \includegraphics[height=.139\linewidth]{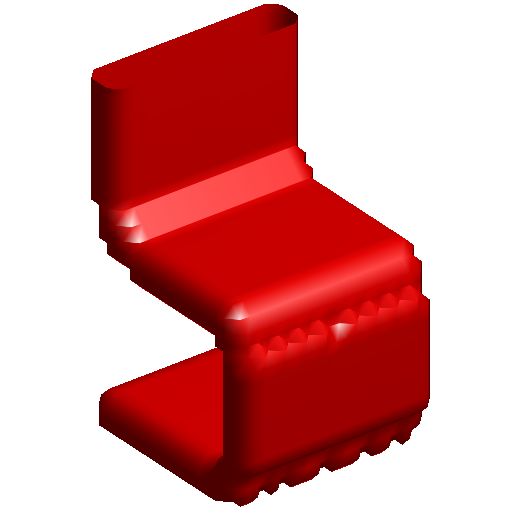} \hspace{-1.9mm}
    \includegraphics[height=.139\linewidth]{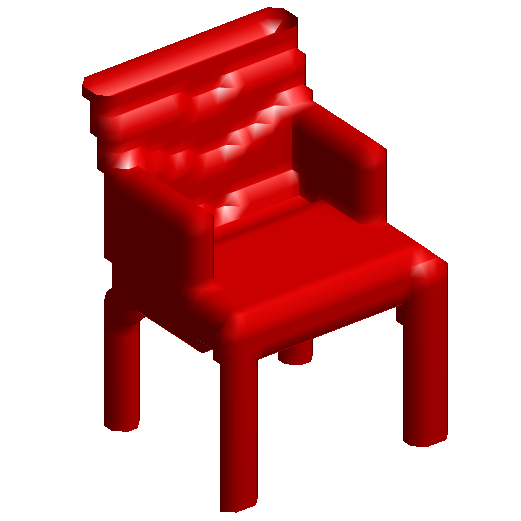} \hspace{-1.9mm}    
    \includegraphics[height=.139\linewidth]{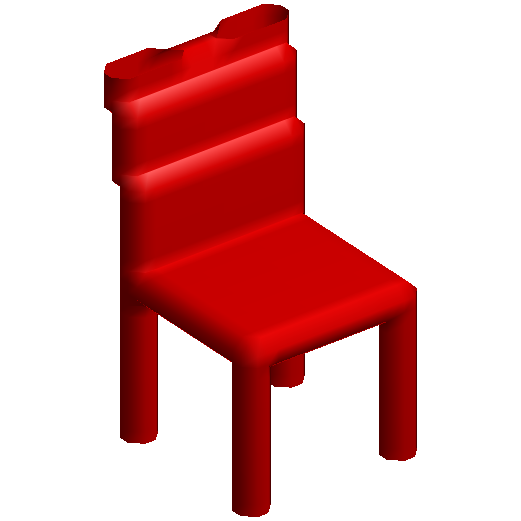}   
     \\
    \rotatebox[origin=l]{90}{\hspace{1mm}\textbf{{\scriptsize corrupted}}}
	\includegraphics[height=.139\linewidth]{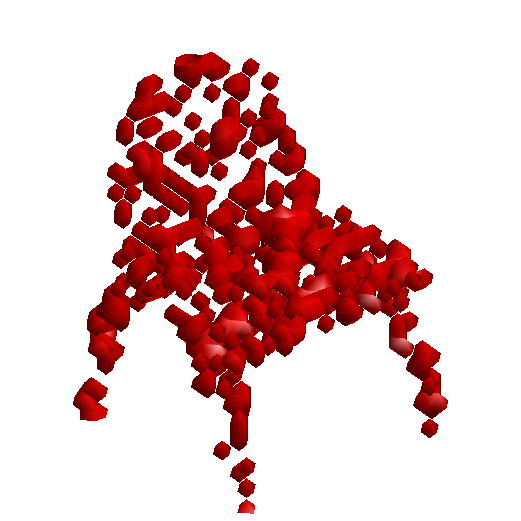} \hspace{-1.8mm}
	\includegraphics[height=.139\linewidth]{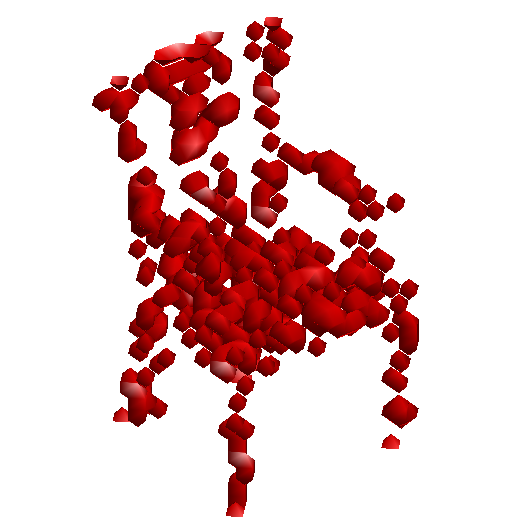} \hspace{-1.8mm}
    \includegraphics[height=.139\linewidth]{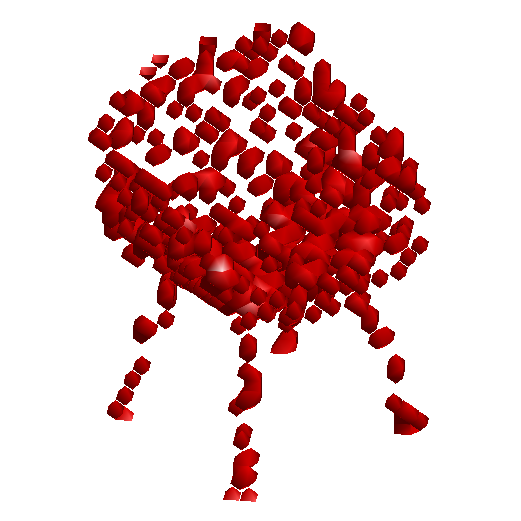} \hspace{-1.9mm}
    \includegraphics[height=.139\linewidth]{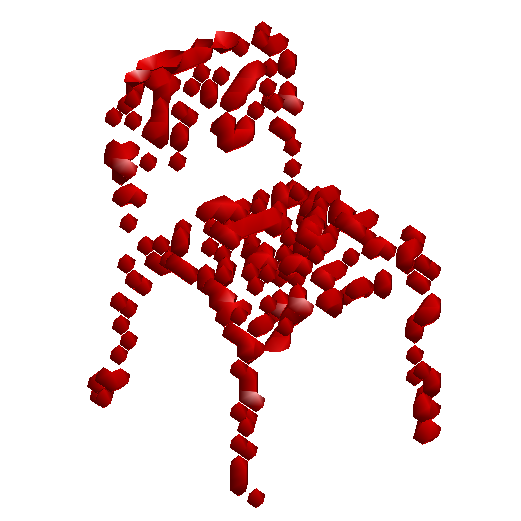} \hspace{-1.9mm}
    \includegraphics[height=.139\linewidth]{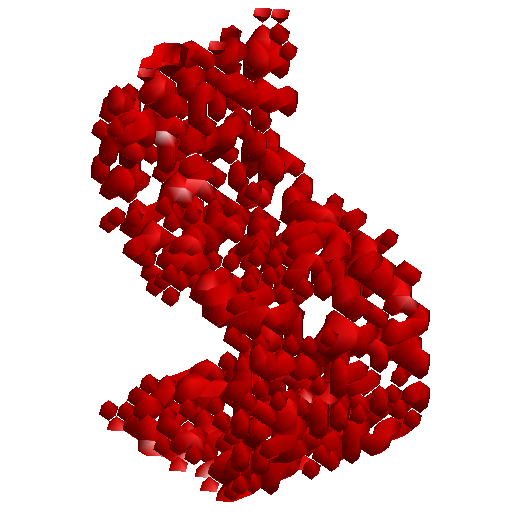} \hspace{-1.9mm}
    \includegraphics[height=.139\linewidth]{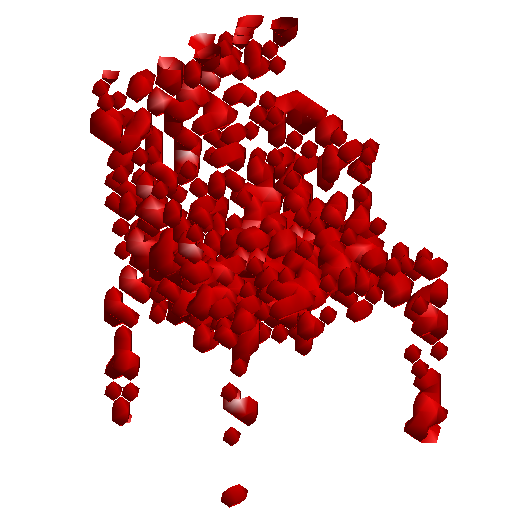} \hspace{-1.9mm}   
    \includegraphics[height=.139\linewidth]{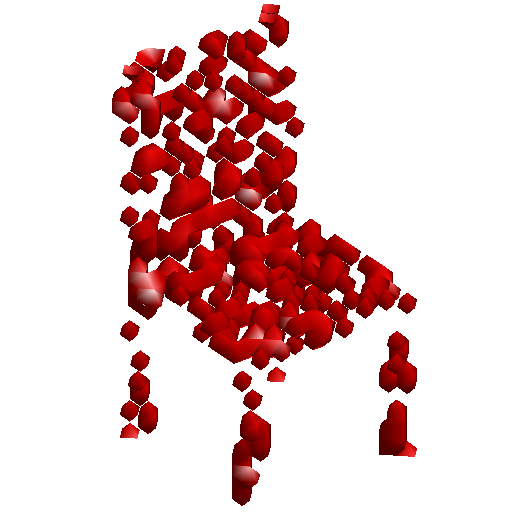} 
     \\
     \rotatebox[origin=l]{90}{\hspace{1mm}\textbf{{\scriptsize recovered}}}
     \includegraphics[height=.139\linewidth]{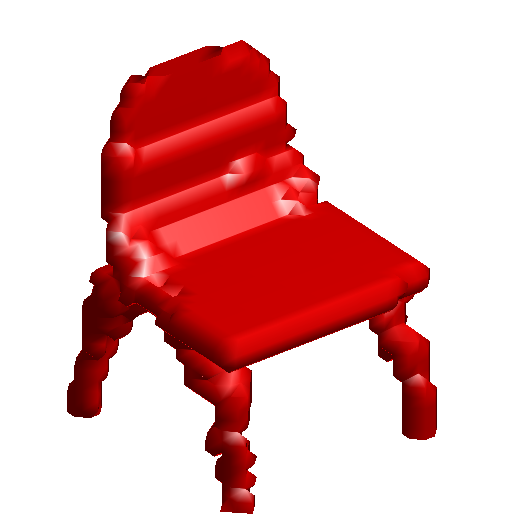} \hspace{-1.8mm}
     \includegraphics[height=.139\linewidth]{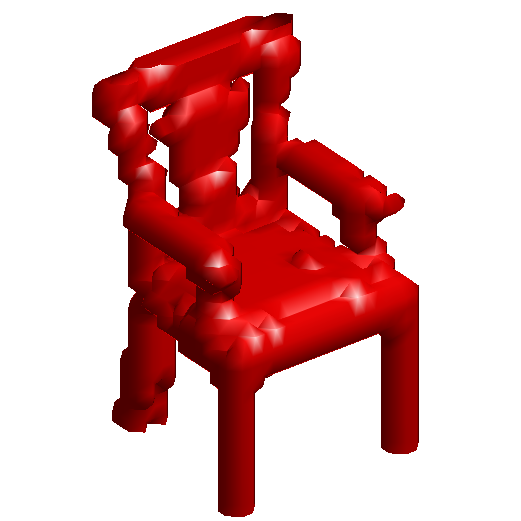} \hspace{-1.8mm}
    \includegraphics[height=.139\linewidth]{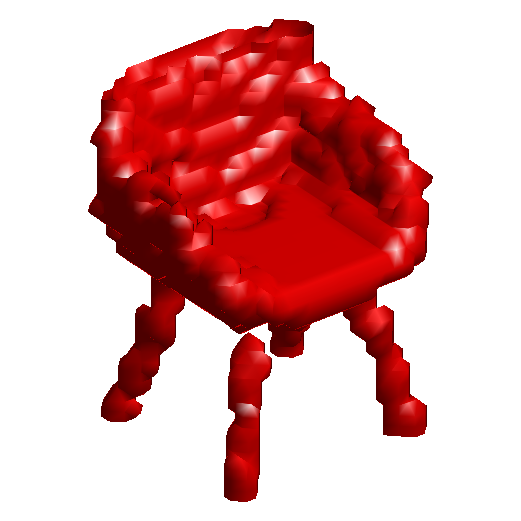} \hspace{-1.9mm}
    \includegraphics[height=.139\linewidth]{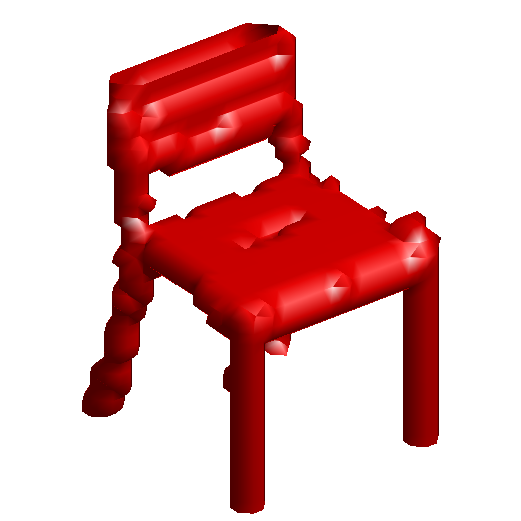} \hspace{-1.9mm}
    \includegraphics[height=.139\linewidth]{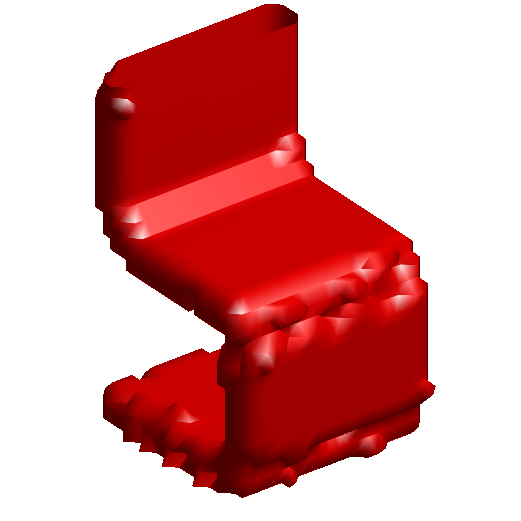} \hspace{-1.9mm}
    \includegraphics[height=.139\linewidth]{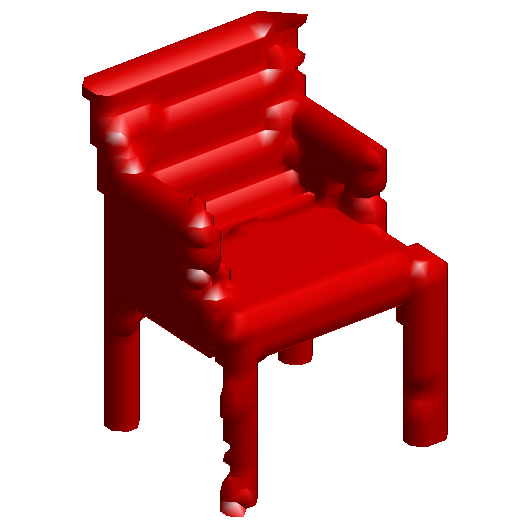} \hspace{-1.9mm}
    \includegraphics[height=.139\linewidth]{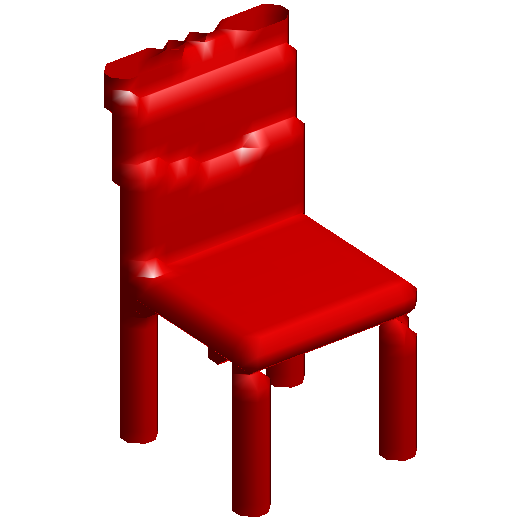}\\
    (a) chair\\
    \vspace{2mm}
    \rotatebox[origin=l]{90}{\hspace{1mm}\textbf{{\scriptsize original}}}
    \includegraphics[height=.139\linewidth]{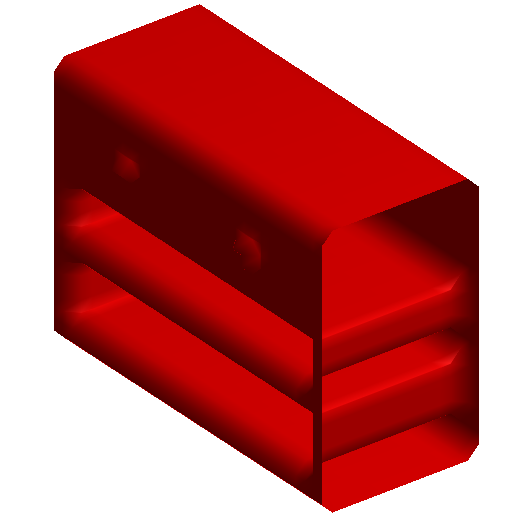} \hspace{-1.8mm}
    \includegraphics[height=.139\linewidth]{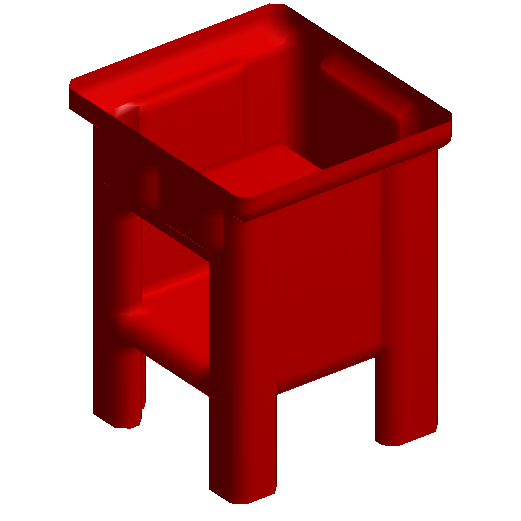} \hspace{-1.8mm}
    \includegraphics[height=.139\linewidth]{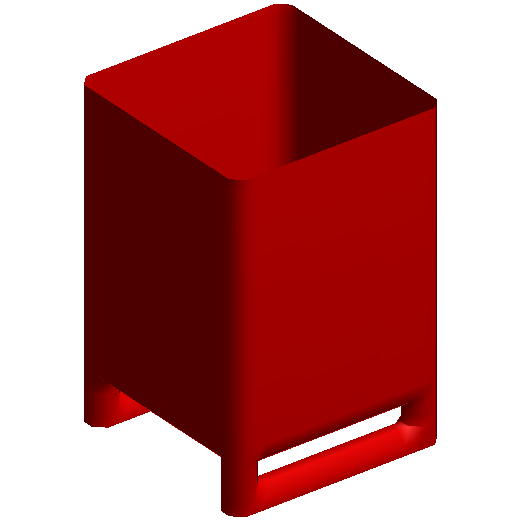} \hspace{-1.8mm}
    \includegraphics[height=.139\linewidth]{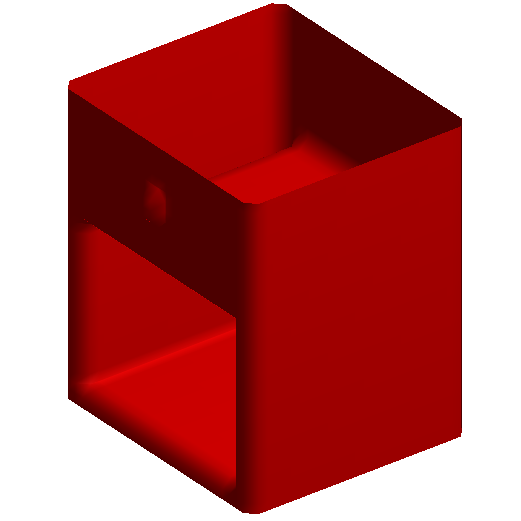} \hspace{-1.9mm}
    \includegraphics[height=.139\linewidth]{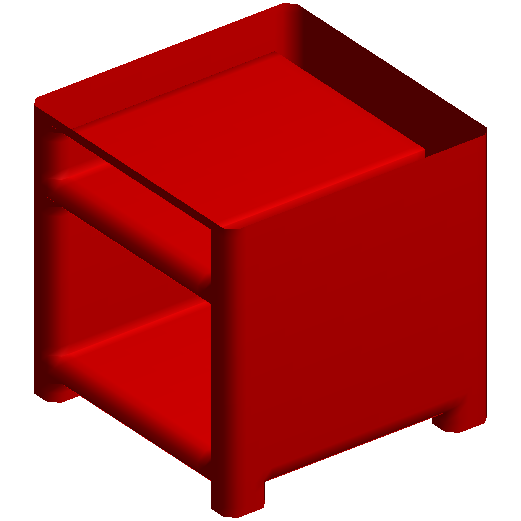} \hspace{-1.9mm}
    \includegraphics[height=.139\linewidth]{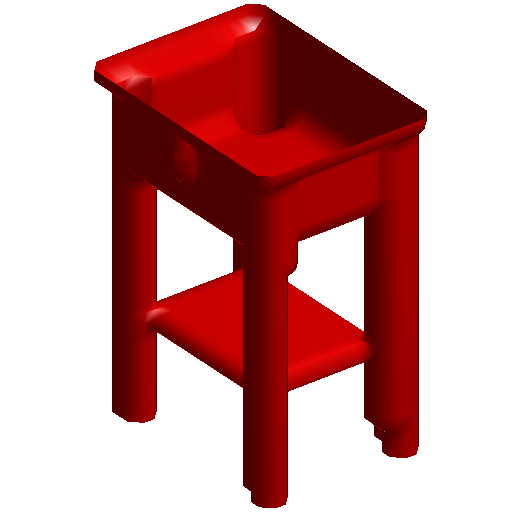} \hspace{-1.9mm}
    \includegraphics[height=.139\linewidth]{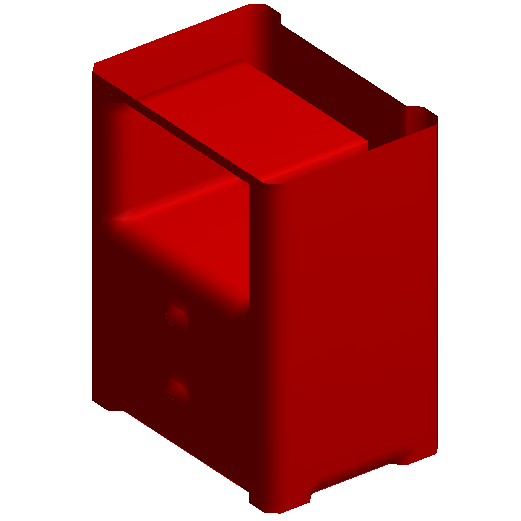} \hspace{-1.9mm}\\
    \rotatebox[origin=l]{90}{\hspace{1mm}\textbf{{\scriptsize corrupted}}}
    \includegraphics[height=.139\linewidth]{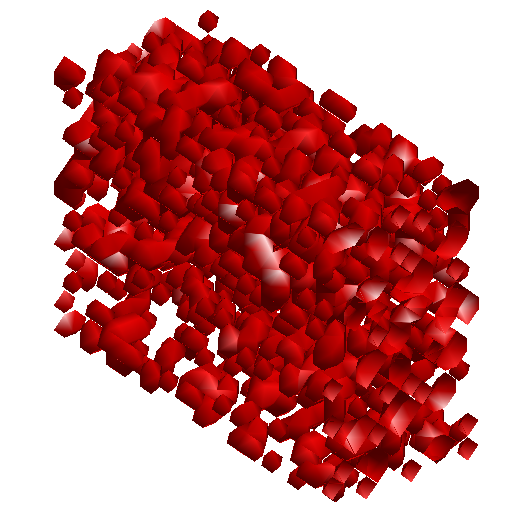} \hspace{-1.8mm}
    \includegraphics[height=.139\linewidth]{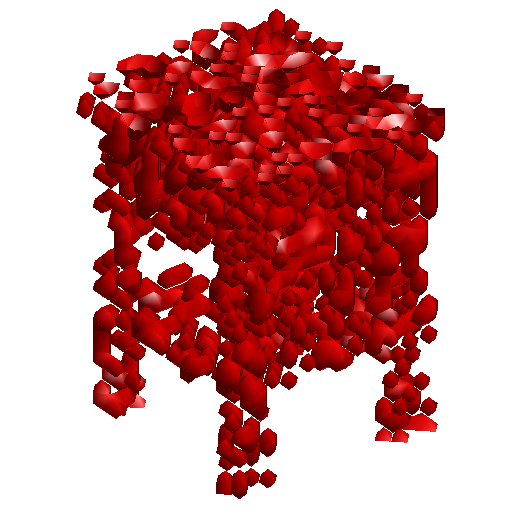} \hspace{-1.8mm}
    \includegraphics[height=.139\linewidth]{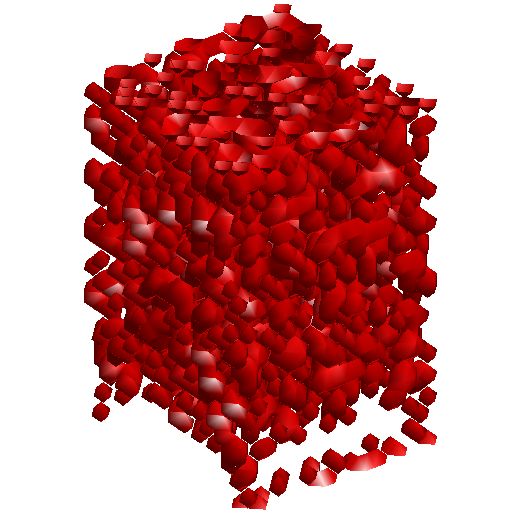} \hspace{-1.8mm}
    \includegraphics[height=.139\linewidth]{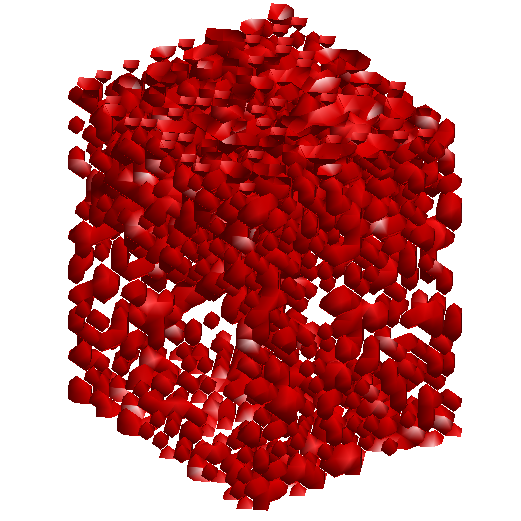} \hspace{-1.9mm}
    \includegraphics[height=.139\linewidth]{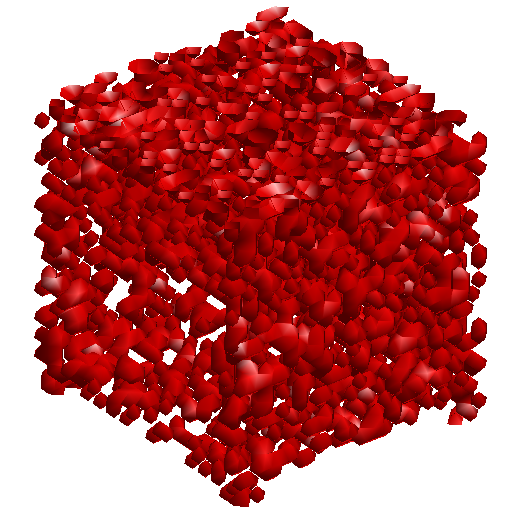} \hspace{-1.9mm}
    \includegraphics[height=.139\linewidth]{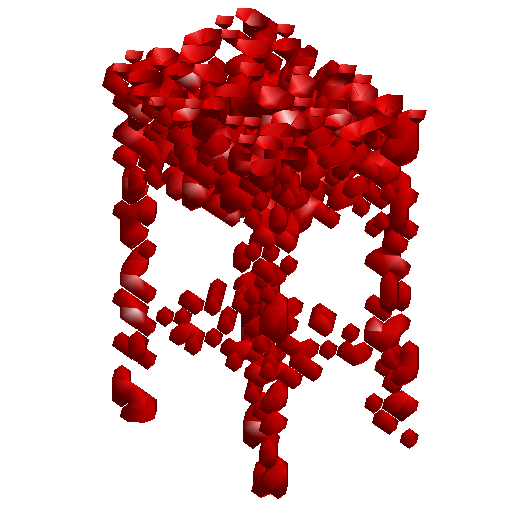} \hspace{-1.9mm}
    \includegraphics[height=.139\linewidth]{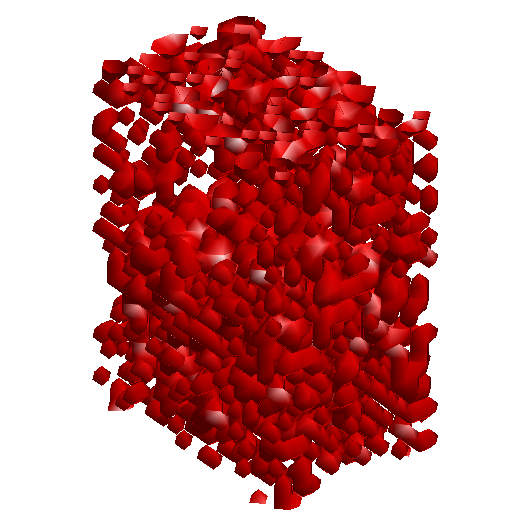} \hspace{-1.9mm}\\
    \rotatebox[origin=l]{90}{\hspace{1mm}\textbf{{\scriptsize recovered}}}
    \includegraphics[height=.139\linewidth]{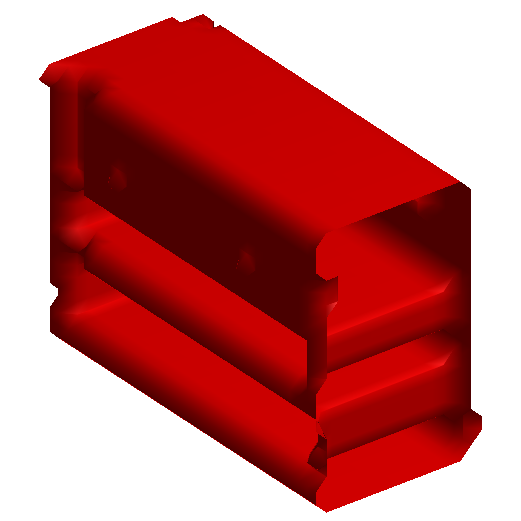} \hspace{-1.8mm}
    \includegraphics[height=.139\linewidth]{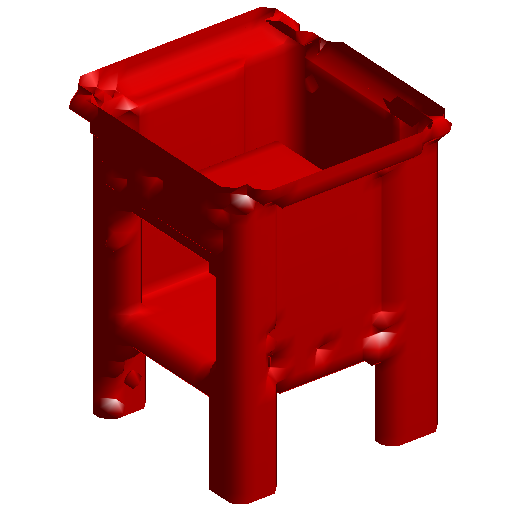} \hspace{-1.8mm}
    \includegraphics[height=.139\linewidth]{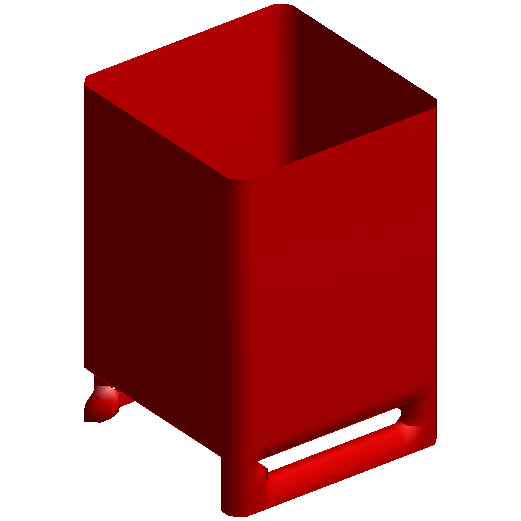} \hspace{-1.8mm}
    \includegraphics[height=.139\linewidth]{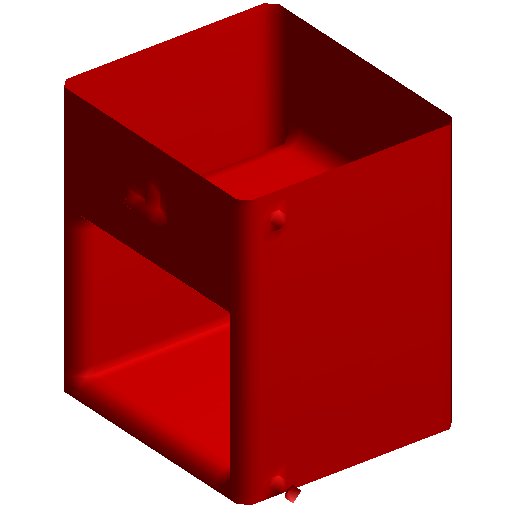} \hspace{-1.8mm}
    \includegraphics[height=.139\linewidth]{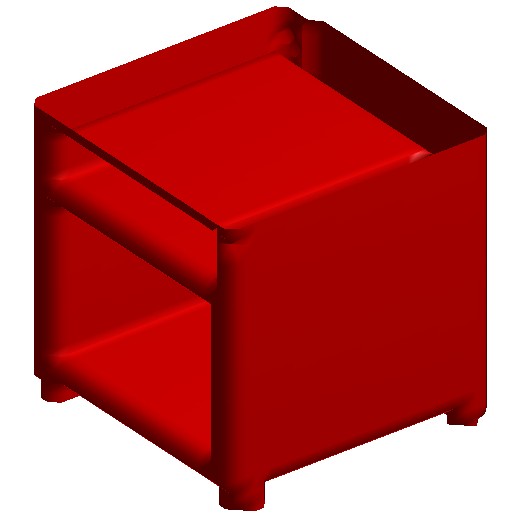} \hspace{-1.8mm}
    \includegraphics[height=.139\linewidth]{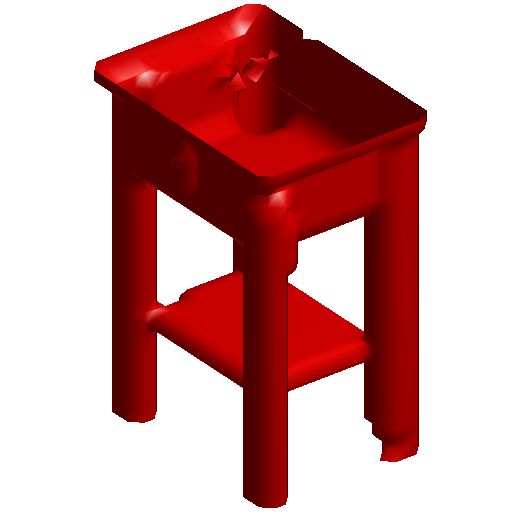} \hspace{-1.8mm}
    \includegraphics[height=.139\linewidth]{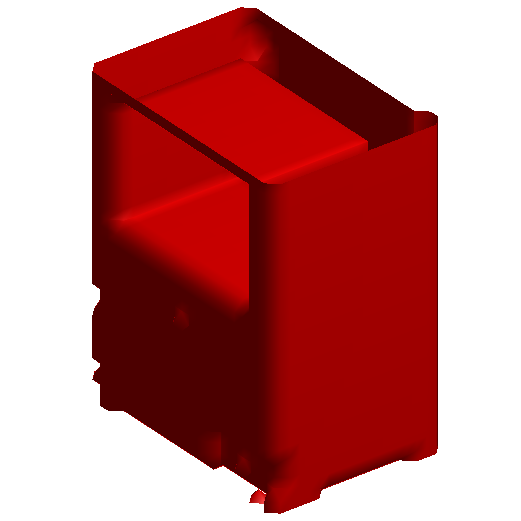} \hspace{-1.8mm}\\
	(b) night stand\\	
    \vspace{2mm}
	\rotatebox[origin=l]{90}{\hspace{1mm}\textbf{{\scriptsize original}}}
    \includegraphics[height=.139\linewidth]{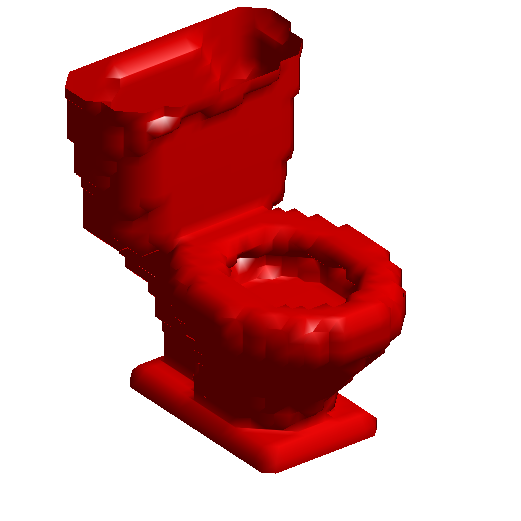} \hspace{-1.8mm}
    \includegraphics[height=.139\linewidth]{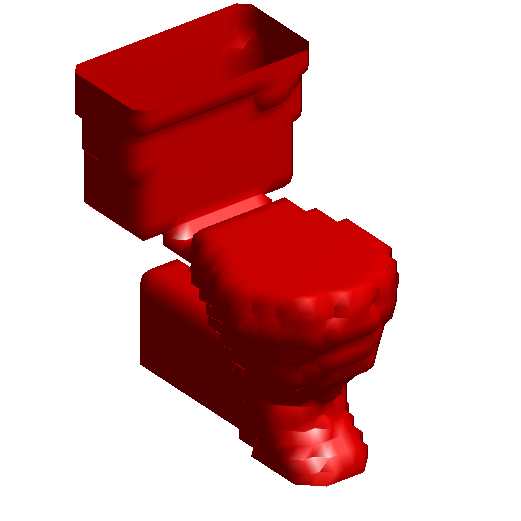} \hspace{-1.8mm}
    \includegraphics[height=.139\linewidth]{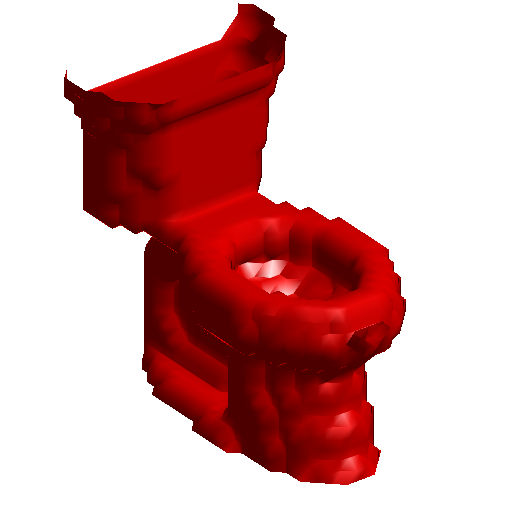} \hspace{-1.8mm}
    \includegraphics[height=.139\linewidth]{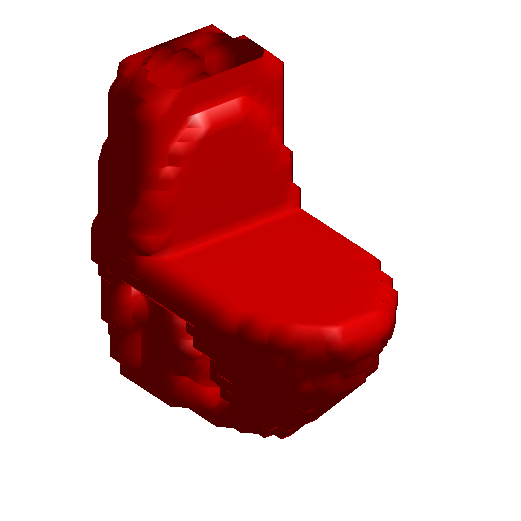} \hspace{-1.9mm}
    \includegraphics[height=.139\linewidth]{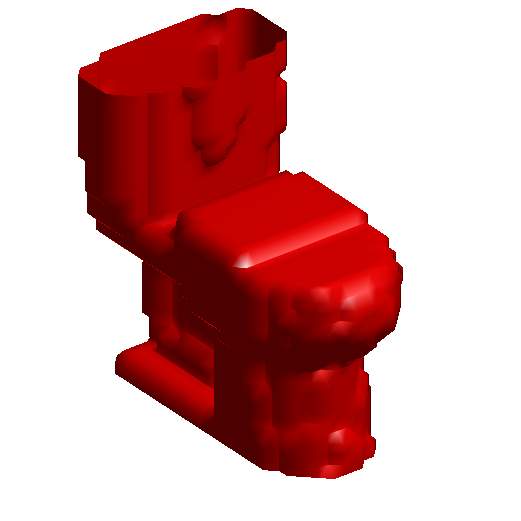} \hspace{-1.9mm}
    \includegraphics[height=.139\linewidth]{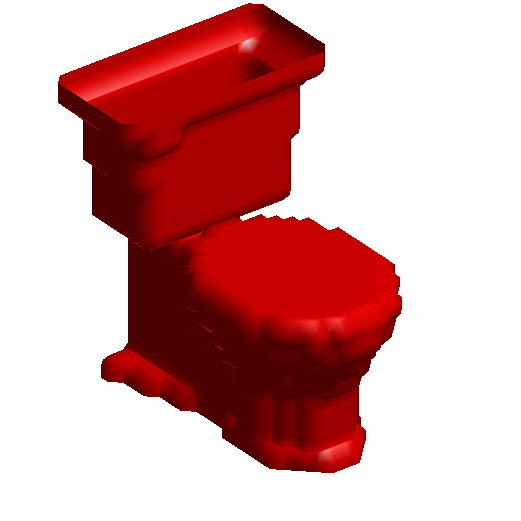} \hspace{-1.9mm}
     \includegraphics[height=.138\linewidth]{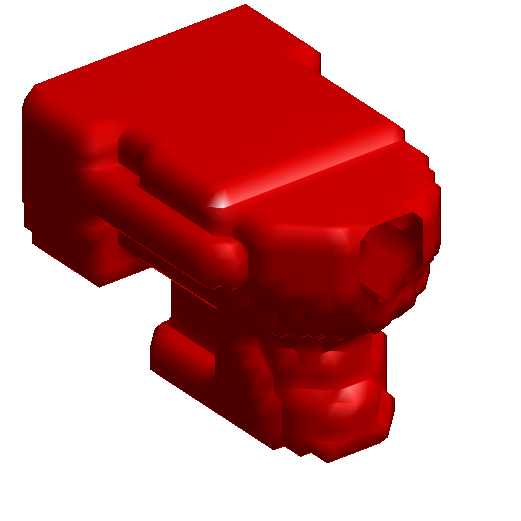} \hspace{-1.9mm}\\
     \rotatebox[origin=l]{90}{\hspace{1mm}\textbf{{\scriptsize corrupted}}}
    \includegraphics[height=.139\linewidth]{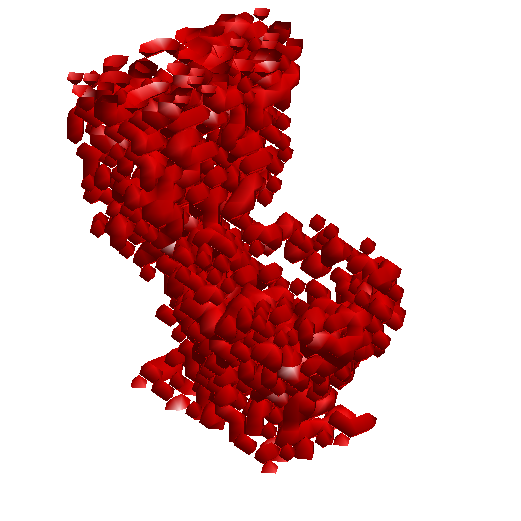} \hspace{-1.8mm}
    \includegraphics[height=.139\linewidth]{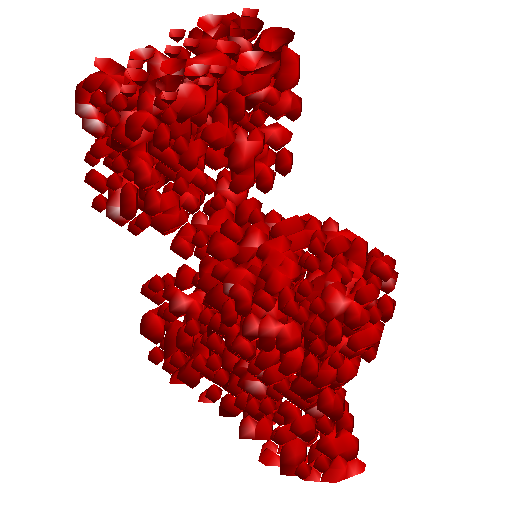} \hspace{-1.8mm}
    \includegraphics[height=.139\linewidth]{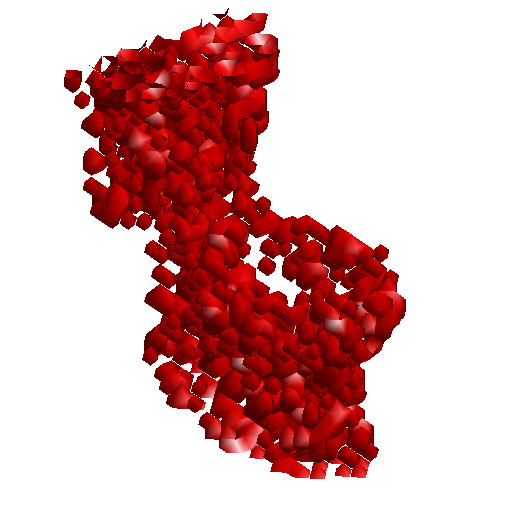} \hspace{-1.8mm}
    \includegraphics[height=.139\linewidth]{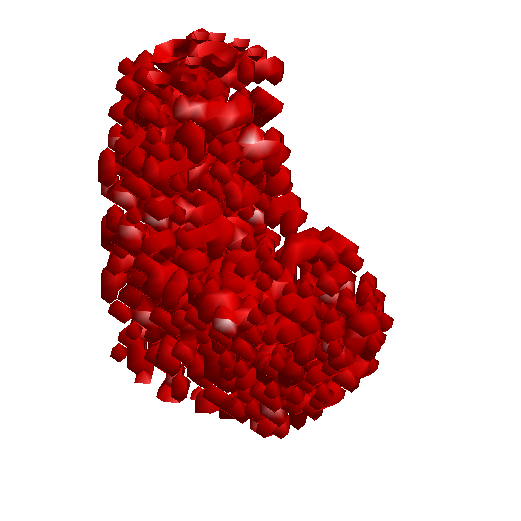} \hspace{-1.9mm}
    \includegraphics[height=.139\linewidth]{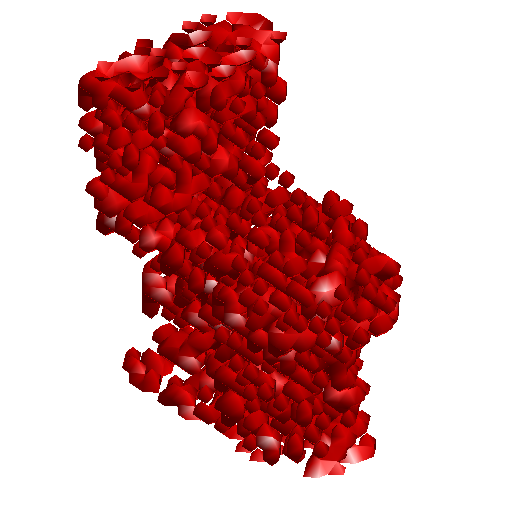} \hspace{-1.9mm}
    \includegraphics[height=.139\linewidth]{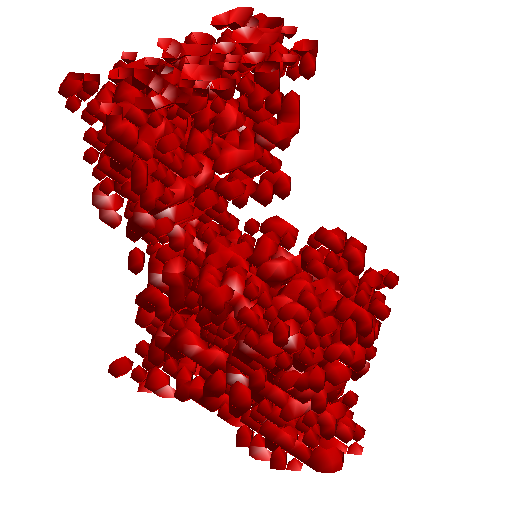} \hspace{-1.9mm}
    \includegraphics[height=.139\linewidth]{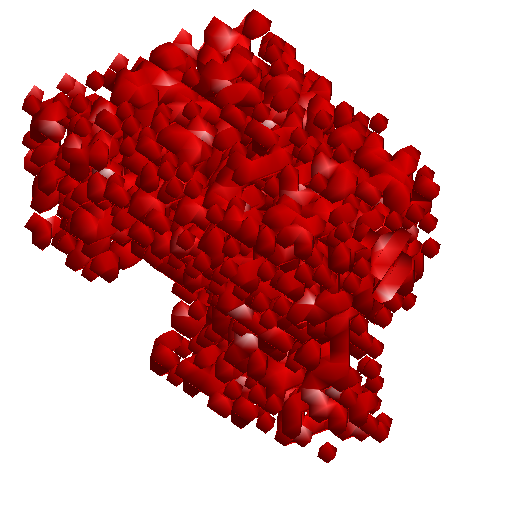} \hspace{-1.9mm}\\
    \rotatebox[origin=l]{90}{\hspace{1mm}\textbf{{\scriptsize recovered}}}
    \includegraphics[height=.139\linewidth]{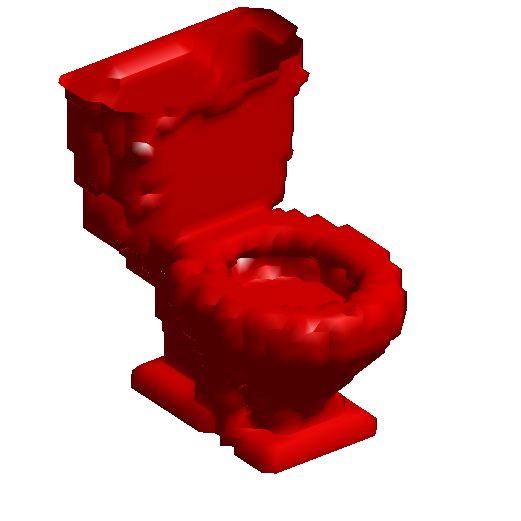} \hspace{-1.8mm}
    \includegraphics[height=.139\linewidth]{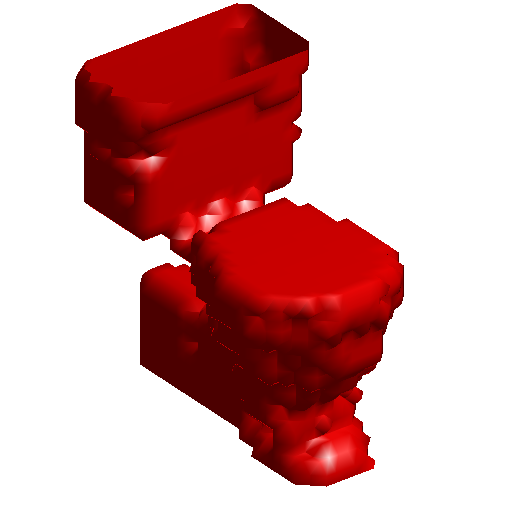} \hspace{-1.8mm}
    \includegraphics[height=.139\linewidth]{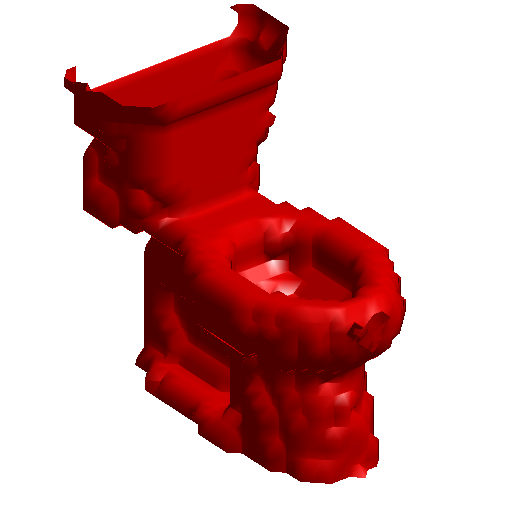} \hspace{-1.8mm}
    \includegraphics[height=.139\linewidth]{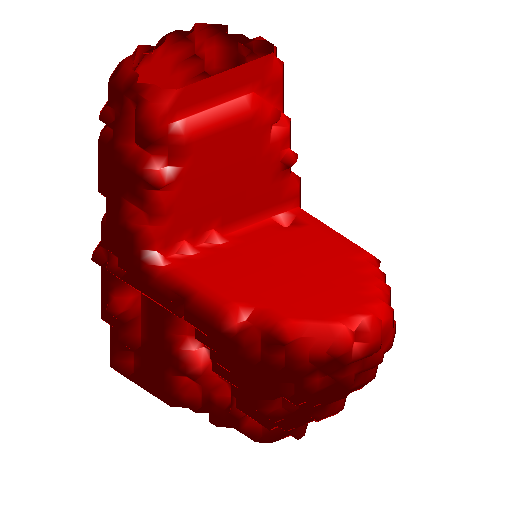} \hspace{-1.8mm}
    \includegraphics[height=.139\linewidth]{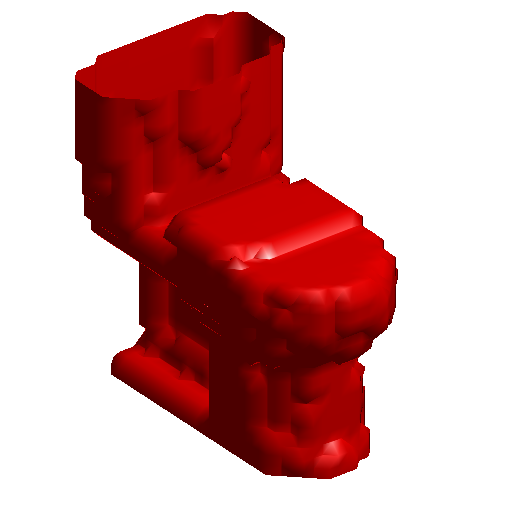} \hspace{-1.8mm}
    \includegraphics[height=.139\linewidth]{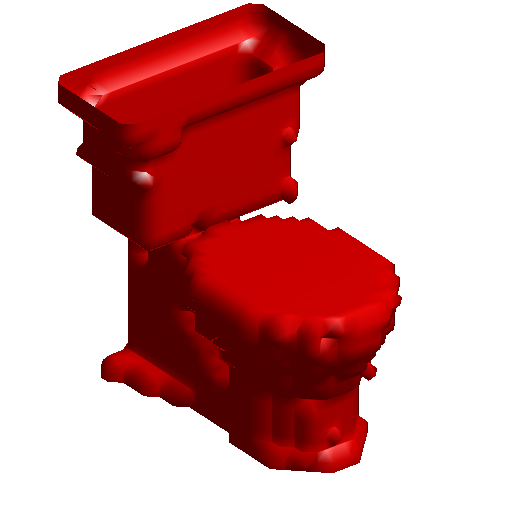} \hspace{-1.8mm}
     \includegraphics[height=.139\linewidth]{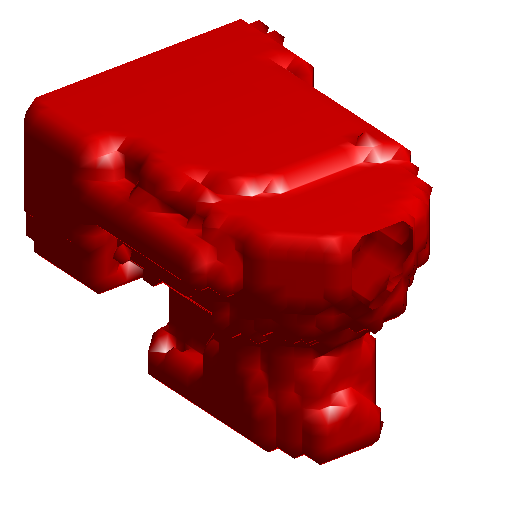} \hspace{-1.8mm}\\	
	(c) toilet	\\
	\vspace{2mm}
	\rotatebox[origin=l]{90}{\hspace{1mm}\textbf{{\scriptsize original}}} \hspace{0.8mm}
    \includegraphics[height=.136\linewidth]{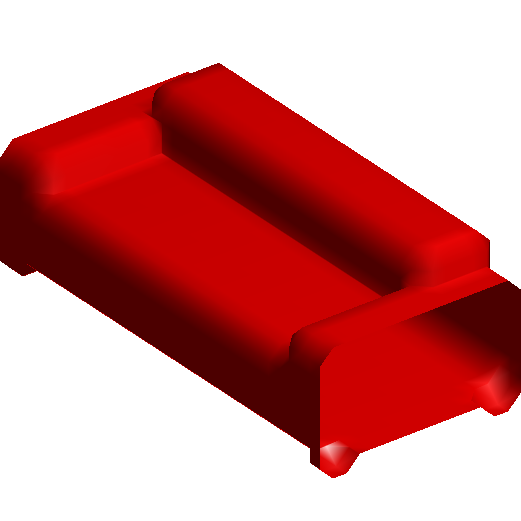} \hspace{-1.8mm}
    \includegraphics[height=.136\linewidth]{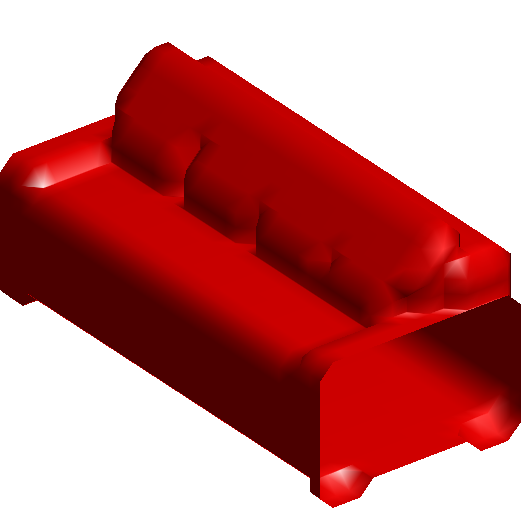} \hspace{-1.8mm}
     \includegraphics[height=.136\linewidth]{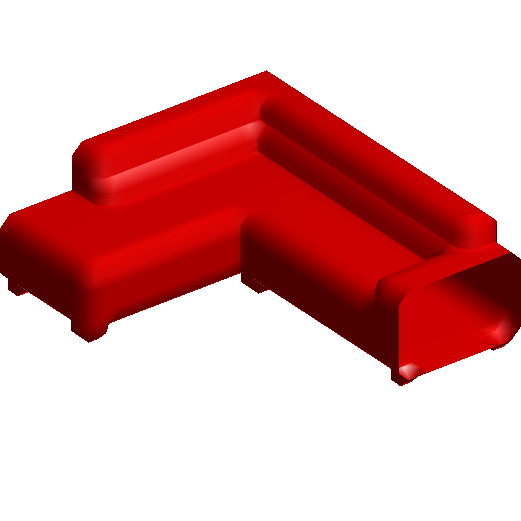} \hspace{-1.8mm}
     \includegraphics[height=.136\linewidth]{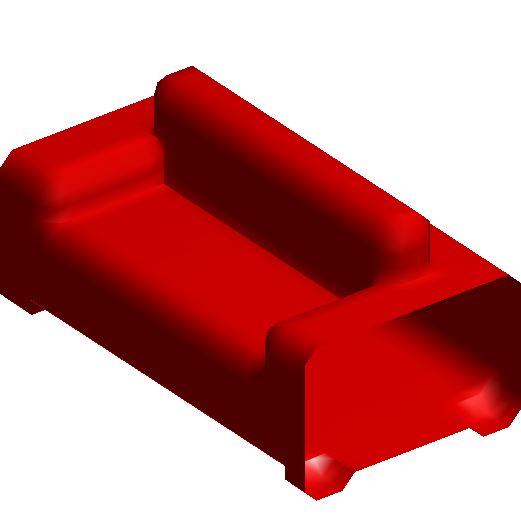} \hspace{-1.8mm}
     \includegraphics[height=.136\linewidth]{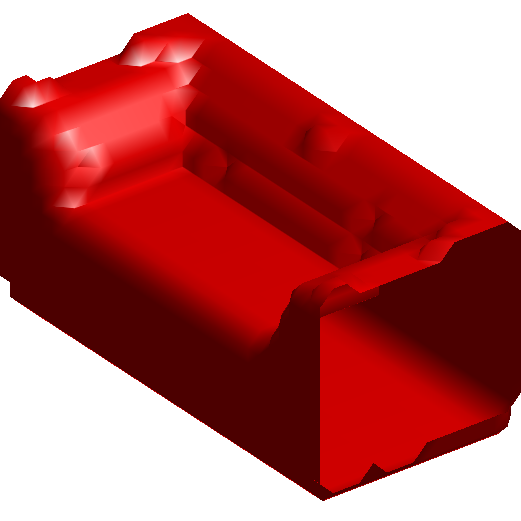} \hspace{-1.8mm}
      \includegraphics[height=.136\linewidth]{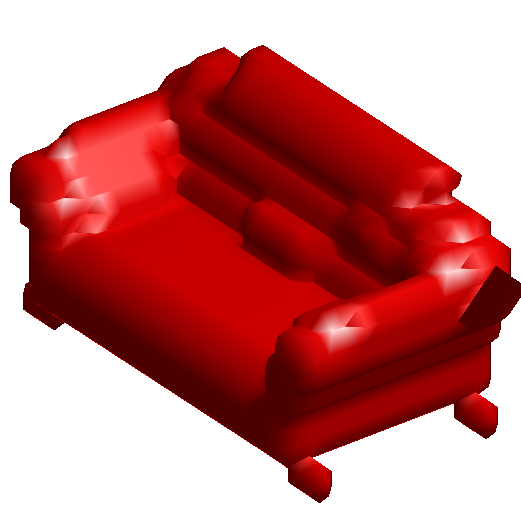} \hspace{-1.8mm}
      \includegraphics[height=.136\linewidth]{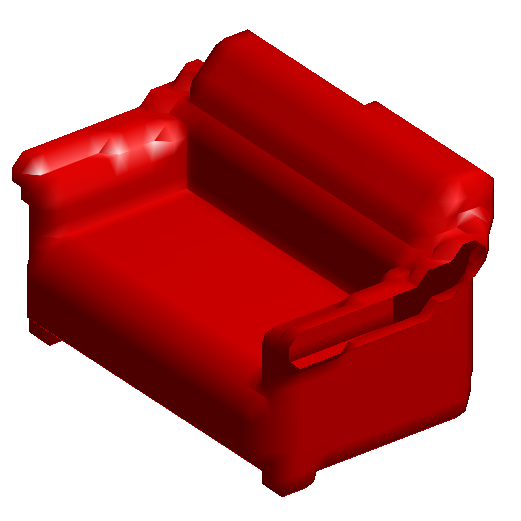} \hspace{-1.8mm}\\
       \rotatebox[origin=l]{90}{\hspace{1mm}\textbf{{\scriptsize corrupted}}} \hspace{0.8mm}
     \includegraphics[height=.136\linewidth]{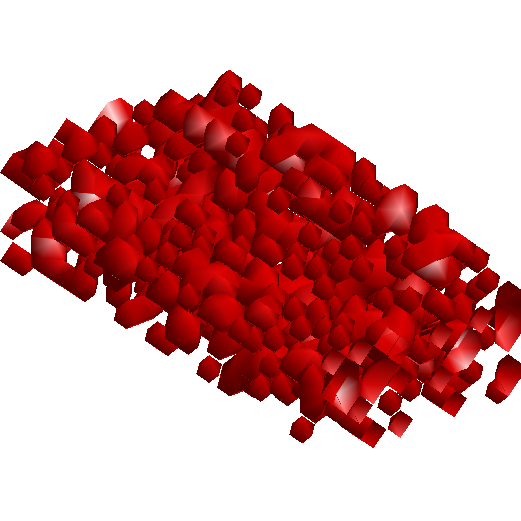} \hspace{-1.8mm}
     \includegraphics[height=.136\linewidth]{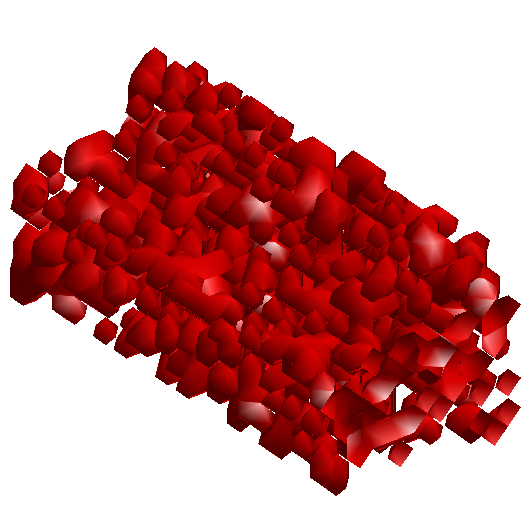} \hspace{-1.8mm}
     \includegraphics[height=.136\linewidth]{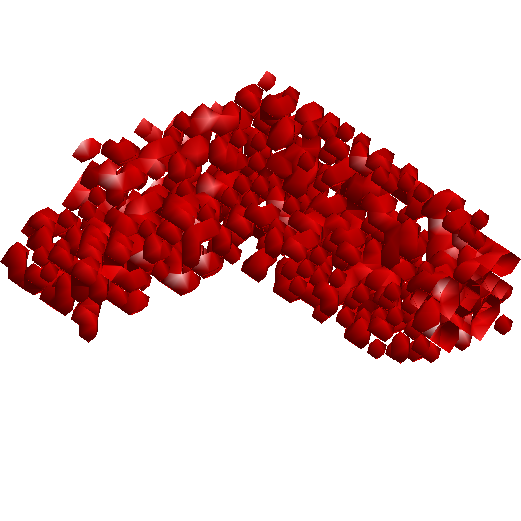} \hspace{-1.8mm}
     \includegraphics[height=.136\linewidth]{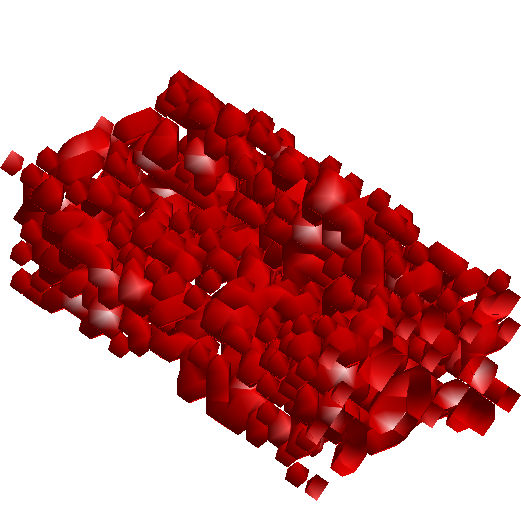} \hspace{-1.8mm}
     \includegraphics[height=.136\linewidth]{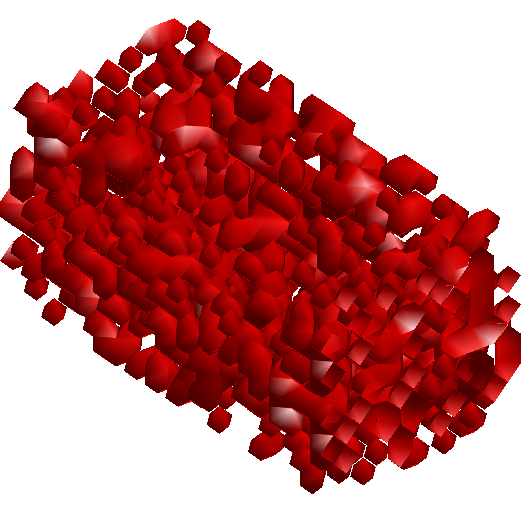} \hspace{-1.8mm}
     \includegraphics[height=.136\linewidth]{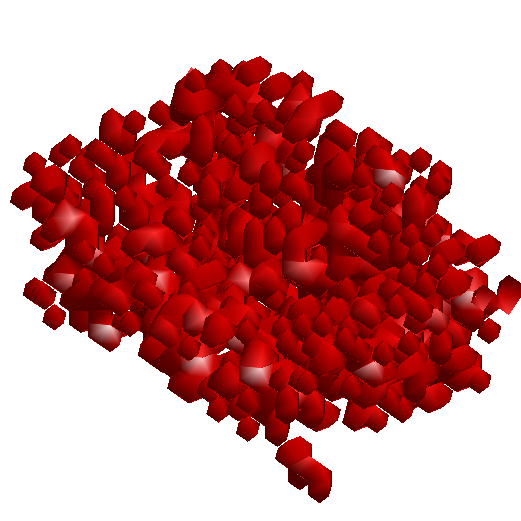} \hspace{-1.8mm}
     \includegraphics[height=.136\linewidth]{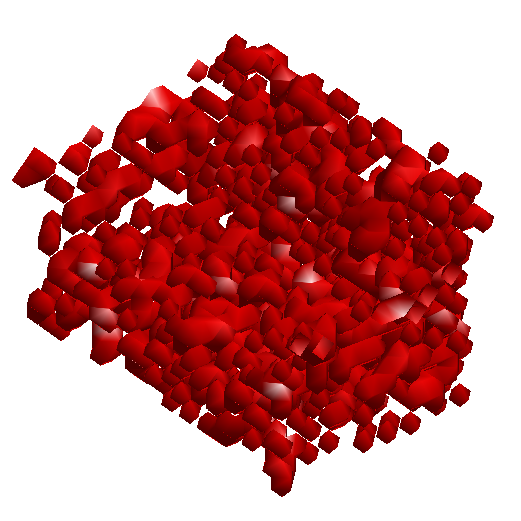} \hspace{-1.8mm}\\
      \rotatebox[origin=l]{90}{\hspace{1mm}\textbf{{\scriptsize recovered}}} \hspace{0.8mm}
     \includegraphics[height=.136\linewidth]{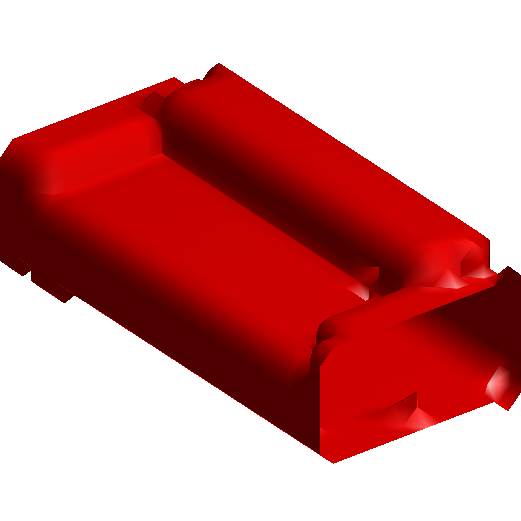} \hspace{-1.8mm}
     \includegraphics[height=.136\linewidth]{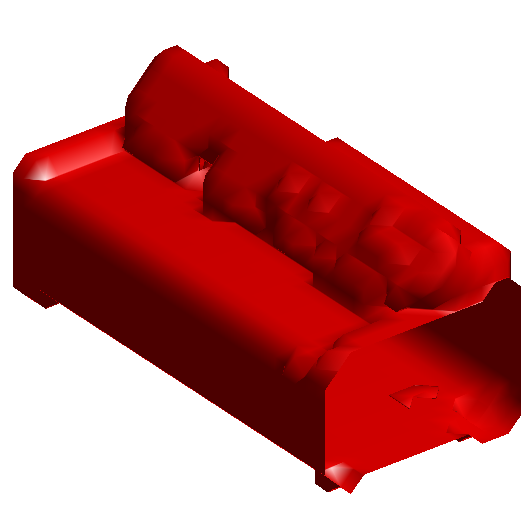} \hspace{-1.8mm}
     \includegraphics[height=.136\linewidth]{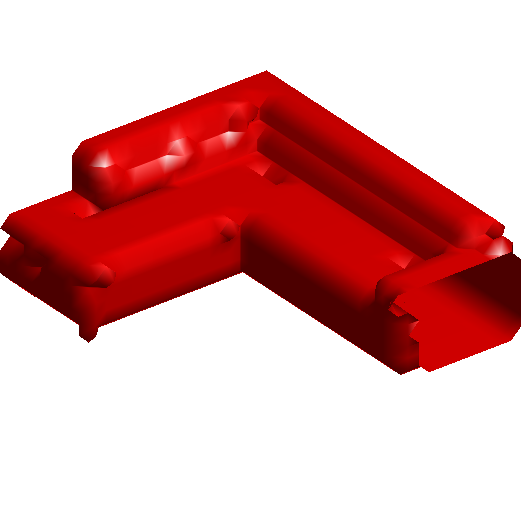} \hspace{-1.8mm}
     \includegraphics[height=.136\linewidth]{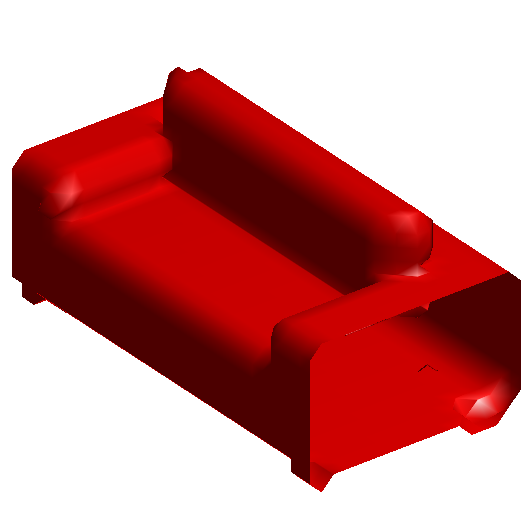} \hspace{-1.8mm}
     \includegraphics[height=.136\linewidth]{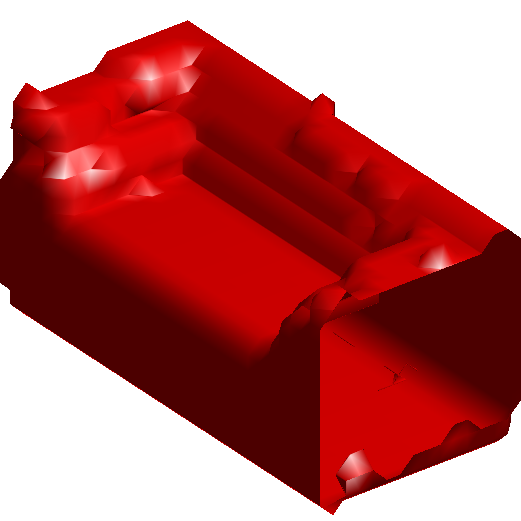} \hspace{-1.8mm}
     \includegraphics[height=.136\linewidth]{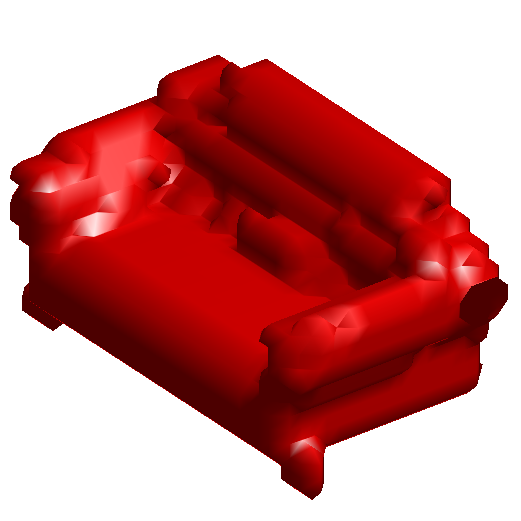} \hspace{-1.8mm}
     \includegraphics[height=.136\linewidth]{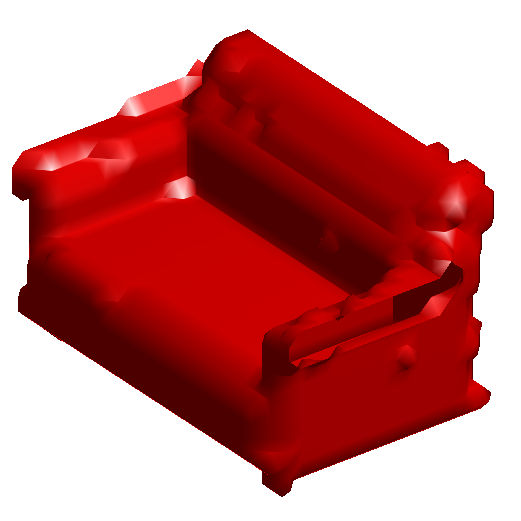} \hspace{-1.8mm}\\
     	(d) sofa	\\
	\caption{3D object recovery by sampling from the conditional generative VoxelNet models. In each category, the first row displays the original 3D objects, the second row shows the corrupted 3D objects, and the third row displays the recovered 3D objects by running Langevin dynamics starting from the corrupted objects. (a) chair (b) night stand (c) toilet (d) sofa.}	
	\label{fig:recovery}
\end{figure}

 We measure the recovery error by the average of per-voxel differences between the original testing data and the corresponding recovered data on the corrupted voxels. Table \ref{exp:recovery} displays the numerical comparison results for the 10 categories, where we use our-2 and our-3 to represent the 2-layer and the 3-layer models respectively. Our methods outperform the other baseline methods in terms of recovery error. Figure \ref{fig:recovery} displays some examples of 3D object recovery. For each experiment, the first row displays the original 3D objects, the second row displays the corrupted 3D objects, and the third row displays the recovered 3D objects that are sampled from the learned conditional distributions given the corrupted 3D objects as inputs.

 \begin{table}[h]
\caption{Recovery errors in occlusion experiments}\label{exp:recovery}
\centering
\begin{tabular}{|c|c|c|c|c|}
\hline
 category          & ours-3   & ours-2 &\cite{3dgan}      & \cite{wu20153d}  \\ \hline \hline
bathtub       & \textbf{0.0152} & 0.0316 & 0.0266  & 0.0621  \\ \hline
bed   & \textbf{0.0068} & 0.0071 & 0.0240  & 0.0617  \\ \hline
chair      & \textbf{0.0118} & 0.0126 & 0.0238 & 0.0444 \\ \hline
desk    & \textbf{0.0122} & 0.019& 0.0298& 0.0731    \\ \hline
dresser & \textbf{0.0038} & 0.0076 & 0.0384  & 0.1558 \\ \hline
monitor & \textbf{0.0103} & 0.0133 & 0.0220  & 0.0783  \\ \hline
night stand    & \textbf{0.0080} & 0.0091& 0.0248 & 0.2925  \\ \hline
sofa   & \textbf{0.0068} & 0.0071 & 0.0186   & 0.0563 \\ \hline 
table    & \textbf{0.0051} & 0.0064 & 0.0326 & 0.0340  \\ \hline
toilet   & \textbf{0.0119} & 0.0129& 0.0180 & 0.0977  \\ \hline
\hline
Avg.   & \textbf{0.0092} & 0.0127& 0.0259    & 0.0956  \\ \hline
\end{tabular}
\end{table}
\vspace{-2mm}

 \begin{algorithm}[h]
\caption{Conditional generative VoxelNet for 3D Recovery}
\label{code:3D_recovery}
\begin{algorithmic}[1]
\REQUIRE ~~\\
(1) 3D training data $\{Y_i, i=1,...,n\}$; \\
(2) percentage points of corruption $\varrho$; \\
(3) the number of Langevin steps $K$; \\
(4) the number of learning iterations $T$.

\ENSURE~~\\
(1) estimated parameters $\theta$. \\

\item[]
\STATE Let $t\leftarrow 0$.
\STATE Randomly initialize $\theta^{(t)}$ with Gaussian distribution. 
\STATE Randomly corrupt $\varrho$ percent of voxels of each training example and obtain a corrupted dataset $\{Y'_{i},i=1,...,n\}$ with corresponding uncorrupted voxels $\{\tilde{M}_i, i=1,...,n\}$. 
\REPEAT 
\STATE For each $i$, starting from the corrupted data $Y'_i$, run $K$ steps of Langevin dynamics to obtain 
${\tilde{Y}}_{i}$, where each step follows equation (\ref{eq:LangevinD}). 
\STATE Fixing the uncorrupted parts of voxels $\tilde{Y}_{i}(\tilde{M}_i) \leftarrow Y_i(\tilde{M}_i)$ 
\STATE Update $\theta^{(t+1)} = \theta^{(t)} + \gamma_t L'(\theta^{(t)}) $,  with learning rate $\gamma_t$, where $L'(\theta^{(t)})$ is computed according to equation (\ref{eq:lD2}). 
\STATE Let $t \leftarrow t+1$
\UNTIL $t = T$
\end{algorithmic}
\end{algorithm}

\subsection{3D object super resolution}
\label{Exp:sr}

\begin{figure}
	\centering
	\rotatebox[origin=l]{90}{\hspace{1mm}\textbf{{\scriptsize original}}}
	\includegraphics[height=.146\linewidth]{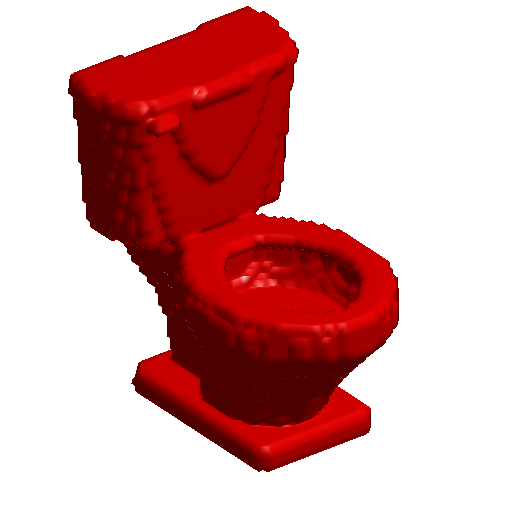}\hspace{-1.8mm}
	\includegraphics[height=.146\linewidth]{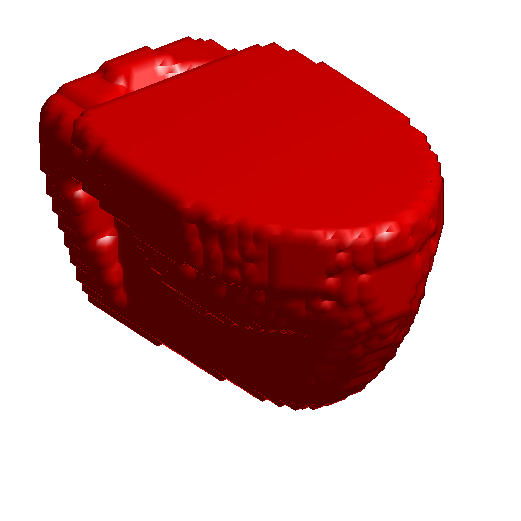}\hspace{-1.1mm}
	\includegraphics[height=.146\linewidth]{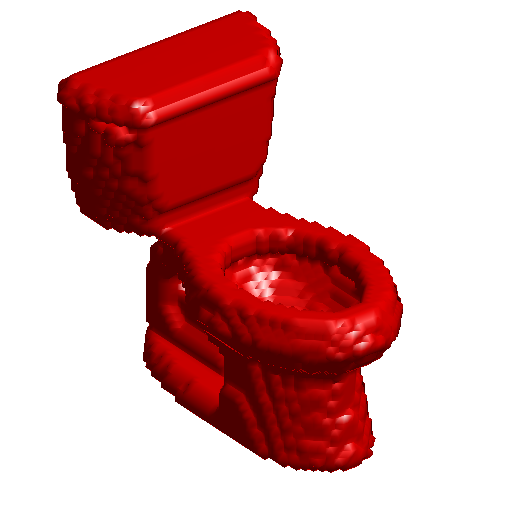}\hspace{-1.8mm}
	\includegraphics[height=.146\linewidth]{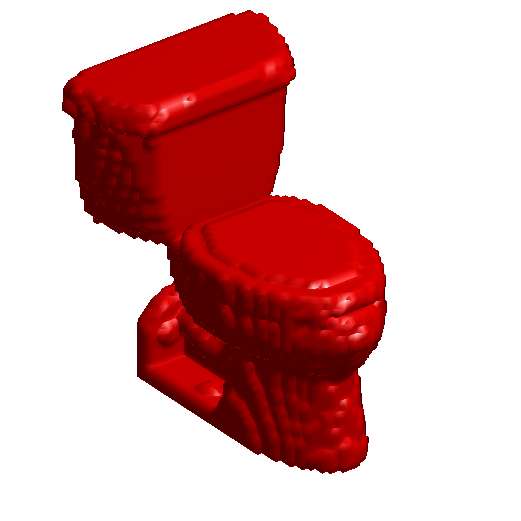}\hspace{-1.8mm}
	\includegraphics[height=.146\linewidth]{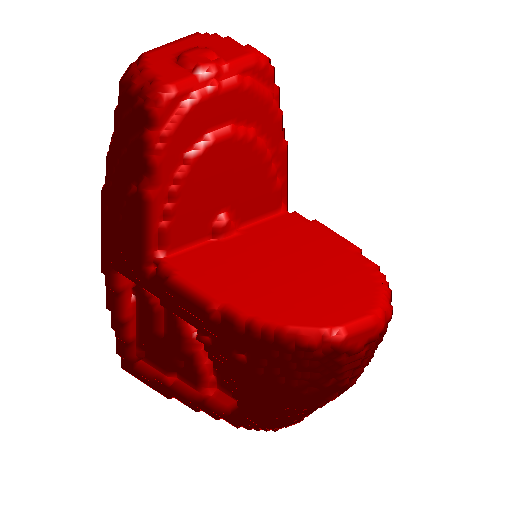}\hspace{-1.8mm}
	\includegraphics[height=.146\linewidth]{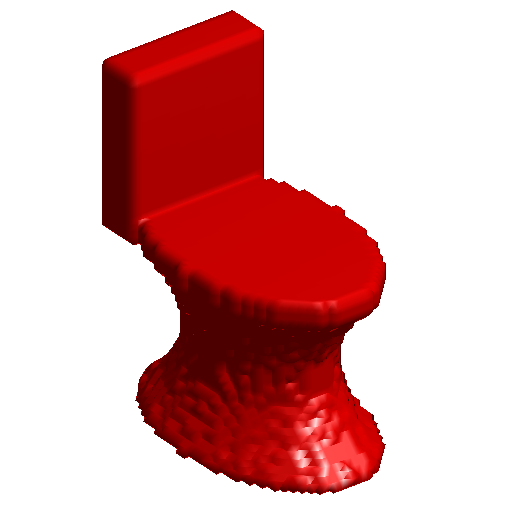}\hspace{-2.2mm}
	\includegraphics[height=.146\linewidth]{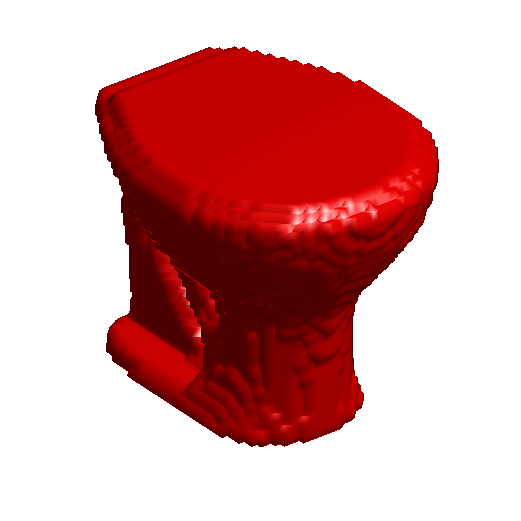}
	\\
	 \rotatebox[origin=l]{90}{\hspace{1mm}\textbf{{\scriptsize low res.}}}
	\includegraphics[height=.146\linewidth]{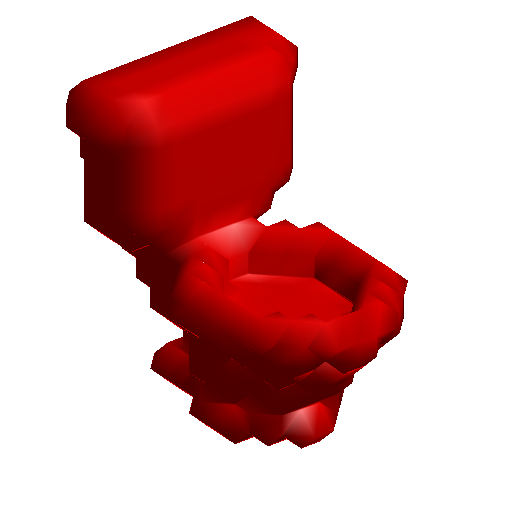}\hspace{-1.4mm}
    \includegraphics[height=.146\linewidth]{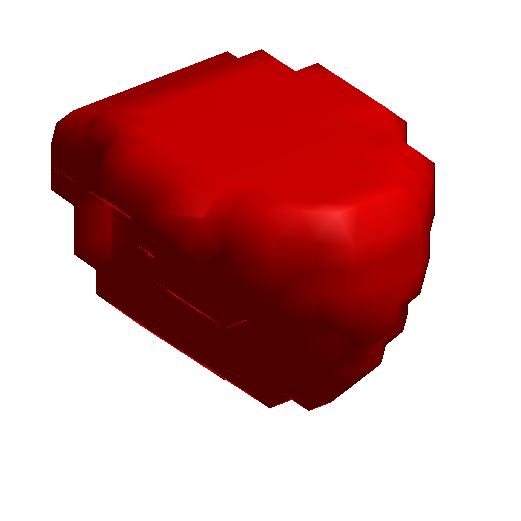}\hspace{-.4mm}
    \includegraphics[height=.146\linewidth]{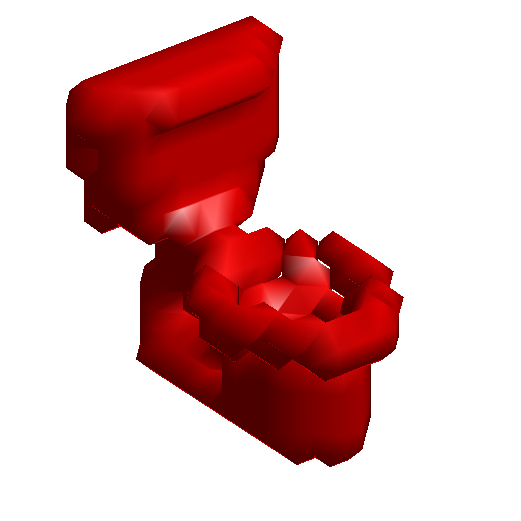}\hspace{-2.5mm}
    \includegraphics[height=.146\linewidth]{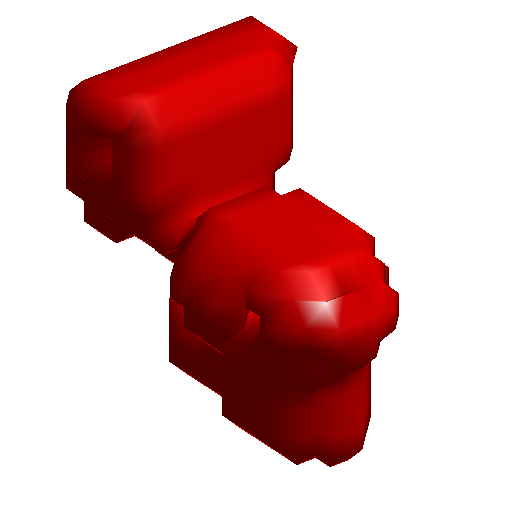}	\hspace{-2.9mm}
    \includegraphics[height=.146\linewidth]{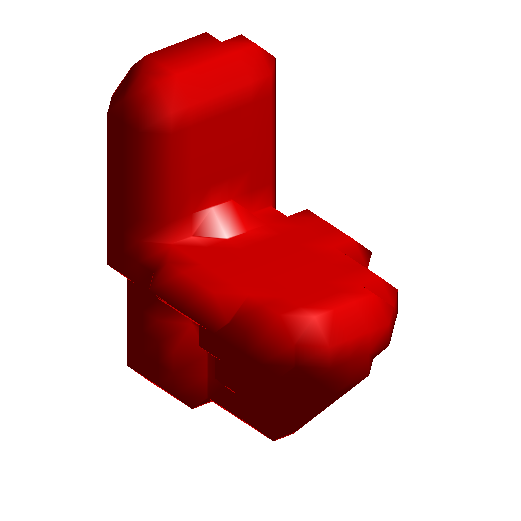}	\hspace{-2.5mm}
    \includegraphics[height=.146\linewidth]{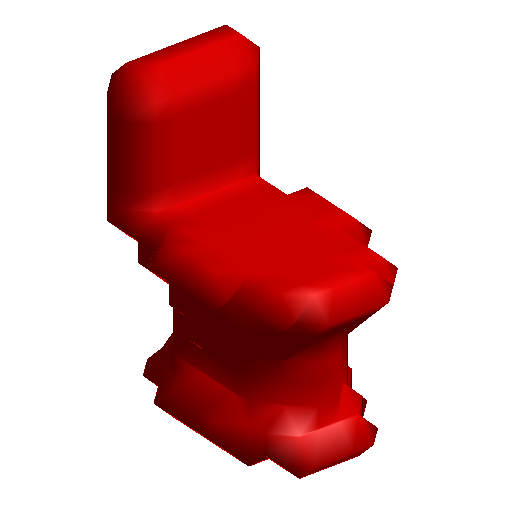}	\hspace{-2.6mm}
    \includegraphics[height=.146\linewidth]{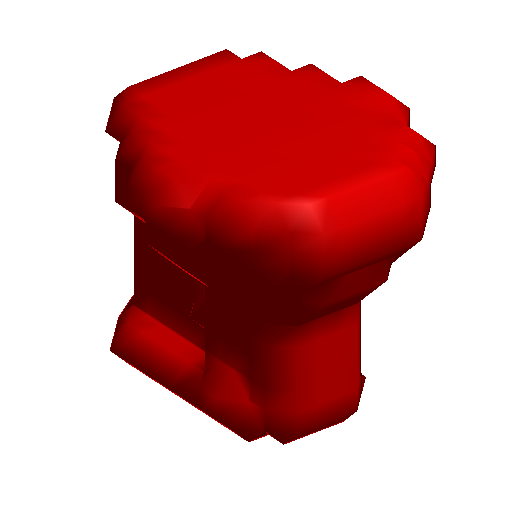}
   	\\
   	\rotatebox[origin=l]{90}{\hspace{1mm}\textbf{{\scriptsize high res.}}}
	\includegraphics[height=.146\linewidth]{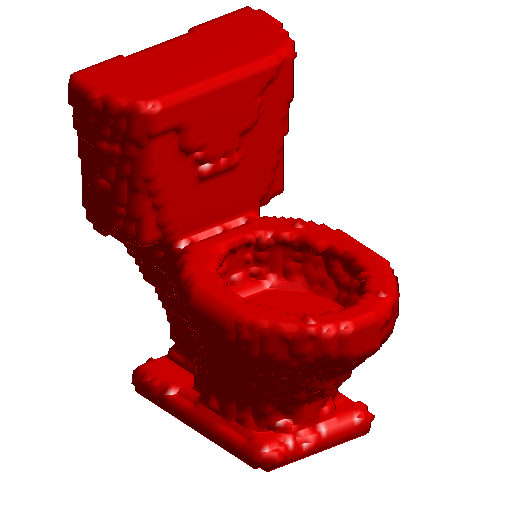}\hspace{-1.8mm}
	\includegraphics[height=.146\linewidth]{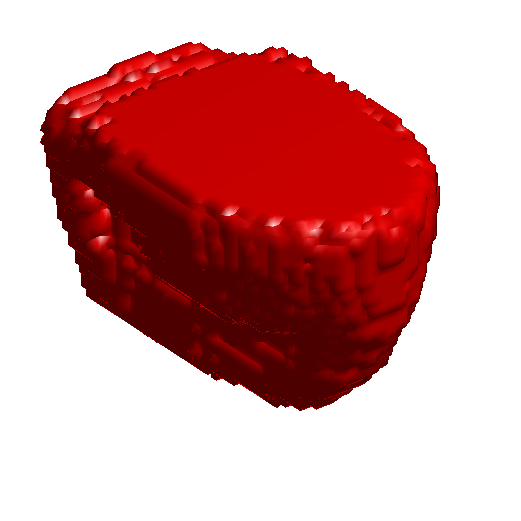}\hspace{-1.1mm}
	\includegraphics[height=.146\linewidth]{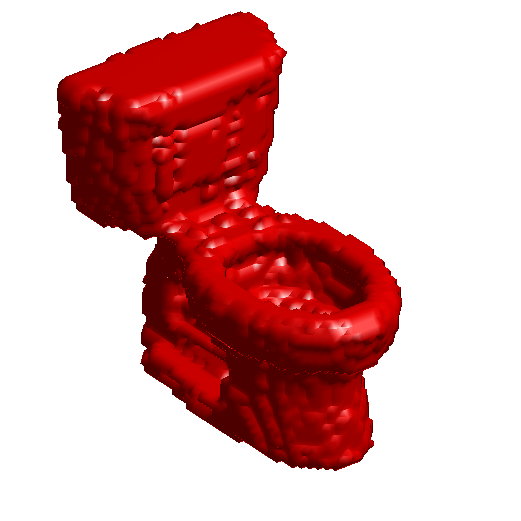}\hspace{-1.8mm}
	\includegraphics[height=.146\linewidth]{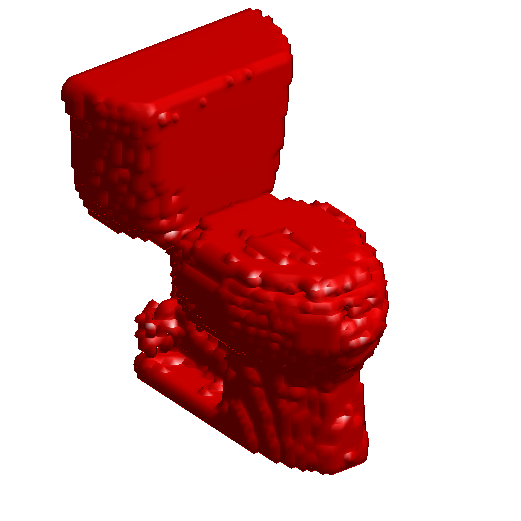}\hspace{-1.8mm}
	\includegraphics[height=.146\linewidth]{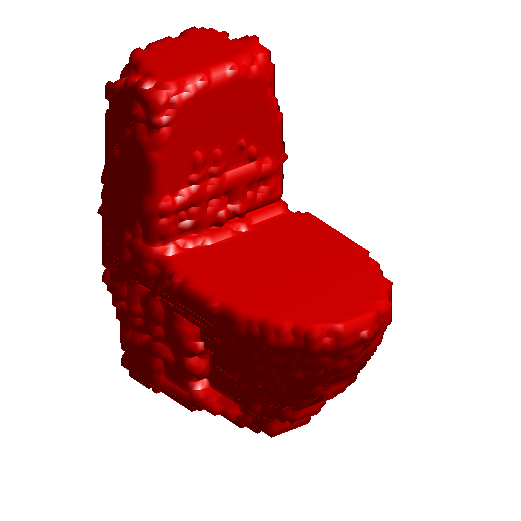}\hspace{-1.8mm}
	\includegraphics[height=.146\linewidth]{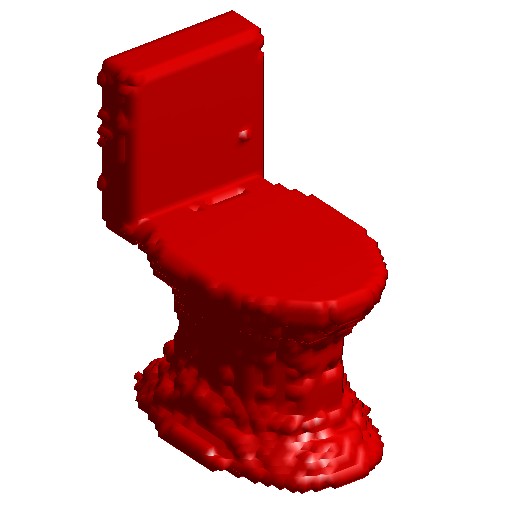}\hspace{-2.2mm}
	\includegraphics[height=.146\linewidth]{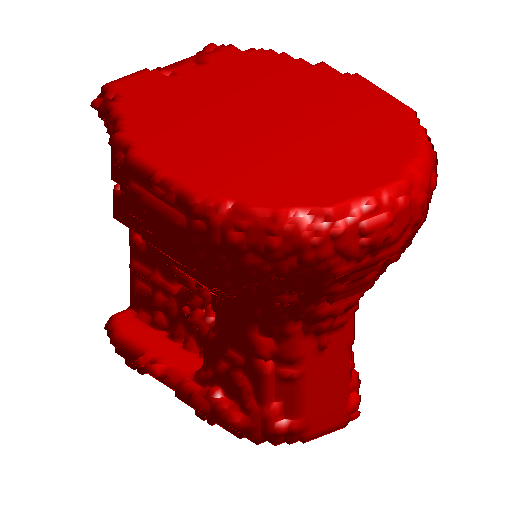}\\
	(a) toilet\\ \vspace{2mm}	

	\rotatebox[origin=l]{90}{\hspace{1mm}\textbf{{\scriptsize original}}}
	\includegraphics[height=.134\linewidth]{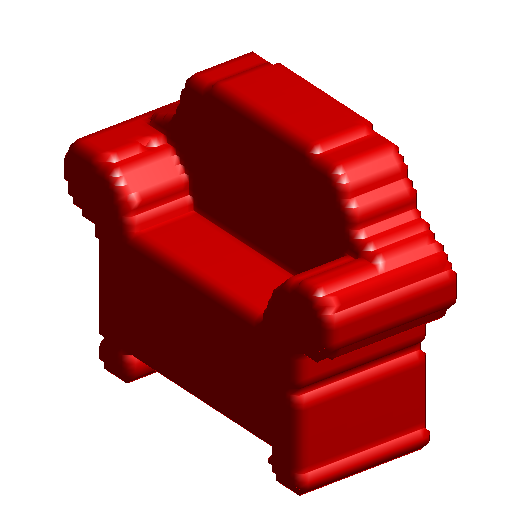}\hspace{-0.9mm}
	\includegraphics[height=.134\linewidth]{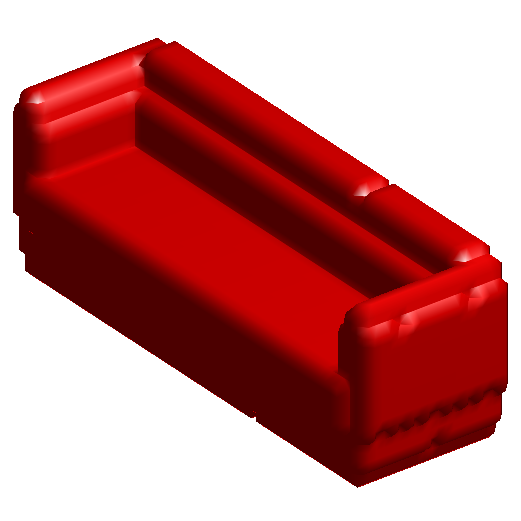}\hspace{-0.8mm}
	\includegraphics[height=.134\linewidth]{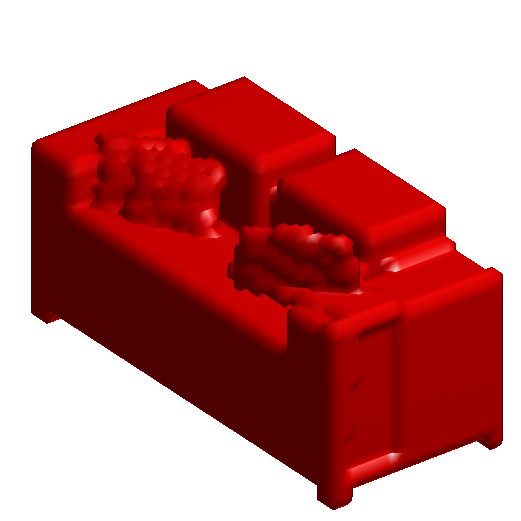}\hspace{-0.8mm}
	\includegraphics[height=.134\linewidth]{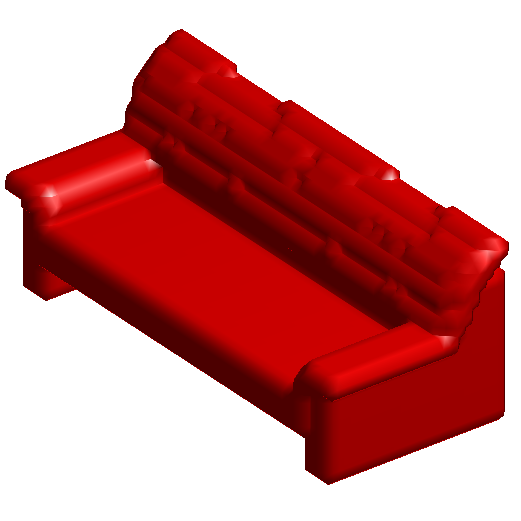}\hspace{-0.8mm}
	\includegraphics[height=.134\linewidth]{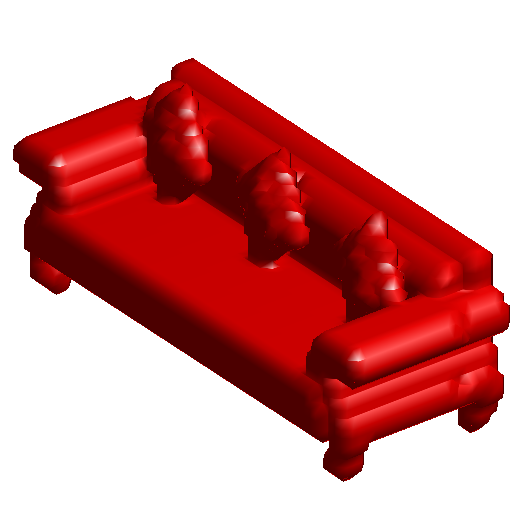}\hspace{-0.8mm}
	\includegraphics[height=.134\linewidth]{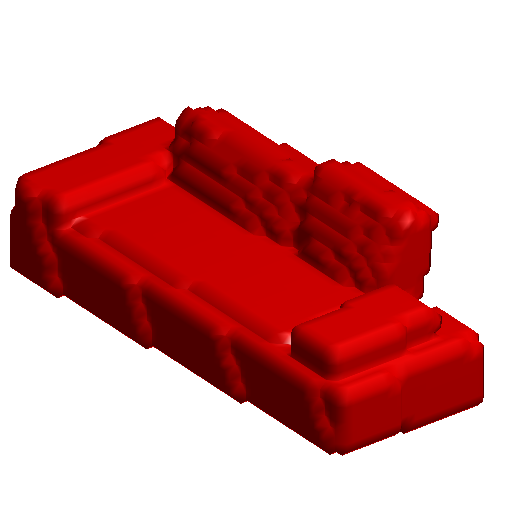}\hspace{-0.8mm}
	\includegraphics[height=.134\linewidth]{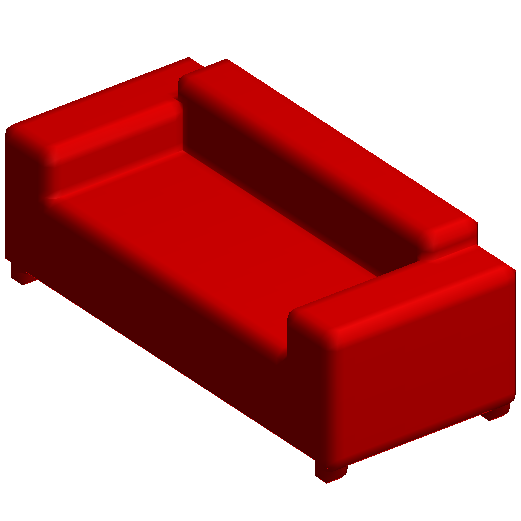}
	\\
	 \rotatebox[origin=l]{90}{\hspace{1mm}\textbf{{\scriptsize low res.}}}
	\includegraphics[height=.134\linewidth]{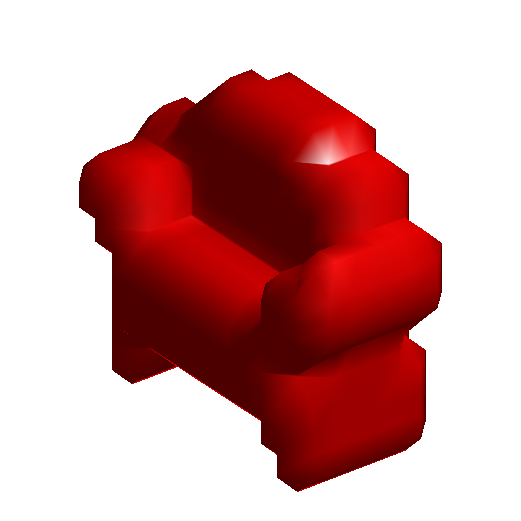}\hspace{-0.9mm}
	\includegraphics[height=.134\linewidth]{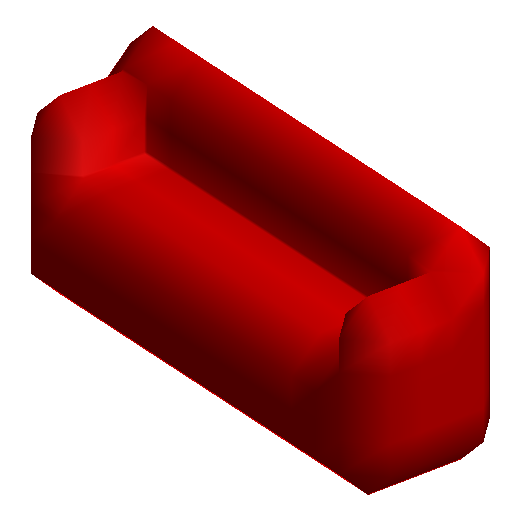}\hspace{-0.8mm}
	\includegraphics[height=.134\linewidth]{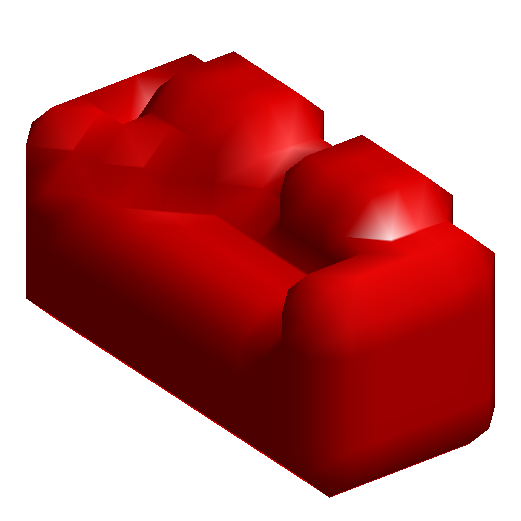}\hspace{-0.8mm}
	\includegraphics[height=.134\linewidth]{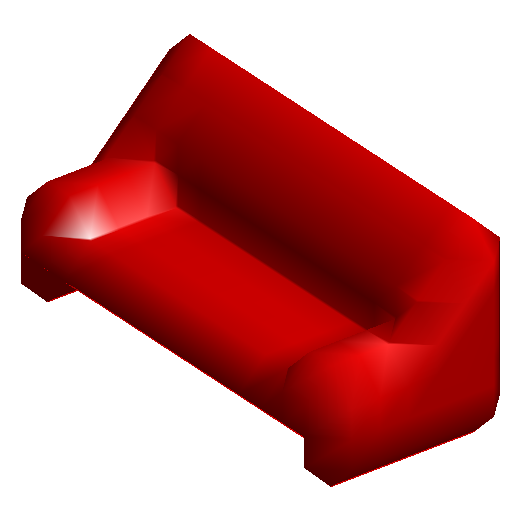}\hspace{-0.8mm}
	\includegraphics[height=.134\linewidth]{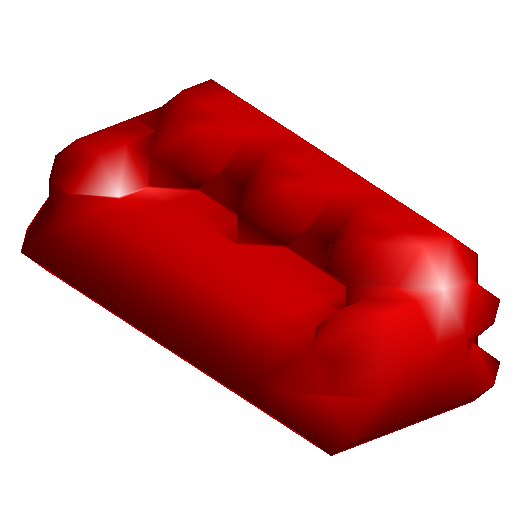}\hspace{-0.8mm}
	\includegraphics[height=.134\linewidth]{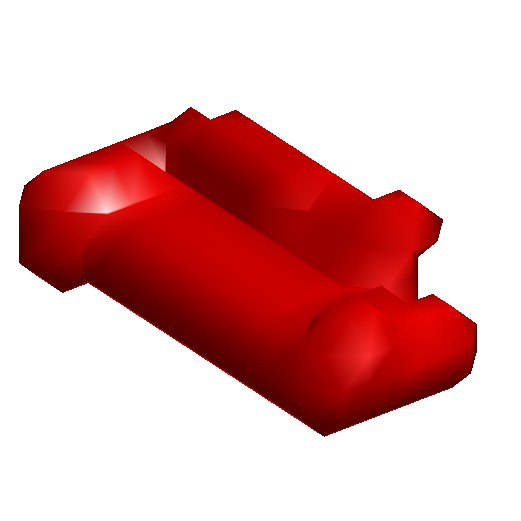}\hspace{-0.8mm}
	\includegraphics[height=.134\linewidth]{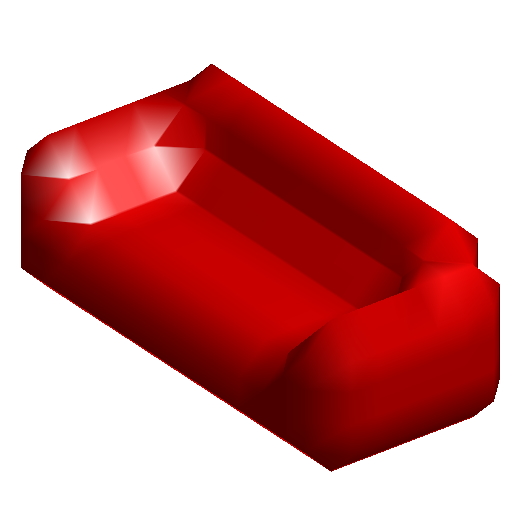}
   	\\
   	\rotatebox[origin=l]{90}{\hspace{1mm}\textbf{{\scriptsize high res.}}}
	\includegraphics[height=.134\linewidth]{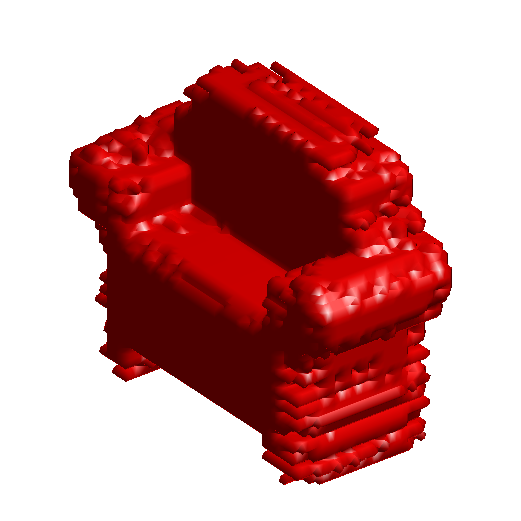}\hspace{-0.9mm}
	\includegraphics[height=.134\linewidth]{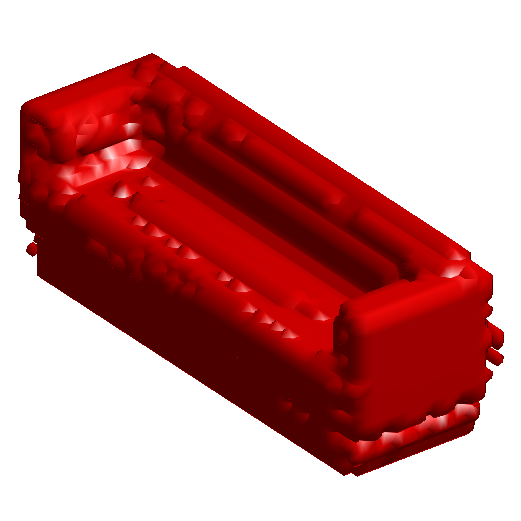}\hspace{-0.8mm}
	\includegraphics[height=.134\linewidth]{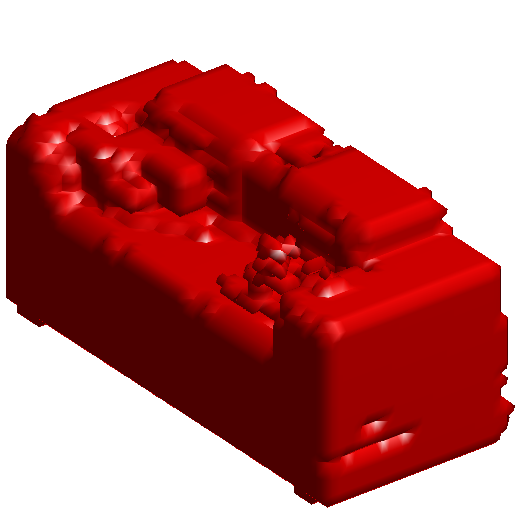}\hspace{-0.8mm}
	\includegraphics[height=.134\linewidth]{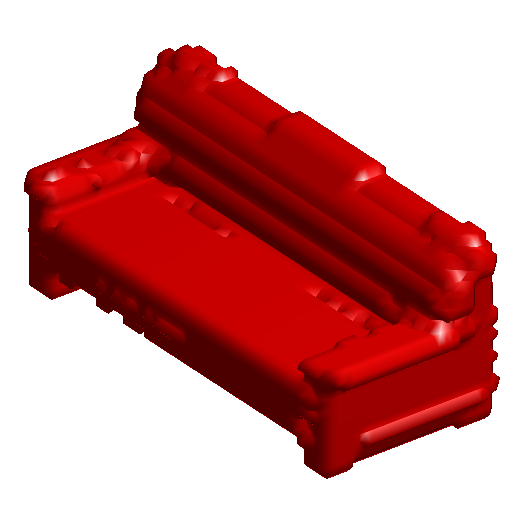}\hspace{-0.8mm}
	\includegraphics[height=.134\linewidth]{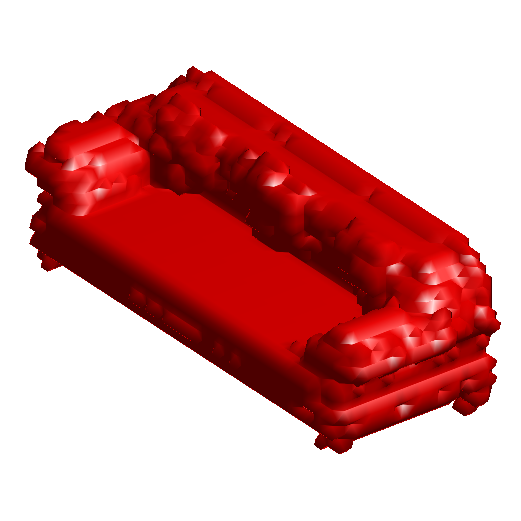}\hspace{-0.8mm}
	\includegraphics[height=.134\linewidth]{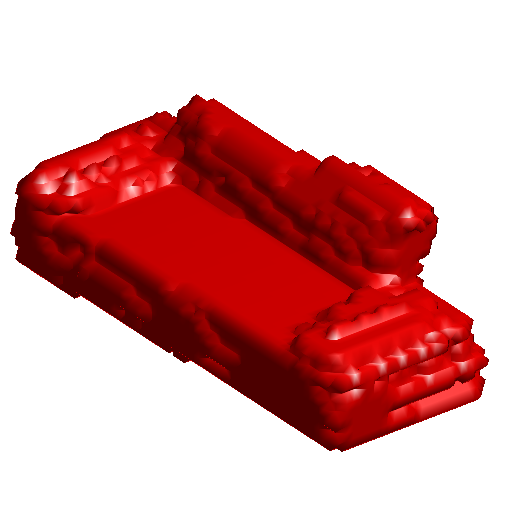}\hspace{-0.8mm}
	\includegraphics[height=.134\linewidth]{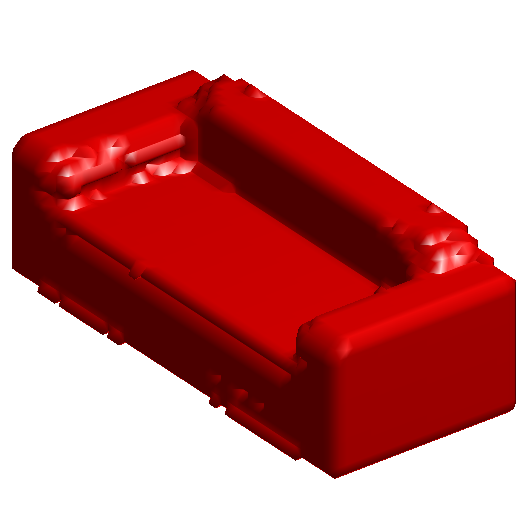}\\
	(b) sofa\\ \vspace{2mm}

		\caption{3D object super resolution by conditional generative VoxelNet. The first row displays some original 3D objects ($64 \times 64 \times 64$ voxels). The second row shows the corresponding low resolution 3D objects ($16 \times 16 \times 16$ voxels). The last row displays the corresponding super resolution results which are obtained by sampling from the conditional generative VoxelNet by running 10 steps of Langevin dynamics initialized with the objects shown in the second row.}	
	\label{exp:superResolution}
\end{figure}

 \begin{algorithm}[h]
\caption{Conditional generative VoxelNet for 3D Super resolution}
\label{code:3D_Super_resolution}
\begin{algorithmic}[1]
\REQUIRE ~~\\
(1) 3D training data $\{Y_i, i=1,...,n\}$; \\
(2) the number of Langevin steps $K$; \\
(3) the number of learning iterations $T$.

\ENSURE~~\\
(1) estimated parameters $\theta$. \\

\item[]
\STATE Let $t\leftarrow 0$. 
\STATE Randomly initialize $\theta^{(t)}$ with Gaussian distribution. 
\STATE Down-scale each training example via $Y_{\text{low}i}=CY_i$ and obtain a low resolution dataset $\{{Y_\text{low}}_{i},i=1,...,n\}$.
\STATE Compute up-scaled data $Y_{\text{high}i}^{'} \leftarrow C^{-}Y_{\text{low}i}$, for $i = 1, ..., n$. 
\REPEAT 
\STATE For $i=1,...,n$, initialize via $\tilde{Y_i}\leftarrow Y_{\text{high}i}^{'}$, run $K$ steps of modified Langevin dynamics, where each step follows $\tY_i \leftarrow \tY_i +(I-C^{-}C)\Delta Y_i$, where $\Delta Y_i$ is the change of $\tY_i$ by runing one step of equation (\ref{eq:LangevinD}). 

\STATE Update $\theta^{(t+1)} = \theta^{(t)} + \gamma_t L'(\theta^{(t)}) $,  with learning rate $\gamma_t$, where $L'(\theta^{(t)})$ is computed according to equation (\ref{eq:lD2}). 
\STATE Let $t \leftarrow t+1$
\UNTIL $t = T$
\end{algorithmic}
\end{algorithm}

\begin{figure*}
\centering
\begin{minipage}[b]{.34\textwidth}
  \centering

     \includegraphics[width=.25\linewidth]{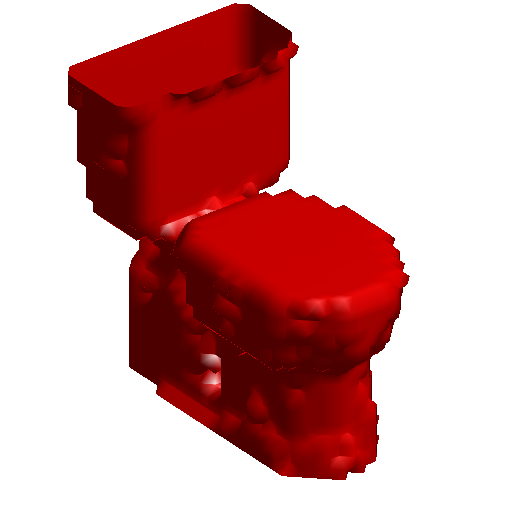}\hspace{-2mm} 
     \includegraphics[width=.25\linewidth]{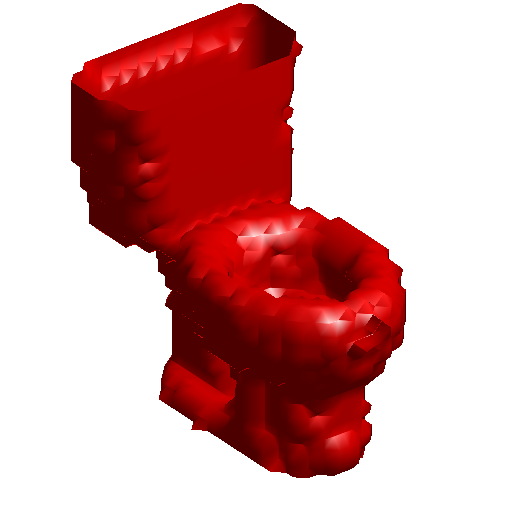}\hspace{-2mm} 
     \includegraphics[width=.25\linewidth]{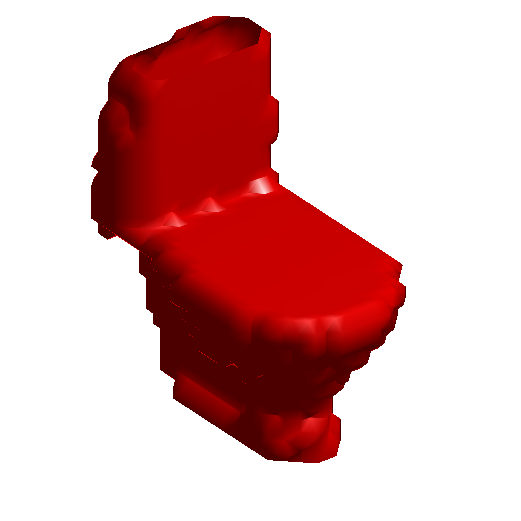}\hspace{-2mm} 
     \includegraphics[width=.25\linewidth]{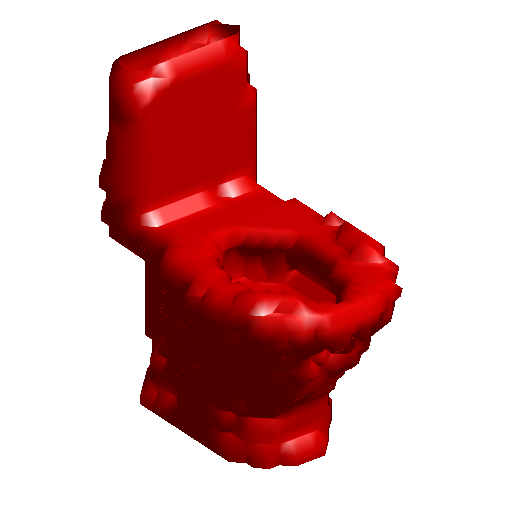}\hspace{-2mm}  \\	

     \includegraphics[width=.25\linewidth]{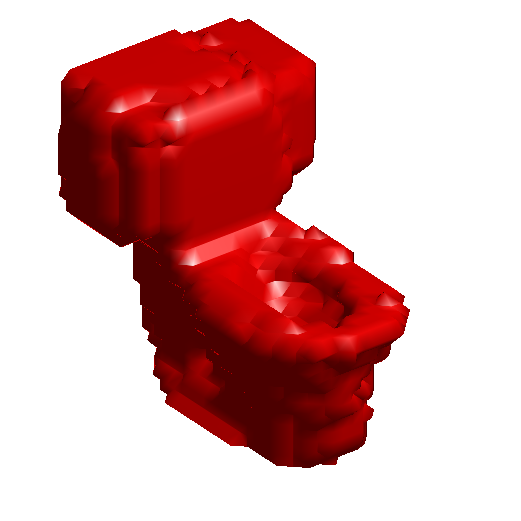}\hspace{-2mm} 
     \includegraphics[width=.25\linewidth]{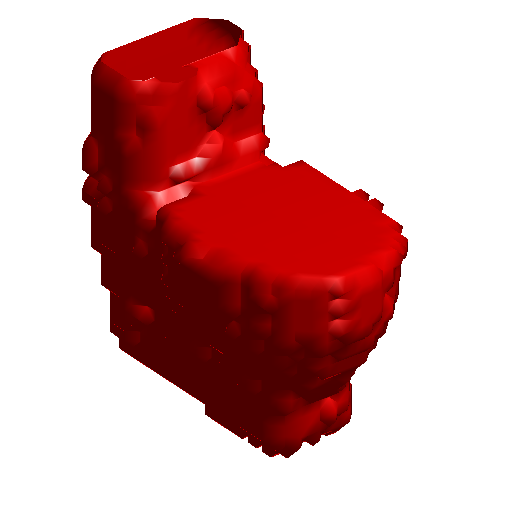}\hspace{-2mm} 
     \includegraphics[width=.25\linewidth]{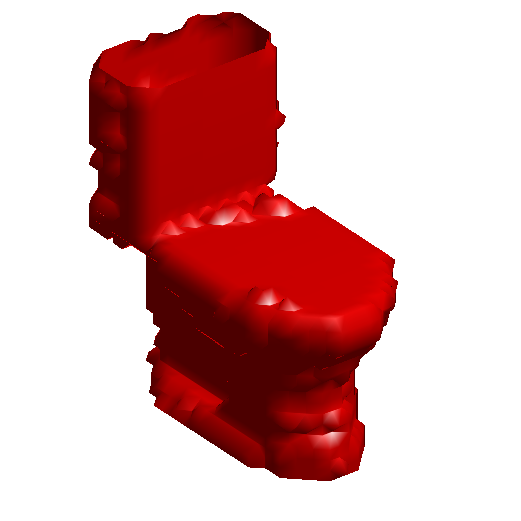}\hspace{-2mm} 
     \includegraphics[width=.25\linewidth]{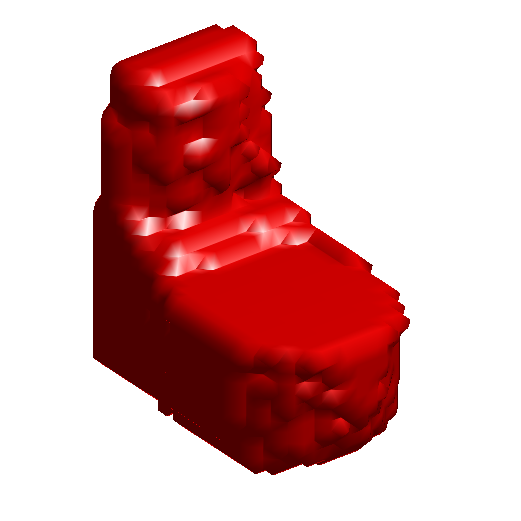}\hspace{-2mm}  \\	     
     (a) toilet \\ \vspace{1mm}

	\includegraphics[width=.23\linewidth]{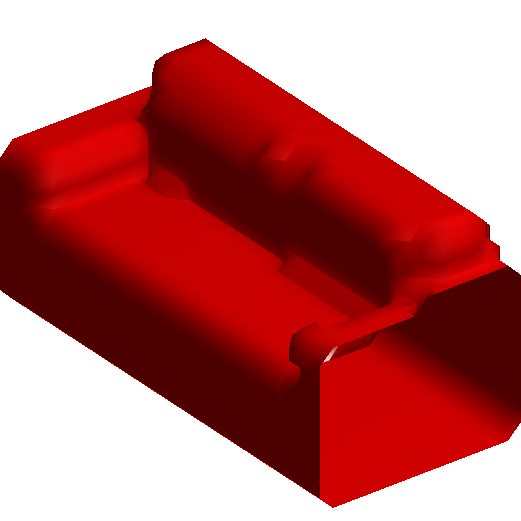}\hspace{-0.8mm}
	\includegraphics[width=.23\linewidth]{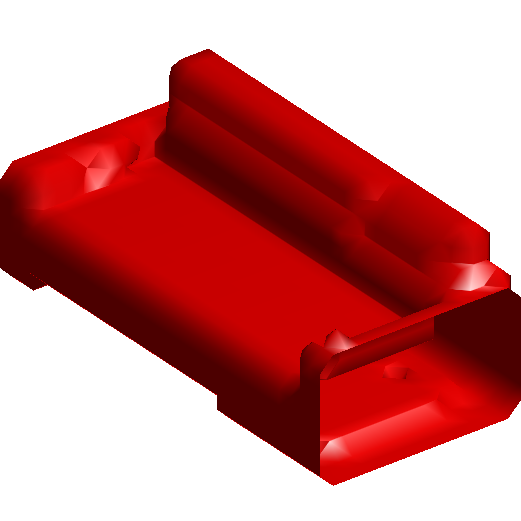}\hspace{-0.8mm}
	\includegraphics[width=.23\linewidth]{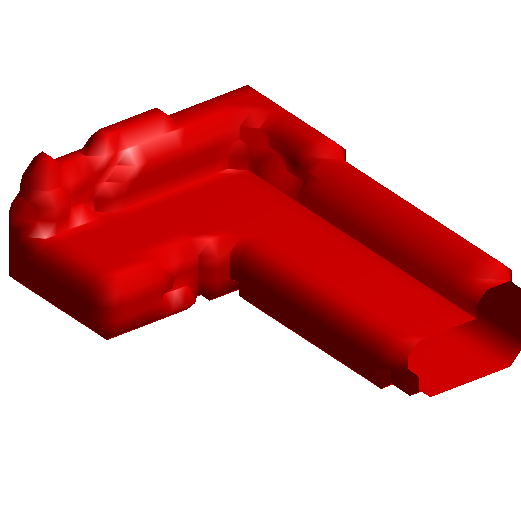}\hspace{-0.8mm}
	\includegraphics[width=.23\linewidth]{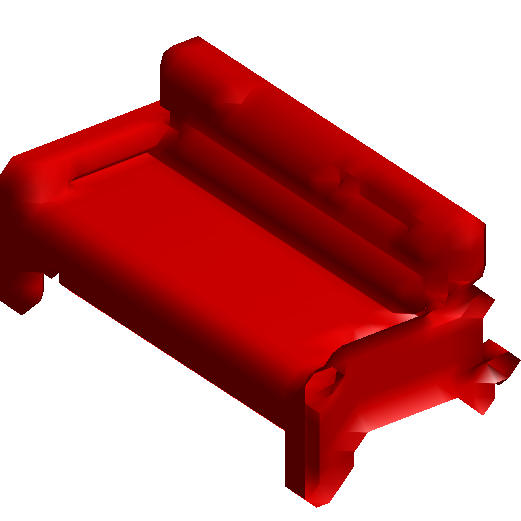}\hspace{-0.8mm} \\

	\includegraphics[width=.23\linewidth]{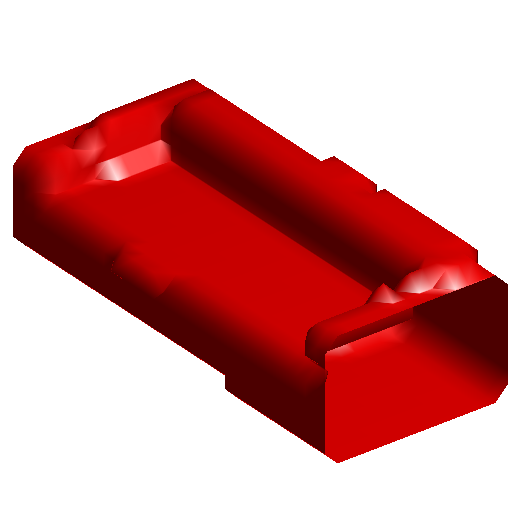}\hspace{-0.8mm}
	\includegraphics[width=.23\linewidth]{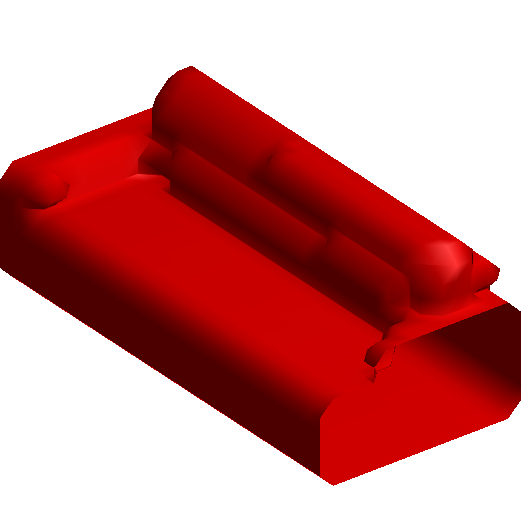}\hspace{-0.8mm}
	\includegraphics[width=.23\linewidth]{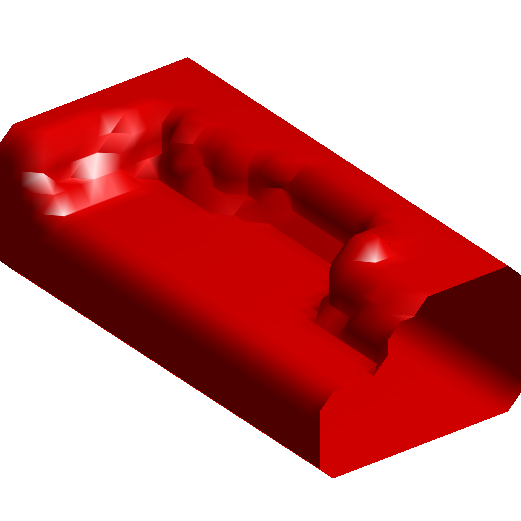}\hspace{-0.8mm}
	\includegraphics[width=.23\linewidth]{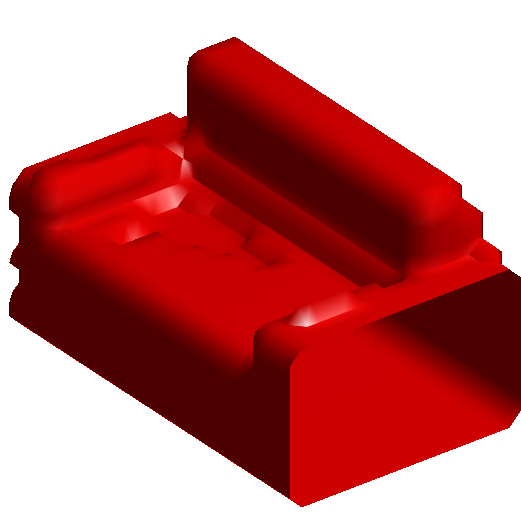}\hspace{-0.8mm}\\
	(b) sofa\\
  \caption{Synthesis by 3D generators. (a) toilet (b) sofa. The 3D generators are trained by the generative VoxelNet via MCMC teaching. The 3D shapes are generated by first sampling $Z$ from $\mathcal{N}(0,I_d)$ and then mapping $Z$ to 3D shape space via $g(Z;\alpha)$.}
  \label{fig:synthesis_generator}
\end{minipage}%
~
\begin{minipage}[b]{.64\textwidth}
  \centering
  \rotatebox[origin=l]{90}{\hspace{3mm}\textbf{{\footnotesize (1)}}}
  	\includegraphics[width=.14\linewidth]{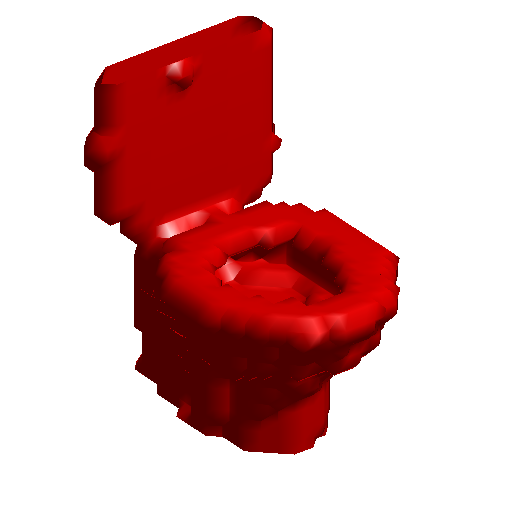}\hspace{-3.5mm} 
	\includegraphics[width=.14\linewidth]{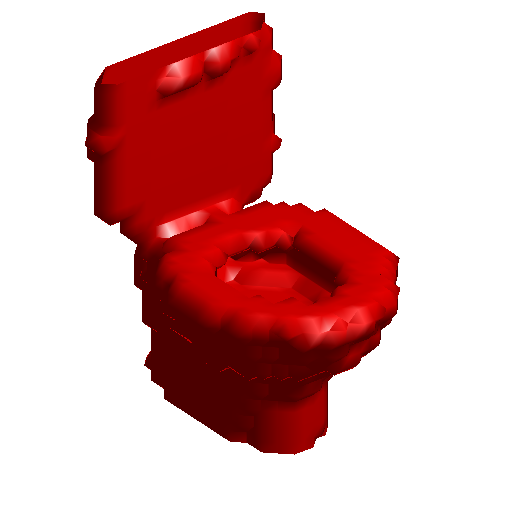}\hspace{-3.5mm} 
	\includegraphics[width=.14\linewidth]{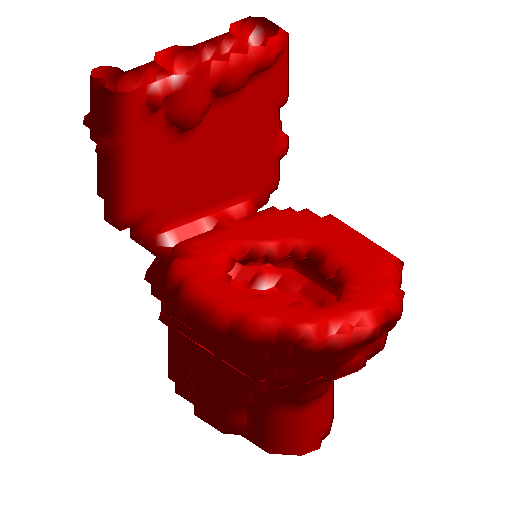}\hspace{-3.5mm} 
    \includegraphics[width=.14\linewidth]{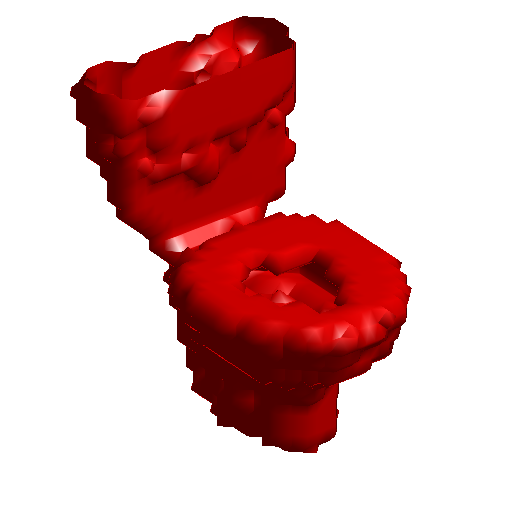}\hspace{-3.5mm} 
    \includegraphics[width=.14\linewidth]{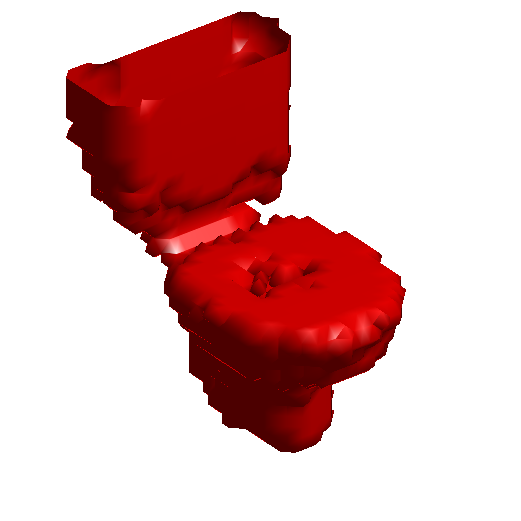}\hspace{-3.5mm} 
    \includegraphics[width=.14\linewidth]{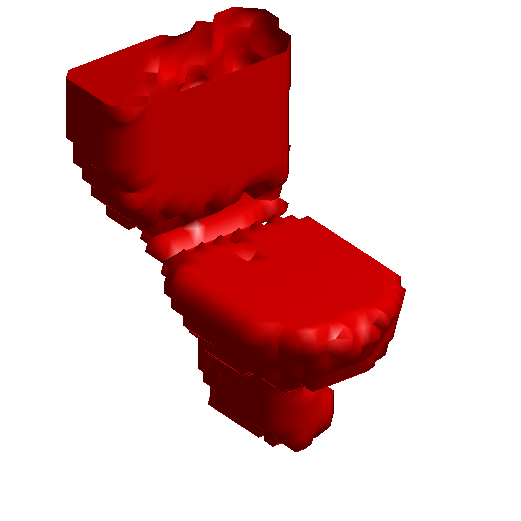}\hspace{-3.5mm} 
    \includegraphics[width=.14\linewidth]{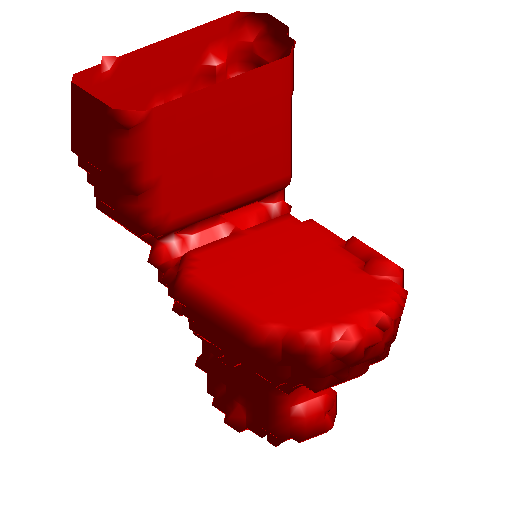}\hspace{-3.5mm} 
    \includegraphics[width=.14\linewidth]{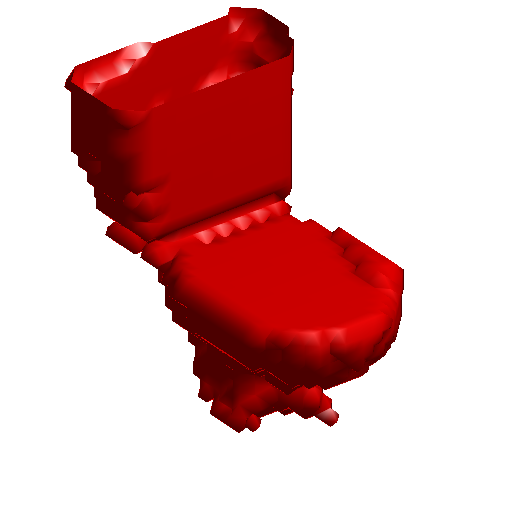}\\
    
      \rotatebox[origin=l]{90}{\hspace{3mm}\textbf{{\footnotesize (2)}}}
  	\includegraphics[width=.14\linewidth]{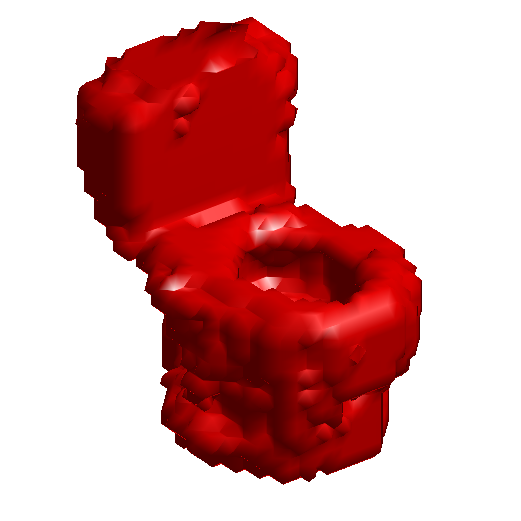}\hspace{-3.5mm} 
	\includegraphics[width=.14\linewidth]{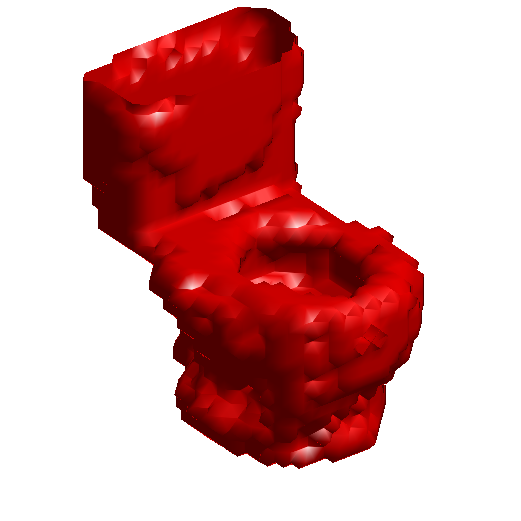}\hspace{-3.5mm} 
	\includegraphics[width=.14\linewidth]{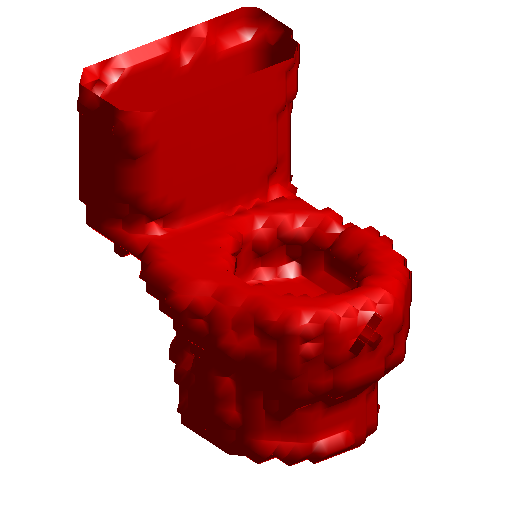}\hspace{-3.5mm} 
    \includegraphics[width=.14\linewidth]{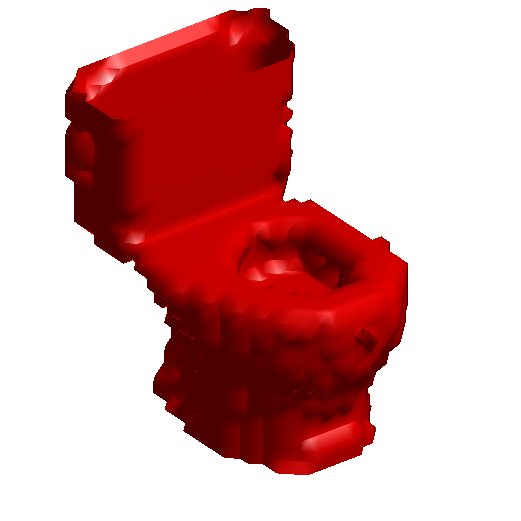}\hspace{-3.5mm} 
    \includegraphics[width=.14\linewidth]{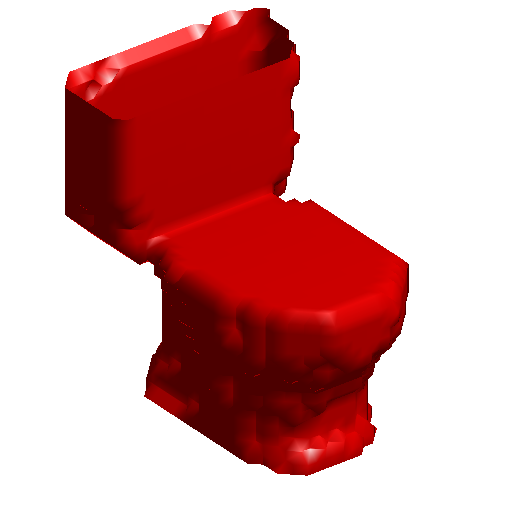}\hspace{-3.5mm} 
    \includegraphics[width=.14\linewidth]{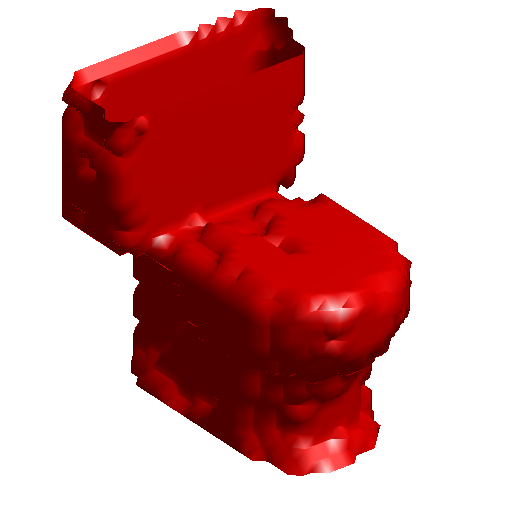}\hspace{-3.5mm} 
    \includegraphics[width=.14\linewidth]{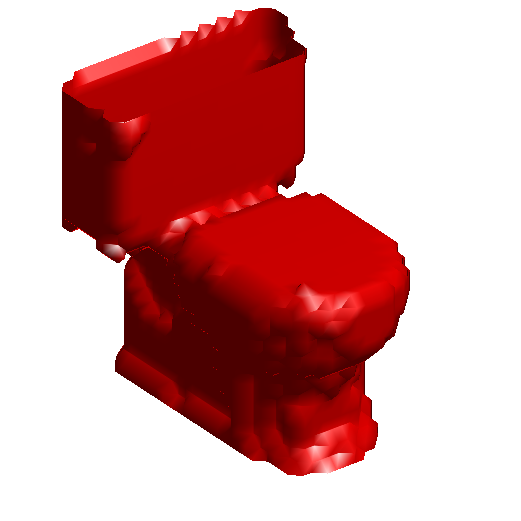}\hspace{-3.5mm} 
    \includegraphics[width=.14\linewidth]{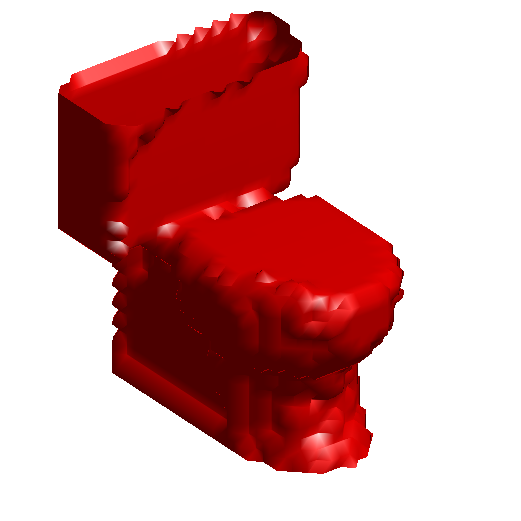}\\    
    (a) toilet \\ \vspace{1mm}
    \rotatebox[origin=l]{90}{\hspace{3mm}\textbf{{\footnotesize (1)}}}
    	\includegraphics[width=.12\linewidth]{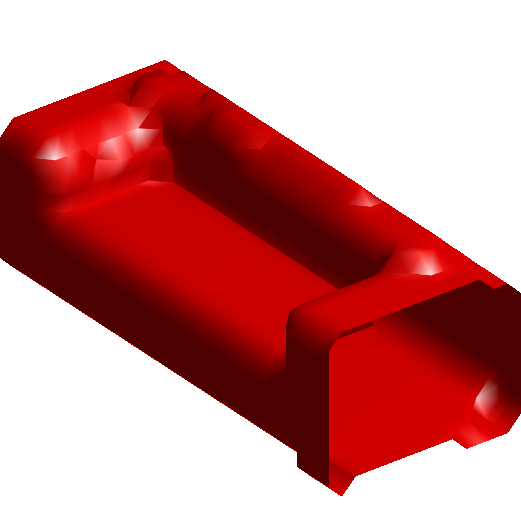} \hspace{-1.7mm} 
	\includegraphics[width=.12\linewidth]{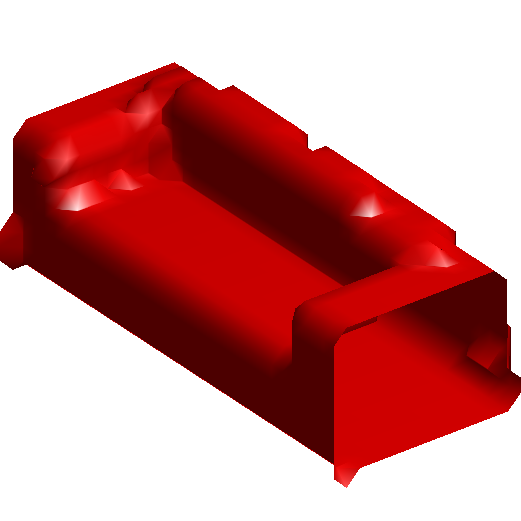} \hspace{-1.7mm}  
	\includegraphics[width=.12\linewidth]{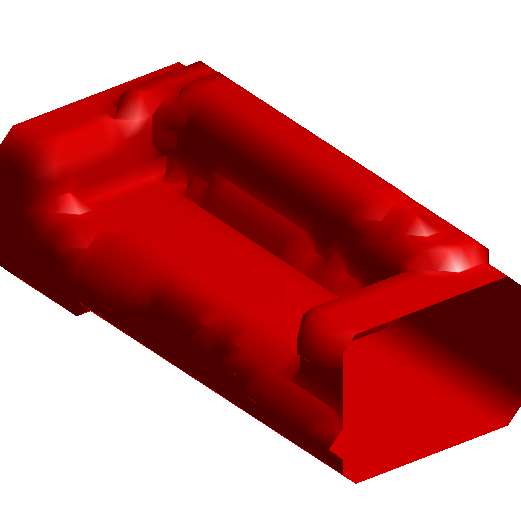} \hspace{-1.7mm} 
	\includegraphics[width=.12\linewidth]{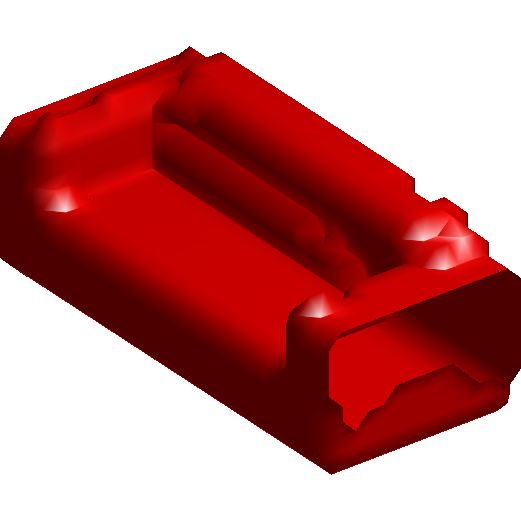} \hspace{-1.7mm}  
	\includegraphics[width=.12\linewidth]{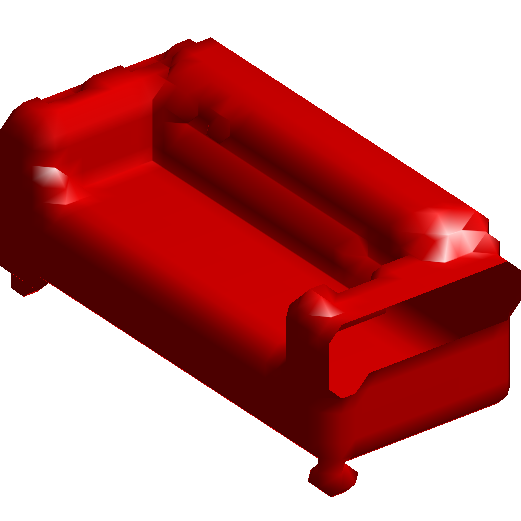} \hspace{-1.7mm} 
	\includegraphics[width=.12\linewidth]{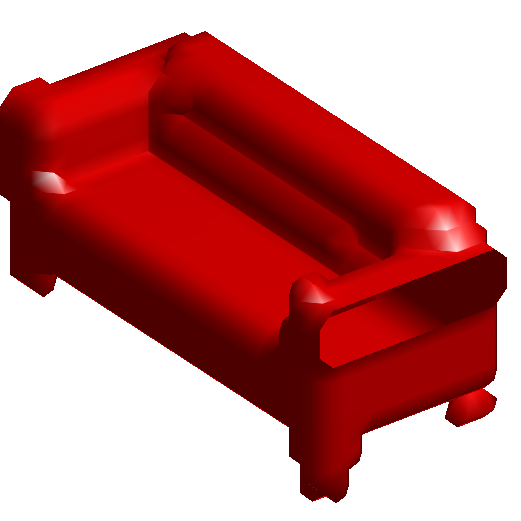} \hspace{-1.7mm} 
	\includegraphics[width=.12\linewidth]{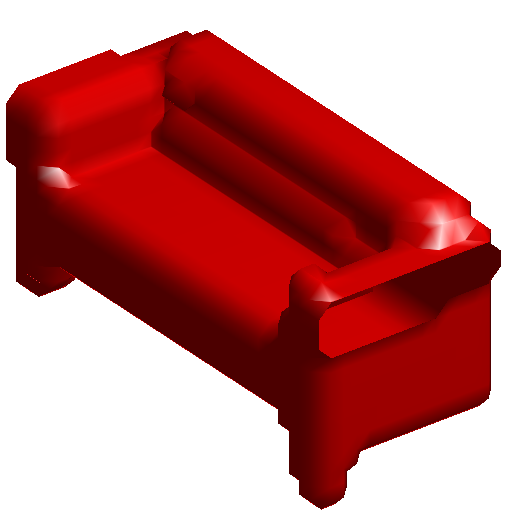} \hspace{-1.7mm} 
	\includegraphics[width=.12\linewidth]{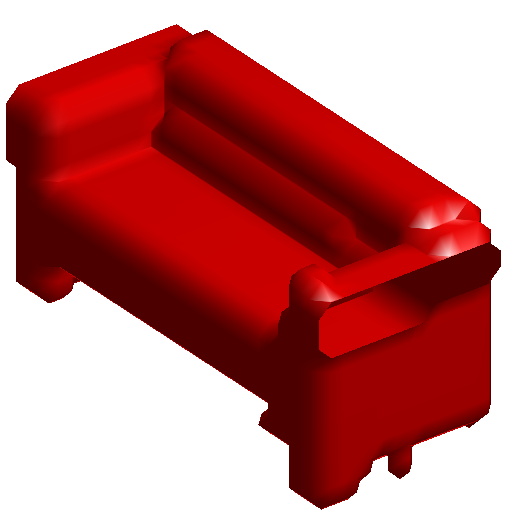}
	
    \rotatebox[origin=l]{90}{\hspace{3mm}\textbf{{\footnotesize (2)}}}
    	\includegraphics[width=.12\linewidth]{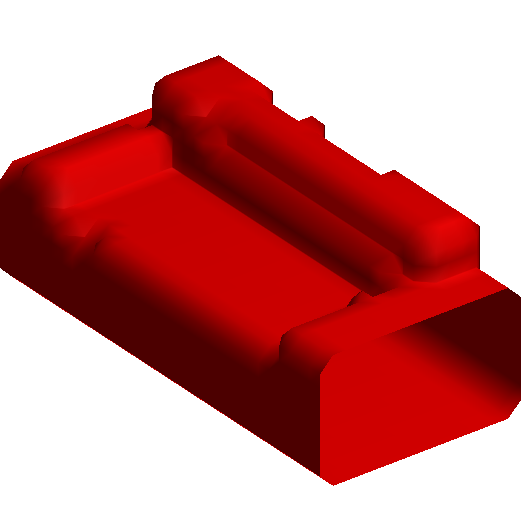} \hspace{-1.7mm} 
	\includegraphics[width=.12\linewidth]{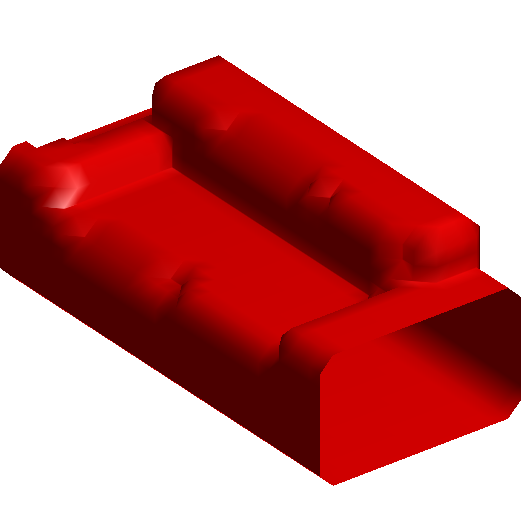} \hspace{-1.7mm}  
	\includegraphics[width=.12\linewidth]{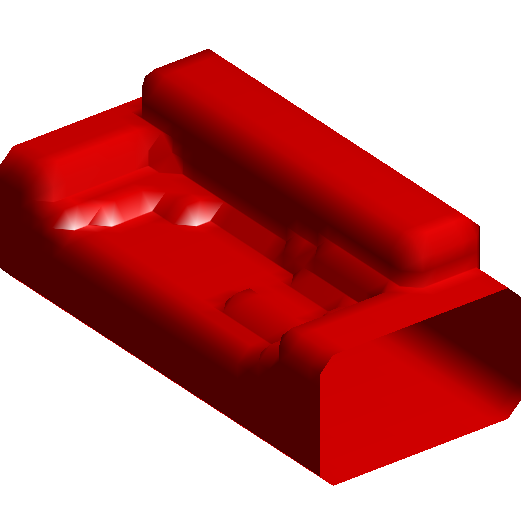} \hspace{-1.7mm} 
	\includegraphics[width=.12\linewidth]{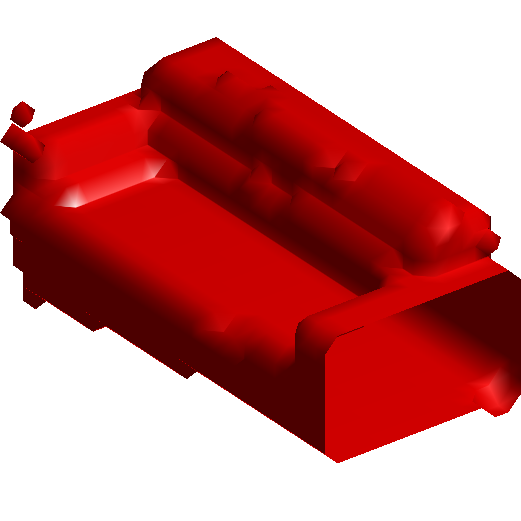} \hspace{-1.7mm}  
	\includegraphics[width=.12\linewidth]{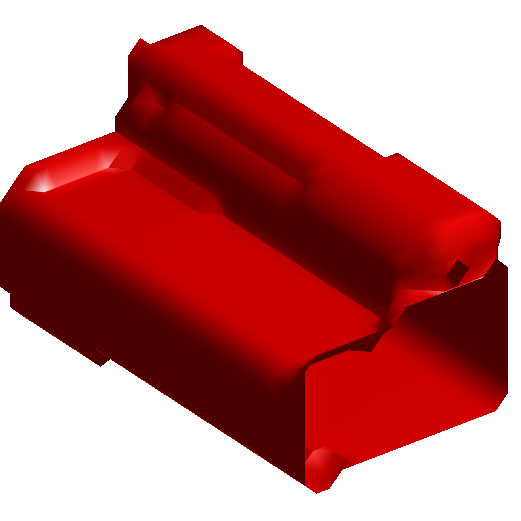} \hspace{-1.7mm} 
	\includegraphics[width=.12\linewidth]{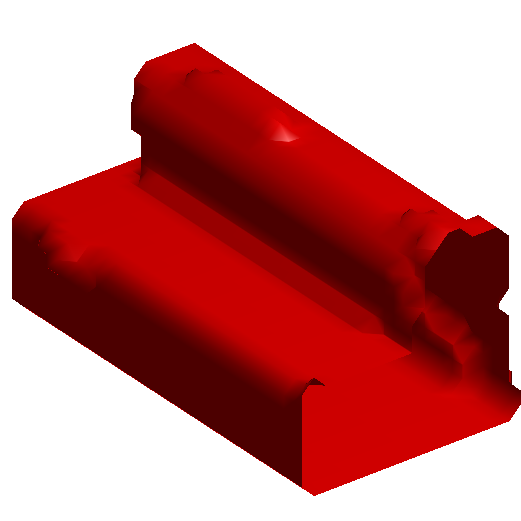} \hspace{-1.7mm} 
	\includegraphics[width=.12\linewidth]{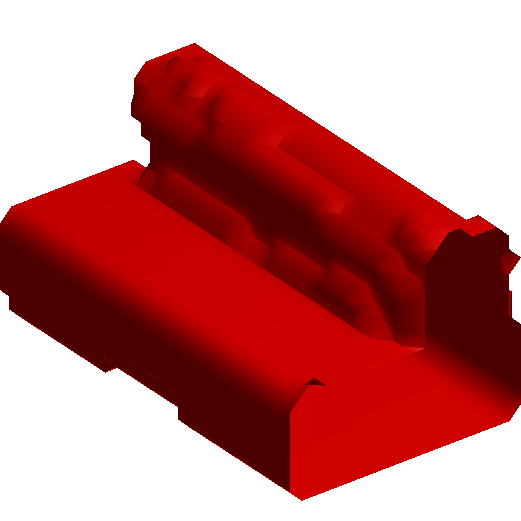} \hspace{-1.7mm} 
	\includegraphics[width=.12\linewidth]{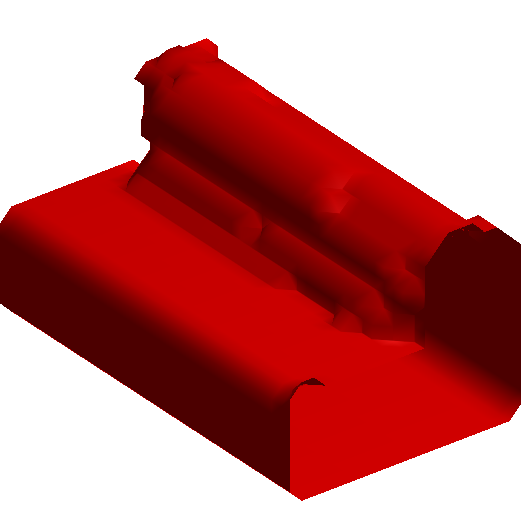}	\\
	(b) sofa \\	
  \caption{Interpolation between latent vectors of the 3D objects on the two ends. (a) toilet (b) sofa. The 3D generators are trained via MCMC teaching on each category. For each row in each category, the 3D shapes are generated from the learned generator model $g(Z; \alpha)$ with linear interpolation in latent space (i.e., $(1-\rho)\cdot Z_0 +\rho \cdot Z_1$), where $Z_0$ and $Z_1$ are latent vectors of the 3D objects on the two ends, $\rho$ is discretized into 8 values within $[0,1]$. }
  \label{fig:interpolation}
\end{minipage}
\end{figure*}

We test the conditional generative VoxelNet on the 3D object super resolution task. Similar to Experiment \ref{Exp:objectRecovery}, we can perform super resolution on a low resolution 3D object by sampling from a conditional generative VoxelNet $p(Y_{\rm high}|Y_{\rm low}; \theta)$, where $Y_{\rm high}$ denotes a high resolution version of $Y_{\rm low}$. 
The sampling of the conditional model $p(Y_{\rm high}|Y_{\rm low}; \theta)$ is accomplished by the Langevin dynamics initialized with the given low resolution 3D object that needs to be super-resolutioned. In the learning stage, we learn the conditional model from the fully observed training 3D objects as well as their low resolution versions. To specialize the learned model to this super resolution task, in the training process, we down-scale each fully observed training 3D object $Y_{\rm high}$ into a low resolution version $Y_{\rm low}$, which leads to information loss. In each iteration, we first up-scale $Y_{\rm low}$ by expanding each voxel of $Y_{\rm low}$ into a $d \times d \times d$ block (where $d$ is the ratio between the sizes of $Y_{\rm high}$ and $Y_{\rm low}$) of constant values to obtain an up-scaled version  $Y^{'}_{\rm high}$ of $Y_{\rm low}$ (The up-scaled $Y^{'}_{\rm high}$ is not identical to the original high resolution $Y_{\rm high}$ since the high resolution details are lost), and then run Langevin dynamics starting from $Y^{'}_{\rm high}$. The parameters $\theta$ are then updated by gradient ascent according to equation (\ref{eq:lD2}). Figure \ref{exp:superResolution} shows some qualitative results of 3D super resolution, where we use a 2-layer conditional generative VoxelNet. The first layer has 200 $16 \times 16 \times 16$ filters with sub-sampling of 3. The second layer is a fully-connected layer with one single filter. The Langevin step size is $\Delta \tau=0.0001$, and the number of Langevin steps is $K=90$. To be more specific, let $Y_{\rm low} = C Y_{\rm high}$, where $C$ is the down-scaling matrix, e.g., each voxel of $Y_{\rm low}$ is the average of  the corresponding $d \times d \times d$ block of $Y_{\rm high}$. Let $C^{-}$ be the pseudo-inverse of $C$, e.g., $C^{-} Y_{\rm low}$ gives us a high resolution shape by expanding each voxel of $Y_{\rm low}$ into a $d \times d\times d$ block of constant values. Then the sampling of $p(Y_{\rm high} | Y_{\rm low}; \theta)$ is similar to sampling the unconditioned model $p(Y_{\rm high}; \theta)$, except that for each step of the Langevin dynamics, let $\Delta Y$ be the change of $Y$, we update $Y \leftarrow Y +  (I - C^{-} C) \Delta Y$, i.e., we project $\Delta Y$ to the null space of $C$, so that  the low resolution version of $Y$, i.e., $CY$, remains fixed.    From this perspective, super resolution is similar to inpainting, except that the visible voxels are replaced by low resolution voxels. See Algorithm \ref{code:3D_Super_resolution} for a detailed description.

\subsection{Analyzing the learned 3D generator trained by MCMC teaching}

We evaluate a 3D generator trained by a generative VoxelNet via MCMC teaching. The generator network $g(Z; \alpha)$ has 4 layers of volumetric deconvolution with $4 \times 4 \times 4$ kernels, with up-sampling factors $\{1, 2, 2, 2\}$ at different layers respectively. The numbers of channels at different layers are 256, 128, 64, and 1. There is a fully connected layer under the 100 dimensional vector of latent factors $Z$. The output size is $32 \times 32 \times 32$. Batch normalization and ReLU layers are used between adjacent deconvolution layers and tanh non-linearity is added at the bottom layer. We train a generative VoxelNet with the above 3D generator as a sampler in a cooperative training scheme presented in Algorithm \ref{code:3} for the categories of toilet, sofa, and nightstand in ModelNet10 dataset independently. The generative VoxelNet has a 4-layer network, where the first layer has 64 $9 \times 9 \times 9$ filters, the second layer has 128 $7 \times 7 \times 7$ filters, the third layer has 256 $4 \times 4 \times 4$ filters, and the fourth layer is a fully connected layer with a single filter. The sub-sampling factors are $\{2, 2, 2, 1\}$. ReLU layers are added between adjacent convolutional layers. 
 
We use Adam for optimization of the generative VoxelNet model with $\beta_1=0.4$ and $ \beta_2=0.999$, and 3D generator with $\beta_1=0.6$ and $\beta_2=0.999$. The learning rates for generative VoxelNet and 3D generator are 0.001 and 0.0003 respectively. The number of parallel chains is 50, and the mini-batch size is 50. The training data are scaled into the range of $[-1, 1]$. The synthesized data are re-scaled back into $[0, 1]$ for visualization. We run $K=20$ steps of Langevin dynamics with a step size $\Delta \tau=0.09$. Figure \ref{fig:synthesis_generator} shows some examples of 3D objects generated by the 3D generators trained by the generative VoxelNet via MCMC teaching on categories toilet and sofa. 
 
\begin{figure}[h]
	\centering
	\includegraphics[height=.139\linewidth]{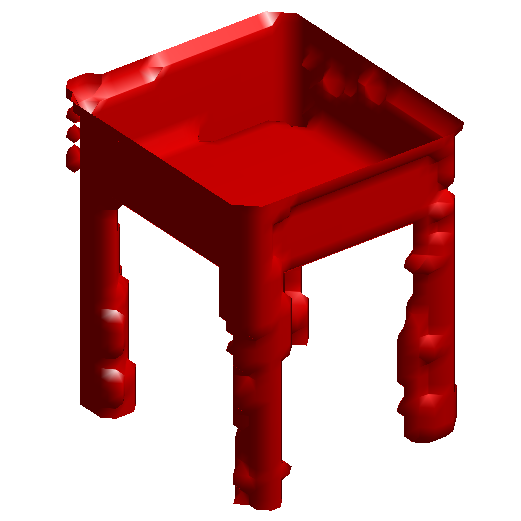} 
	\includegraphics[height=.139\linewidth]{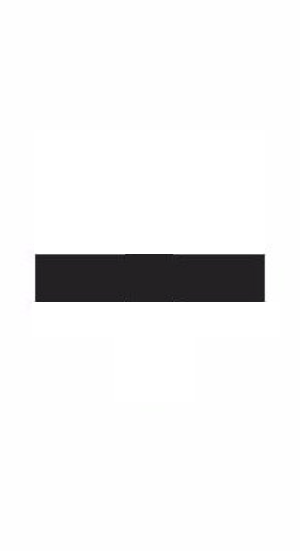} 
	\includegraphics[height=.139\linewidth]{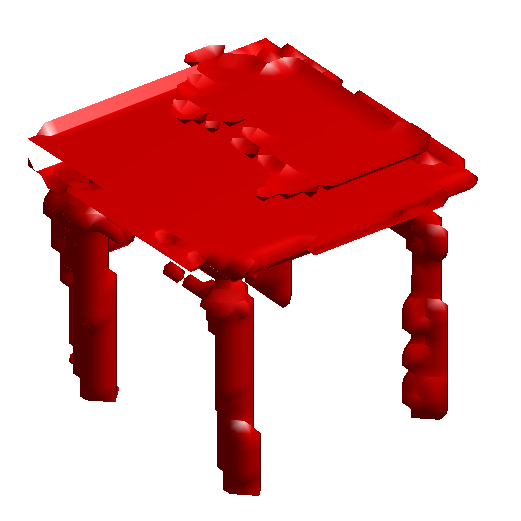} \hspace{-2mm}
	\includegraphics[height=.139\linewidth]{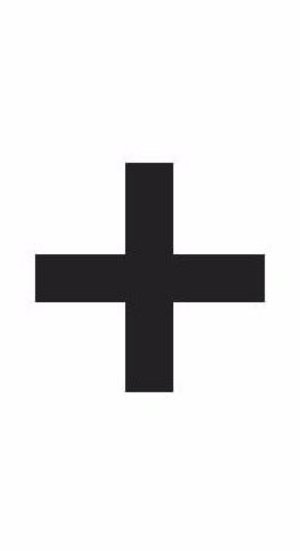} 
	\includegraphics[height=.139\linewidth]{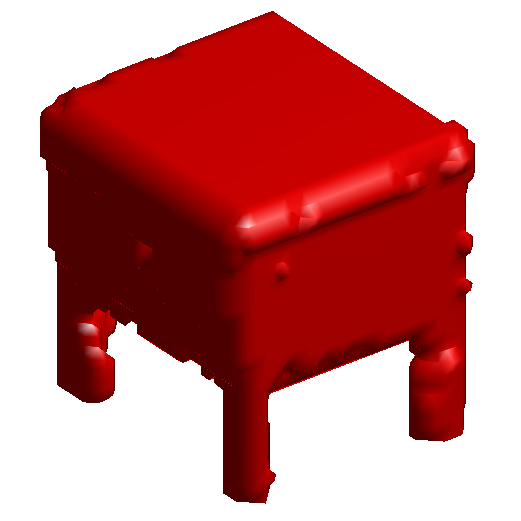} 
	\includegraphics[height=.139\linewidth]{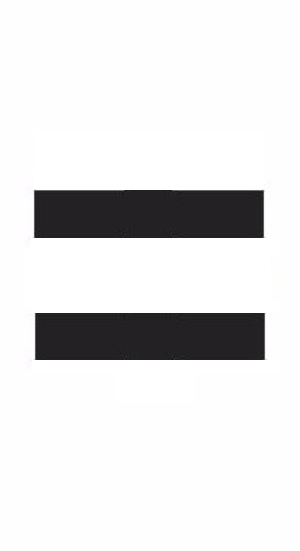} 
	\includegraphics[height=.139\linewidth]{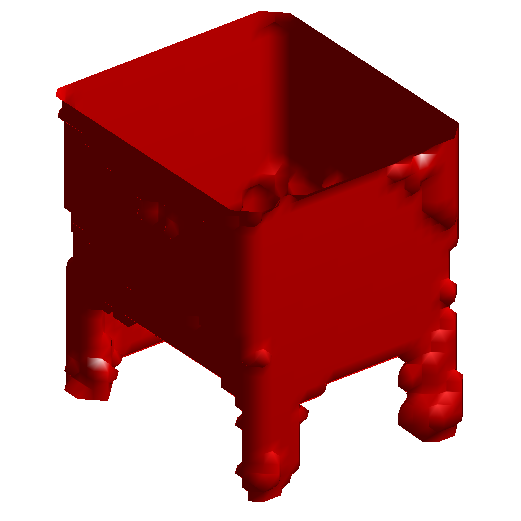} \hspace{-3mm}	\\
	\includegraphics[height=.139\linewidth]{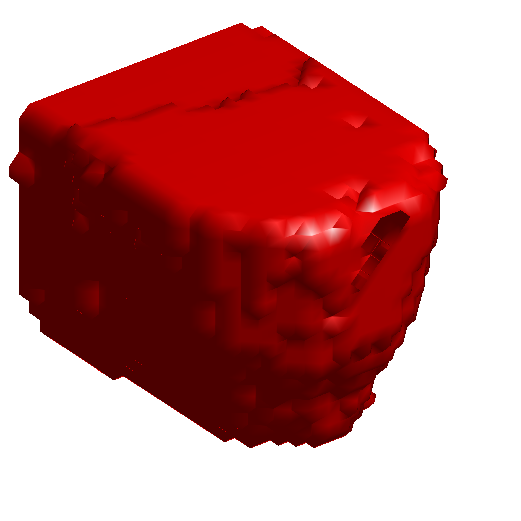} \hspace{-3mm}
	\includegraphics[height=.139\linewidth]{./figures/output_arith/minus.png} 
	\includegraphics[height=.139\linewidth]{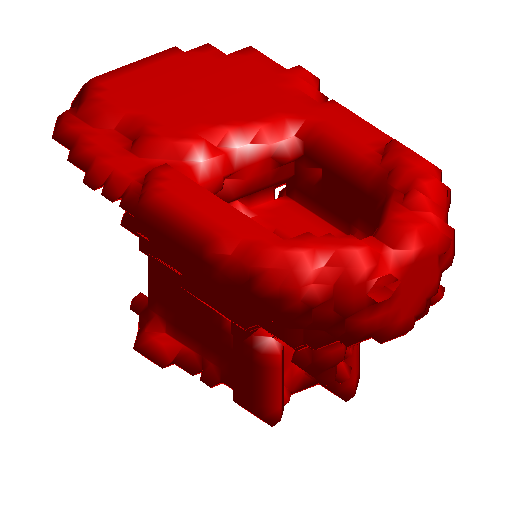} \hspace{-2mm}
	\includegraphics[height=.139\linewidth]{./figures/output_arith/plus.png} 
	\includegraphics[height=.139\linewidth]{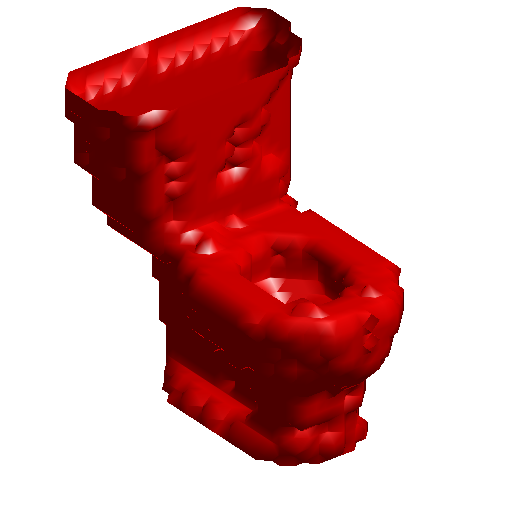} 
	\includegraphics[height=.139\linewidth]{./figures/output_arith/equal.png} 
	\includegraphics[height=.139\linewidth]{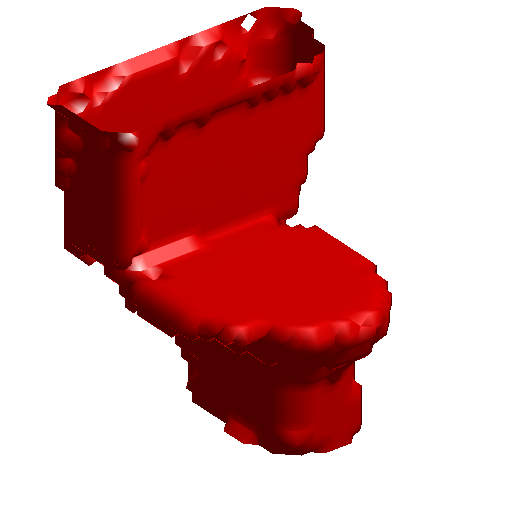}  \hspace{-3mm}
	\caption{3D shape arithmetic in the latent space. Each row of images show that the obtained ``part'' vector can be added to a new object. The latent space is of 100 dimensions. The size of 3D shape is $32 \times 32 \times 32$ voxels. }
	\label{fig:arith}
\end{figure}

We show results of interpolating between two latent vectors of $Z$ in Figure \ref{fig:interpolation}. For each row in each category, the 3D objects at the two ends are generated from $Z$ vectors that are randomly sampled from $\mathcal{N}(0,I_d)$. Each object in the middle is obtained by first interpolating the $Z$ vectors of the two end objects, and then generating the objects using the 3D generator. We observe smooth transitions in 3D shape structures and that most intermediate objects are also physically plausible.
This experiment demonstrates that the learned 3D generator embeds the 3D object distribution into a smooth low dimensional manifold. Another way to investigate the learned 3D generator is to show shape arithmetic in the latent space. As shown in Figure \ref{fig:arith}, the 3D generator is able to encode semantic knowledge of 3D shapes in its latent space such that arithmetic can be performed on $Z$ vectors for visual concept manipulation of 3D shapes. For example, the second row of images show the obtained ``toilet lid'' vector can be added to another toilet.

\subsection{3D object classification}

We evaluate the feature maps learned by our generative VoxelNet model. We perform a classification experiment on ModelNet10 dataset. We first train a single model on all categories of the training set in an unsupervised manner. The network architecture and learning configuration are the same as the one used for synthesis in Section \ref{Exp:objectSynthesis}. 
Then we use the model as a feature extractor. Specifically, for each input 3D object, we use the model to extract its first and second layers of feature maps, apply max pooling of kernel sizes $4 \times 4 \times 4$ and $2 \times 2\times 2$ respectively, and concatenate the outputs as a feature vector of length 8,100. We train a multinomial logistic regression classifier using labeled data based on the extracted feature vectors. We evaluate the classification accuracy of the classifier on the testing data using the one-versus-all rule. For the purpose of comparison, Table \ref{exp:classification} lists 12 published results on this dataset obtained by other baseline methods. Our method outperforms the other methods in terms of classification accuracy on this dataset.
 
\begin{table}[h]
\caption{3D object classification on ModelNet10 dataset}\label{exp:classification}
\centering
\begin{tabular}{|l|c|}
\hline
          Method   & Accuracy\\ \hline \hline
     Geometry Image \cite{sinha2016deep} & 88.4$\%$\\ \hline      
      
      PANORAMA-NN    \cite{sfikas2017exploiting} & 91.1$\%$ \\ \hline
    ECC  \cite{simonovsky2017dynamic} & 90.0$\%$\\ \hline    
       3D ShapeNets \cite{wu20153d}  & 83.5$\%$ \\ \hline
        DeepPano \cite{shi2015deeppano} & 85.5$\%$ \\ \hline
 SPH  \cite{kazhdan2003rotation}     & 79.8$\%$    \\ \hline
  LFD  \cite{chen2003visual}  & 79.9$\%$  \\ \hline
 VConv-DAE   \cite{sharma2016vconv}   & 80.5$\%$   \\ \hline
 VoxNet \cite{maturana2015voxnet} &92.0$\%$ \\ \hline
 3D-GAN  \cite{3dgan}  & 91.0$\%$  \\ 
 \hline
 3D-WINN  \cite{huang20193d}  & 91.9$\%$  \\ 
  \hline
 Primitive GAN  \cite{khan2019unsupervised}  & 92.2$\%$  \\ \hline
  generative VoxelNet (ours) & \textbf{92.4}$\%$\\ \hline 
\end{tabular}
\end{table}
\vspace{-3mm}

\subsection{3D multi-grid modeling and sampling}
\begin{figure}
\centering
\hspace{0.6mm}
\rotatebox[origin=l]{90}{\hspace{2mm}\textbf{{\tiny 16$\times$16$\times$16}}}
\includegraphics[trim=40 0 40 0, clip, height=.17\linewidth]{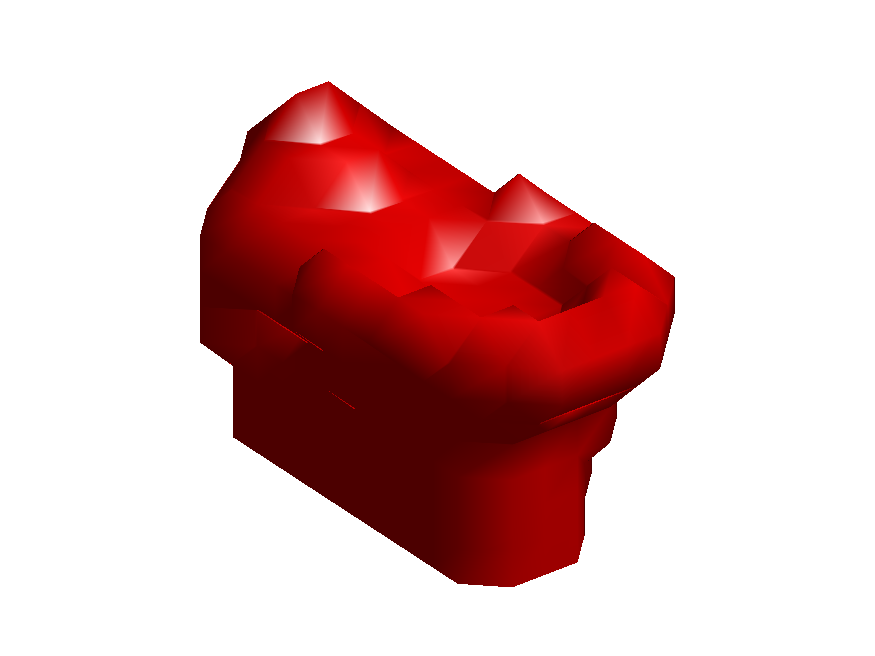} 
\includegraphics[trim=40 0 40 0, clip, height=.17\linewidth]{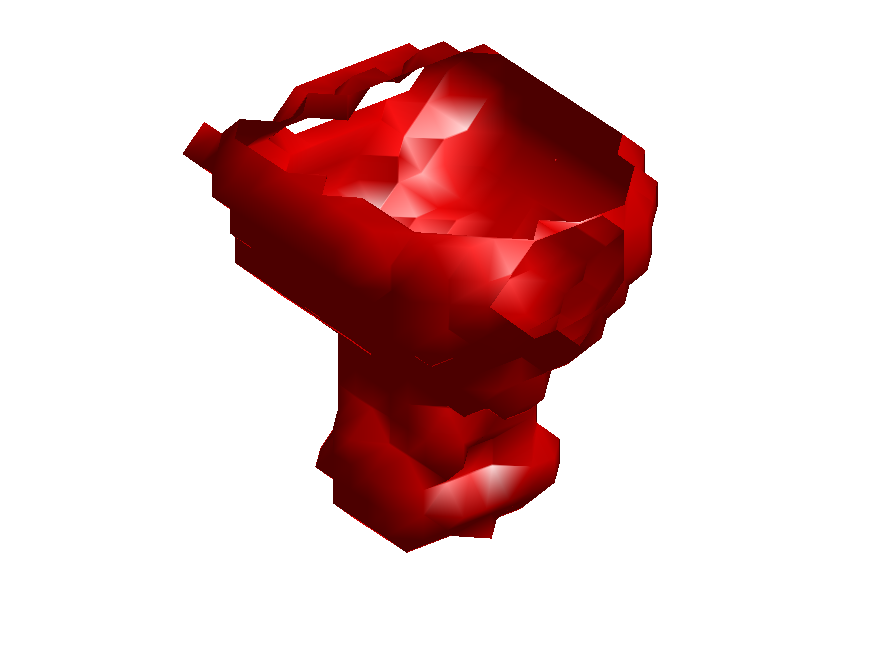}
\includegraphics[trim=40 0 40 0, clip, height=.17\linewidth]{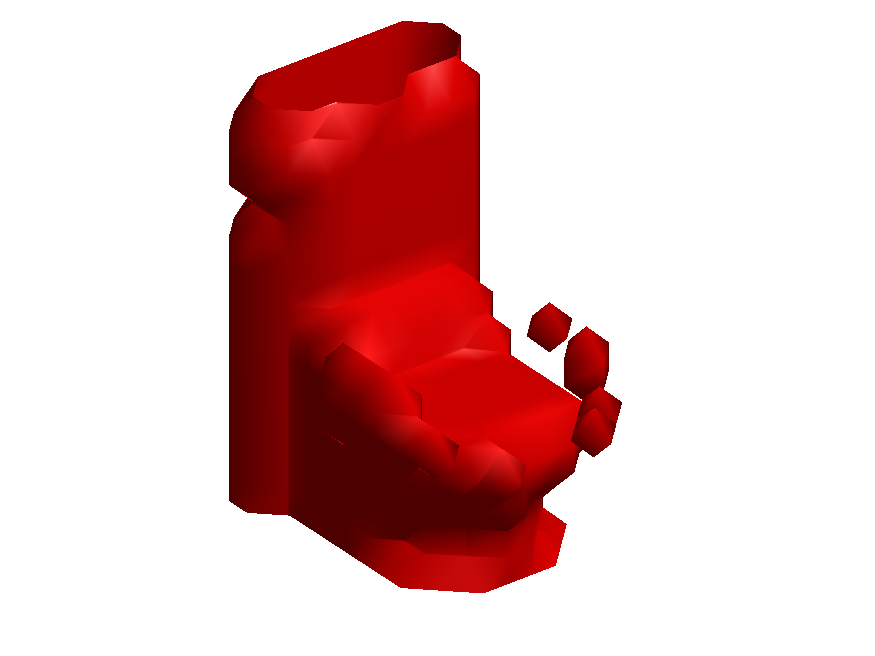}
\includegraphics[trim=40 0 40 0, clip, height=.17\linewidth]{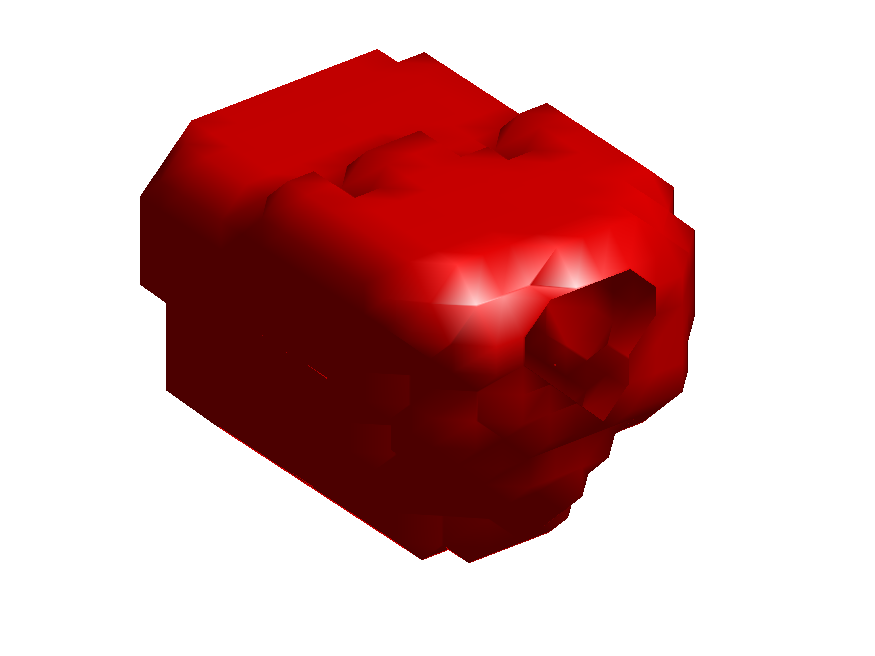}
\includegraphics[trim=40 0 40 0, clip, height=.17\linewidth]{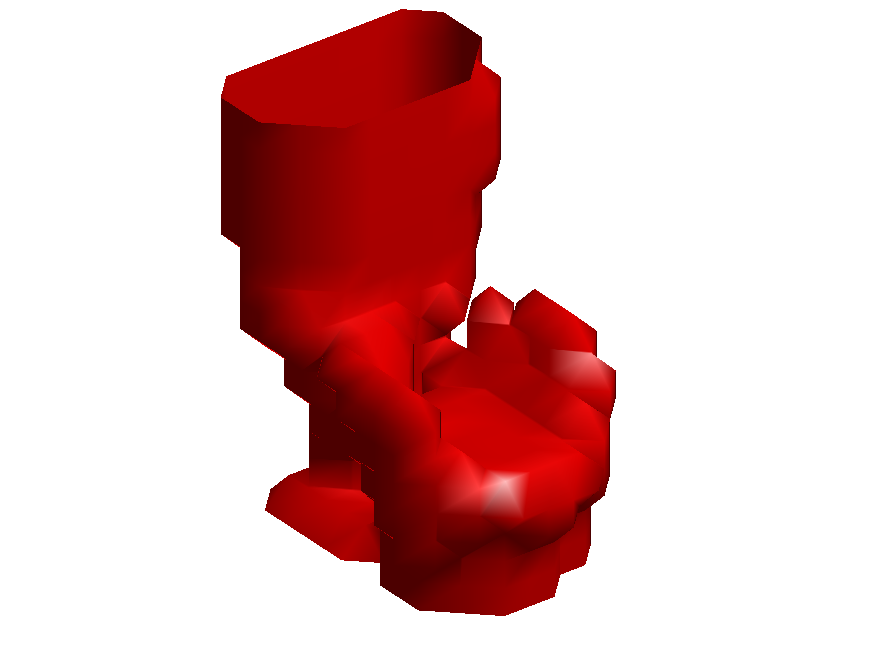}

\hspace{0.6mm}
\rotatebox[origin=l]{90}{\hspace{2mm}\textbf{{\tiny 32$\times$32$\times$32}}}
\includegraphics[trim=40 0 40 0, clip, height=.17\linewidth]{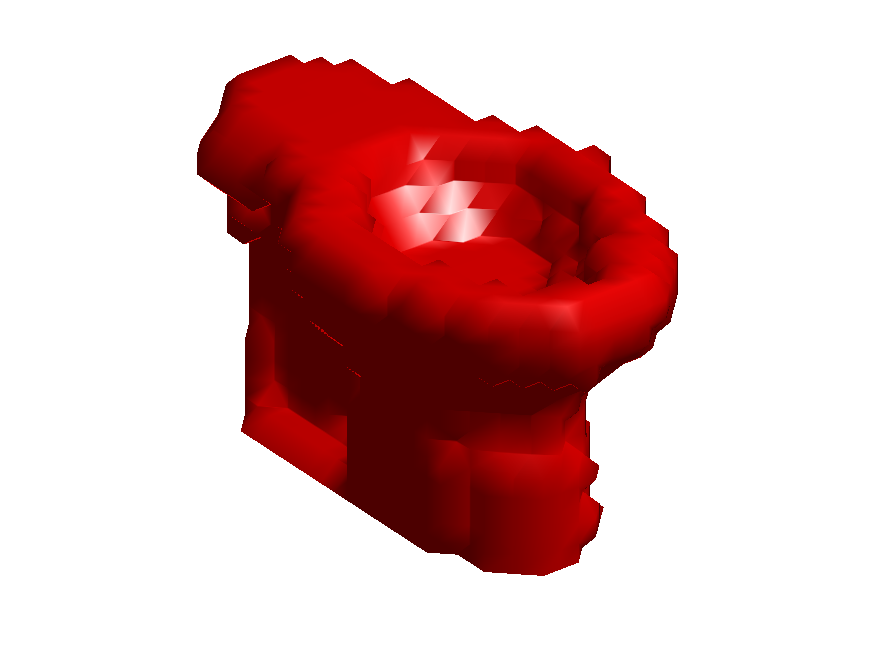} 
\includegraphics[trim=40 0 40 0, clip, height=.17\linewidth]{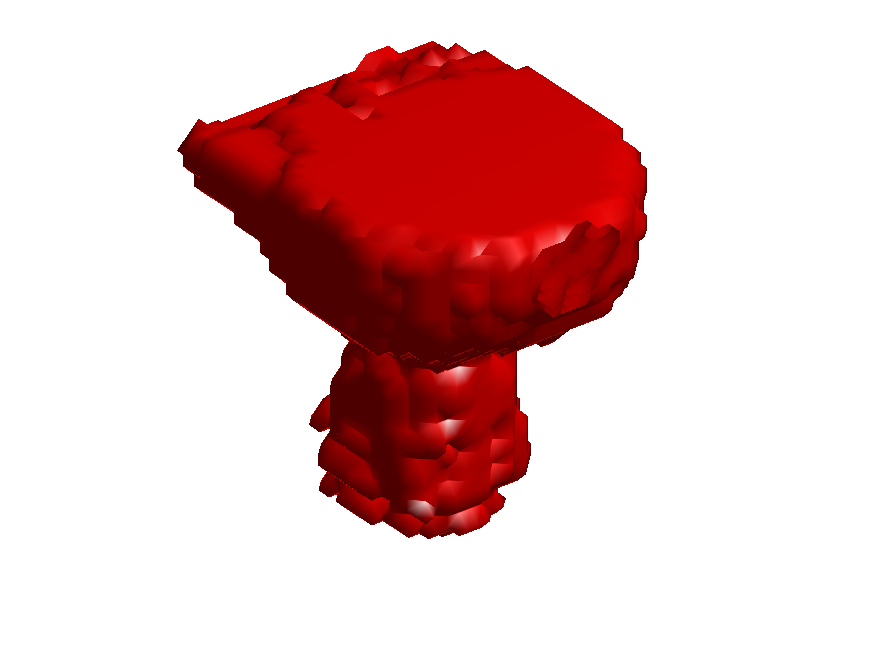}
\includegraphics[trim=40 0 40 0, clip, height=.17\linewidth]{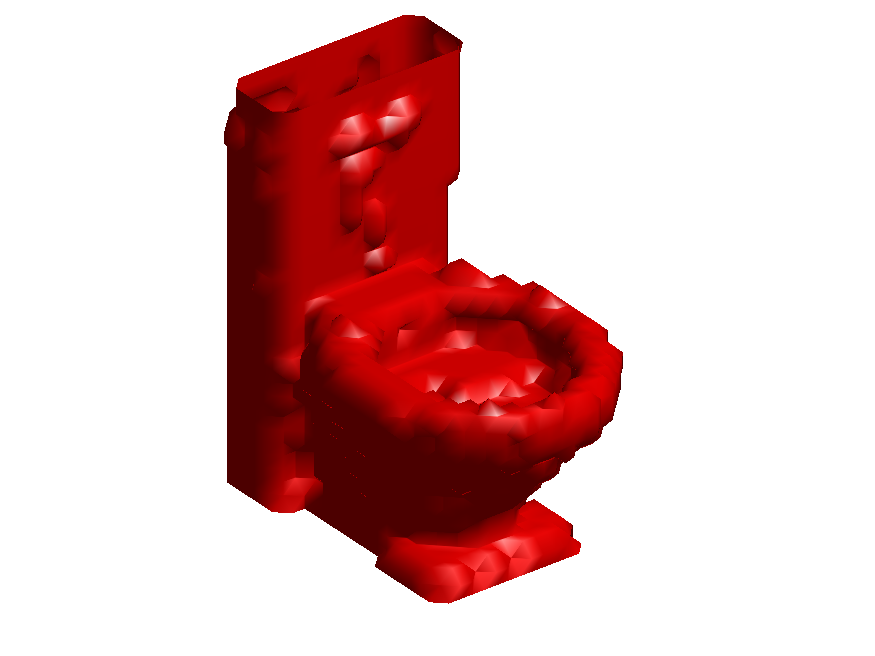}
\includegraphics[trim=40 0 40 0, clip, height=.17\linewidth]{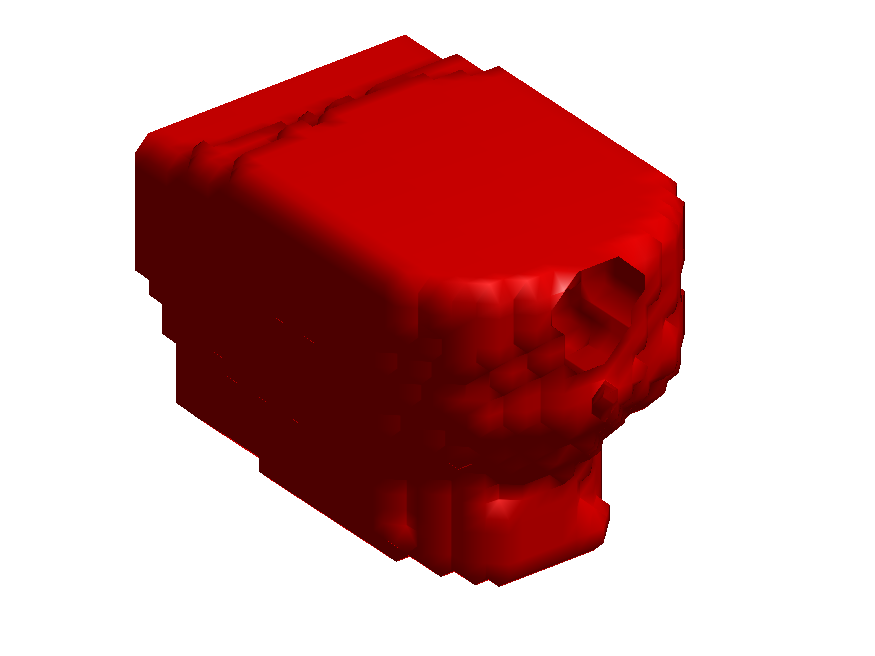}
\includegraphics[trim=40 0 40 0, clip, height=.17\linewidth]{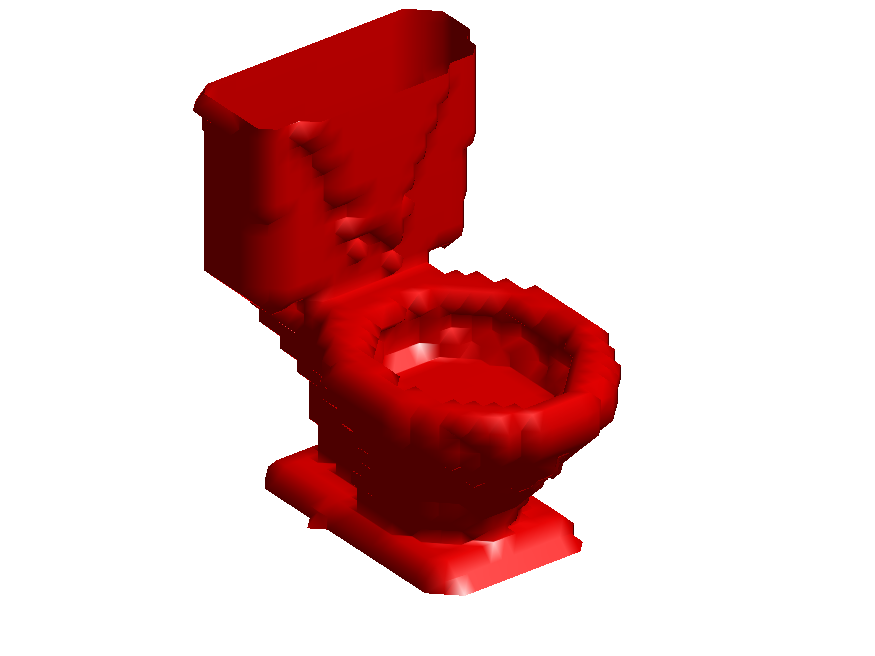}

\hspace{0.6mm}
\rotatebox[origin=l]{90}{\hspace{2mm}\textbf{{\tiny 64$\times$64$\times$64}}}
\includegraphics[trim=40 0 40 0, clip, height=.17\linewidth]{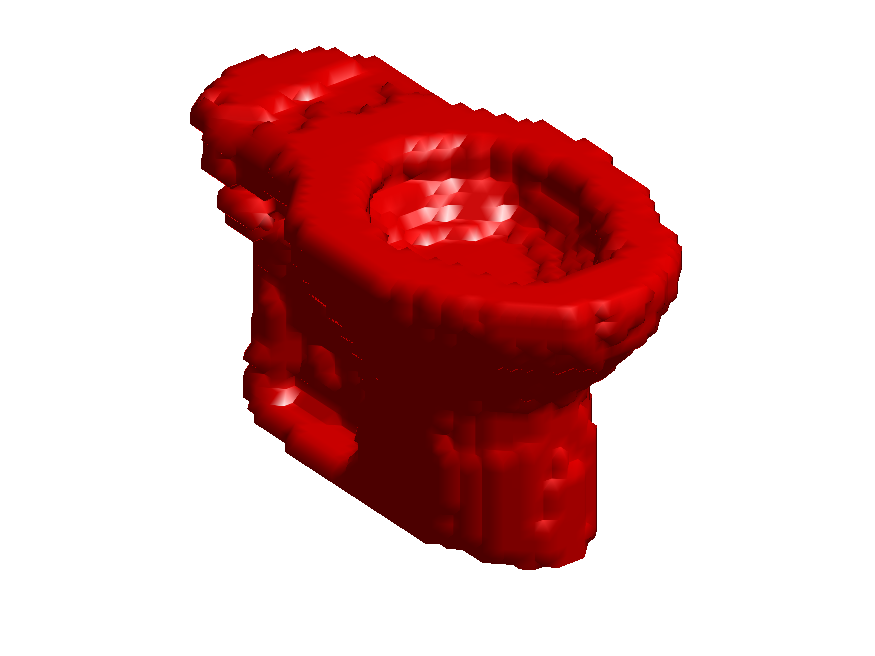} 
\includegraphics[trim=40 0 40 0, clip, height=.17\linewidth]{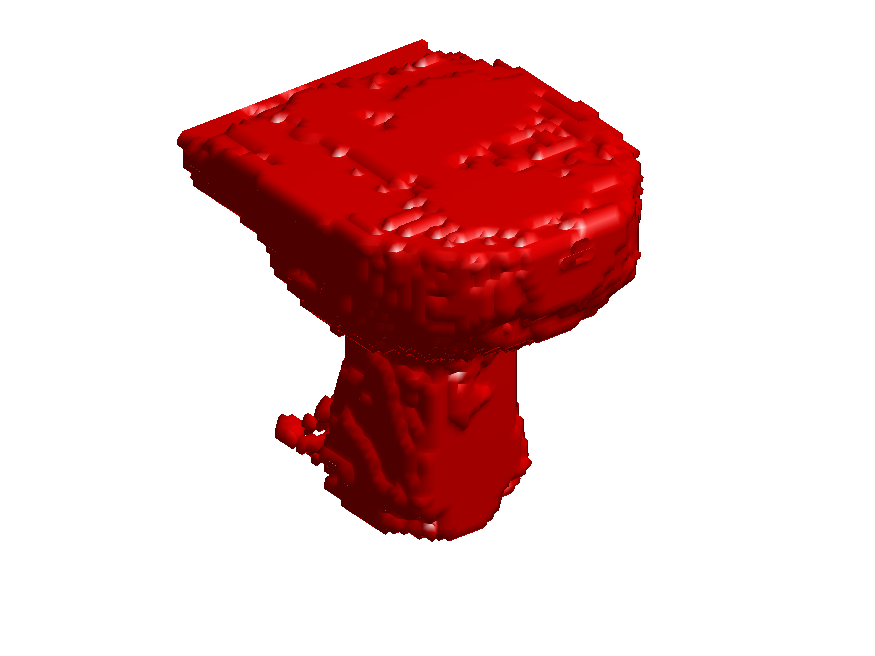}
\includegraphics[trim=40 0 40 0, clip, height=.17\linewidth]{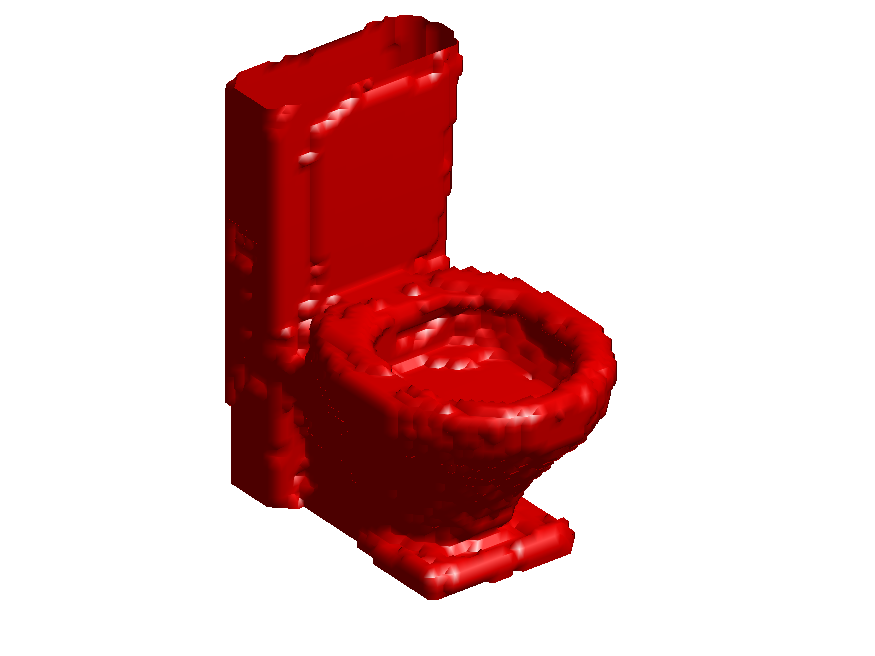}
\includegraphics[trim=40 0 40 0, clip, height=.17\linewidth]{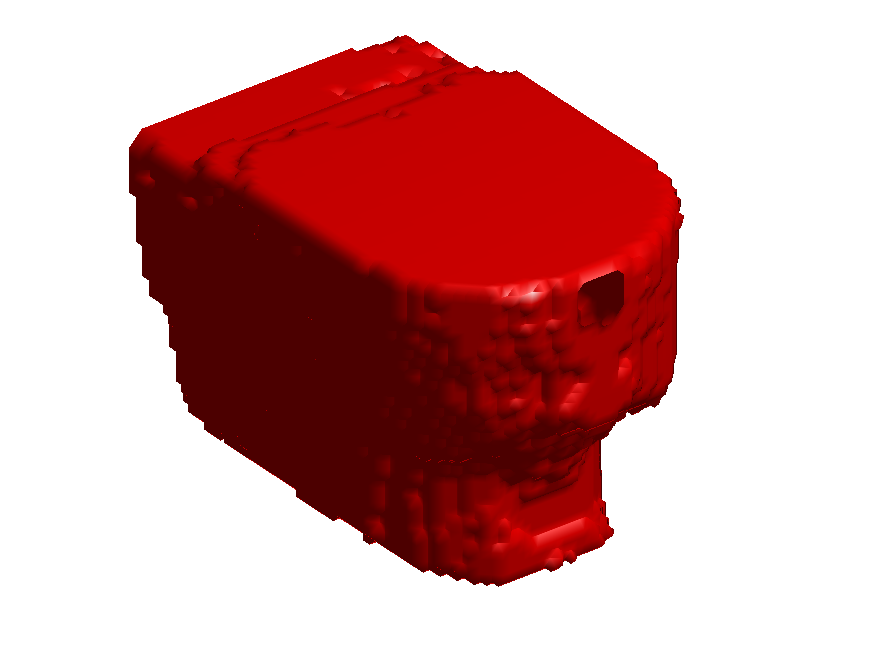}
\includegraphics[trim=40 0 40 0, clip, height=.17\linewidth]{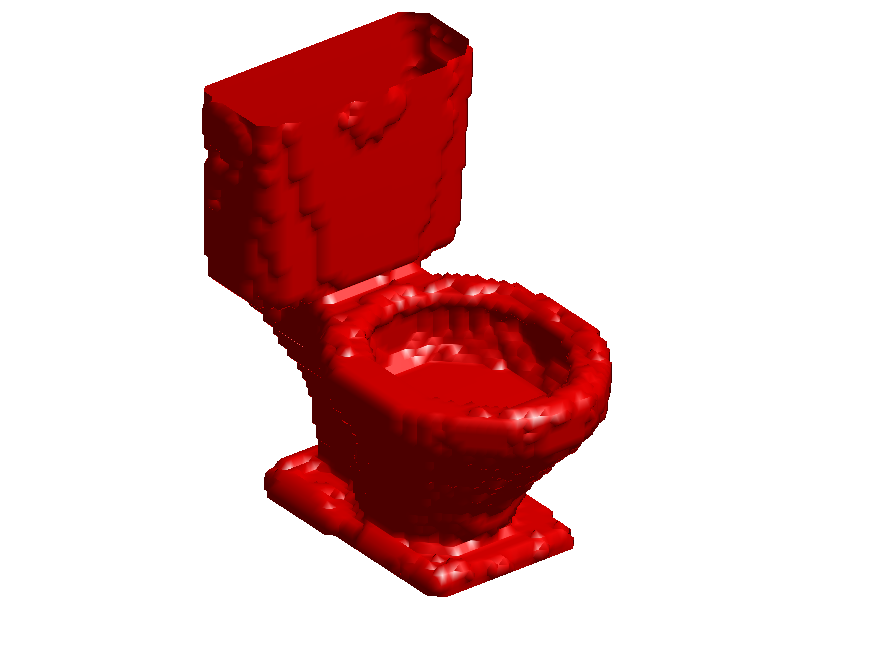}

\hspace{0.6mm}
\rotatebox[origin=l]{90}{\hspace{2mm}\textbf{{\tiny 128$\times$128$\times$128}}}
\includegraphics[trim=40 0 40 0, clip, height=.17\linewidth]{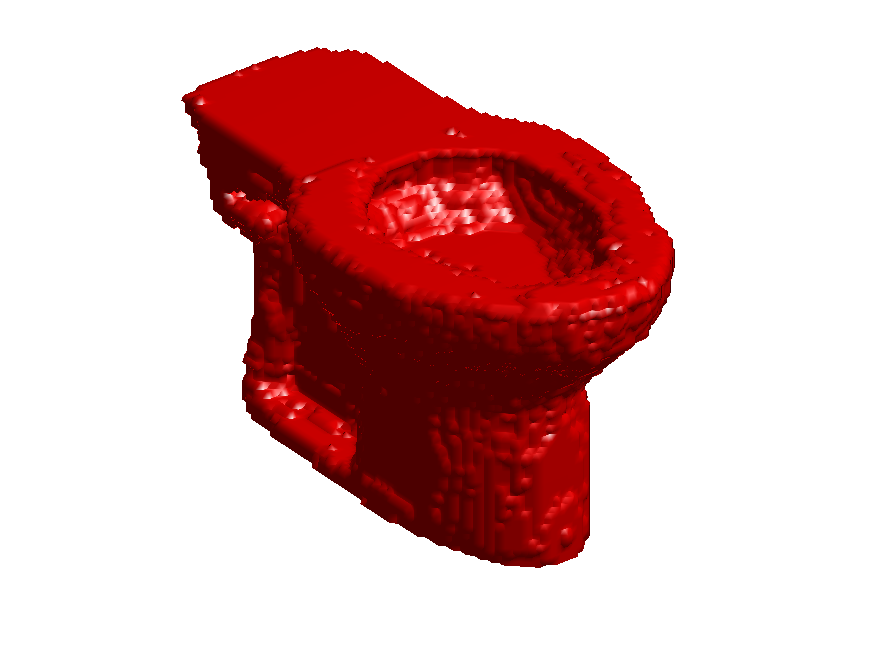} 
\includegraphics[trim=40 0 40 0, clip, height=.17\linewidth]{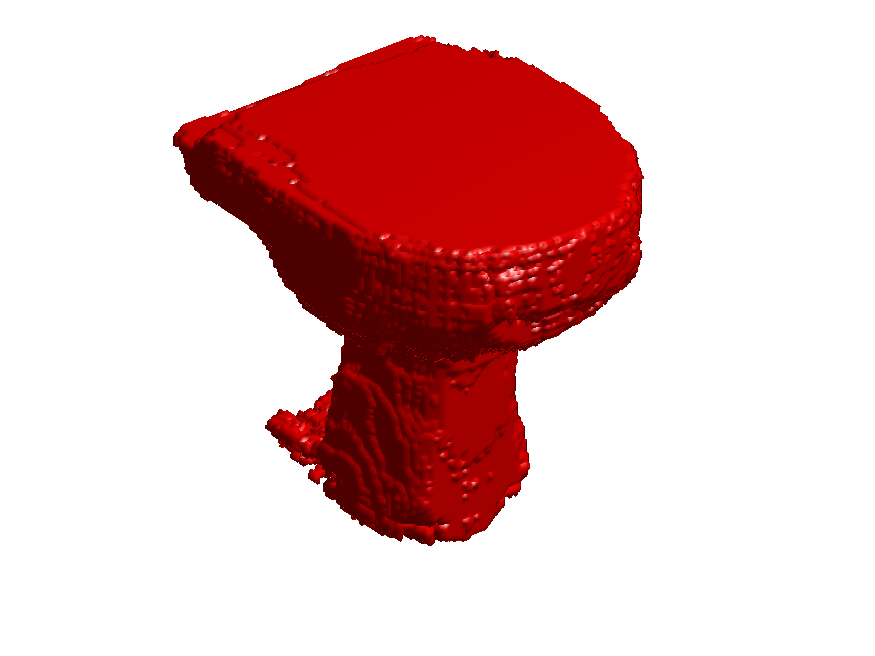}
\includegraphics[trim=40 0 40 0, clip, height=.17\linewidth]{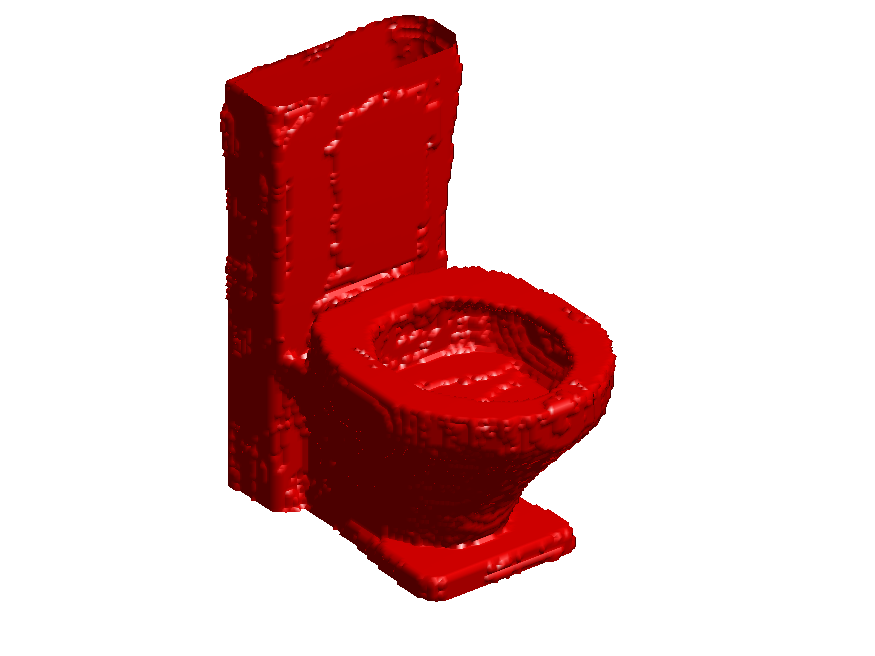}
\includegraphics[trim=40 0 40 0, clip, height=.17\linewidth]{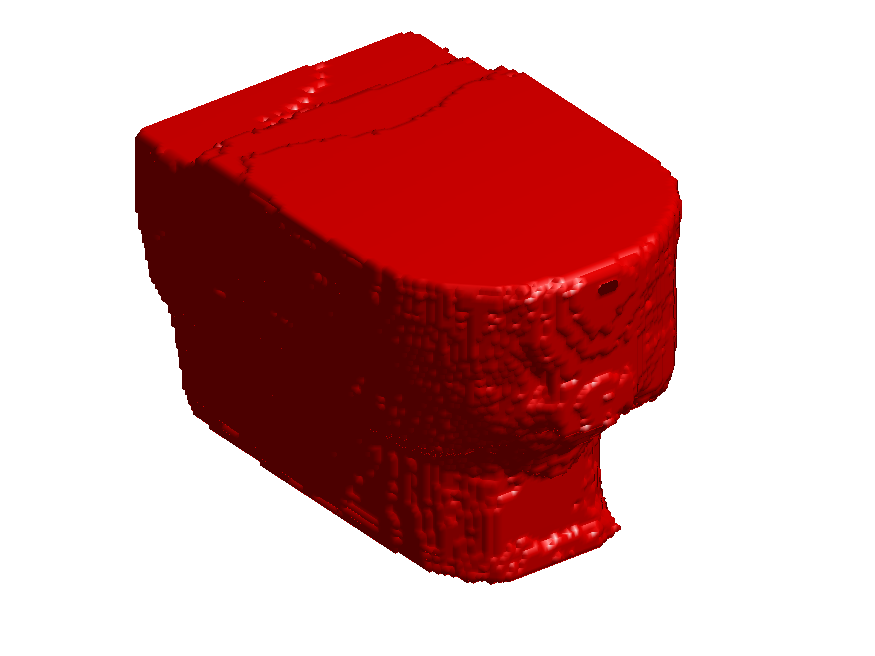}
\includegraphics[trim=40 0 40 0, clip, height=.17\linewidth]{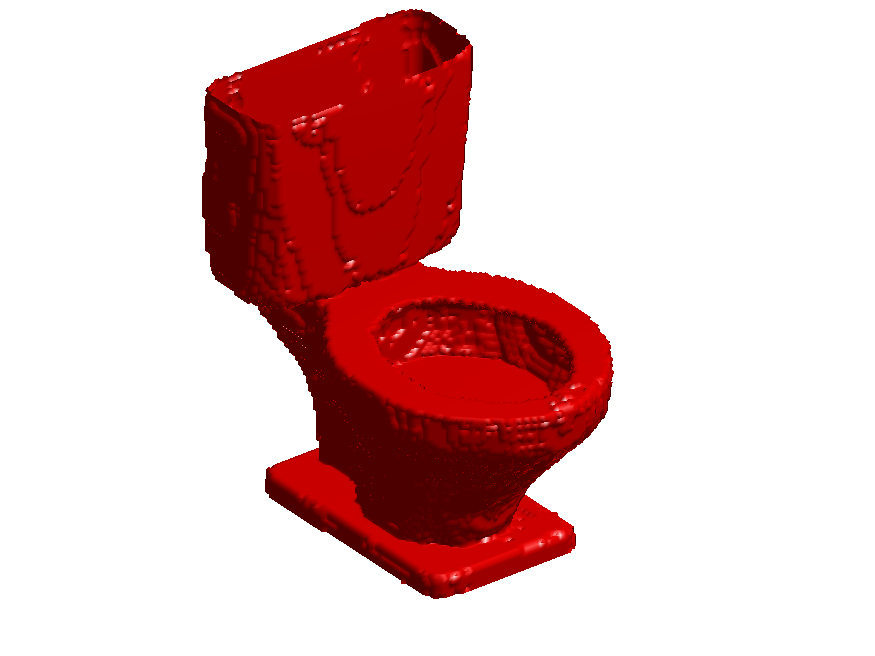}\\
(a) toilet \\ \vspace{2mm}
\rotatebox[origin=l]{90}{\hspace{2mm}\textbf{{\tiny 16$\times$16$\times$16}}}
\includegraphics[trim=25 0 25 0, clip, height=.15\linewidth]{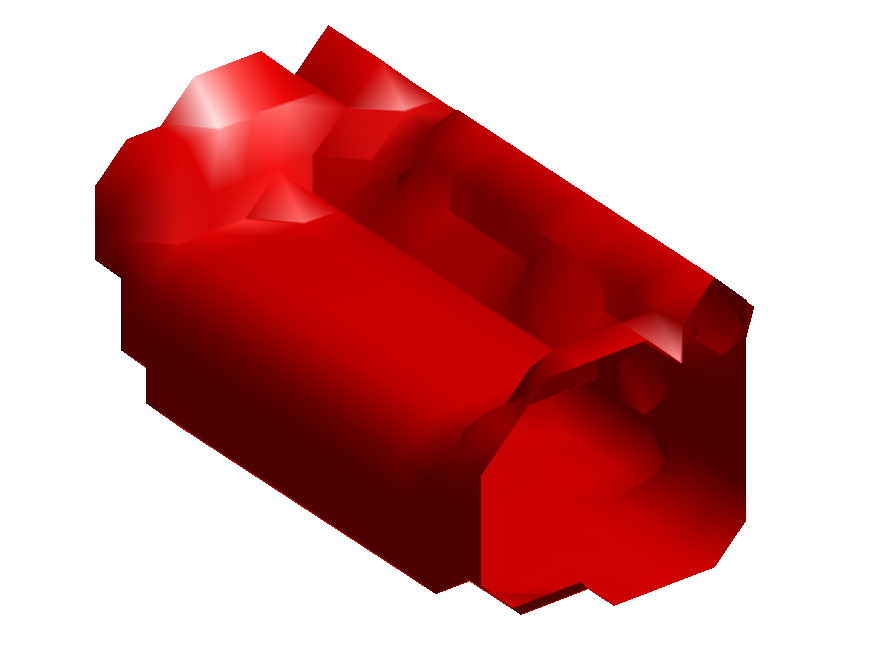} 
\includegraphics[trim=25 0 25 0, clip, height=.15\linewidth]{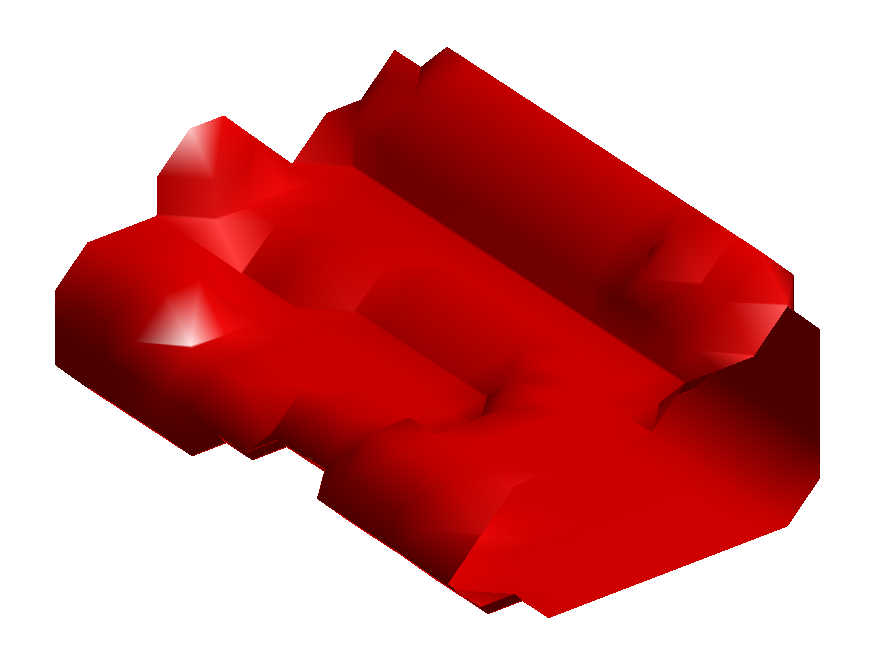}
\includegraphics[trim=25 0 25 0, clip, height=.15\linewidth]{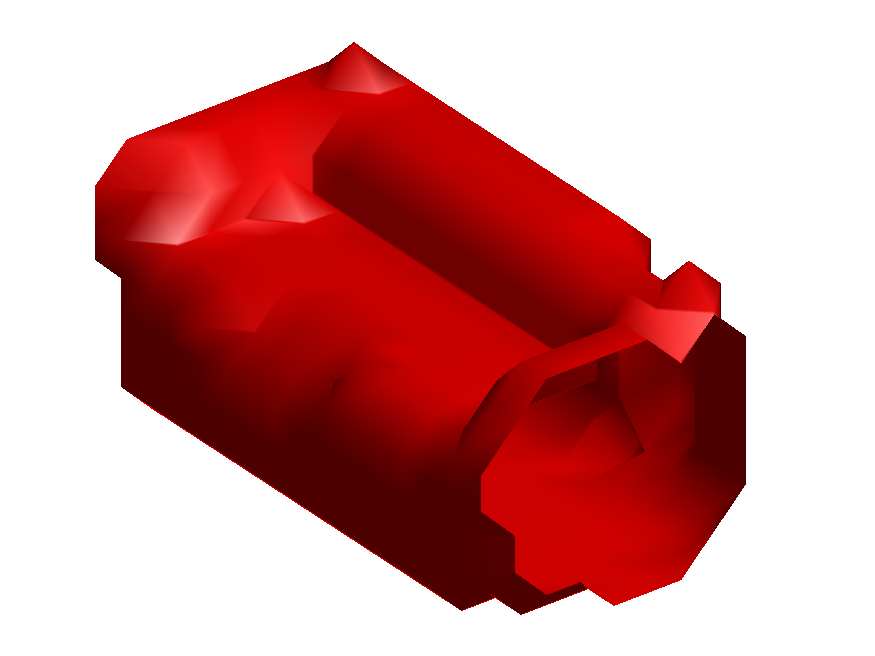}
\includegraphics[trim=25 0 25 0, clip, height=.15\linewidth]{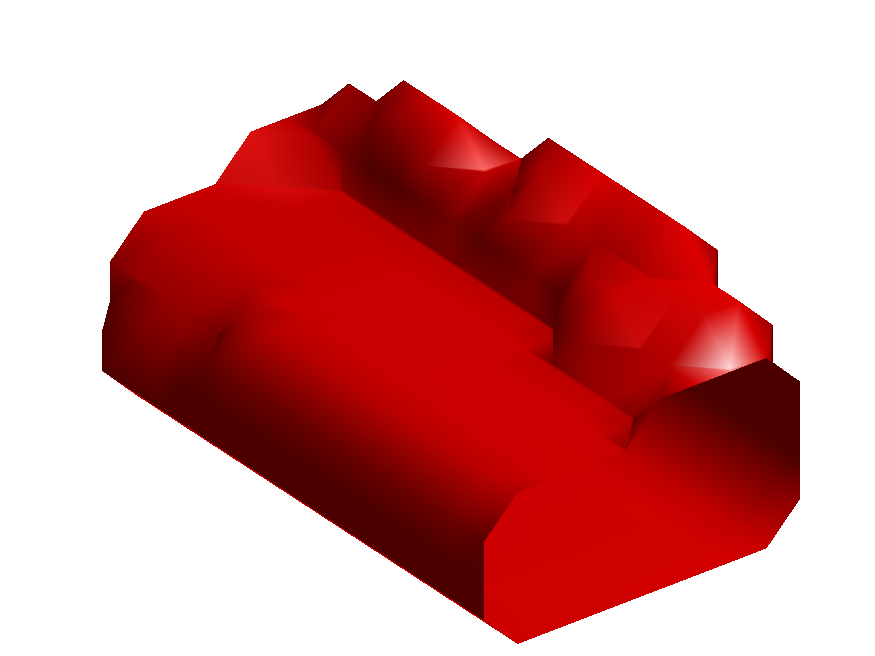}
\includegraphics[trim=25 0 25 0, clip, height=.15\linewidth]{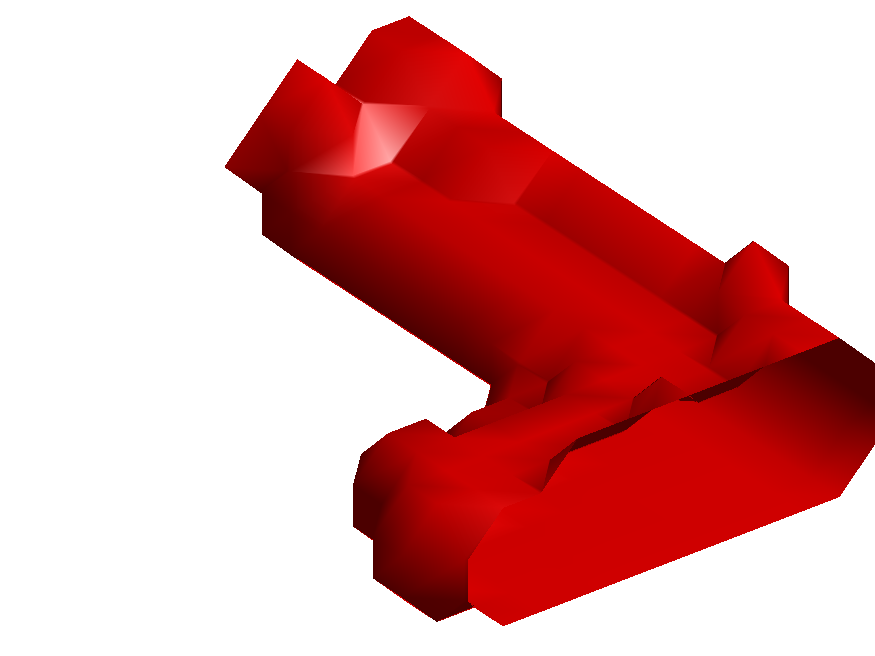}\\

\rotatebox[origin=l]{90}{\hspace{2mm}\textbf{{\tiny 32$\times$32$\times$32}}}
\includegraphics[trim=25 0 25 0, clip, height=.15\linewidth]{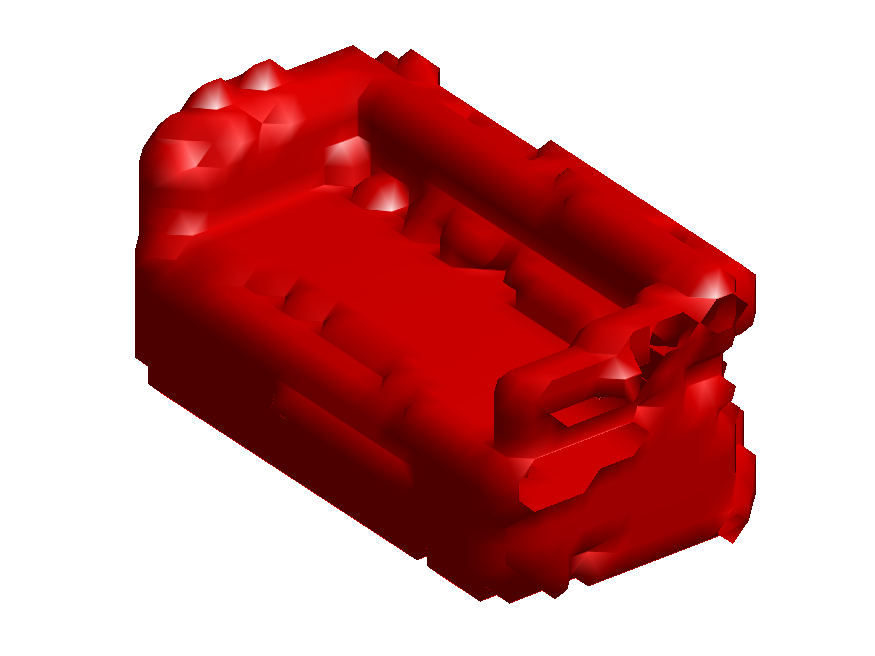} 
\includegraphics[trim=25 0 25 0, clip, height=.15\linewidth]{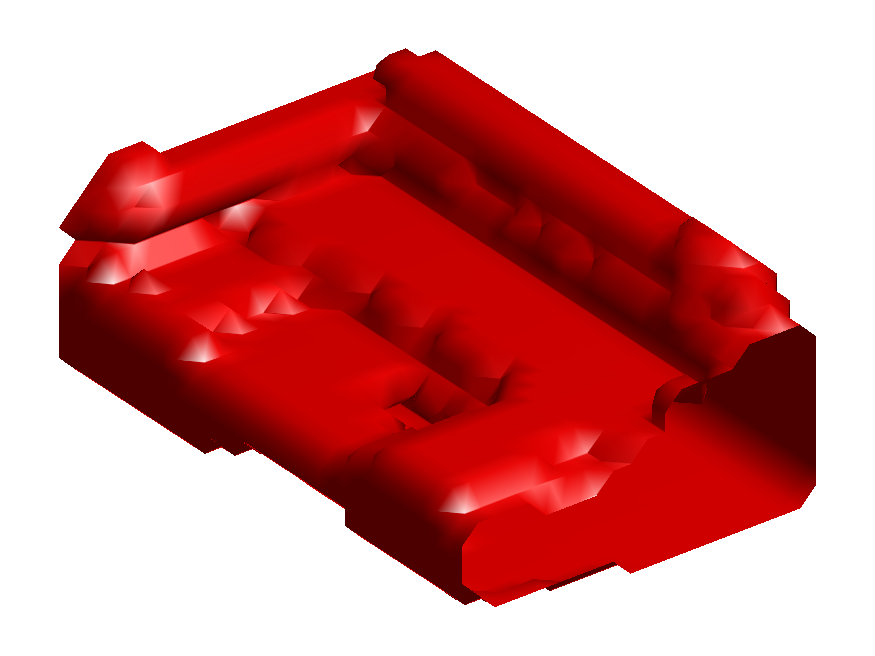}
\includegraphics[trim=25 0 25 0, clip, height=.15\linewidth]{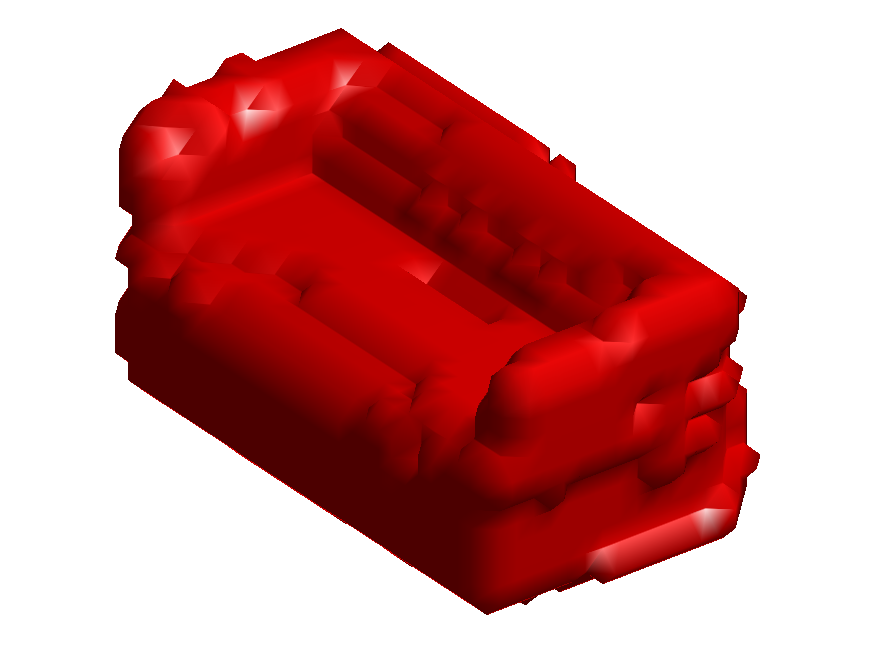}
\includegraphics[trim=25 0 25 0, clip, height=.15\linewidth]{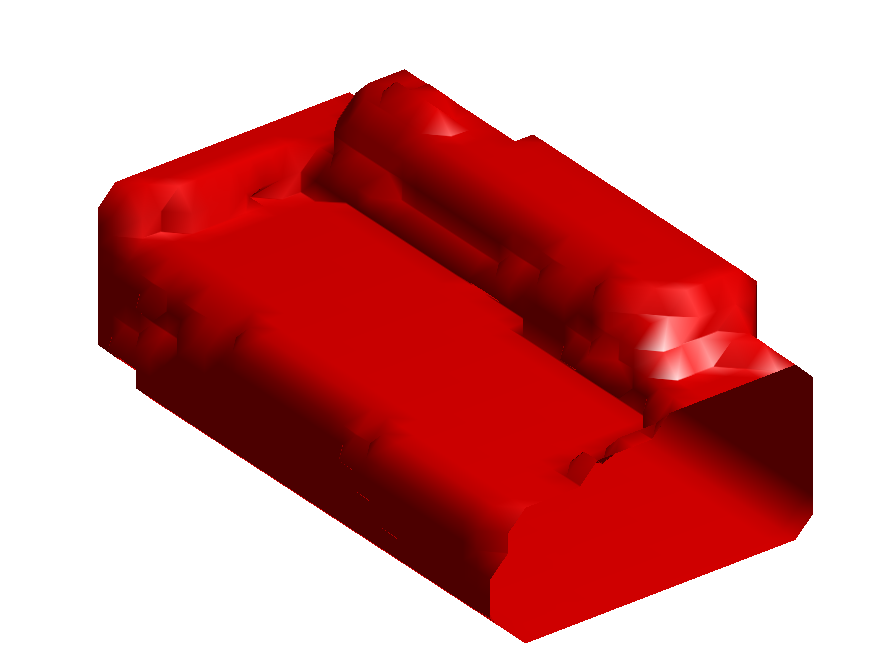}
\includegraphics[trim=25 0 25 0, clip, height=.15\linewidth]{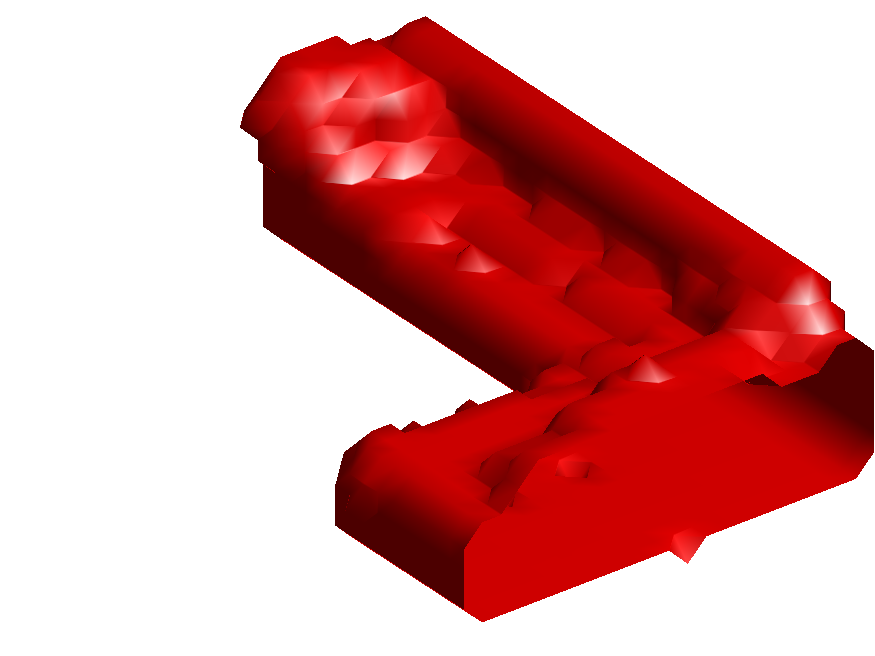}\\

\rotatebox[origin=l]{90}{\hspace{2mm}\textbf{{\tiny 64$\times$64$\times$64}}}
\includegraphics[trim=25 0 25 0, clip, height=.15\linewidth]{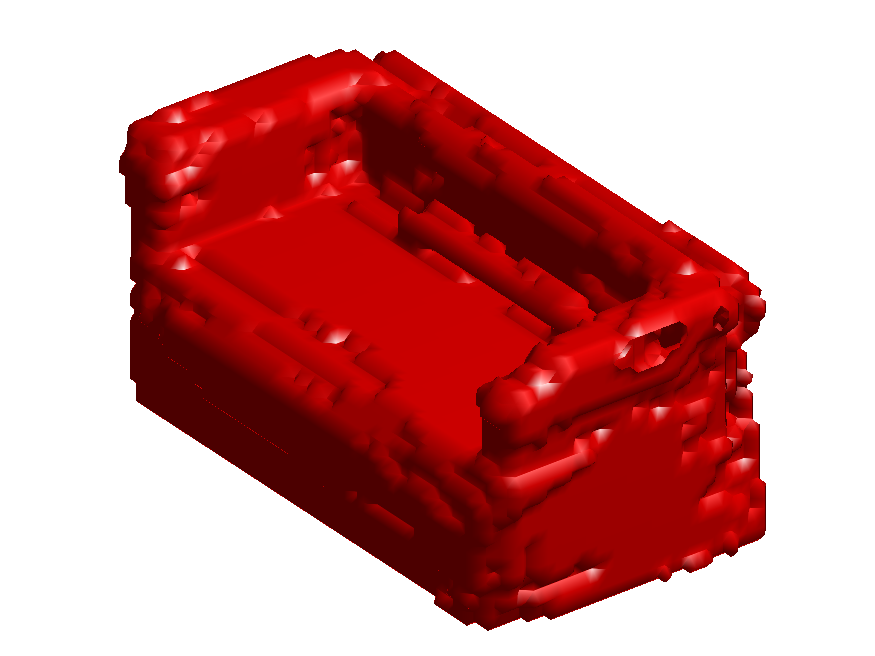} 
\includegraphics[trim=25 0 25 0, clip, height=.15\linewidth]{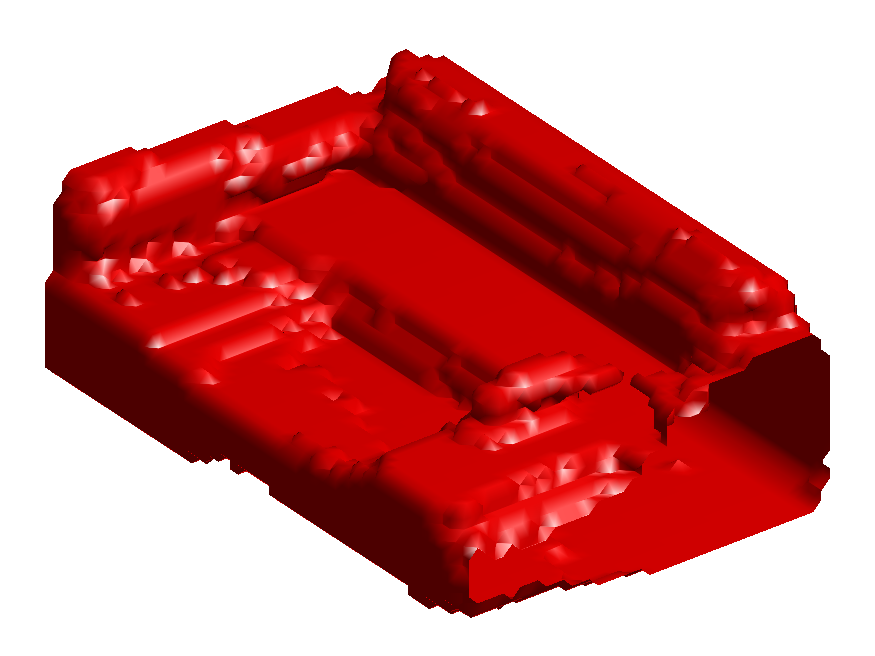}
\includegraphics[trim=25 0 25 0, clip, height=.15\linewidth]{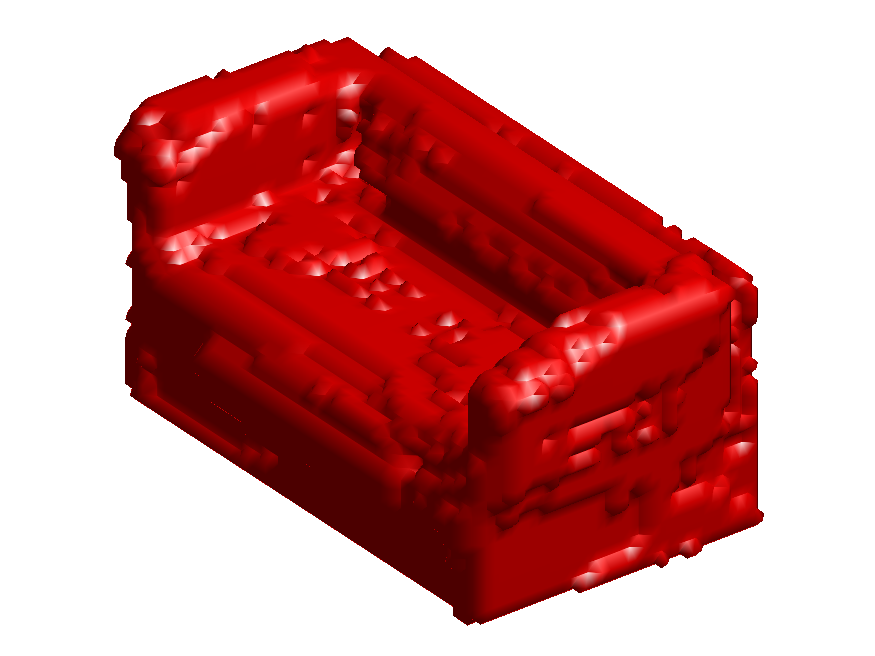}
\includegraphics[trim=25 0 25 0, clip, height=.15\linewidth]{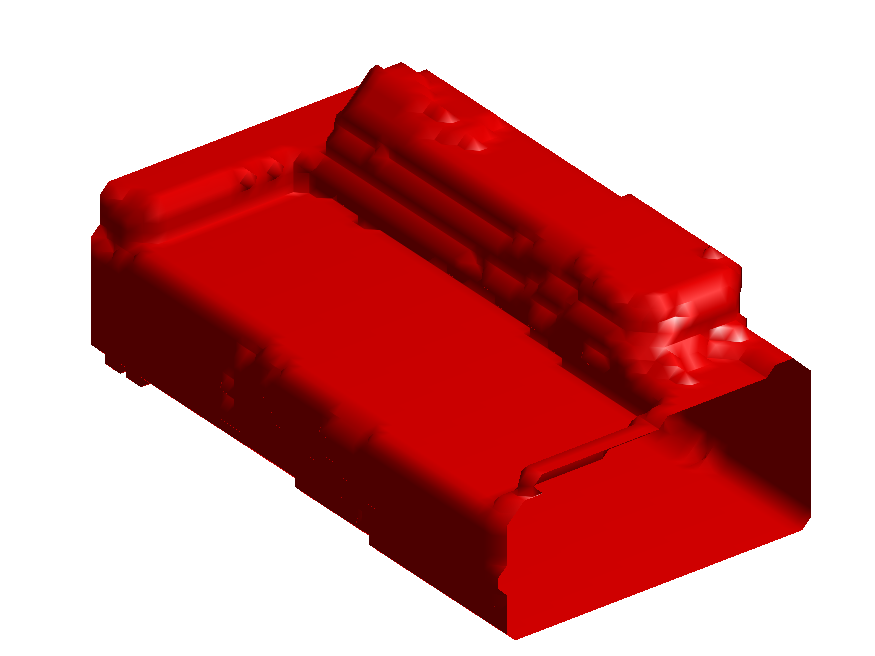}
\includegraphics[trim=25 0 25 0, clip, height=.15\linewidth]{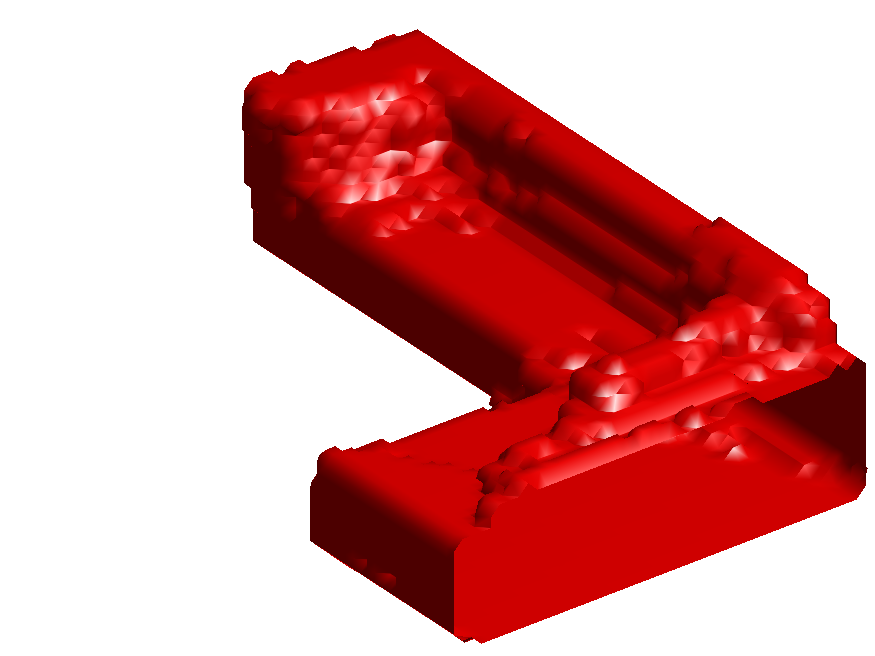}\\

\rotatebox[origin=l]{90}{\textbf{{\tiny 128$\times$128$\times$128}}}
\includegraphics[trim=25 0 25 0, clip, height=.15\linewidth]{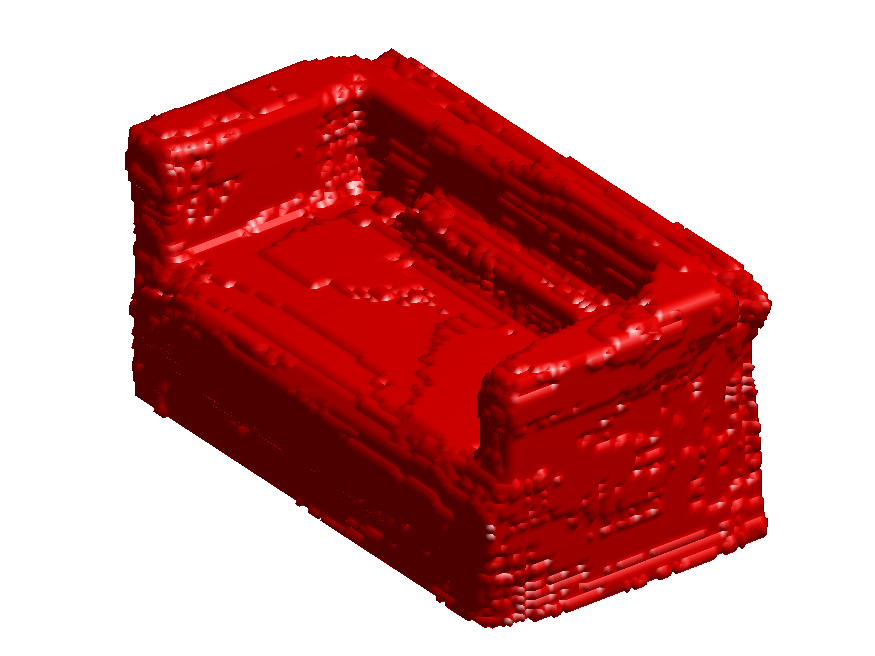} 
\includegraphics[trim=25 0 25 0, clip, height=.15\linewidth]{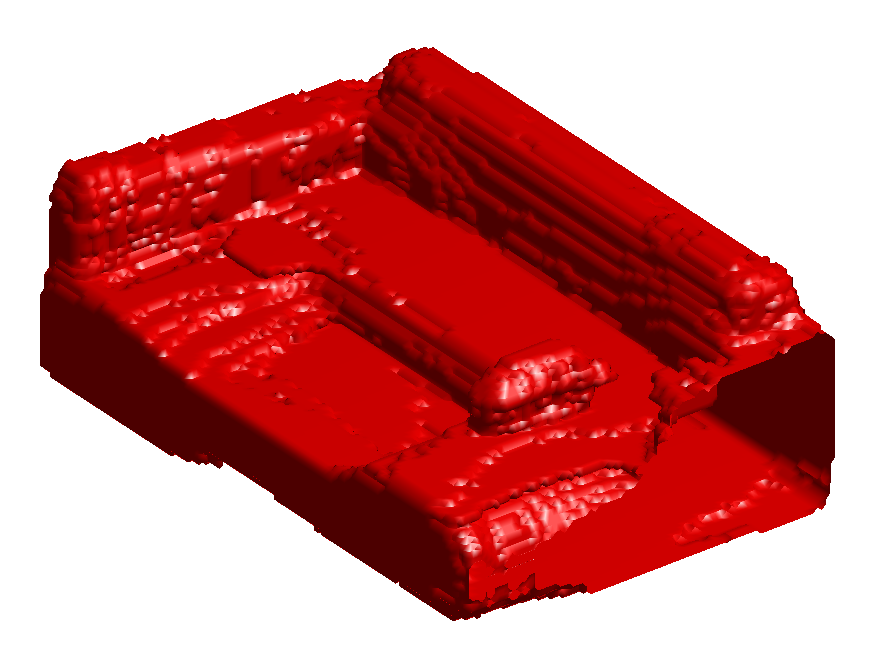}
\includegraphics[trim=25 0 25 0, clip, height=.15\linewidth]{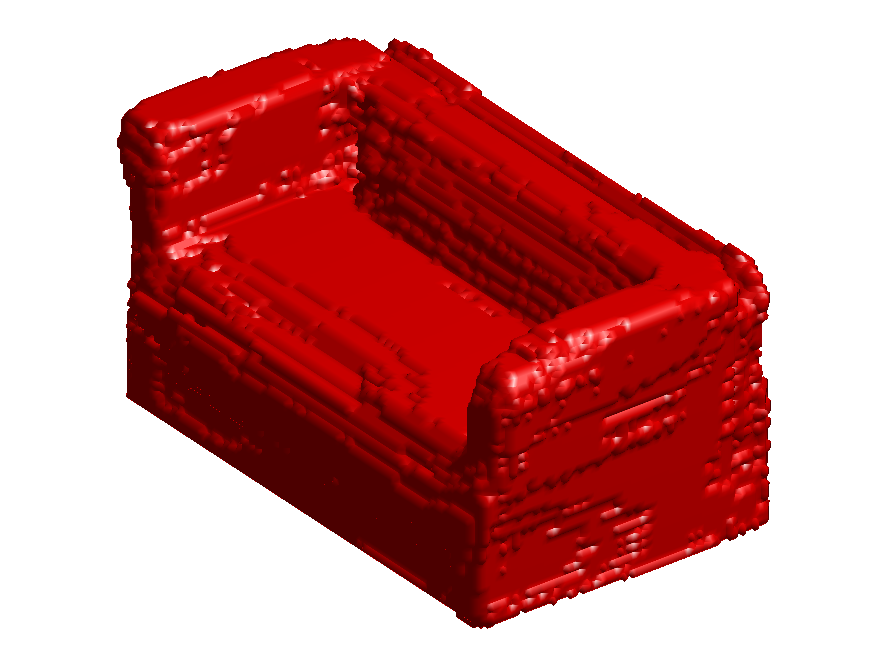}
\includegraphics[trim=25 0 25 0, clip, height=.15\linewidth]{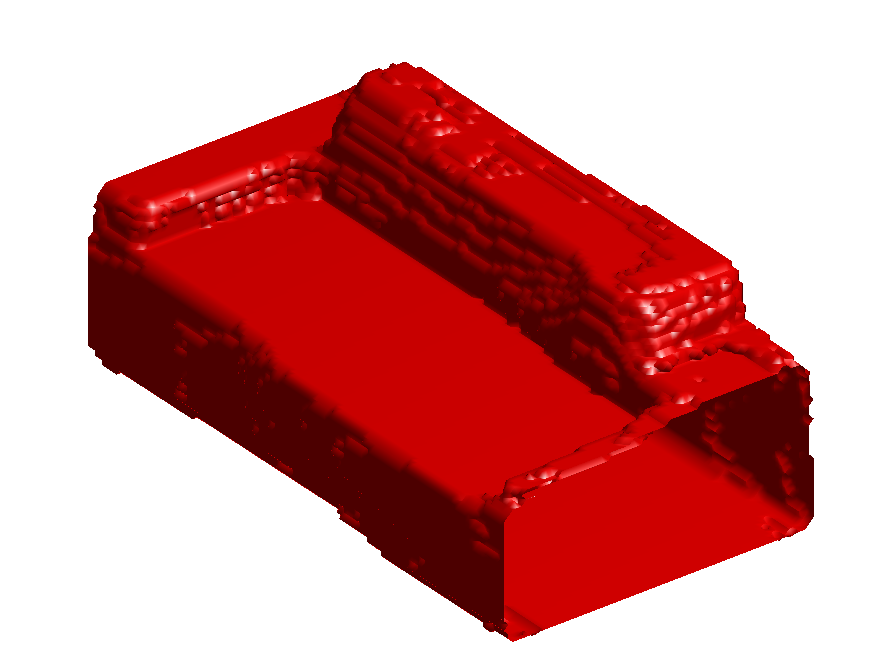}
\includegraphics[trim=25 0 25 0, clip, height=.15\linewidth]{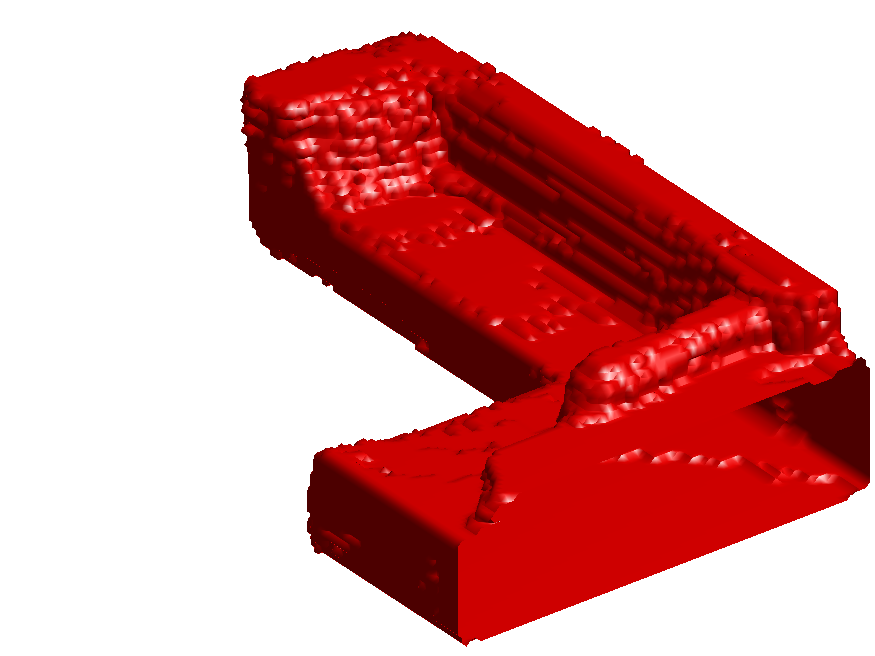}\\
(b) sofa \\
\caption{Synthesized 3D shapes at multi-grids by 3D multi-grid sampling. (a) toilet (b) sofa. From top to bottom: $16 \times 16 \times 16$ grid, $32 \times 32 \times 32$ grid, $64 \times 64 \times 64$ grid and $128 \times 128 \times 128$ grid. Synthesized example at each grid is obtained by 20 steps Langevin sampling initialized from the synthesized examples at the previous coarser grid, starting from the $1 \times 1 \times 1$ grid.}
\label{fig:mul}
\end{figure}

We conduct qualitative experiments to evaluate the proposed 3D multi-grid sampling algorithm on the toilet and sofa categories in ModelNet10 dataset. We learn the models at 5 grids: $4 \times 4 \times 4$, $16 \times 16 \times 16$, $32 \times 32 \times 32$, $64 \times
64 \times
64$ and $128 \times
128 \times
128$, which we refer to as grid1, grid2, grid3, grid4 and grid5, respectively. The training 3D shapes are of size $128 \times 128 \times 128$ voxels. Since the models of the 5 grids act on 3D shapes of different resolutions, we design a specific network structure per grid: grid1 has a 2-layer network, where the first layer has 128 $4 \times 4 \times 4$ filters with sub-sampling of 1 and the second layer is a fully-connected layer with 1 channel output; grid2 has a 2-layer network, where the first layer has 256 $8 \times 8 \times 8$ filters with a sub-sampling factor 2 and the second layer is a fully-connected layer with one single filter; grid3 has a 3-layer network, where the first layer has 256 $16 \times 16 \times 16$ filters with a sub-sampling factor 3, the second layer has 128 $6 \times 6 \times 6$ filters with a sub-sampling factor 2, and the third layer is a fully-connected layer with one single filter; grid4 has the same network structure as that of grid3; grid5 has a similar 3-layer network as that of grid4, except that the sub-sampling factor in the first layer is 4 and the filters in the second layer are of size $8 \times 8 \times 8$. For all grids, ReLU layers are added between adjacent layers. At each iteration, we run $K = 20$ steps of Langevin dynamics for each grid with a step size $\Delta {\tau} = 0.01$. All networks are trained simultaneously with mini-batches of size 40 and an initial learning rate of 0.001. Figure \ref{fig:mul} displays the synthesized 3D shapes at different grids for toilet and sofa categories in ModelNet10 dataset. For each category, we show 5 examples of the 3D multi-grid sampling results. Each column corresponds to one example, in which synthesized results shown from top to bottom are generated by models from low resolution grid (i.e., $16 \times 16 \times 16$ voxels) to high resolution grid (i.e., $128 \times 128 \times 128$ voxels), respectively. 

\begin{figure}
\centering
\includegraphics[trim=40 0 40 0, clip, height=.17\linewidth]{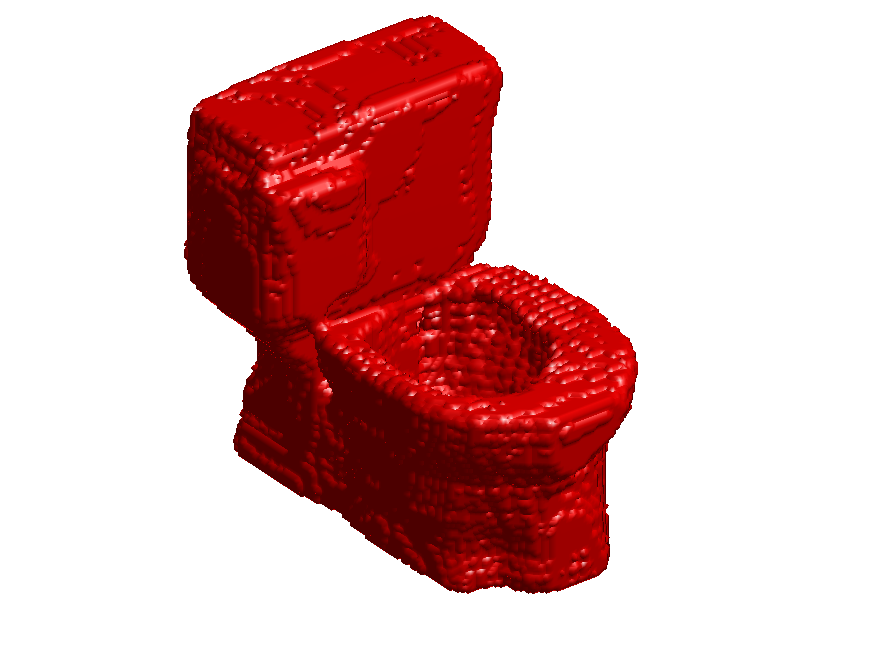} 
\includegraphics[trim=40 0 40 0, clip, height=.17\linewidth]{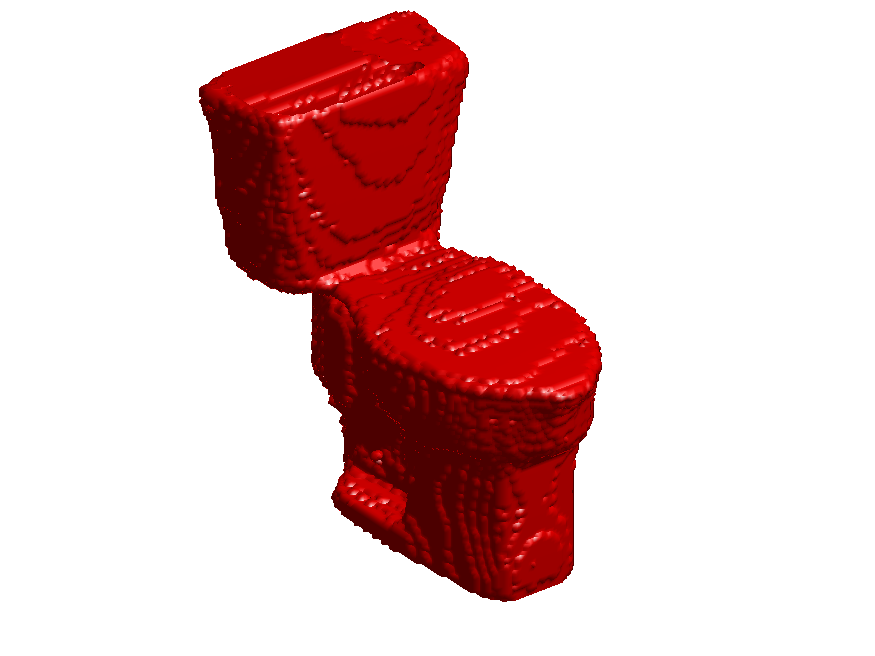} 
\includegraphics[trim=40 0 40 0, clip, height=.17\linewidth]{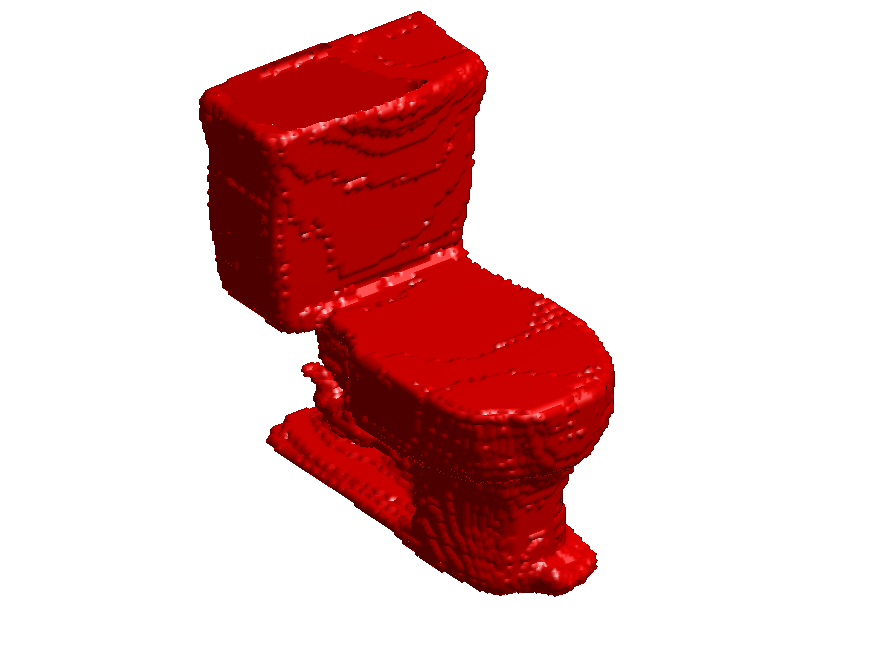} 
\includegraphics[trim=40 0 40 0, clip, height=.17\linewidth]{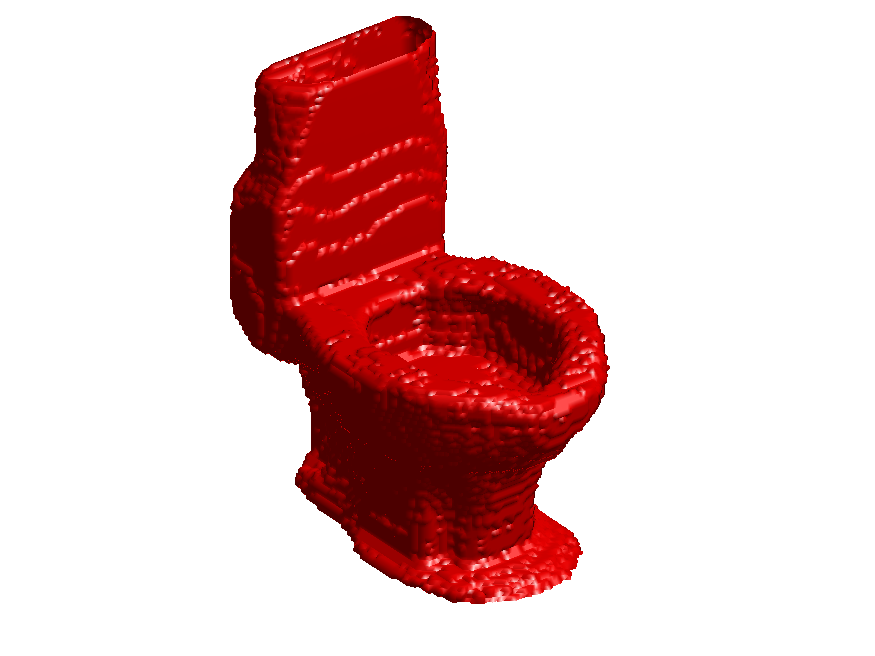}
\includegraphics[trim=40 0 40 0, clip, height=.17\linewidth]{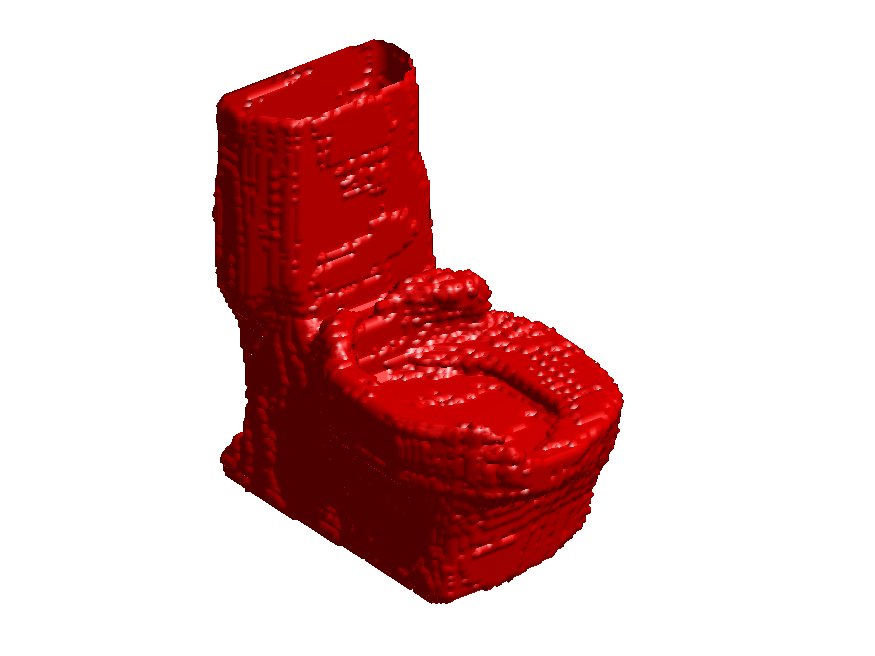} \\
\includegraphics[trim=40 0 40 0, clip, height=.17\linewidth]{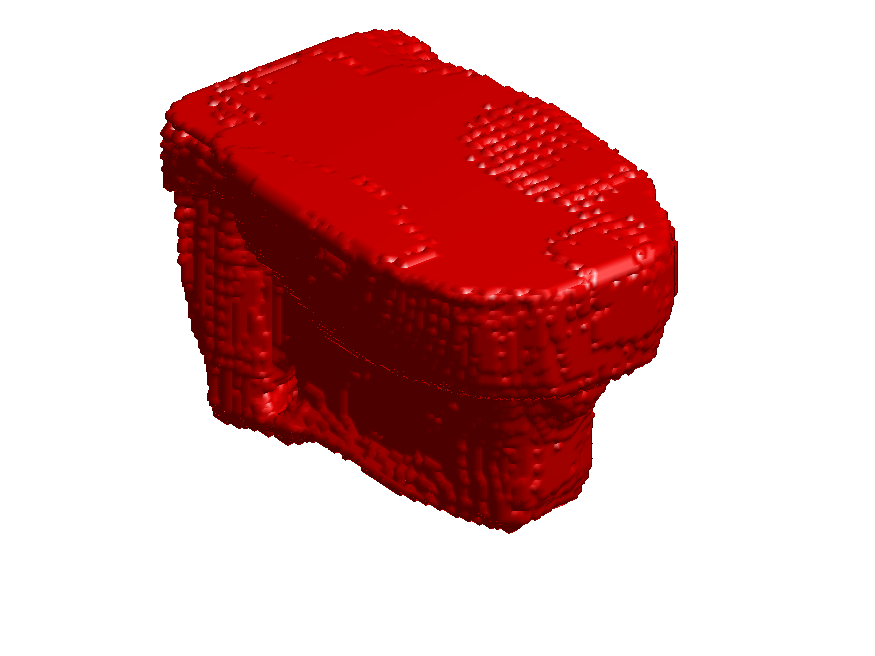} 
\includegraphics[trim=40 0 40 0, clip, height=.17\linewidth]{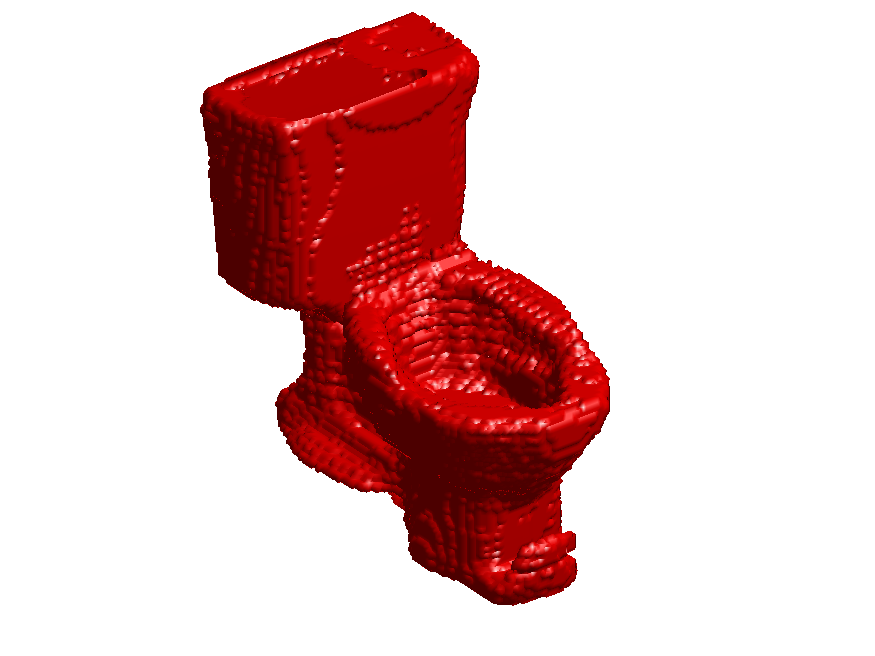} 
\includegraphics[trim=40 0 40 0, clip, height=.17\linewidth]{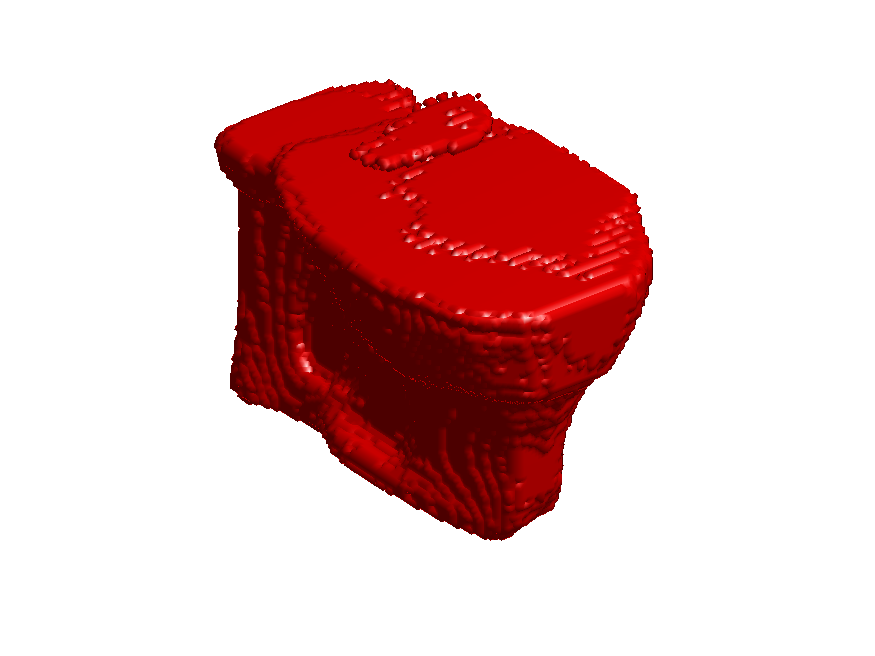} 
\includegraphics[trim=40 0 40 0, clip, height=.17\linewidth]{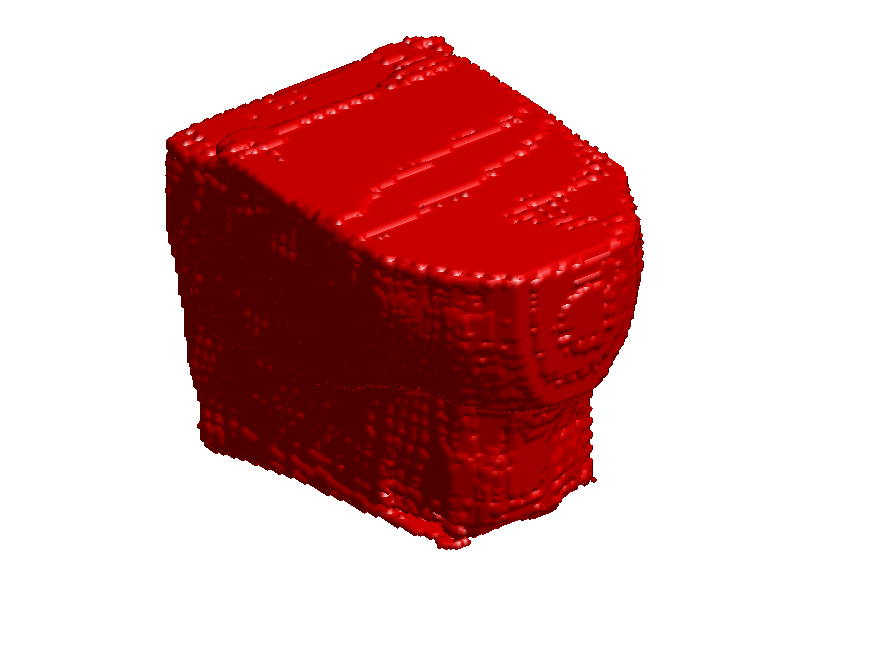}
\includegraphics[trim=40 0 40 0, clip, height=.17\linewidth]{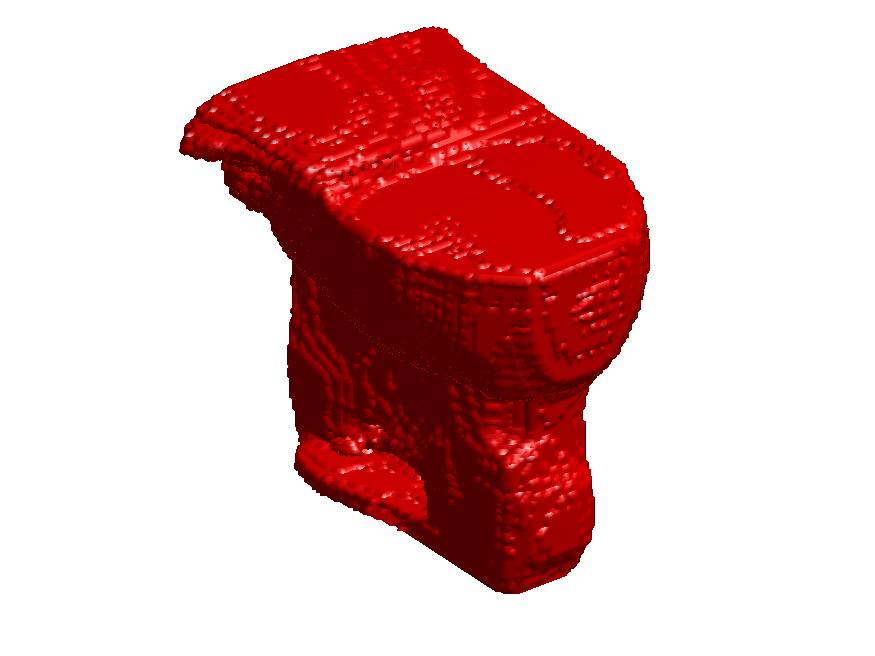}\\
(a) toilet \\ \vspace{1mm}
\includegraphics[trim=45 0 40 0, clip, height=.15\linewidth]{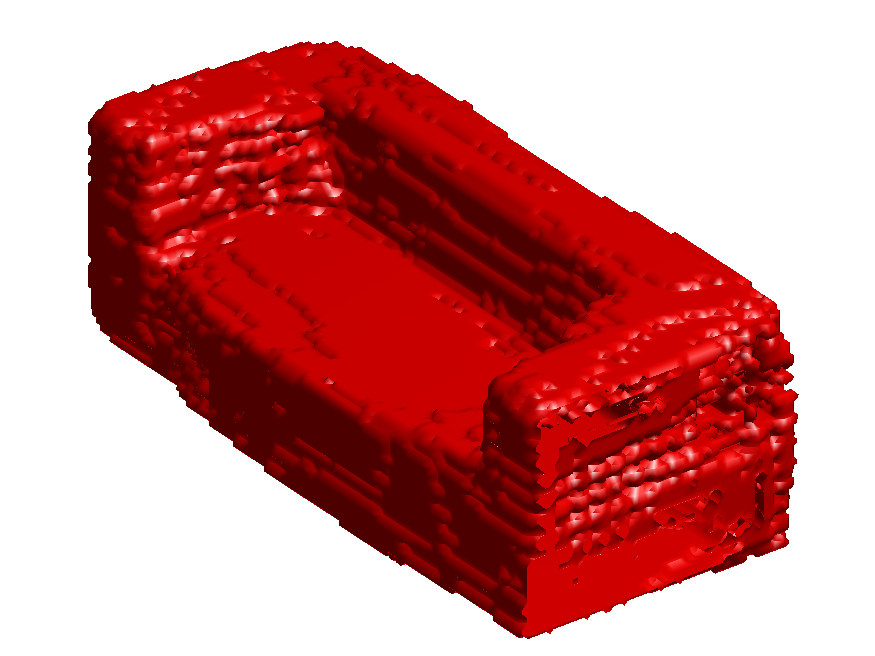} 
\includegraphics[trim=45 0 23 0, clip, height=.15\linewidth]{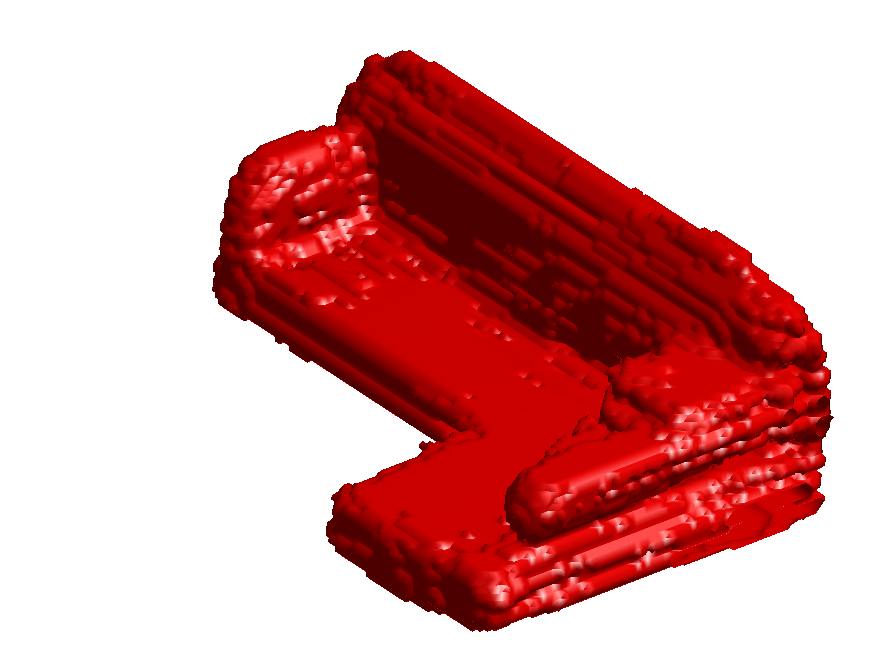} 
\includegraphics[trim=25 0 25 0, clip, height=.15\linewidth]{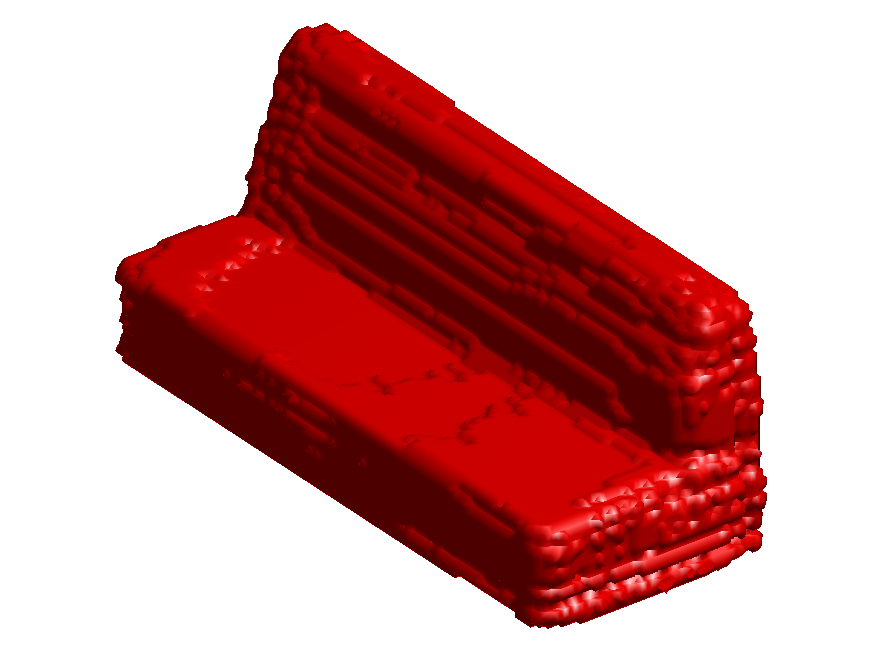} 
\includegraphics[trim=25 0 25 0, clip, height=.15\linewidth]{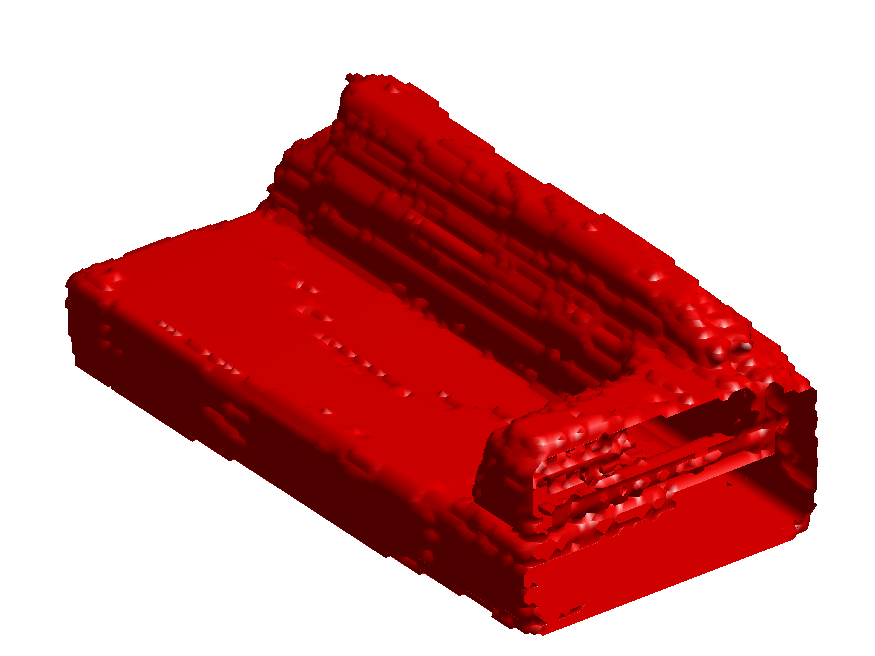}
\includegraphics[trim=25 0 8 0, clip, height=.14\linewidth]{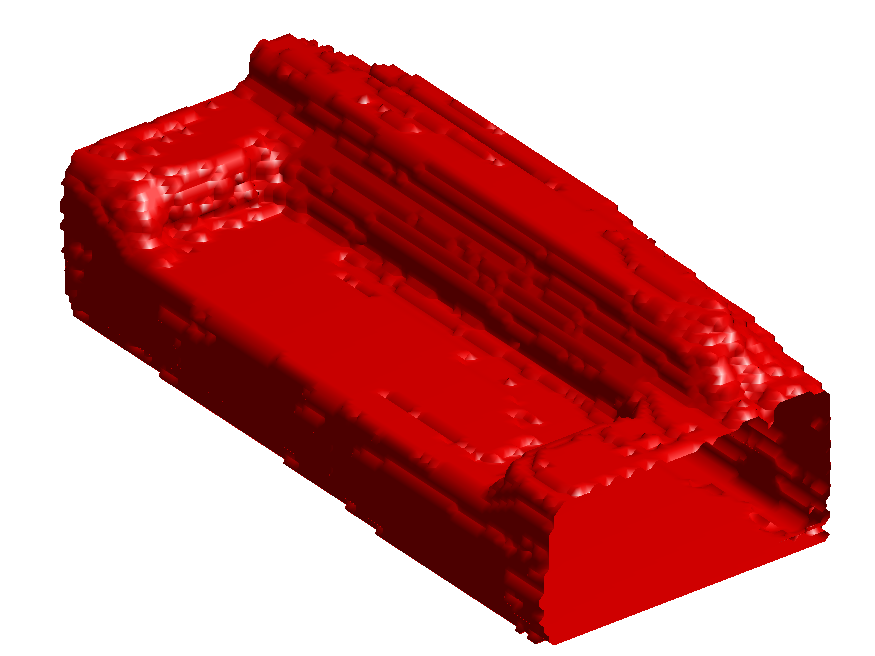}\\
\includegraphics[trim=45 0 40 0, clip, height=.15\linewidth]{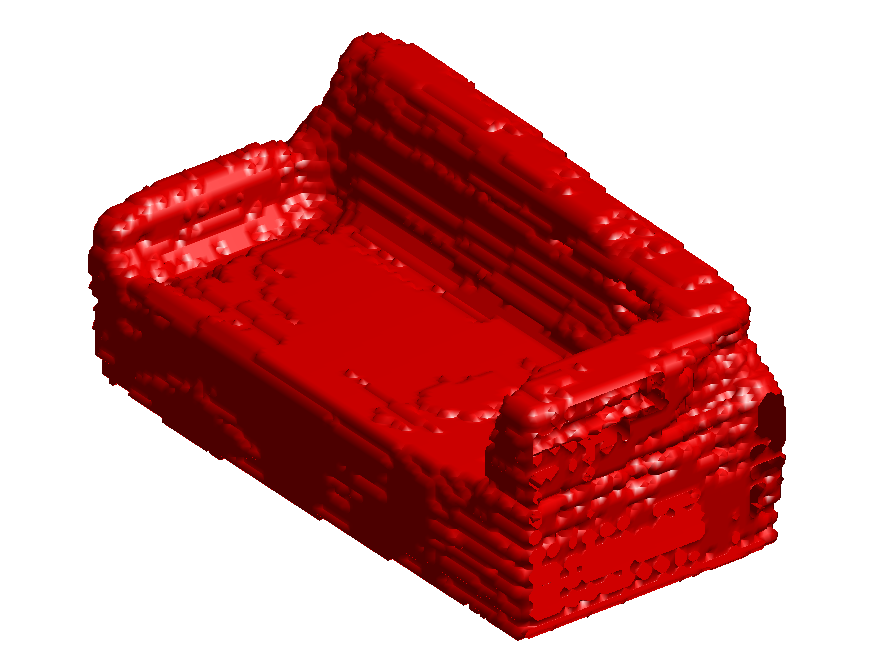} 
\includegraphics[trim=35 0 25 0, clip, height=.14\linewidth]{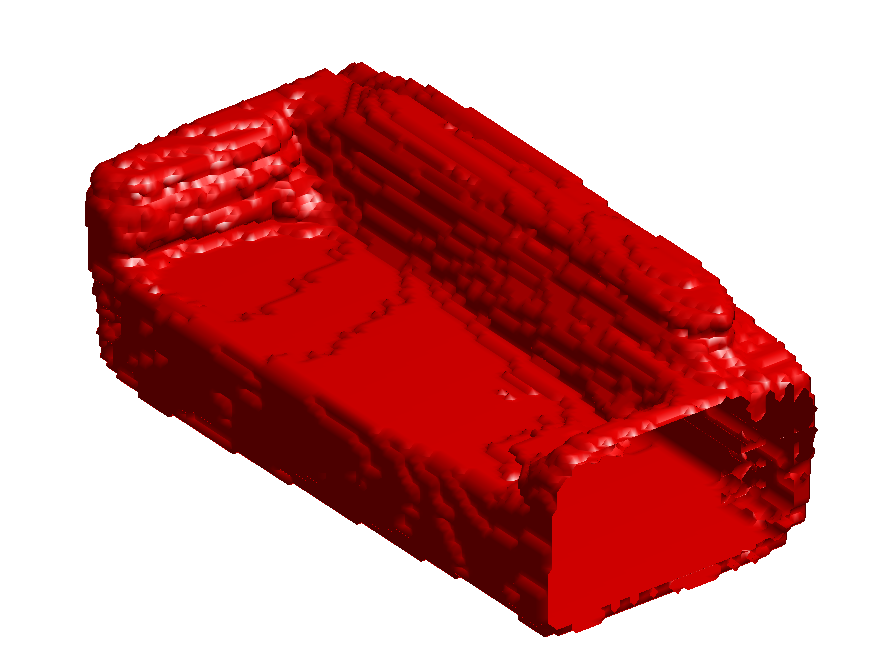} 
\includegraphics[trim=15 0 18 0, clip, height=.14\linewidth]{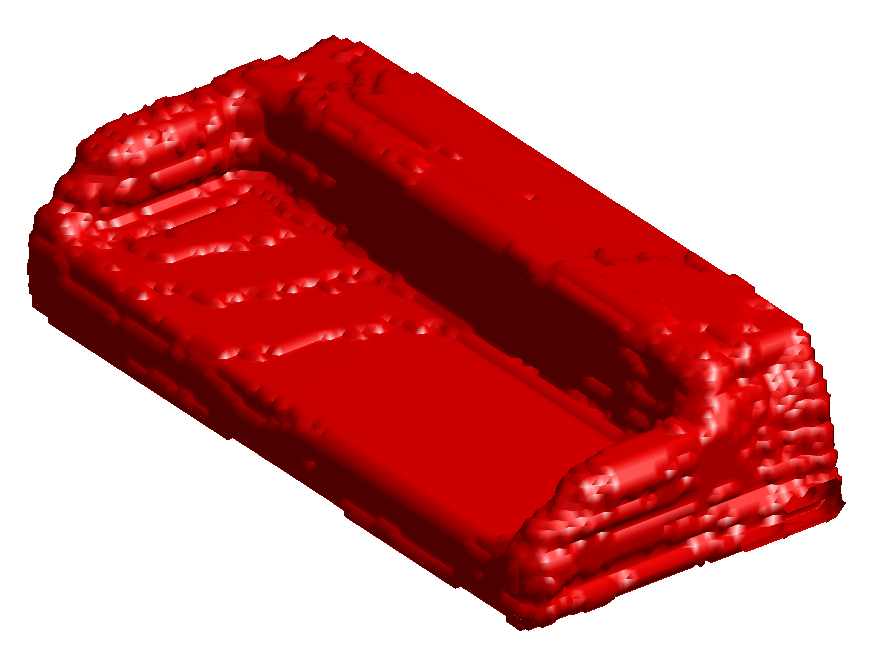} 
\includegraphics[trim=25 0 25 0, clip, height=.15\linewidth]{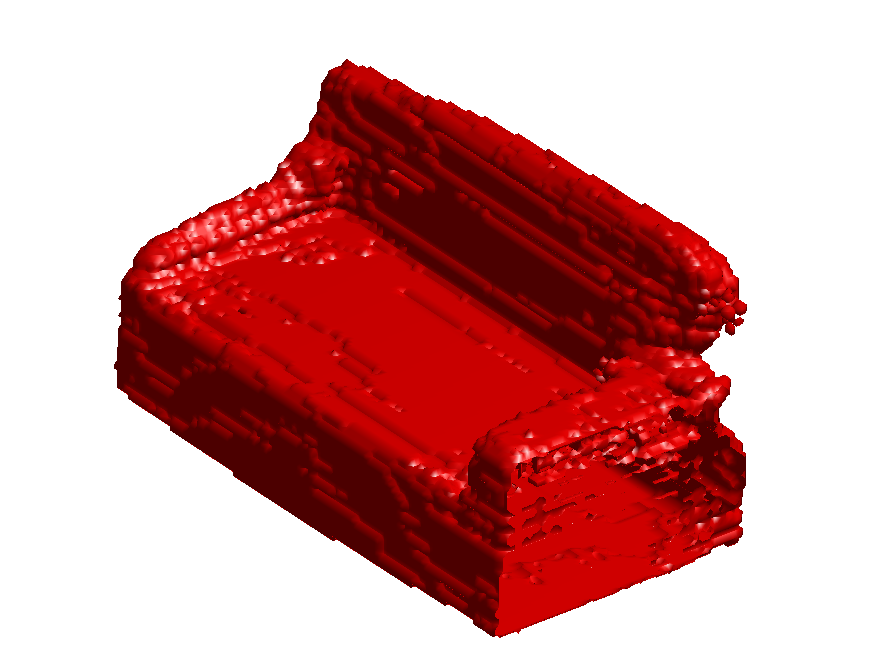}
\includegraphics[trim=25 0 25 0, clip, height=.15\linewidth]{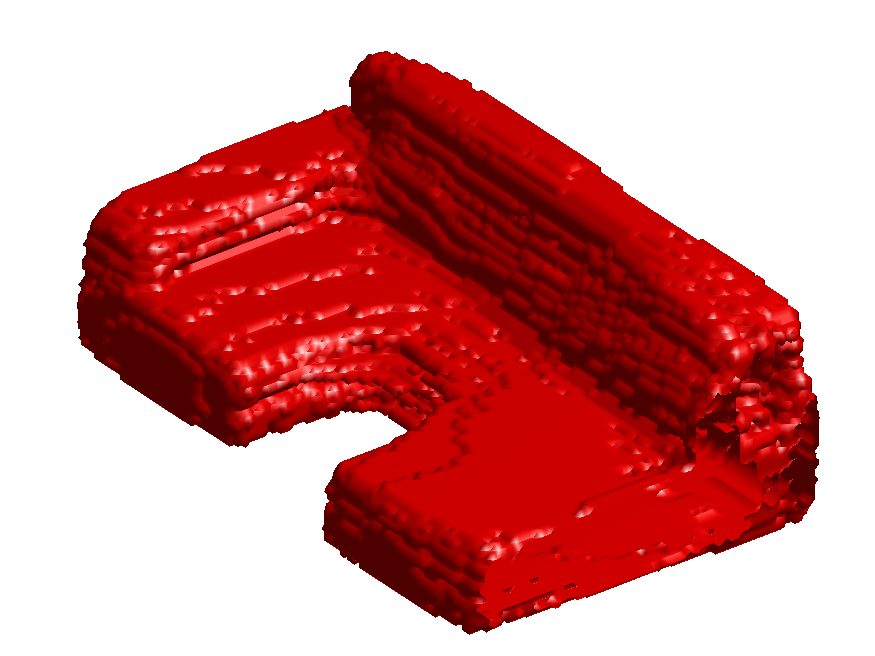}\\
(b) sofa \\
\caption{Generating high resolution 3D objects with $128 \times 128 \times 128$ voxels by the generative VoxelNet. (a) toilet (b) sofa.}
\label{fig:ebm128}
\end{figure}

For comparison purpose, we also train a generative VoxelNet with a high resolution of $128 \times 128 \times 128$ voxels as a single-grid baseline. The model has a 5-layer network, where the first four layers have filter sizes $\{16, 8, 4, 4\}$, sub-sampling factors $\{4, 2, 2, 2\}$, and numbers of output channels $\{128, 128, 256, 512\}$, and the last layer is a fully-connected layer with one filter. The batch size is 20. We run $K=20$ Langevin steps with a step size $\Delta \tau=0.0025$.
Figure \ref{fig:ebm128} displays some synthesized examples with a resolution of $128 \times 128 \times 128$ voxels from toilet and sofa categories. To quantitatively compare the quality of the 3D shapes generated by the multi-grid sampling strategy and the original one, we calculate the ``Fréchet Inception Distance'' \cite{heusel2017gans} (FID) score based on a reference network, which is a state-of-the-art discriminative neural network  for 3D object classification\cite{qi2016volumetric}. FID is a metric that computes the distance between feature vectors extracted from observed and synthesized 3D objects by the reference network,
\begin{eqnarray} 
\text{FID}=||\tilde{\mu} - \mu ||^2 + \text{Tr}\left(  \tilde{\Sigma} + \Sigma - 2(\tilde{\Sigma} \Sigma)^{1/2}\right) , \nonumber 
\end{eqnarray} 
where $V \sim \mathcal{N}(\mu, \Sigma)$ and $\tilde{V} \sim \mathcal{N}(\tilde{\mu}, \tilde{\Sigma})$ are the feature vectors of the reference network for observed and generated samples respectively. A lower FID score means better quality of the synthesized 3D objects.  We compares the multi-grid sampling strategy with the original one using single-grid sampling on synthesizing  3D shapes with a resolution of $128 \times 128 \times 128$ voxels in Table \ref{exp:fid_128}, where FID scores and computation times (in seconds) are reported. We generate 300 examples to compute the FID score.  Although the computation time for each training epoch of the multi-grid model is longer than that of the single-grid one, the former model usually needs much less epochs to converge and is more stable than the latter one. In general, the multi-grid method outperforms the single-grid one in terms of sampling stability, computational efficiency, and synthesis quality.

The experiment demonstrates that the proposed multi-grid sampling method can learn realistic models of 3D objects in a coarse-to-fine scheme, and  we hope that this experiment will push forward the development of efficient MCMC algorithms for learning high resolution deep 3D volumetric energy-based models. 

\begin{table}[h]
\caption{A comparison of Fréchet Inception Distance (FID) scores and computation times (seconds per epoch $\times$ number of epochs) between single-grid and multi-grid models on generating $128 \times 128 \times 128$ 3D shapes. The running times were recorded in a PC with an
Intel i7-6700k CPU and a Titan Xp GPU.}\label{exp:fid_128}
\centering
\begin{tabular}{|c|cc|cc|}
\hline
      \multirow{2}{*}{Method}   & \multicolumn{2}{c|}{FID $\downarrow$} & \multicolumn{2}{c|}{Time $\downarrow$ (sec / epoch $\times$ $\sharp$ of epochs)} \\ \cline{2-5} 
             & toilet & sofa & toilet & sofa \\ \hline \hline
     single-grid & 18.477 & 15.551 & 348.96$\times$1240  & 632.04$\times$680\\ \hline      
      
      multi-grid &10.850 & 7.857 & 792.90$\times$280 & 1650.12$\times$70 \\ \hline
\end{tabular}
\end{table}

\section{Conclusion}
We propose the generative VoxelNet model for volumetric object synthesis, and the conditional generative VoxelNet for 3D object recovery and 3D object super resolution. The proposed model is a deep energy-based model parameterized by a VoxelNet, which can be trained by an ``analysis by synthesis'' scheme. The training of the model can be interpreted as a mode seeking and mode shifting process, and the zero temperature limit has an adversarial interpretation. We also propose a 3D multi-grid energy-based modeling and sampling algorithm to train multi-scaled generative VoxelNets for high resolution 3D shape synthesis. The generative VoxelNet can also be useful in training a 3D generator network via MCMC teaching. The 3D generator trained in such a cooperative scheme is capable of synthesizing meaningful 3D shapes and 3D shape embedding. Experiments demonstrate that our models are able to generate realistic 3D shape patterns and are useful for 3D shape analysis, such as 3D shape recovery, 3D shape super resolution, and 3D shape classification.

\ifCLASSOPTIONcompsoc
  \section*{Acknowledgments}
\else
  \section*{Acknowledgment}
\fi

 The work is supported by NSF DMS-2015577, DARPA SIMPLEX N66001-15-C-4035, ONR MURI N00014-16-1-2007, DARPA ARO W911NF-16-1-0579, DARPA N66001-17-2-4029, and XSEDE grant ASC180018. We gratefully acknowledge the support of NVIDIA Corporation with the donation of the Titan Xp GPU used for this research. We thank Erik Nijkamp for his help with coding. We thank Siyuan Huang for helpful discussions.   

\ifCLASSOPTIONcaptionsoff
  \newpage
\fi

\bibliographystyle{IEEEtran}
\bibliography{mybibfile}

\end{document}